\documentclass[review]{elsarticle}

\usepackage{color}
\usepackage{subfigure}
\usepackage{bm}
\usepackage{amsmath}
\usepackage{amssymb}
\usepackage{algorithm} 
\usepackage{algorithmic} 
\usepackage{float}
\usepackage{stfloats}
\usepackage{algorithm} 
\usepackage{algorithmic} 
\usepackage{tabularx}
\usepackage{multirow}
\usepackage{subfigure}
\usepackage{pifont}
\usepackage{epstopdf}

\biboptions{numbers,sort&compress}
\usepackage{amsmath}

\usepackage{lineno,hyperref}
\modulolinenumbers[5]

\journal{Pattern Recognition}

\begin{document}

\begin{frontmatter}

\title{Hypergraph Modelling for Geometric Model Fitting}

\author[1]{Guobao Xiao}
\ead{x-gb@163.com}
\author[1]{Hanzi Wang\corref{correspondingauthor}}
\cortext[correspondingauthor]{Corresponding author. Tel./fax: +86 5922580063.}
\ead{wang.hanzi@gmail.com}
\author[1]{Taotao Lai}
\ead{laitaotao@gmail.com}
\author[2]{David Suter}
\ead{david.suter@adelaide.edu.au}
\address[1]{Fujian Key Laboratory of Sensing and Computing for Smart City, School of Information Science and Engineering, Xiamen University, China}
\address[2]{School of Computer Science, The University of Adelaide, Australia}



\begin{abstract}
In this paper, we propose a novel hypergraph based method {(called HF)} to fit and segment multi-structural data. {The proposed HF formulates the geometric model fitting problem as a hypergraph partition problem based on a novel hypergraph model.} In the hypergraph model, vertices represent data points and hyperedges denote model hypotheses. The hypergraph, with large and ``data-determined'' {degrees} of hyperedges, can express the complex relationships {between} model hypotheses and data points. In addition, we develop a robust hypergraph partition algorithm to detect sub-hypergraphs for model fitting. HF can effectively and efficiently estimate the number {of,} and the parameters of{,} model instances in multi-structural data heavily corrupted with outliers simultaneously. Experimental results show the advantages of the proposed method over previous methods on {both} synthetic data and real images.
\end{abstract}


\end{frontmatter}


\section{Introduction}
\begin{figure}[t]
\centering
 \centerline{\includegraphics[width=0.75\textwidth]{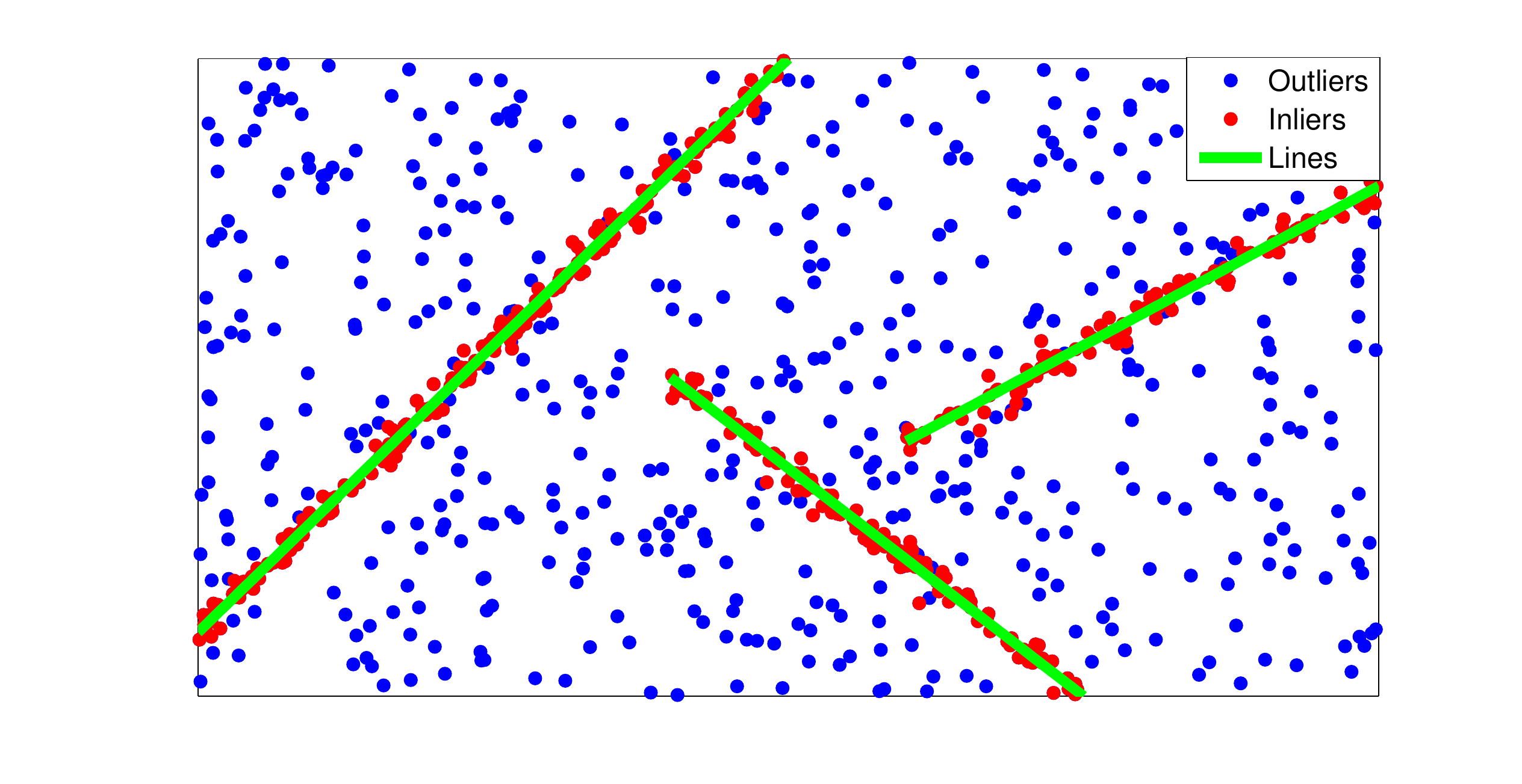}}
\caption{An example of the problem we addressed in this paper.}
\label{fig:addressproblem}
\end{figure}
Many computer vision tasks (such as optical flow calculation, motion segmentation{,} homography or fundamental matrix estimation) use robust statistical methods. To illustrate the problem addressed in this paper, we {first} consider {a} standard line fitting problem. As shown in Fig.~\ref{fig:addressproblem}, we {deal with data with} an unknown number of {model instances (lines) and} unknown ratios of inliers and outliers. We adopt the paradigm that these ``structures" (i.e., lines) can be estimated by a {robust} fitting method. Of course, the ``structures" are not restricted to lines{. We can} also consider {as ``structures":  homographies, fundamental matrixes}, etc.

A number of robust {fitting} methods \cite{Haifeng2003robust,pearl:ijcv12,mittal2012generalized,pham2014random} have been proposed in recent years. The main steps of these robust {fitting} methods include: (1) Generate putative hypotheses; (2) Verify these hypotheses according to a robust criterion; (3) Output the estimated model parameters corresponding to the best verified hypothesis~\cite{Wong20131755}. One of the most popular robust methods is RANSAC \cite{fischler1981random}, which can deal with a large number of outliers effectively. However, RANSAC is {originally} designed to fit data involving a single model and {it requires a} user-specified threshold to dichotomize inliers from outliers. Some {other} methods have been proposed to enhance RANSAC, e.g., LO-RANSAC \cite{chum2003locally}, PROSAC \cite{chum2005matching}, Cov-RANSAC \cite{raguram2009exploiting}, and QDEGSAC \cite{frahm2006ransac}. In addition, we note that some recently proposed robust fitting methods {(}such as J-linkage \cite{toldo2008robust}, KF \cite{chin2009robust}, PEARL \cite{pearl:ijcv12}, AKSWH \cite{wang2012simultaneously} and T-linkage~\cite{Magri_2014_CVPR}{)} claim that they are able to robustly fit models under severe noise. These fitting methods have their own advantages, but {the} fitting results {are} still far from being perfect for many real-world problems, due to the limitations of speed or accuracy.

Recently, hypergraphs have been introduced to {solve} some computer vision tasks, e.g., \cite{parag2011supervised,jain2013efficient,liu2012efficient,ochs2012higher,pulak2014clustering}. A hypergraph contains higher order similarities instead of pairwise similarities, which {can} be beneficial to overcome the above-mentioned limitations. In the hypergraph, each vertex represents a data point and each hyperedge connects a small group of vertices. Parag and Elgammal \cite{parag2011supervised} proposed an effective data point labeling method in which the data point labeling problem is equivalent to the hypergraph labeling problem. Liu and Yan \cite{liu2012efficient} proposed to use a random consensus graph to efficiently fit structures in data. Ochs and Brox \cite{ochs2012higher} developed a method based on spectral clustering on hypergraphs for motion segmentation. Jain and Govindu \cite{jain2013efficient} extended higher-order clustering to plane segmentation using hypergraphs for RGBD data.

However, the previous hypergraph based works \cite{parag2011supervised,jain2013efficient,liu2012efficient,ochs2012higher} only {consider} the {\em smallest} {justified} degrees of hyperedges (for a hyperedge, its degree is defined as the number of vertices connected by the hyperedge) {in} a {\em uniform} hypergraph. {That is, the hyperedges of a hypergraph in these works only connect a small number of vertices.} {Using large degrees of hyperedges can yield better clustering accuracy because it can gather more information of the relationships between vertices; which has been demonstrated by the theoretical analysis and comprehensive experiments in~\cite{pulak2014clustering}.} From this aspect, the {\em largest} possible degrees (i.e., all vertices that represents inlier data points belonging to the same structure are connected by the same hyperedge) would be {the} best. But {there are two potential problems:} (i) Complexity might be an issue; (ii) Determining all inliers can be difficult. In this paper, we show that a tractable method can be devised. Obviously, since the number of inliers associated to each structure is different, the degree of each hyperedge is varying. Therefore, we devise a scheme {having the following} features: the degree of a hyperedge is ``data-driven" ({hence}, varying), and much larger ({as close as possible to include all inliers}) than the one {used} in the previous hypergraph based works~\cite{parag2011supervised,jain2013efficient,liu2012efficient,ochs2012higher,pulak2014clustering}.

\begin{figure}[t]
\begin{minipage}{.45\textwidth}
\centering
\begin{tabular}[width=1.0\textwidth]{|c|c|c|c|c|}

\hline
& $e_1$ & $e_2$ & $e_3$ \\
\hline
$v_1$ & 1 & 0 & 0  \\
\hline
$v_2$ & 1 & 1 & 0  \\
\hline
$v_3$ & 0 & 1 & 0  \\
\hline
$v_4$ & 0 & 1 & 1  \\
\hline
$v_5$ & 0 & 0 & 1  \\
\hline
$v_6$ & 0 & 0 & 1  \\
\hline
$v_7$ & 1 & 0 & 1  \\
\hline
$v_8$ & 1 & 0 & 0  \\
\hline
\end{tabular}
\begin{center} (a)   \end{center}
\end{minipage}
\begin{minipage}{.55\textwidth}
\centerline{\includegraphics[width=1.2\textwidth]{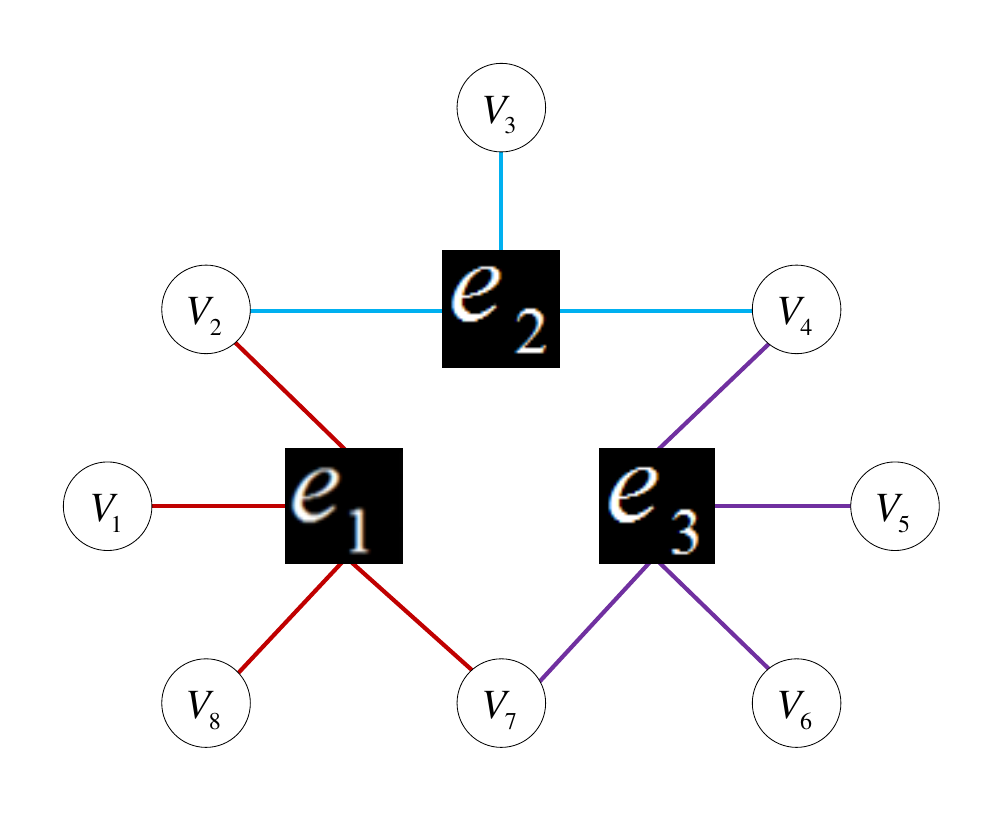}}
\medskip\medskip
 \centerline{(b)}\medskip
\end{minipage}
\caption{An example of hypergraph modelling. (a) A set of data points $V=\{v_1,v_2,v_3,\ldots,v_{8}\}$ and a set of model hypotheses $E=\{e_1,e_2,e_3\}$. The entry $(v_i,e_j)$ is set to $1$ if $v_i$ is an inlier data point of the model hypothesis $e_j$, and $0$ otherwise. (b) A hypergraph that expresses the complex relationships {between model hypotheses and data points}. The hypergraph includes $3$ hyperedges $\{e_1,e_2,e_3\}$ and $8$ vertices $\{v_1,v_2,v_3,\ldots,v_{8}\}$ which are respectively shown in rectangles and circles in (b).}
\label{fig:hypergraphexample}
\end{figure}

In this paper, we propose a robust hypergraph based fitting method (called HF), which formulates geometric model fitting as a hypergraph partition problem, to fit and segment multi-structural data with outliers. In the hypergraph, each vertex represents a data point and each hyperedge denotes a model hypothesis. HF can decide an appropriate degree of hyperedges in a simple and effective way for each hypergraph. More specifically, HF generates a set of potential hyperedges for a hypergraph, and expands these hyperedges {as close as possible to} the {\em largest} {justified} degrees using an inlier scale estimate, where the {\em largest} {justified} degree of a hyperedge is the number of the inlier data points belonging to the corresponding model hypothesis. Once a hyperedge is generated, HF can obtain the inlier data points belonging to the corresponding model hypothesis. Thus this hypergraph can effectively express the complex relationships {between} model hypotheses and data points, as shown in Fig.~\ref{fig:hypergraphexample}. To reduce complexity of a hypergraph, we prune the hypergraph by removing {some hyperedges with low weighting scores and the vertices without being connected by any remaining hyperedge.} In addition, we develop a robust hypergraph partition algorithm, which can effectively detect sub-hypergraphs with each sub-hypergraph representing a potential model instance in {the} data, for model fitting. Overall, HF can effectively and efficiently estimate the number {of,} and the parameters of{,} model instances in data{. The method can} deal with multi-structural data heavily corrupted with outliers. Experimental results on both synthetic and real data show that HF can achieve better results than the other competing methods.

{The proposed method (HF) has four main advantages over previous works}. First, the constructed {hypergraph in this paper} may include large and ``data-determined'' degrees of hyperedges (i.e., each hyperedge may connect a large and variable number of vertices in the {hypergraph}), which can yield better accuracy in hypergraph partitioning. Second, the {hypergraph} can be directly used to fit models, rather than constructing a pairwise affinity matrix by which the projection from a hypergraph to an induced graph may cause information-loss (shown in \cite{pu2012hypergraph,bulo2009game}). Third, HF can simultaneously detect all structures in multi-structural data, instead of using the sequential ``fit-and-remove'' procedure. {Fourth, HF clusters data points by the proposed sub-hypergraph detection algorithm, which does not totally depend on the inlier scale estimate to dichotomize inliers from outliers.} Thus the proposed method is more effective and computationally efficient over previous works.

{Note that the proposed HF uses some similar techniques to those used in AKSWH. However, HF has some significant differences to AKSWH: HF uses the inlier scale estimate to construct hypergraph modelling, and estimates model instances in data by using a novel sub-hypergraph detection algorithm, which introduces a new spectral clustering algorithm to deal with model fitting problems. In contrast, AKSWH estimates model instances by an agglomerative clustering algorithm. Benefiting from these improvements, HF achieves better results than AKSWH on both speed and accuracy. More importantly,  the performance of AKSWH is dependent on the accuracy of the inlier scale estimate. The reason behind this is that AKSWH derives the inlier and outlier dichotomy according to the corresponding inlier scales of the estimated model hypotheses, and it achieves bad results when residual values from model hypotheses to data points are very small and close. In contrast, for HF, although the constructed hypergraph is also based on the inlier scale estimate, data points are clustered by the proposed sub-hypergraph detection algorithm, whose performance is not very sensitive to the inlier scale estimate. This will be shown in the experiments for 3D-motion segmentation (see Sec.~\ref{sec:3Dmotionsegmentation}), where the residual values based on subspace clustering are very small and close, and thus inliers and outliers cannot be effectively distinguished by an inlier scale estimate for a parameter space based fitting method (e.g., AKSWH).}
 \begin{figure}[t]
\centering
\begin{minipage}[t]{.64\textwidth}
  \centering
 \centerline{\includegraphics[width=1.15\textwidth]{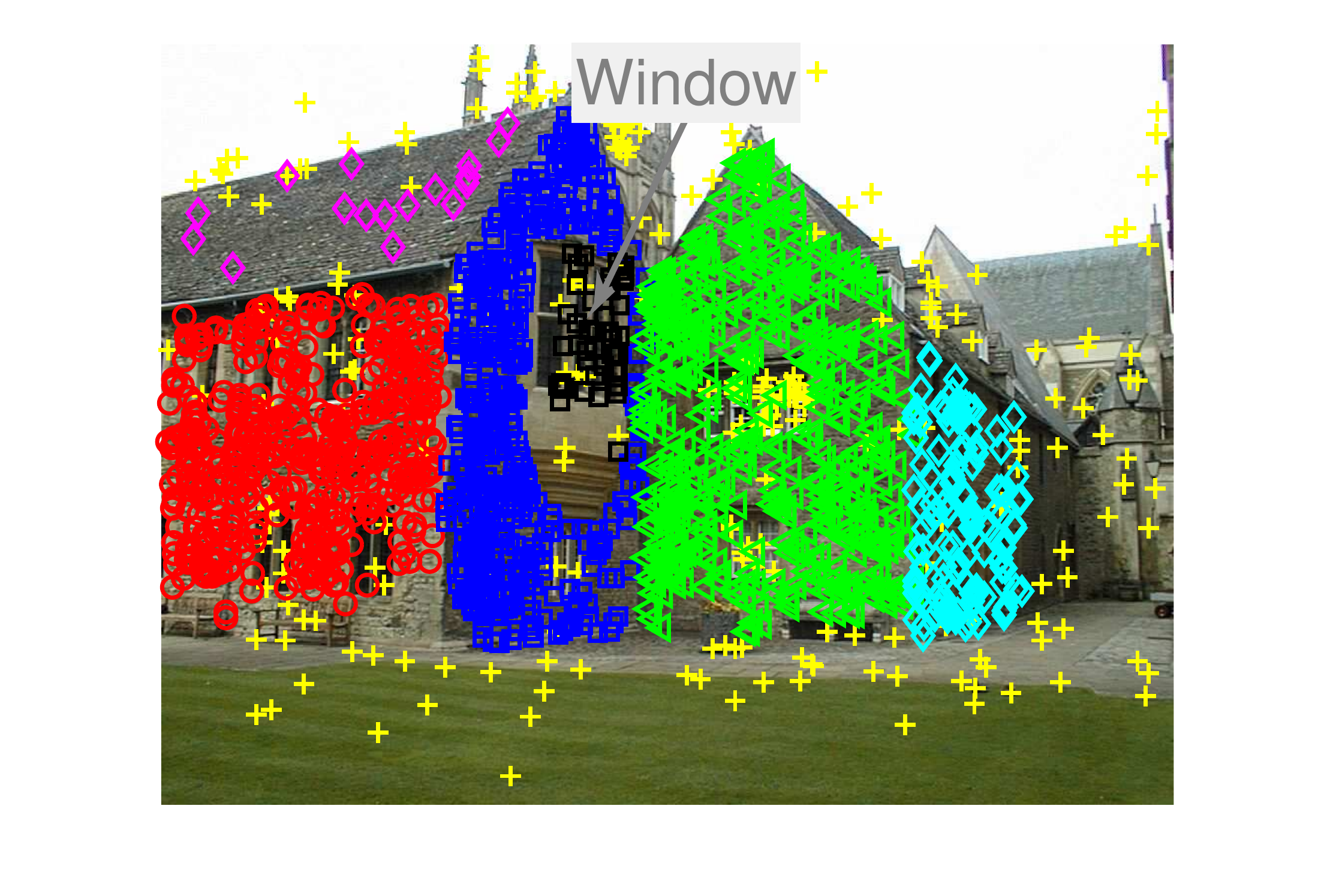}}
\end{minipage}
\hfill
\caption{An example of selecting {the ground truth} number of model instances in Merton College $3$, which may cause ambiguity. The data points belonging to the ``window" may be considered as one structure or outliers depending on different definitions of a structure.}
\label{fig:correctnumber}
\end{figure}

It is worth pointing out that: we aim to design a fitting method that cannot only estimate the parameters of model instances in data but also automatically decide the number of model instances as if {the ground truth number always has a well-defined meaning}. {For} synthetic data, {the ground truth number of model instances has a precise meaning since one can decide the number of model instances during designing synthetic data}. However, {for real images}, the {ground truth} number of model instances may be ambiguous. For example, as shown in Fig.~\ref{fig:correctnumber} (the ``Window" in Merton College $3$ from the Oxford Visual Geometry Group \footnote{\url{http://www.robots.ox.ac.uk/\~vgg/data/data-mview.html}}), ``what is important" is usually something to do with the size of the structures (size being either number of data items or spatial extent of those items or both) and also with the function of {objects}. Clearly, in such cases, the {ground truth} number of model instances in data can only be decided when one knows the purpose of {segmentation}. Of course, one can generally determine the number of model instances following some common (and heuristic) rules, e.g., selecting the most significant (larger in spatial extent, larger in data population, for example) model instances. Therefore, in this work we evaluate the proposed method and several other competing fitting methods according to the ground truth (usually this coincides with {``significant"} in the above senses) of standard {datasets}, though of course the decision as to what constitutes a structure is subjective and problem dependent. As a result, it is naive to expect (and even more to claim to have produced) a method that will always agree with every human judgement of what should be the {ground truth number of model instances} in all cases. We only claim that {the proposed method} usually extracts the right structures in some sense, and that this {generally} agrees with the ground truth of standard {datasets}.

The rest of the paper is organized as follows: In Sec.~\ref{sec:IIHG}, we construct hypergraphs for geometric model fitting. In Sec.~\ref{sec:sub-hypergraph detection}, we develop a novel hypergraph partition algorithm, and based on which, we propose a hypergraph based fitting method in Sec.~\ref{sec:overall algorithm}. In Sec.~\ref{sec:Experiment}, we present the experimental results obtained by the proposed method and several other competing methods on both synthetic and real data. We draw conclusions in Sec.~\ref{sec:conclusion}.

\section{Hypergraphs}
\label{sec:IIHG}
In this paper, {the} geometric model fitting {problem} is formulated as a hypergraph partition problem. Therefore, for each {dataset}, we construct a hypergraph to effectively express the relationships {between} model hypotheses and data points.
%
%

\subsection{Hypergraph Modelling}
\label{sec:hypergraphconstruction}
For hypergraph modelling, we regard each data point as a vertex and each model hypothesis as a hyperedge {in} a hypergraph $G=(V,E,\omega)$ (as shown in Fig.~\ref{fig:hypergraphexample}). Assume that there are $n$ data points and $m$ model hypotheses, and thus, the generated hypergraph contains $n$ vertices and $m$ hyperedges. Let $V=\{v_1,v_2,\ldots,v_n\}$ represent $n$ vertices and $E=\{e_1,e_2,\ldots,e_m\}$ denote $m$ hyperedges. We assign each hyperedge a positive weight value $\omega(e)$. When $v\in e$, the hyperedge $e$ is incident with vertex $v$. An $|V|\times|E|$ incident matrix $\mathbf{H}$, satisfying entries $h(v,e)=1$ if $v\in e$ and 0 otherwise, can be used to represent the relationships between vertices and hyperedges in the hypergraph $G$. 

We aim to construct a hypergraph to express the relationships {between} model hypotheses and data points, that is, we can directly determine whether a data point is one of inlier data points of a particular model hypothesis from a hypergraph. Therefore, to construct a hypergraph, we devise a scheme that contains two main parts, i.e., hyperedge generation and {hyperedge expansion}. (i) {\em Hyperedge Generation}. We firstly sample a number of minimal subsets, based on which we estimate model hypotheses using the Direct Linear Transformation algorithm~\cite{hartley2003multiple}. A minimal subset is composed {of} the minimum number of data points, which are necessary to estimate a model hypothesis (e.g., 2 data points for line fitting and 4 data points for homography fitting). Since each hypothesis is associated to a hyperedge, we can directly generate a number of potential hyperedges, which connect {the vertices that correspond to the associated minimal sampled subsets}.
(ii) {\em {Hyperedge Expansion}}. We expand each potential hyperedge to connect as more vertices as possible---that is, we expand the degree of each potential hyperedge {as close as possible to} the largest {justified} value. We do this by using a robust inlier noise scale estimator. Scale estimation is very important for robust model fitting because that it can be used to dichotomize inliers from outliers. Accordingly, we use it to determine whether a hyperedge $e$ is incident with a vertex $v$ or not. Based on this, each hyperedge is expanded from the smallest {justified} degree (i.e., {each hyperedge only connects the vertices that correspond to its minimal sampled subset) as close as possible to} the largest {justified} degree (i.e., {each hyperedge connects the vertices that correspond to} all the inlier data points decided by using the inlier scale estimator). In this paper, we adopt the Iterative Kth Ordered Scale Estimator (IKOSE) proposed in \cite{wang2012simultaneously} to estimate the inlier scale (here, we adopt IKOSE because of its efficiency and simplicity of implementation. Of course, we can also adopt other inlier noise scale estimators, e.g., ALKS~\cite{lee1998robust}, MSSE~\cite{bab1999robust} and TSSE~\cite{wang2004robust}, instead of IKOSE).
\begin{figure}[t]
\centering
\begin{minipage}[t]{.32\textwidth}
 \centering
  \centerline{\includegraphics[width=1.0\textwidth]{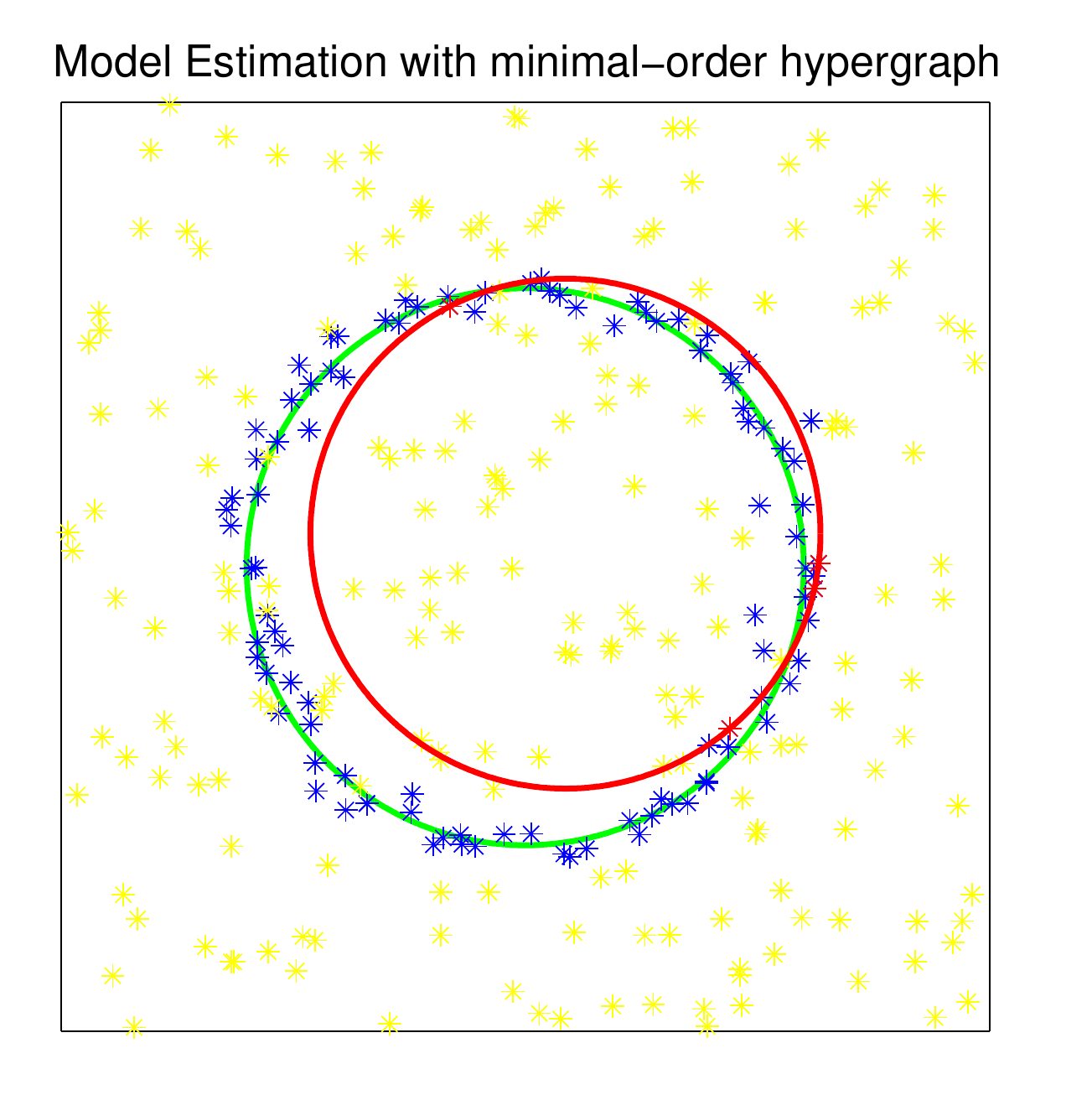}}
  \begin{center} (a)   \end{center}
\end{minipage}
\begin{minipage}[t]{.32\textwidth}
  \centering
   \centerline{\includegraphics[width=1.0\textwidth]{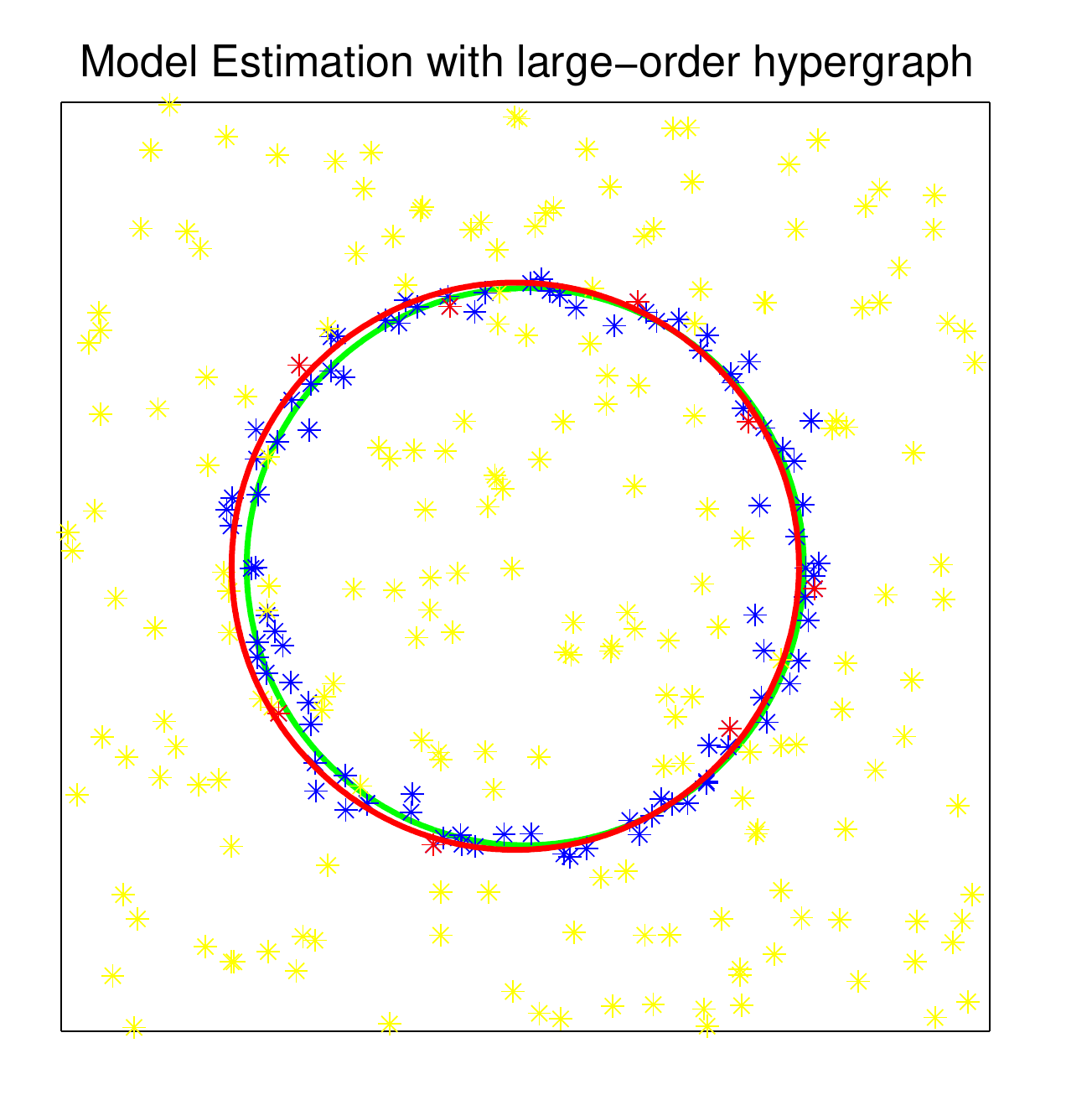}}
  \centerline{(b) }\medskip
\end{minipage}
\begin{minipage}[t]{.32\textwidth}
  \centering
 \centerline{\includegraphics[width=1.0\textwidth]{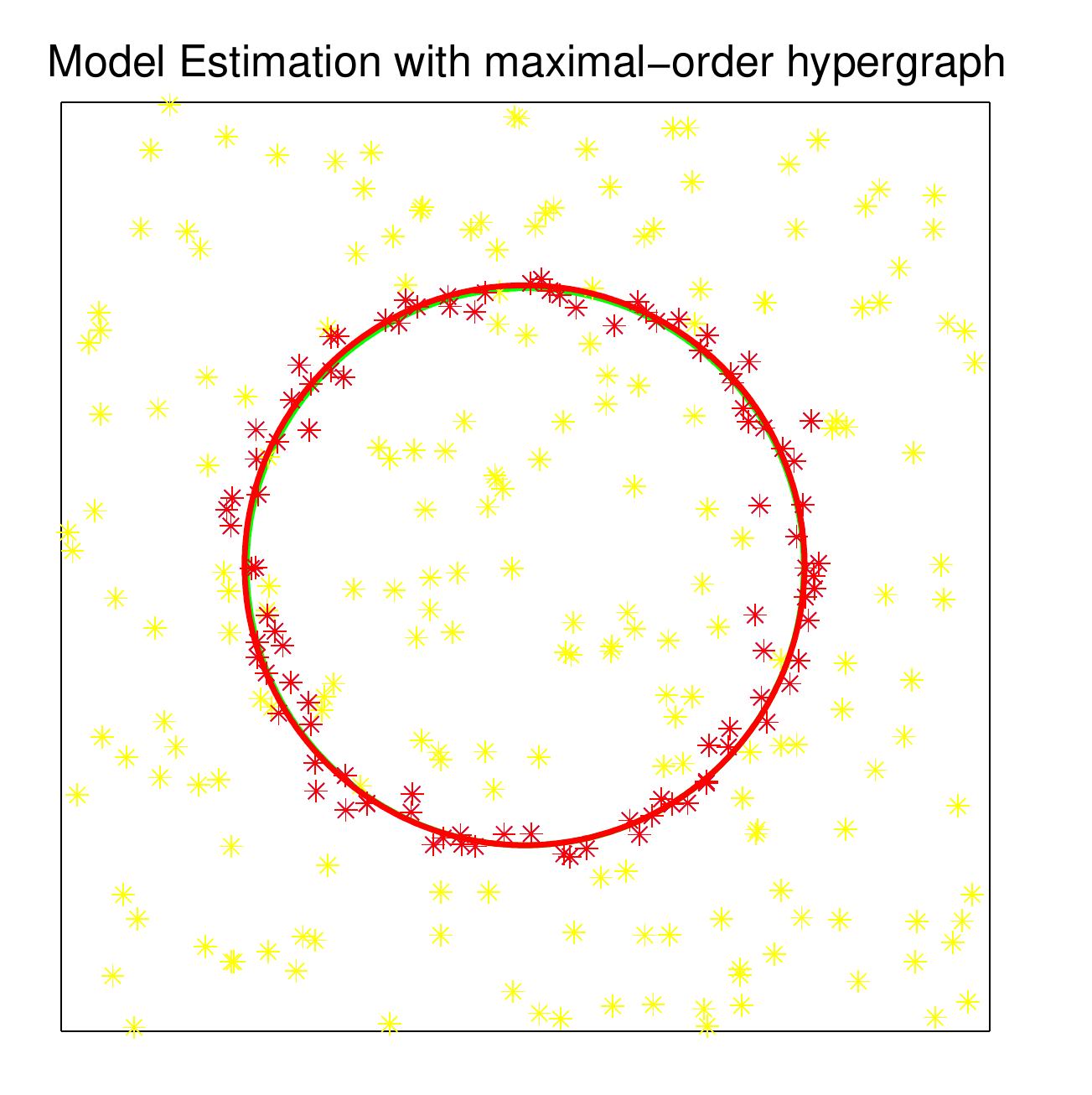}}
  \begin{center} (c)   \end{center}
\end{minipage}
\caption{{Segmentation results obtained by NCut} for circle fitting. (a), (b) and (c) {show} the NCut segmentation on a $4$-uniform, $8$-uniform and nonuniform hypergraph, respectively. {The data points with blue, yellow and red color are inliers, outliers and subsets connected by a hyperedge, respectively. }}
\label{fig:modelsegmentation}
\end{figure}

We argue that expanding degrees of hyperedges is necessary and reasonable for improving the fitting accuracy. Purkait et al.~\cite{pulak2014clustering} proved that NCut~\cite{zhou2007learning} trusts larger degrees of hyperedges (i.e., each hyperedge connects more number of vertices in a hypergraph) more than smaller degrees of hyperedges. Fig.~\ref{fig:modelsegmentation} shows the NCut segmentation results for circle fitting on a $4$-uniform, $8$-uniform and nonuniform (i.e., different hyperedges may connect variable number of vertices) hypergraph{, from} which we can see that the eight data points (in the $8$-uniform hypergraph) constrain a circle better than the four data points (in the $4$-uniform). However, using the largest {justified} degrees of hyperedges (in the nonuniform hypergraph) can fit the circle more accurately. Ideally, a hyperedge (in the nonuniform hypergraph) connects all vertices that represent inlier data points of the corresponding model hypothesis. {Though}, of course, we cannot guarantee that all outliers are not wrongly recognized as inlier data points of a particular model hypothesis, especially for {data} with {a} large number of outliers. However, {when} vertices that represent most of inlier data points belonging to a model hypothesis are connected by the corresponding hyperedge and {when} most of outliers are removed by hypergraph pruning (in Sec.~\ref{sec:hypergraphpruning}), then the effectiveness of the hypergraph with the largest {justified} degrees of hyperedges will not be affected by {a} few outliers. This will be further verified in experiments (see Sec.~\ref{sec:SyntheticData}).


We can see that, in addition to the advantage that the degrees of hyperedges in the {hypergraph} constructed by {the proposed method} are much larger than the ones in the {hypergraph} constructed by the previous works \cite{jain2013efficient,liu2012efficient,ochs2012higher,parag2011supervised}, {the hypergraph} constructed by {the proposed method} has an important {attribute} that {it allows} hyperedges of arbitrary degrees{. In contrast,} the {hypergraphs} constructed by the previous works \cite{jain2013efficient,liu2012efficient,ochs2012higher,parag2011supervised} have used a fixed degree (and {moreover}, mostly {of} the smallest {justified} degree). This is because {the} hyperedges of the {hypergraph} constructed by {the proposed method} connect the corresponding vertices based on the estimated inlier scale and different model hypotheses may have different numbers of inlier data points{. That is,} the constructed {hypergraph} contains ``data-driven" ({hence}, varying) degrees of hyperedges.
\subsection{Hypergraph Weighting}
\label{sec:hypergraphweighting}
Large hypergraphs are {computationally} costly to deal with, {and} many hyperedges are less important: therefore, we introduce {a weighting function} to capture this {degree of importance,} so that we can identify and prune the less significant hyperedges {whose weight scores are small}. For the model fitting problem, ideally, the hyperedges corresponding to the true ``structures'' in {the} data (i.e., the hyperedges connecting more numbers of vertices corresponding to inliers in {the} data), should have {weight scores that are as high} as possible; in contrast, {the weight scores of} the other hyperedges {should be} as low one as possible.

Inspired by \cite{wang2012simultaneously}, we assign each hyperedge a weight based on the non-parametric kernel density estimate techniques \cite{wand1994kernel}. The weight of a hyperedge $e_i$ can be written as
\begin{align}
\label{equ:weight}
\omega_{e_i}=\frac{1}{n}\sum_{j=1}^n\frac{\mathbf{KM}(\hat{r}_{v_j}^{e_i}/{\hat{h}_{e_i}})}{\tilde{s}_{e_i}\hat{h}_{e_i}},
\end{align}
where $n$ is the number of vertices; $\tilde{s}_{e_i}$ is the {estimated} inlier scale of the corresponding hypothesis; $\hat{r}_{v_j}^{e_i}$ is the residual derived from the corresponding hypothesis and data point; $\mathbf{KM}(\cdot)$ is the popular Epanechnikov kernel~\cite{wand1994kernel}, which is written as follows:
\begin{align}
\label{equ:kernel}
\mathbf{KM}(\lambda) &=\left\{ \begin{array}    {r@{\quad \quad} l}
0.75(1-{\|\lambda\|}^2), & \|\lambda\|\leq 1,\\
0~~~~~~~~,&\|\lambda\|> 1,
\end{array}\right.\;
\end{align}
where $\hat{h}_{e_i}$ is the bandwidth of the $i$-th hyperedge, and it can be defined as \cite{wand1994kernel}
\begin{align}
\label{equ:bandwith}
\hat{h}_{e_i}=\left[\frac{243\int_{-1}^1{\mathbf{KM}(\lambda)}^2d\lambda}{35n\int_{-1}^1{\lambda}^2\mathbf{KM}(\lambda)d\lambda}\right]^{0.2}\tilde{s}_{e_i}.
\end{align}

As discussed in \cite{wang2012simultaneously}, a model hypothesis (here, a corresponding hyperedge) with {more numbers of inlier data points and smaller residuals should} have a higher weight score.
\subsection{Hypergraph Pruning}
\label{sec:hypergraphpruning}
Dealing with a large and high degree hypergraph often has the high computational complexity. Therefore, based on the weighted hyperedges (described above), we can select some significant hyperedges (with higher weight scores) by pruning insignificant hyperedges (with lower weight scores). However, how to select significant hyperedges is not a trivial task, i.e., it is difficult to decide a threshold for making a cut between what to prune and not to prune. A data driven threshold is desirable since hyperedges are usually different for different hypergraphs. In this paper, we adopt the information theoretic approach proposed in \cite{ferraz2007density}, which uses a filtering process to choose data points, to select significant hyperedges. Although one could reasonably ask whether there are better methods for this, for the moment we have only explored this option and it {is shown} to be highly effective in experiments (see Sec.~\ref{sec:Experiment}).

For a set of hyperedges $E=\{e_1,e_2,\ldots,e_m\}$ and the associated weights $\bm{W}=\{\omega_{e_1},\omega_{e_2},\ldots,\omega_{e_m}\}$, where $m$ is the number of the hyperedges, let $\bm{\varpi}_i =max(\bm{W})-\omega_{e_i}$ denote the gap between the weight of $e_i$ and the maximum weight of all hyperedges. Then we normalize $\bm{\varpi}_i$ to obtain the prior probability:
 \begin{align}
\label{equ:probability}
p(\bm{\varpi}_i)= \frac{\bm{\varpi}_i}{\sum_{i=1}^m\bm{\varpi}_i}.
\end{align}
The entropy can be defined as
 \begin{align}
\label{equ:entropy}
{L}= -\sum_{i=1}^mp(\bm{\varpi}_i)\log p(\bm{\varpi}_i).
\end{align}
The significant hyperedges with a higher probability than the entropy ${L}$ can be selected by
 \begin{align}
\label{equ:condition}
\hat{E}=\{e_i|{L}+\log p(\bm{\varpi}_i)<0\}.
\end{align}

After selecting significant hyperedges, we remove insignificant hyperedges and {the vertices that are not connected by any significant hyperedge. We note that the removed vertices usually correspond to outliers in the data.} Thus we can obtain a simplified and more effective hypergraph (with less influence {by} outliers) for model fitting.

\section{Sub-Hypergraph Detection via Hypergraph Partition}
\label{sec:sub-hypergraph detection}
The result of hypergraph partitioning {of a hypergraph} is a set of sub-hypergraphs. Each sub-hypergraph represents a potential model instance in data. For a sub-hypergraph $G_B=(V_B,E_B,\omega_B)$, $V_B$ is a subset of the vertices $V$ and $E_B$ is the hyperedges of $G_B${. Each hyperedge is assigned a positive weight value $\omega_B$}. In this section, we propose a novel hypergraph partition algorithm, which cannot only adaptively estimate the number of sub-hypergraphs, but also be highly robust to multi-structural data with outliers.



\subsection{Hypergraph Partition}
Before we propose the novel hypergraph partition algorithm, we first briefly introduce the hypergraph partition algorithm in \cite{zhou2007learning} because {the proposed algorithm} is one of its variants. We choose this algorithm because it generalizes the well-known ``Normalized cut" pairwise clustering algorithm \cite{shi2000normalized} from simple graphs to hypergraphs, and it has been proven to be very effective \cite{yu2012adaptive,huang2011unsupervised}.

Given a hypergraph $G=(V,E,\omega)$ and the associated incident matrix $H$, {the} degree of a vertex $v \in V$ is defined to be $d(v)=\sum_{e \in E}\omega(e)h(v,e)$, and the degree of a hyperedge $e \in E$ is defined to be $\delta(e)=\sum_{v\in V} h(v,e)$. Based on this, $D_v$, $D_e$ and $W$ are used to represent the diagonal matrices of the vertex degrees, hyperedge degrees, and hyperedge weights, respectively.

The hypergraph $G$ can be partitioned into two parts $A$ and $B$, $A\cup B =V$, $A \cap B =\emptyset$. The hyperedge boundary $\partial A :=\{e\in E|e\cap A \neq \emptyset, e\cap B \neq \emptyset\}$ is a hyperedge set that partitions the hypergraph $G$ into two parts, A and B. A two-way hypergraph partition is then defined as
 \begin{align}
\label{equ:twowaypartition}
S\textup{cut}(A,B)=\sum_{e \in \partial A}\omega (e) \frac{|e\cap A||e\cap B|}{\delta(e)}.
\end{align}

For a hypergraph, a two-way normalized hypergraph partition is written as
 \begin{align}
\label{equ:twowaynormalizedpartition}
NS\textup{cut}(A,B)= S\textup{cut}(A,B)\left(\frac{1}{\textup{vol}(A)}+\frac{1}{\textup{vol}(B)}\right),
\end{align}
where $\textup{vol}(A)$ is the volume of $A$, i.e., $\textup{vol}(A)=\sum_{v \in A}d(v)$, and $\textup{vol}(B)$ is similarly defined.

Then, in~\cite{zhou2007learning}, Eq.~(\ref{equ:twowaynormalizedpartition}) is relaxed into a real-valued optimization problem as {per} Eq.~(\ref{equ:minimization}), which is a NP-complete problem:
 \begin{align}
\label{equ:minimization}
\arg\min_{q\in \mathbb{R}^{|V|}}\sum_{e\in E}\sum_{\{u,v\}\subset e} \frac{\omega (e)}{\delta(e)}\left(\frac{q(u)}{\sqrt{d(u)}}-\frac{q(v)}{\sqrt{d(v)}}\right)^2
= \arg\min_{q\in \mathbb{R}^{|V|}} 2 q^T\triangle  q,
\end{align}
where $q$ is a label vector and the hypergraph Laplacian matrix $\triangle = I- D_v^{-\frac{1}{2}}HWD_e^{-1}H^TD_v^{-\frac{1}{2}}$, where $I$ denotes the identity matrix. The eigenvector associated with the smallest nonzero eigenvalue of $\triangle$ is a theoretical solution of Eq.~(\ref{equ:minimization}).

For a multi-way classification of vertices in the hypergraph, the first $k$ eigenvectors with the $k$ smallest eigenvalues of $\triangle$ can be taken as the representations of the vertices, as in \cite{ng2001spectral}. After that, the $k$-means algorithm is used to obtain final clustering results.
\begin{algorithm}[t] 
\renewcommand{\algorithmicrequire}{\textbf{Input:}}
\renewcommand\algorithmicensure {\textbf{Output:} }
\caption{Hypergraph Partition for Sub-Hypergraph Detection} 
\label{alg:subhypergraphdection} 
\begin{algorithmic}[1] 
\REQUIRE 
A hypergraph $G$ and the largest possible sub-hypergraphs number $C$
\STATE Compute the hypergraph Laplacian matrix $\triangle = I- D_v^{-\frac{1}{2}}HWD_e^{-1}H^TD_v^{-\frac{1}{2}}$.
\vspace{-0.6cm}
\STATE Obtain the eigenvector matrix $Y$ by selecting the $C$ {smallest} eigenvectors of $\triangle$.
\STATE Find the best alignment of $Y$'s columns to recover the rotation matrix $R$ (see \cite{zelnik2004self}).
\STATE Determine the number of sub-hypergraphs $k_0$ by minimizing Eq.~(\ref{equ:costfunction}).
\STATE Assign the vertices of the hypergraph to the $k_0$ sub-hypergraphs according to the alignment results $U$, i.e., {$s_{\hat{m}}=\{v_i\in V|\max_j U_{ij}=U_{i\hat{m}}\}$, $\hat{m}=1,\ldots,k_0$}.
\ENSURE sub-hypergraphs $\hat{S}=\{s_1,s_2,\ldots, s_{k_0}\}$.
\end{algorithmic}
\end{algorithm}
\subsection{Sub-Hypergraph Detection}
\label{sec:sub-hypergraph detection algorithm}
The hypergraph partition algorithm in~\cite{zhou2007learning} (described above) is very effective, but it cannot adaptively estimate the number of sub-hypergraphs, and it conducts the final clustering by the $k$-means algorithm, which is usually sensitive to the initialization. However, one important task in fitting multi-structural data is to automatically estimate the number of the structures in the data, and multi-structural data usually contain outliers. Therefore, we improve that hypergraph partition algorithm by introducing {an} idea from Zelnik-Manor and Perona~\cite{zelnik2004self}. They presented a spectral clustering algorithm which obtains final clustering results by non-maximum suppression and adaptively finds the number of groups by exploiting the structure of eigenvectors.

The spectral clustering algorithm in \cite{zelnik2004self} minimizes the cost of aligning the top eigenvectors, to determine the number of groups. For a normalized affinity matrix, the cost function is defined as
 \begin{align}
\label{equ:costfunction}
E=\sum_{i=1}^n\sum_{j=1}^C \frac{U_{ij}^2}{(\max_j U_{ij})^2},
\end{align}
where $C$ is the largest possible group number, and $U$ is a matrix {derived from} the rotation matrix $R$ of an eigenvector matrix $Y$ ($Y$ consists of the $C$ {top} eigenvectors of the normalized affinity matrix), i.e., $U=YR$ ($R$ is an orthogonal matrix).

By combining the advantages of both the hypergraph partition algorithm in \cite{zhou2007learning} and the spectral clustering algorithm in \cite{zelnik2004self}, we propose a novel hypergraph partition algorithm for sub-hypergraph detection (see Algorithm~\ref{alg:subhypergraphdection}). This novel algorithm can adaptively detect sub-hypergraphs, and it is more robust than the hypergraph partition algorithm in \cite{zhou2007learning} because it obtains final clustering results by non-maximum suppression instead of the $k$-means clustering process (in detecting sub-hypergraphs). Although there is a parameter $C$ used in the proposed hypergraph partition algorithm, it has a clear meaning which means the largest possible number of model instances included in data, and the result of the proposed hypergraph partition algorithm is not sensitive to the value of $C$. We set its value to $10$ (which means that we assume that there are $10$ model instances, at most, included in data), and we do not change the $C$ value in all of the following experiments.
\section{The Proposed Hypergraph based Model Fitting Method}
\label{sec:overall algorithm}
By the hypergraph partition algorithm, we can adaptively obtain $k_0$ sub-hypergraphs. Each partition in a hypergraph would ideally be related to a single model instance in data (for each model instance there is a partitioned sub-hypergraph, and for each partitioned sub-hypergraph there is model instance). However, unless the partition has only one hyperedge, in practice it tends to have multiple models within {a} single partition, which are essentially imperfect estimates of the same model. Thus we need to select {the} best representative (i.e., the hyperedge with the highest weight).

\begin{algorithm}[htb] 
\renewcommand{\algorithmicrequire}{\textbf{Input:}}
\renewcommand\algorithmicensure {\textbf{Output:} }
\caption{Hypergraph based Geometric Model Fitting} 
\label{alg:summarize} 
\begin{algorithmic}[1] 
\REQUIRE 
Data points $X$, the $C$ value and the $K$ value
\STATE Sample many minimal subsets, which are used to generate the corresponding potential hyperedges.
\label{STATE:1}
\STATE Expand the hyperedges according to IKOSE and assign each hyperedge a weight by the approach introduced in Sec.~\ref{sec:hypergraphconstruction}.
\label{STATE:2}
\STATE Select some significant hyperedges by Eq.~(\ref{equ:condition}) and then remove {the vertices without being connected by any significant hyperedges} by hypergraph pruning.
\label{STATE:3}
\STATE Detect sub-hypergraphs $\hat{S}=\{s_1,s_2,\ldots, s_{k_0}\}$ by Algorithm~\ref{alg:subhypergraphdection}.
\label{STATE:4}
\STATE Select {the} best representative of hyperedges in each sub-hypergraph.
\label{STATE:5}
\STATE Eliminate duplicate hyperedges by the mutual information theory~\cite{wang2012simultaneously}.
\label{STATE:6}
\ENSURE The retained hyperedges (model hypotheses) and the vertices (inliers) connected by the associated hyperedges.
\end{algorithmic}
\end{algorithm}
Now, we have all the ingredients developed in the previous sections, based on which, we propose the complete hypergraph based fitting method (HF) (see Algorithm~\ref{alg:summarize}). The proposed HF method well formulates the problem of geometric model fitting as a hypergraph partition problem. HF consists of two main steps, i.e., hypergraph construction (described in Sec.~\ref{sec:IIHG}) and sub-hypergraph detection (described in Sec.~\ref{sec:sub-hypergraph detection}). Besides the parameter $C$, there is the other parameter (i.e., $K$) used in the proposed hypergraph-based fitting method: $K$ is the $K$-th ordered point used in IKOSE to estimate the inlier scale. The value of $K$ in IKOSE has a clear meaning and it does not have significant influence on the performance of model selection by the proposed HF. As for the number of the sampled minimal subsets, one can easily estimate its value when the outlier percentage and the dimension of the model parameters are known \cite{fischler1981random}.

{To characterise the computational complexity, we focus on the verification stage for a hypothesize-and-verify
framework, and we do not consider the time for sampling subsets (i.e., Step~\ref{STATE:1}). For hypergraph modelling (i.e., Step~\ref{STATE:2} and Step~\ref{STATE:3}), the complexity approximately amounts to $O(M)$, where $M$ is the number of hyperedges (i.e., the number of generated model hypotheses). For the sub-hypergraph detection (i.e., Step~\ref{STATE:4}), the computational cost of Algorithm~\ref{alg:subhypergraphdection} is mainly used to compute the Eq.~(\ref{equ:costfunction}). So, the complexity of Step~\ref{STATE:4} is about $O(C*n)$, where $n$ is the number of data points. For Step~\ref{STATE:5} and Step~\ref{STATE:6}, they only deal with a small amount of data, thus they are efficient. Therefore, the total computational complexity of HF is approximately $O(M)$ since the value of $M$ is usually larger than the value of $C*n$.}


\section{Experiments}
\label{sec:Experiment}
We evaluate the proposed method (HF) on synthetic data and real images, and compare it with four state-of-the-art robust model fitting methods, namely, KF \cite{chin2009robust}, RCG \cite{liu2012efficient}, AKSWH \cite{wang2012simultaneously} and T-linkage~\cite{Magri_2014_CVPR}. All of the competing methods can handle multi-structural data and estimate the number of model instances\footnote{For KF and T-linkage, we use the codes published on the web: \url{http://cs.adelaide.edu.au/~tjchin/doku.php?id=code} and \url{http://www.diegm.uniud.it/fusiello/demo/jlk/}, respectively. For RCG and AKSWH, we use the codes provided by the authors.}. Our test environment is MS Windows $7$ with Intel Core i$7$-$3630$ CPU $2.4GHz$ and $16GB$ RAM.

Since all the competing fitting methods operate a hypothesize-and-verify framework and we focus on the verification stage{; to} be fair, we first generate {model hypotheses} by {using} the proximity sampling technique \cite{kanazawa2004detection,toldo2008robust} for all these methods. There are $5,000$ model hypotheses generated for line fitting in Sec.~\ref{sec:SyntheticData}, $10,000$ model hypotheses generated for homography based segmentation in Sec.~\ref{sec:homographbasedsegmentation}, and $20,000$ model hypotheses generated for two-view based motion segmentation in Sec.~\ref{sec:motionsegmentation}. Then we fit the model instances of the input multi-structural data by the five competing fitting methods using the same model hypotheses generated from the sampling step. We optimize the parameters of all the competing fitting methods on each dataset for the best performance. 

\subsection{Synthetic Data}
\label{sec:SyntheticData}
\begin{figure}[t]
\centering
\begin{minipage}[t]{.1585\textwidth}
  \centering
  \centerline{\includegraphics[width=1.16\textwidth]{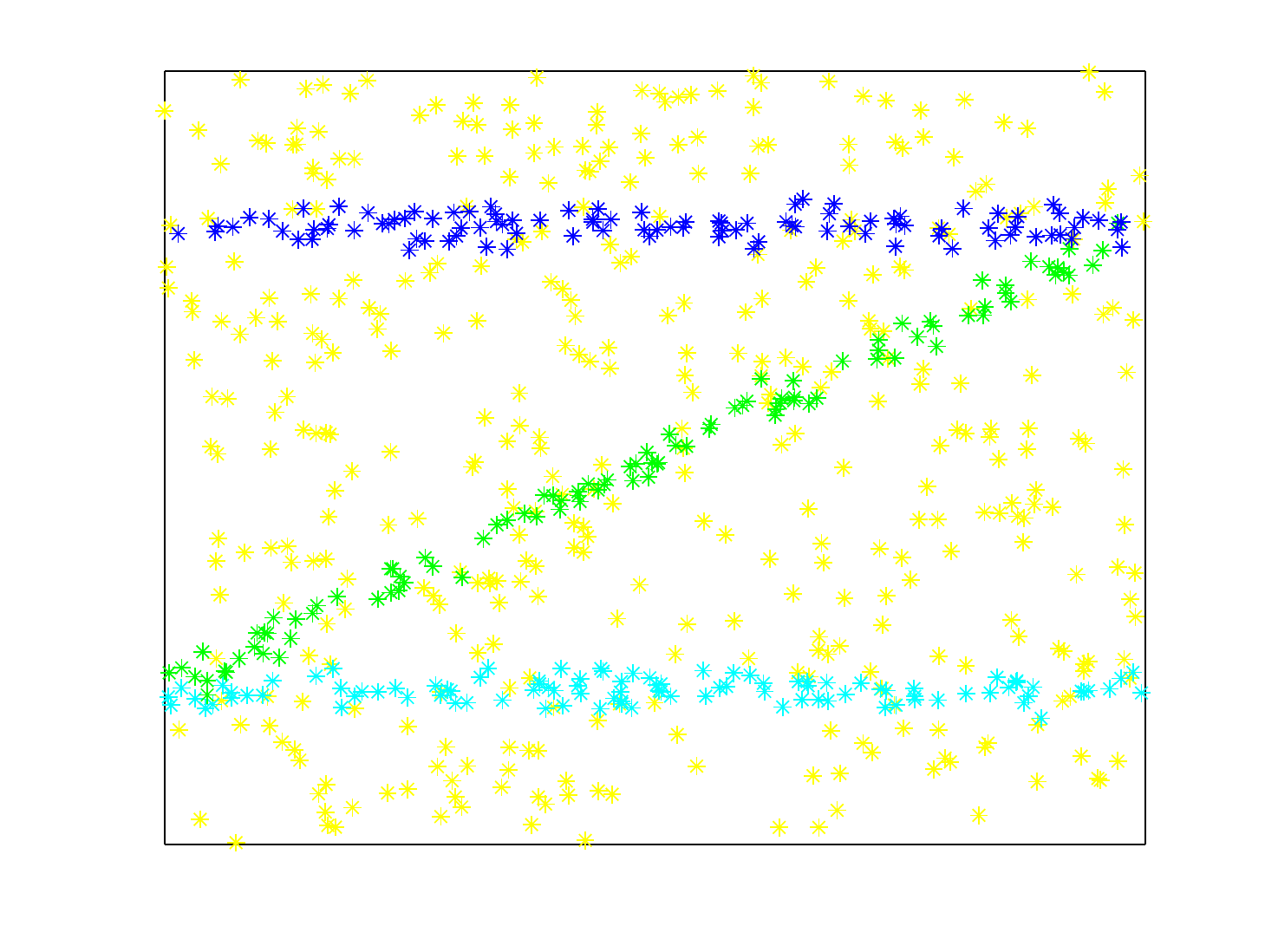}}
  \centerline{\includegraphics[width=1.16\textwidth]{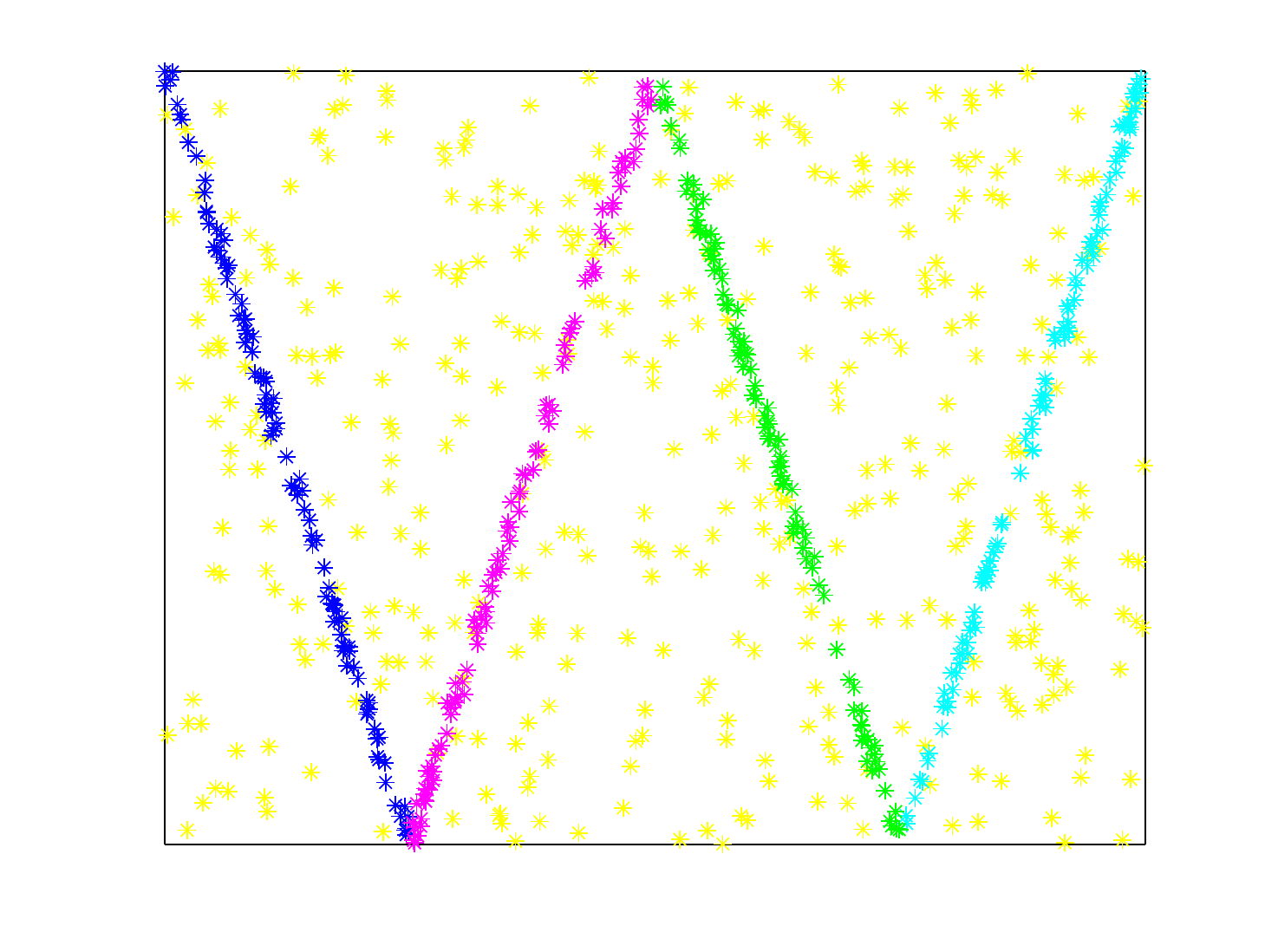}}
  \centerline{\includegraphics[width=1.16\textwidth]{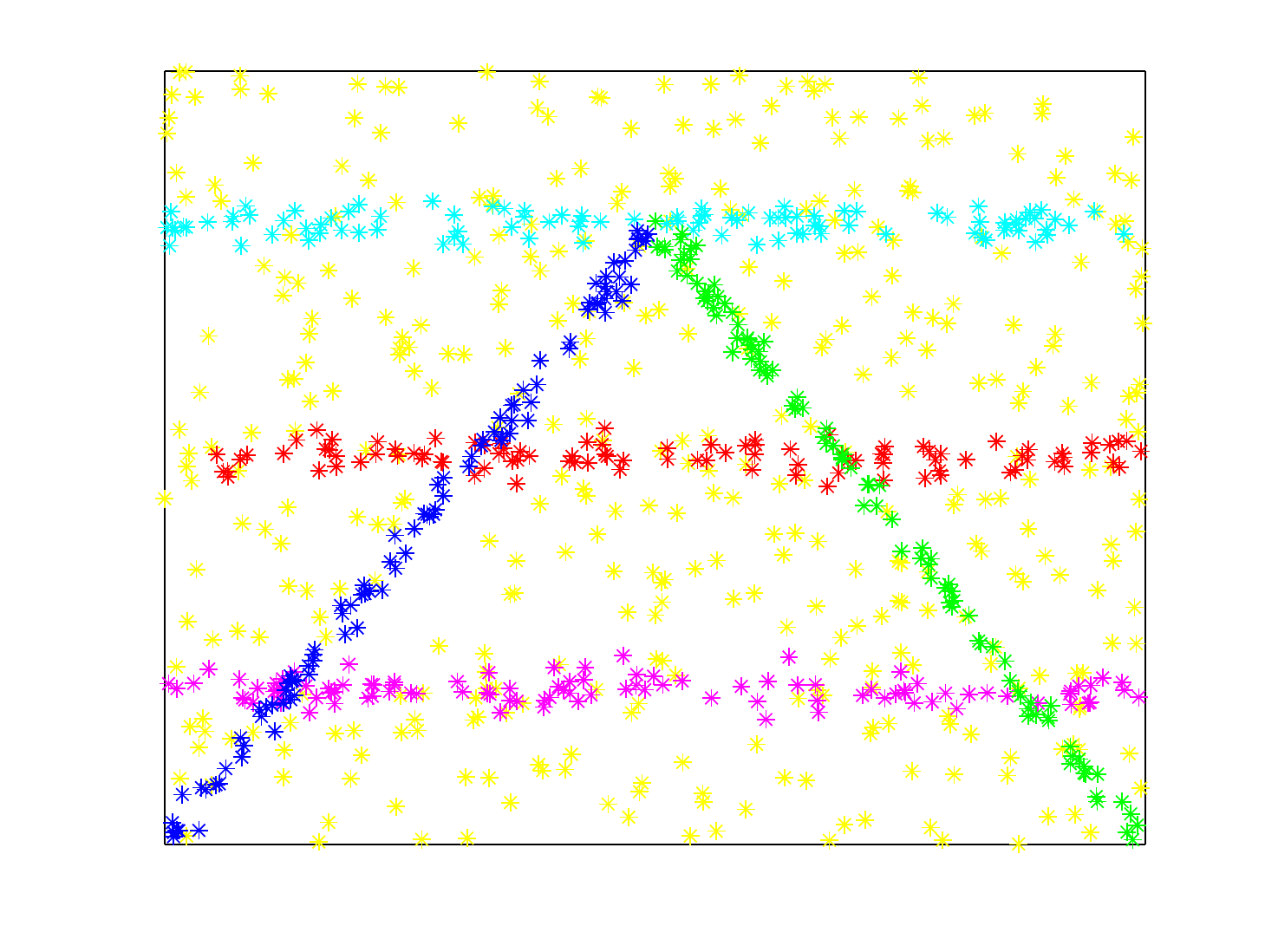}}
  \centerline{\includegraphics[width=1.16\textwidth]{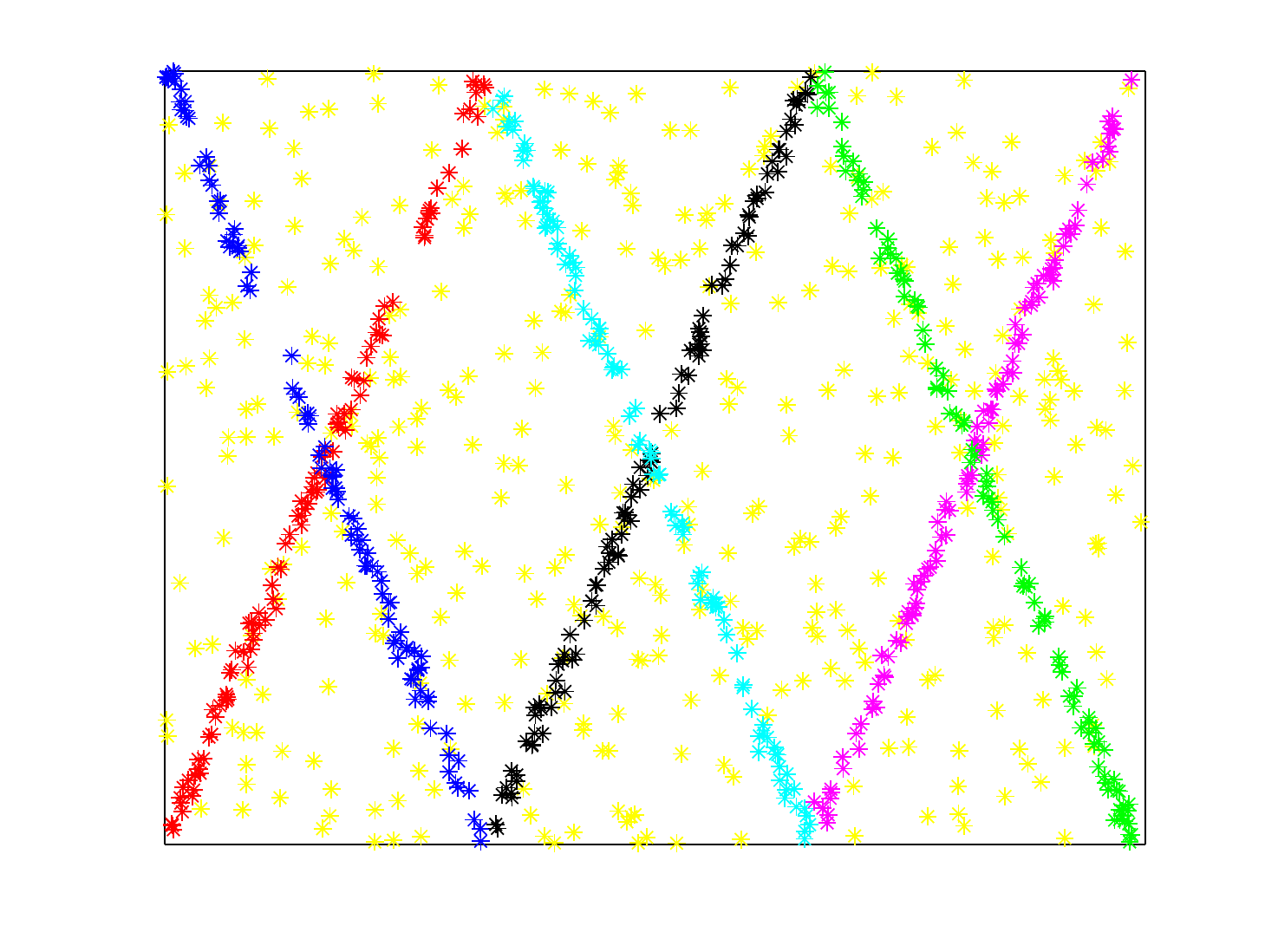}}
  \begin{center} (a)  \end{center}
\end{minipage}
\begin{minipage}[t]{.1585\textwidth}
  \centering
  \centerline{\includegraphics[width=1.16\textwidth]{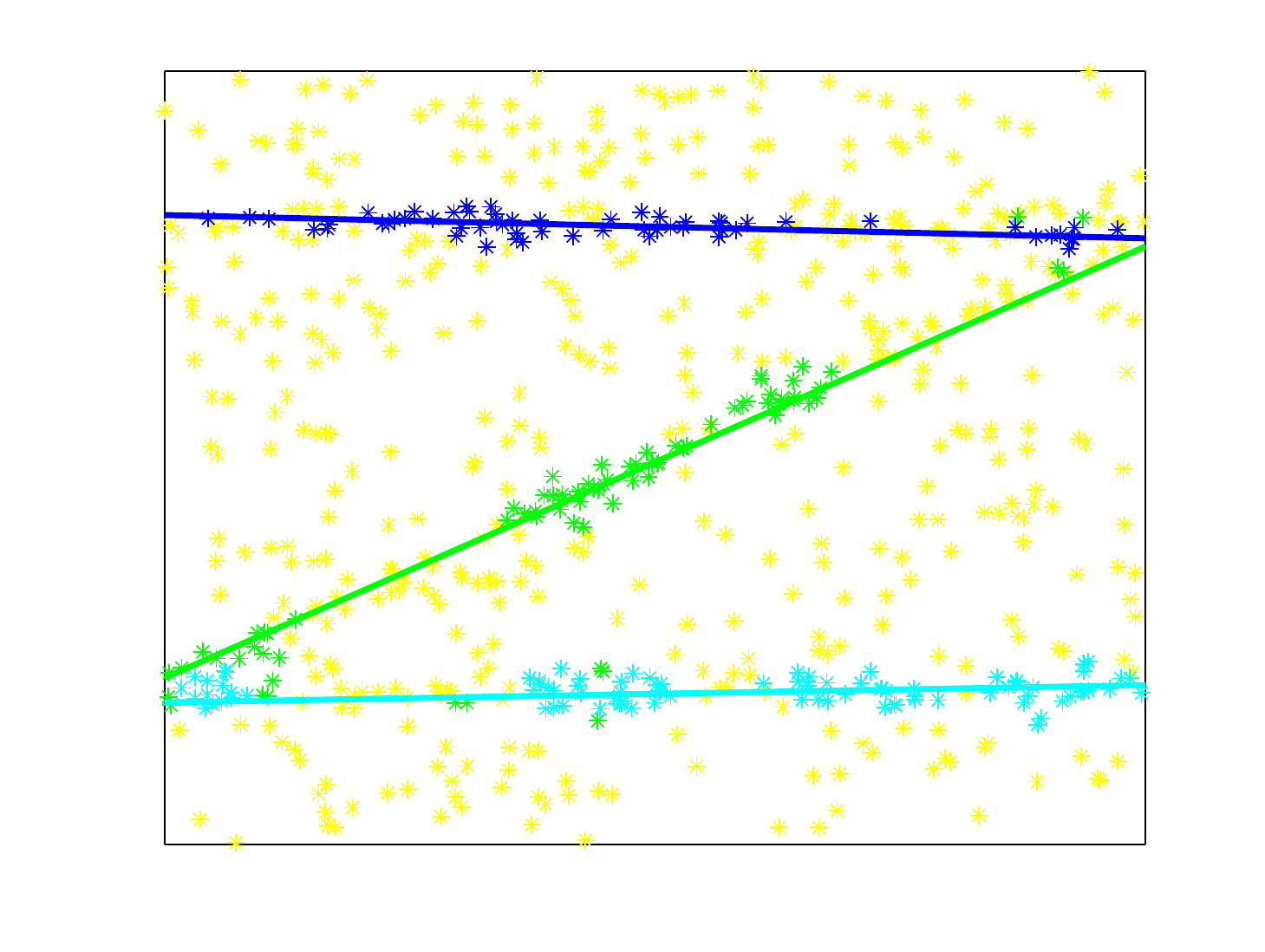}}
  \centerline{\includegraphics[width=1.16\textwidth]{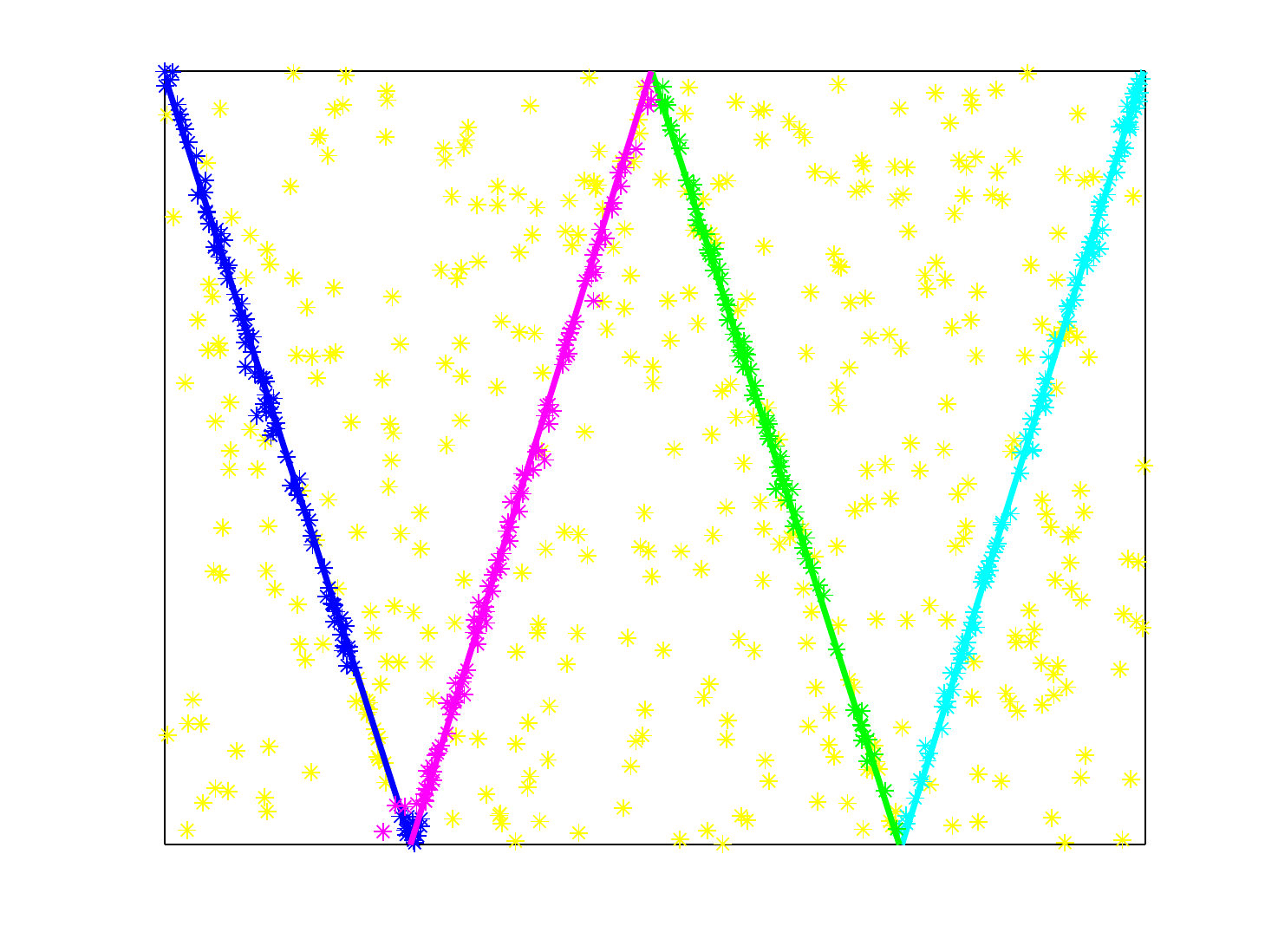}}
  \centerline{\includegraphics[width=1.16\textwidth]{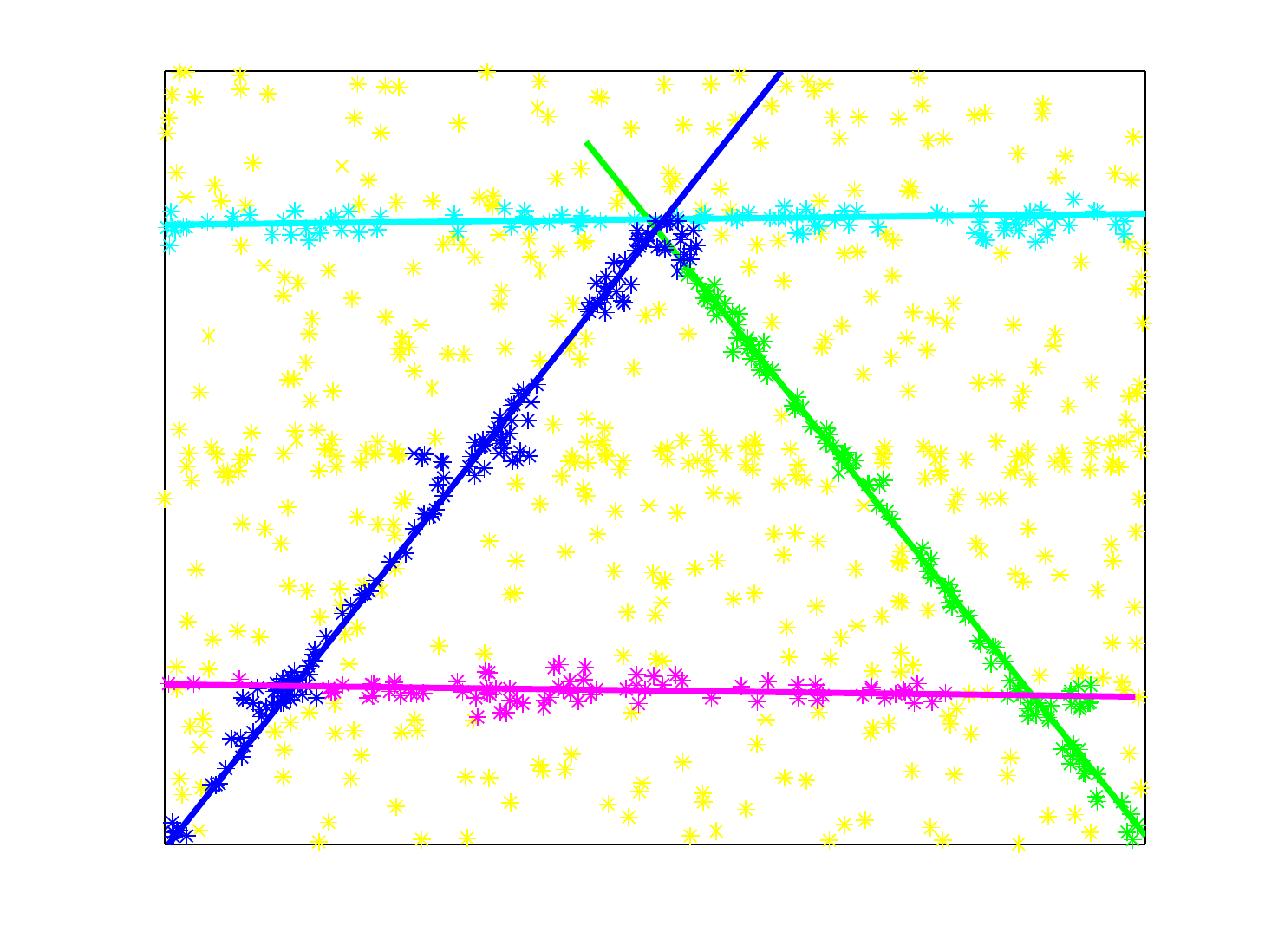}}
  \centerline{\includegraphics[width=1.16\textwidth]{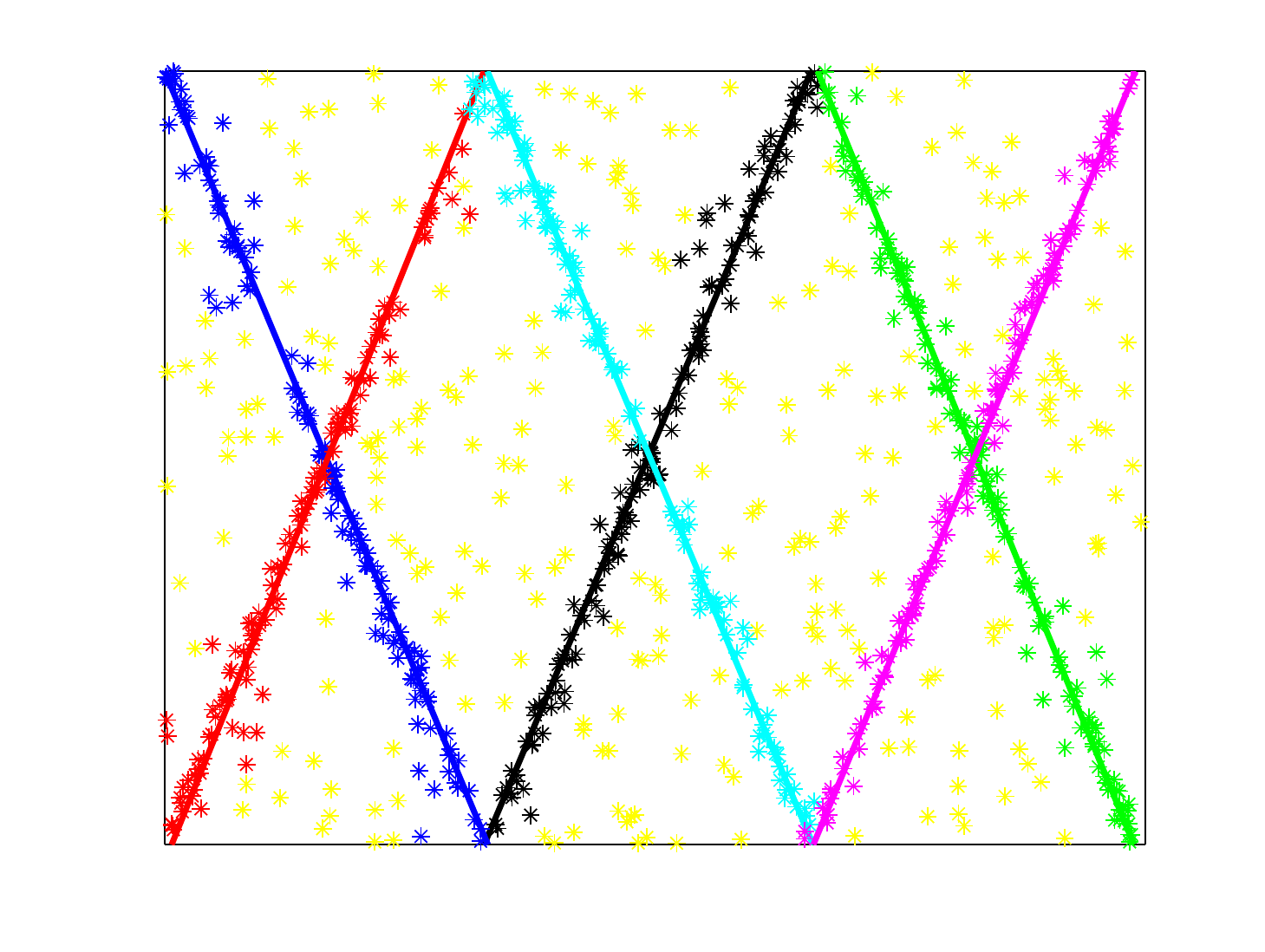}}
  \begin{center} (b) \end{center}
\end{minipage}
\begin{minipage}[t]{.1585\textwidth}
  \centering
  \centerline{\includegraphics[width=1.16\textwidth]{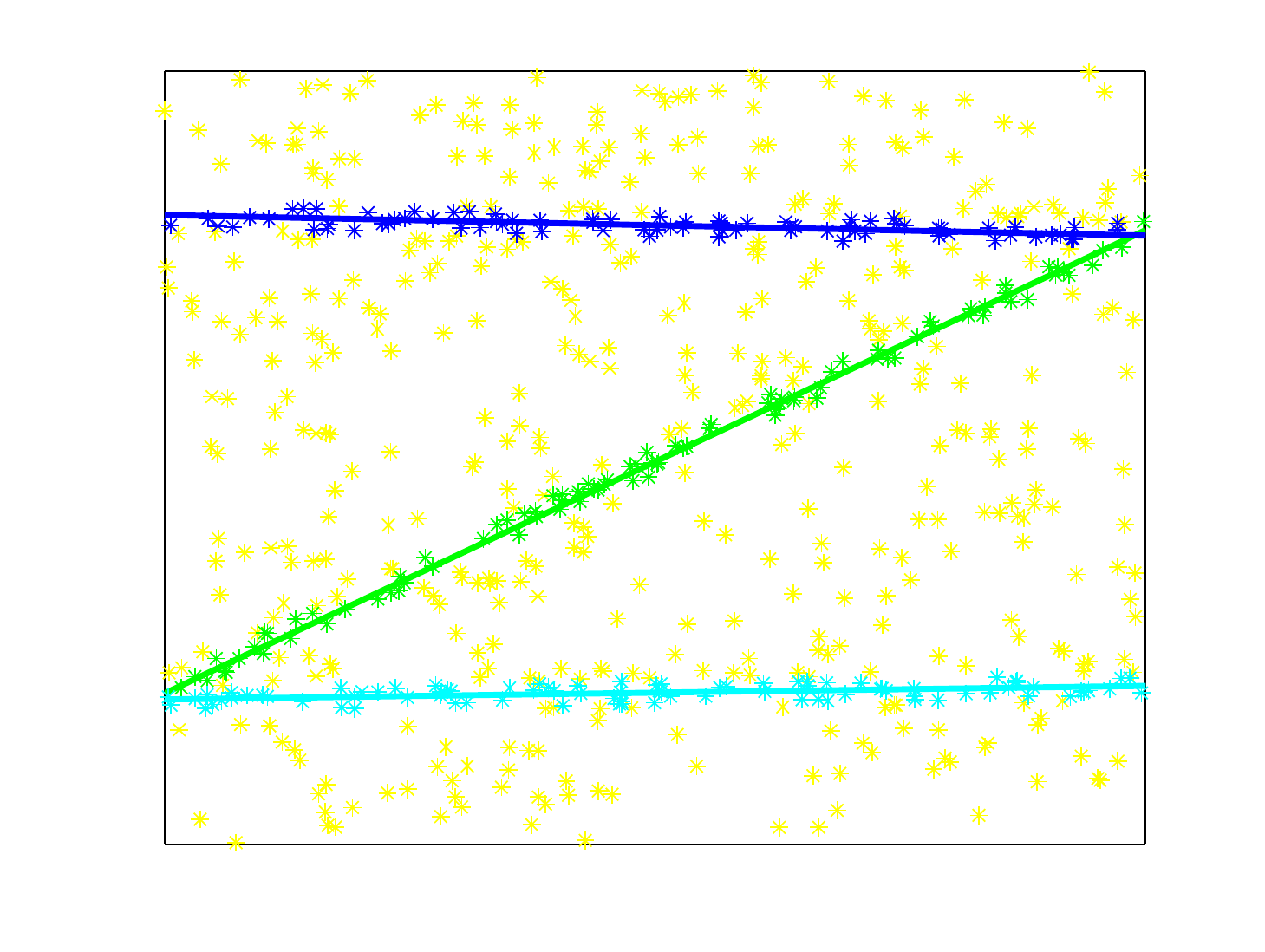}}
  \centerline{\includegraphics[width=1.16\textwidth]{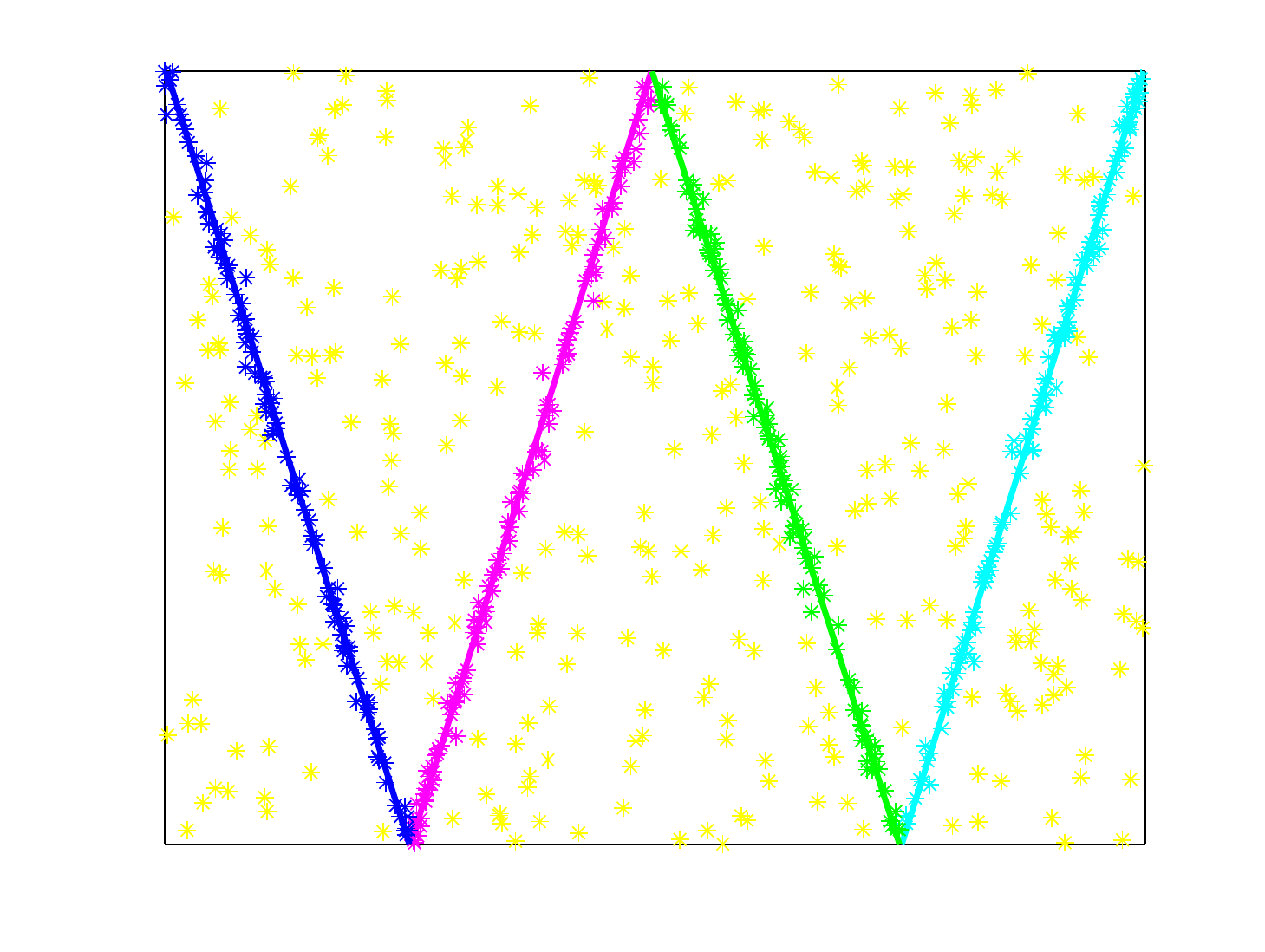}}
  \centerline{\includegraphics[width=1.16\textwidth]{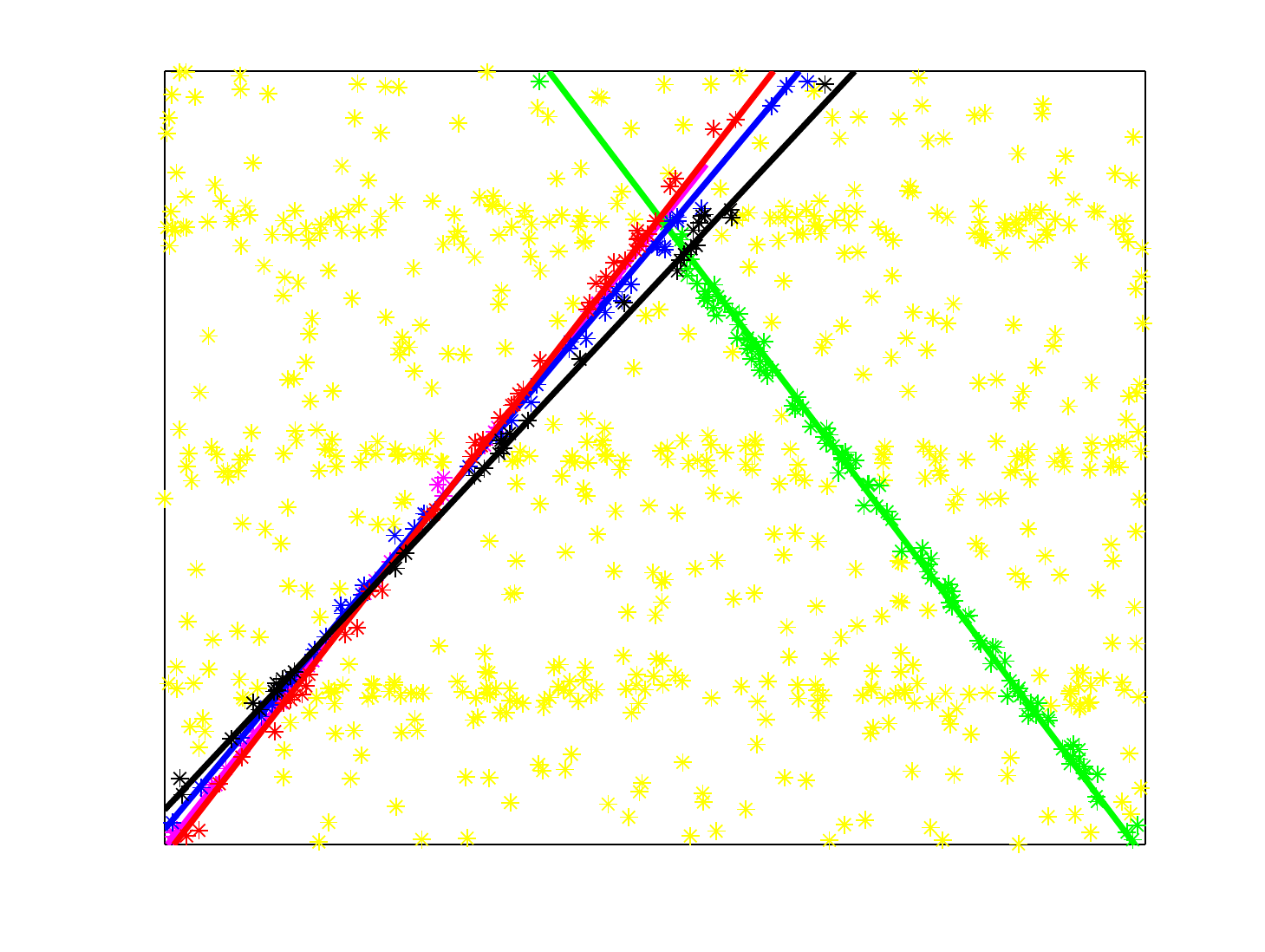}}
  \centerline{\includegraphics[width=1.16\textwidth]{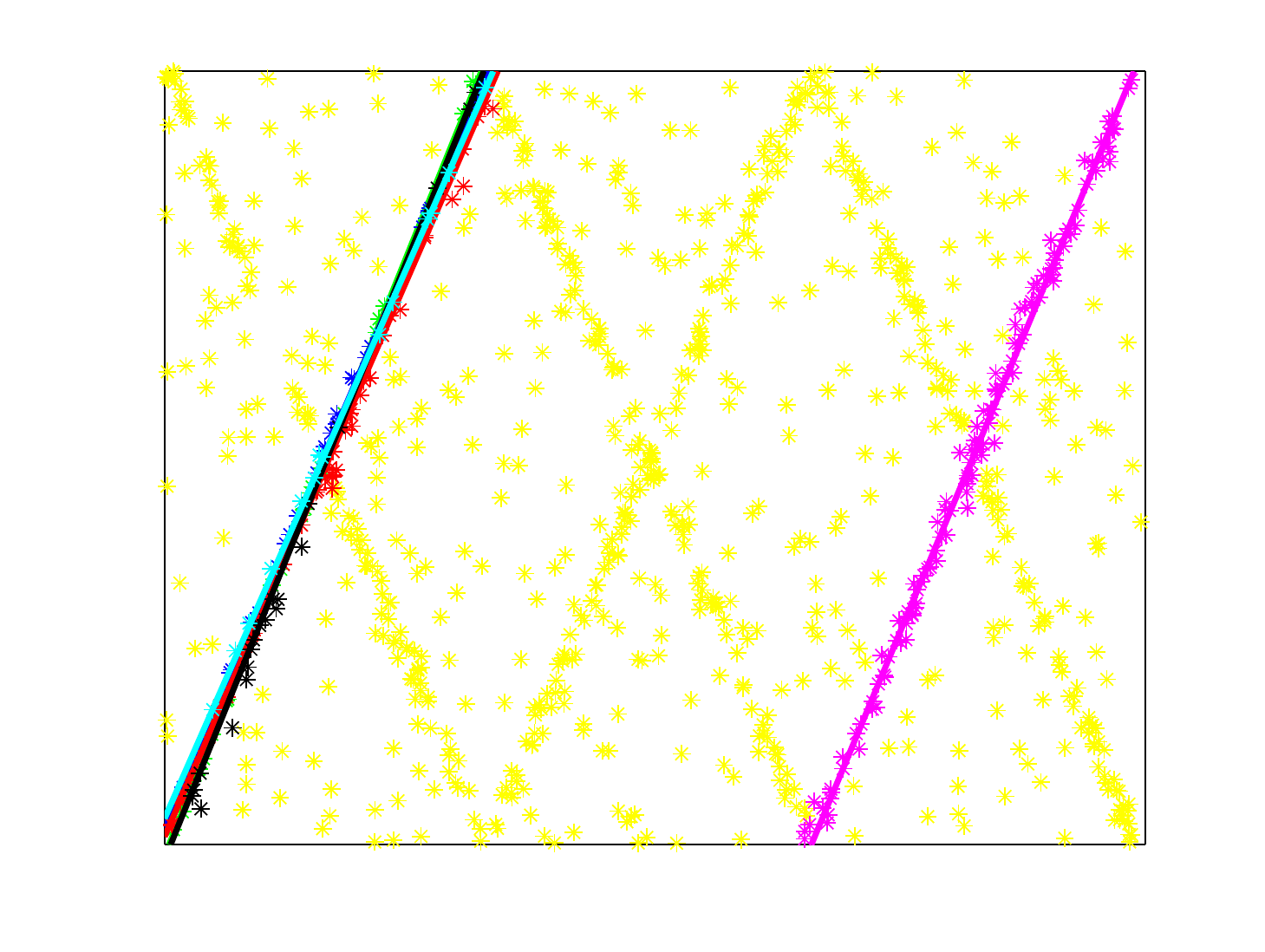}}
  \begin{center} (c) \end{center}
\end{minipage}
\begin{minipage}[t]{.1585\textwidth}
  \centering
  \centerline{\includegraphics[width=1.16\textwidth]{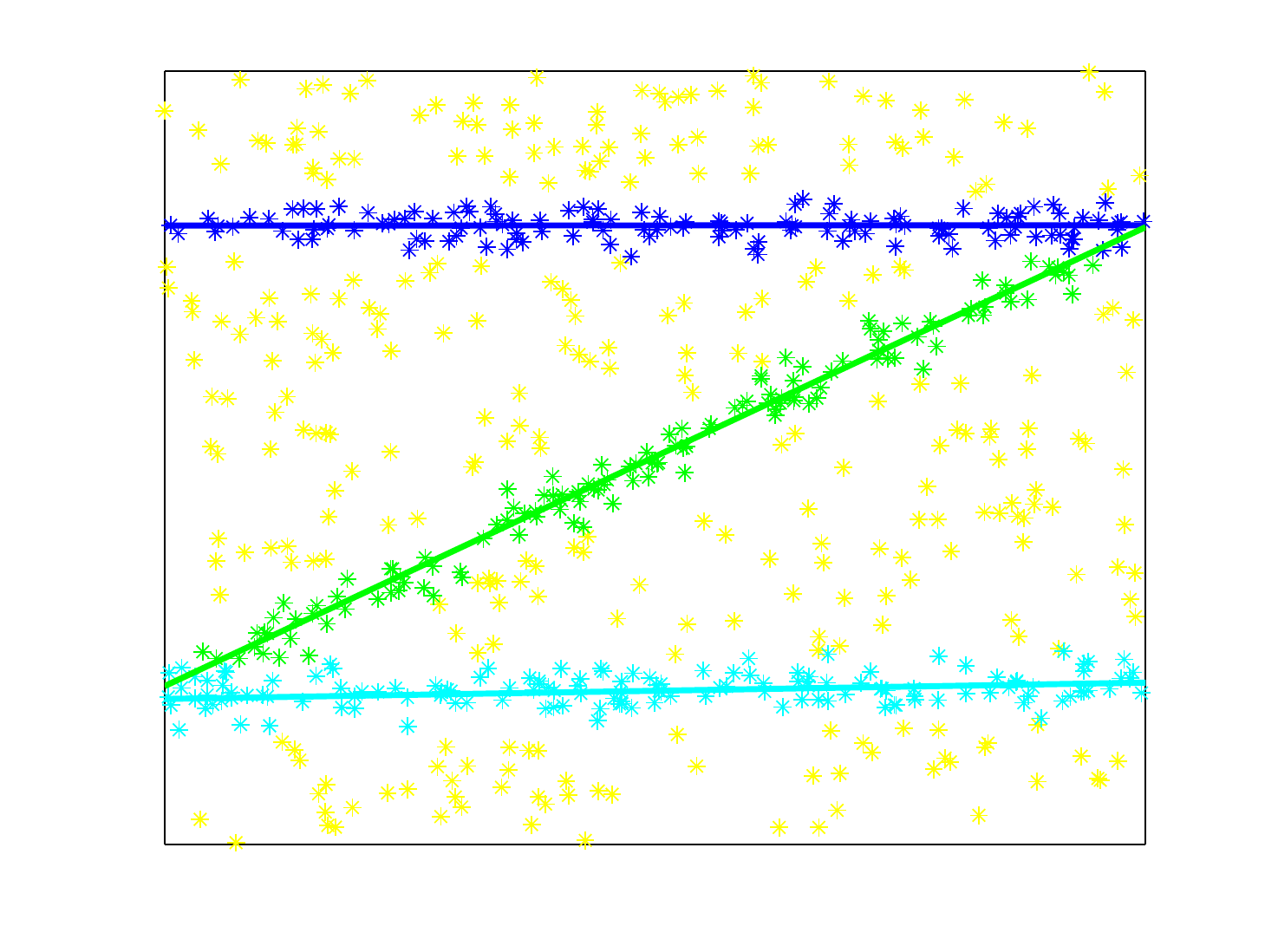}}
  \centerline{\includegraphics[width=1.16\textwidth]{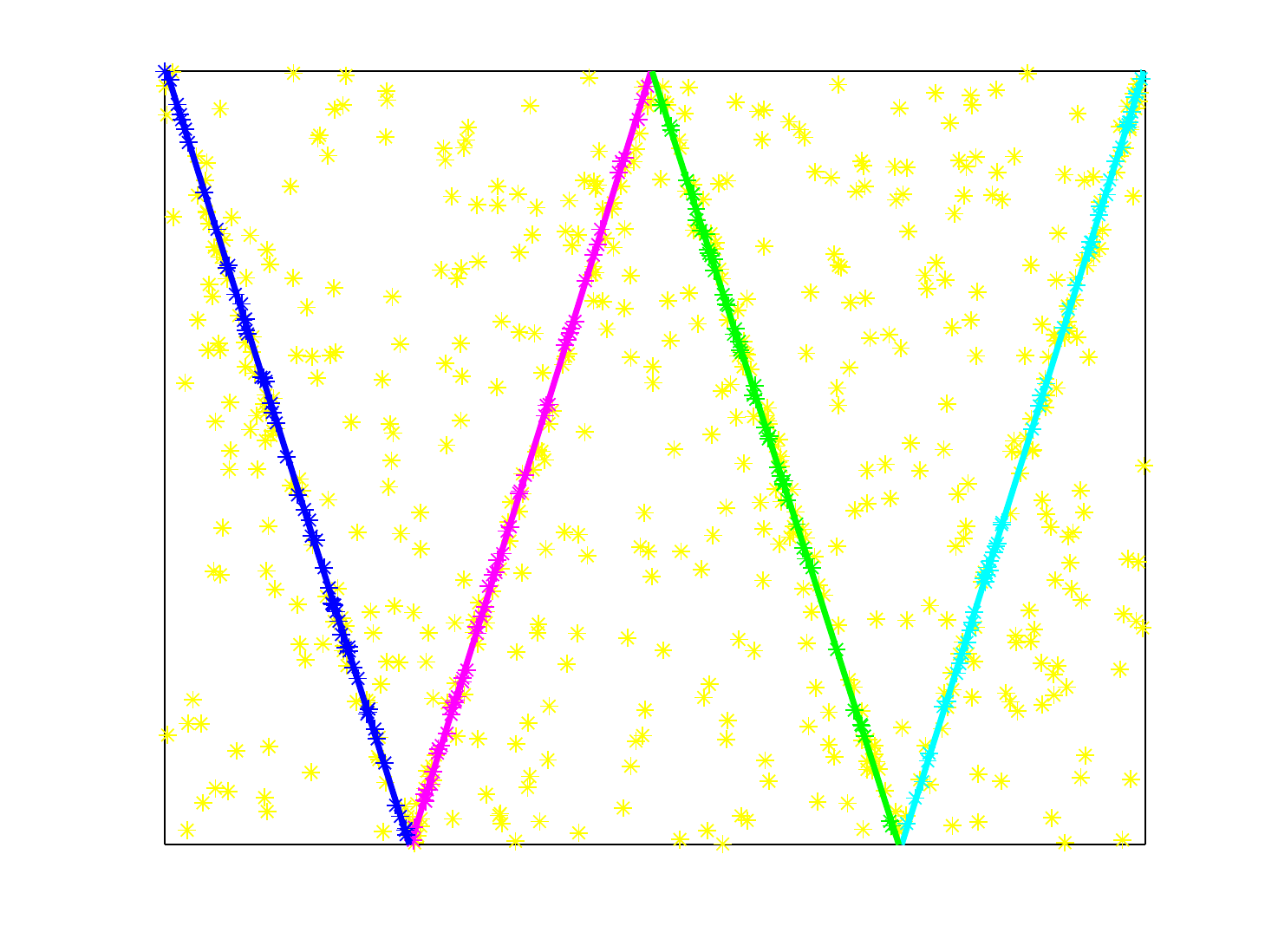}}
  \centerline{\includegraphics[width=1.16\textwidth]{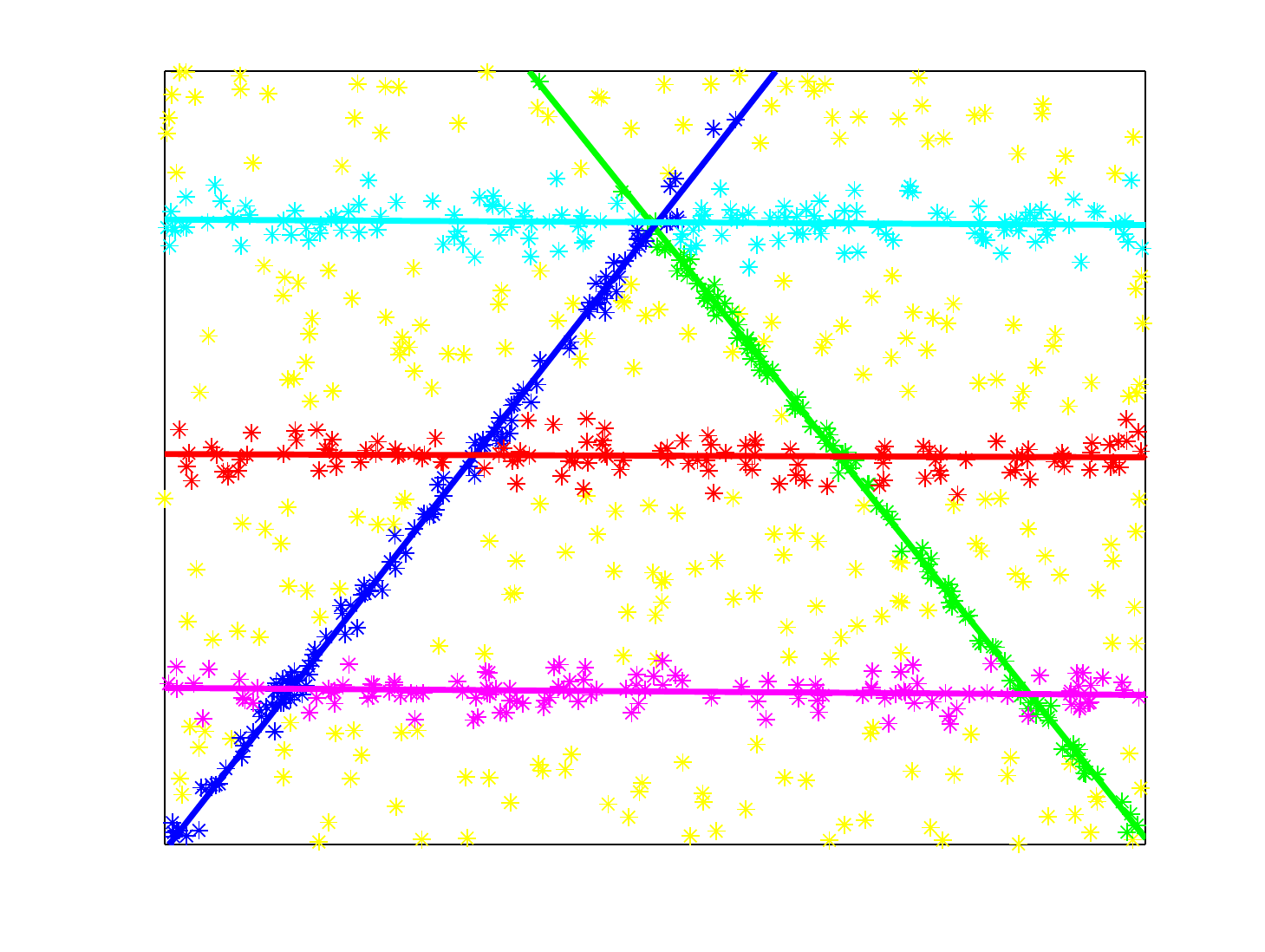}}
  \centerline{\includegraphics[width=1.16\textwidth]{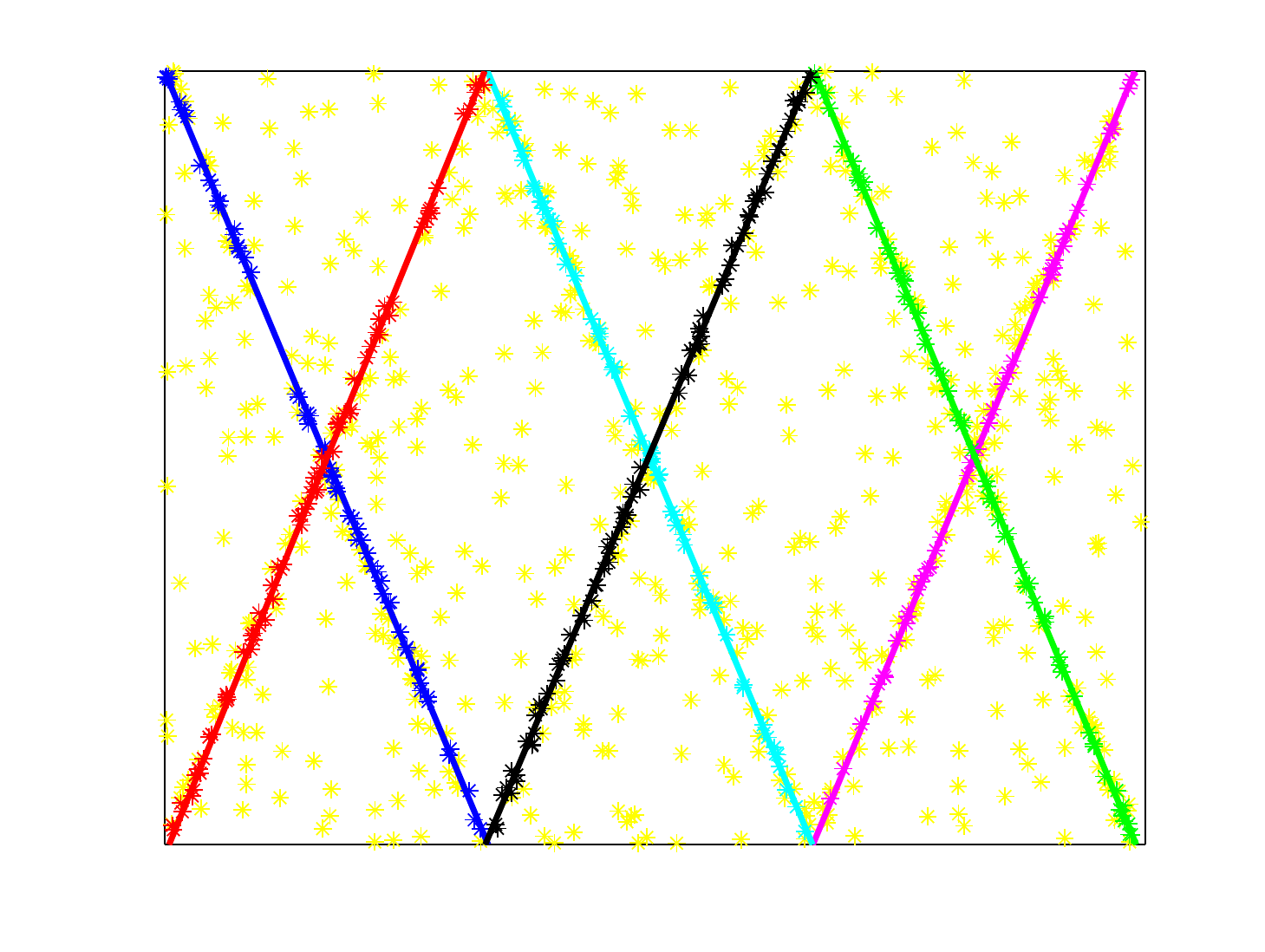}}
  \begin{center} (d)  \end{center}
\end{minipage}
\begin{minipage}[t]{.1585\textwidth}
  \centering
  \centerline{\includegraphics[width=1.16\textwidth]{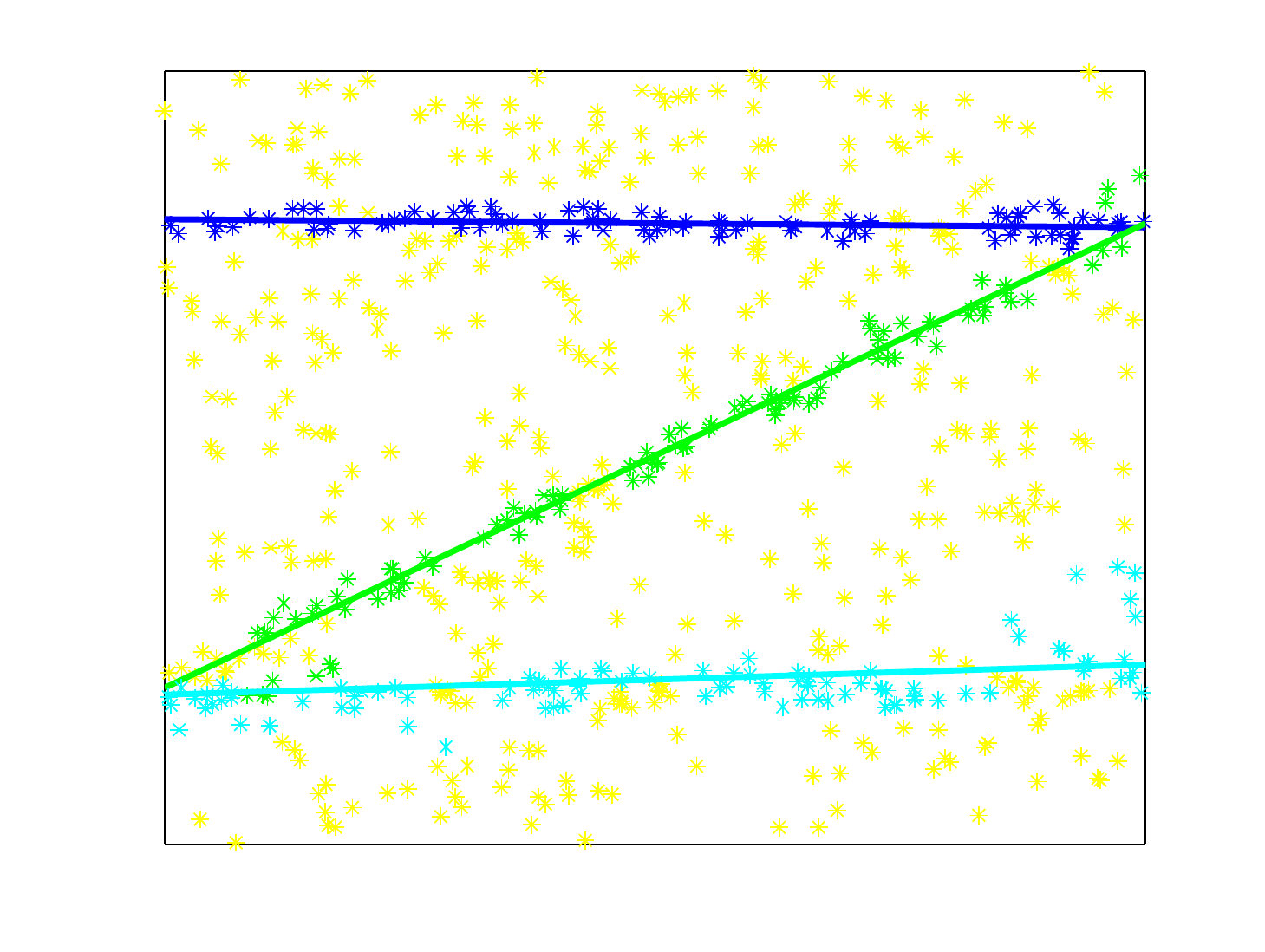}}
  \centerline{\includegraphics[width=1.16\textwidth]{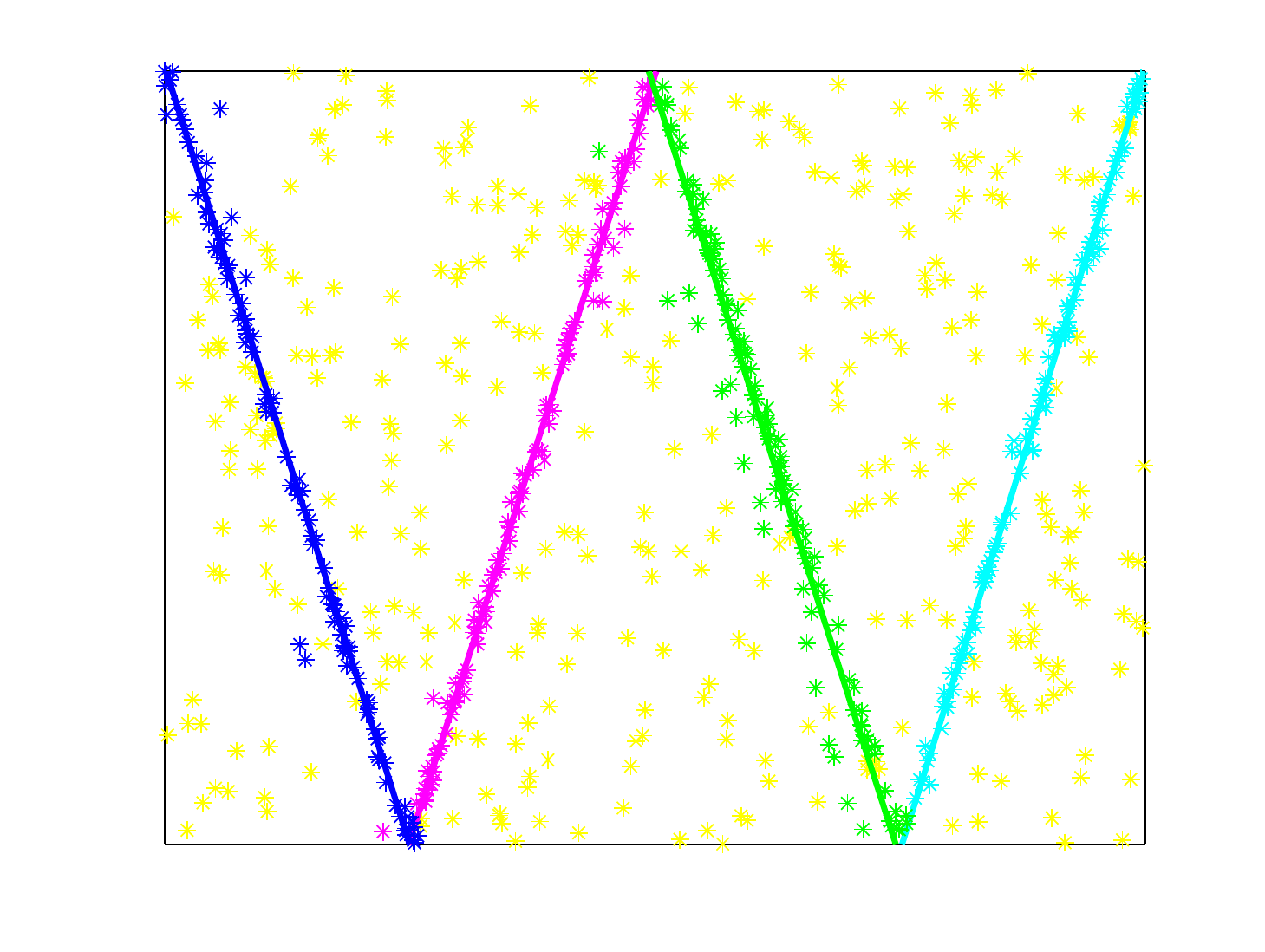}}
  \centerline{\includegraphics[width=1.16\textwidth]{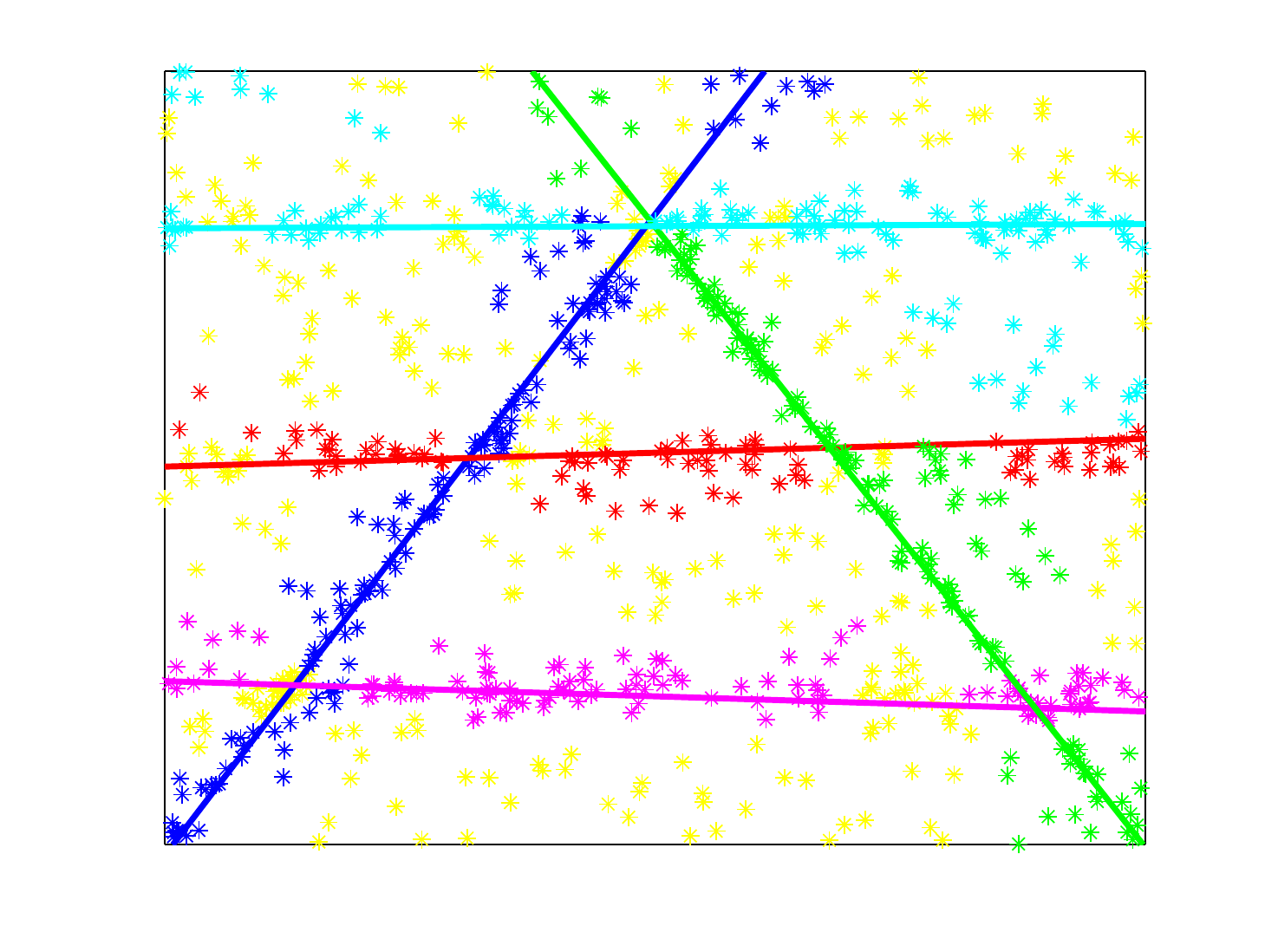}}
  \centerline{\includegraphics[width=1.16\textwidth]{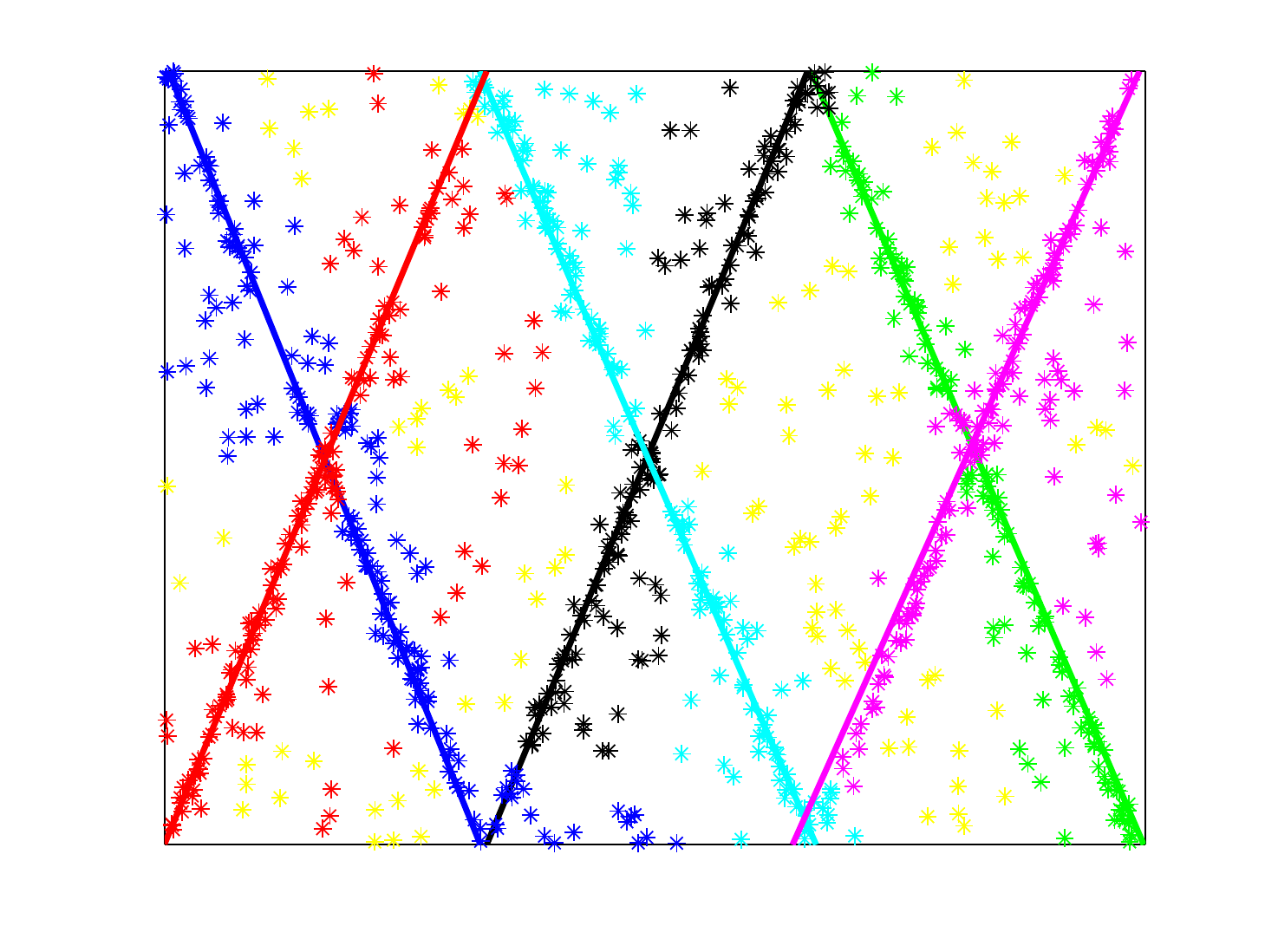}}
  \begin{center} (e)  \end{center}
\end{minipage}
\begin{minipage}[t]{.1585\textwidth}
  \centering
  \centerline{\includegraphics[width=1.16\textwidth]{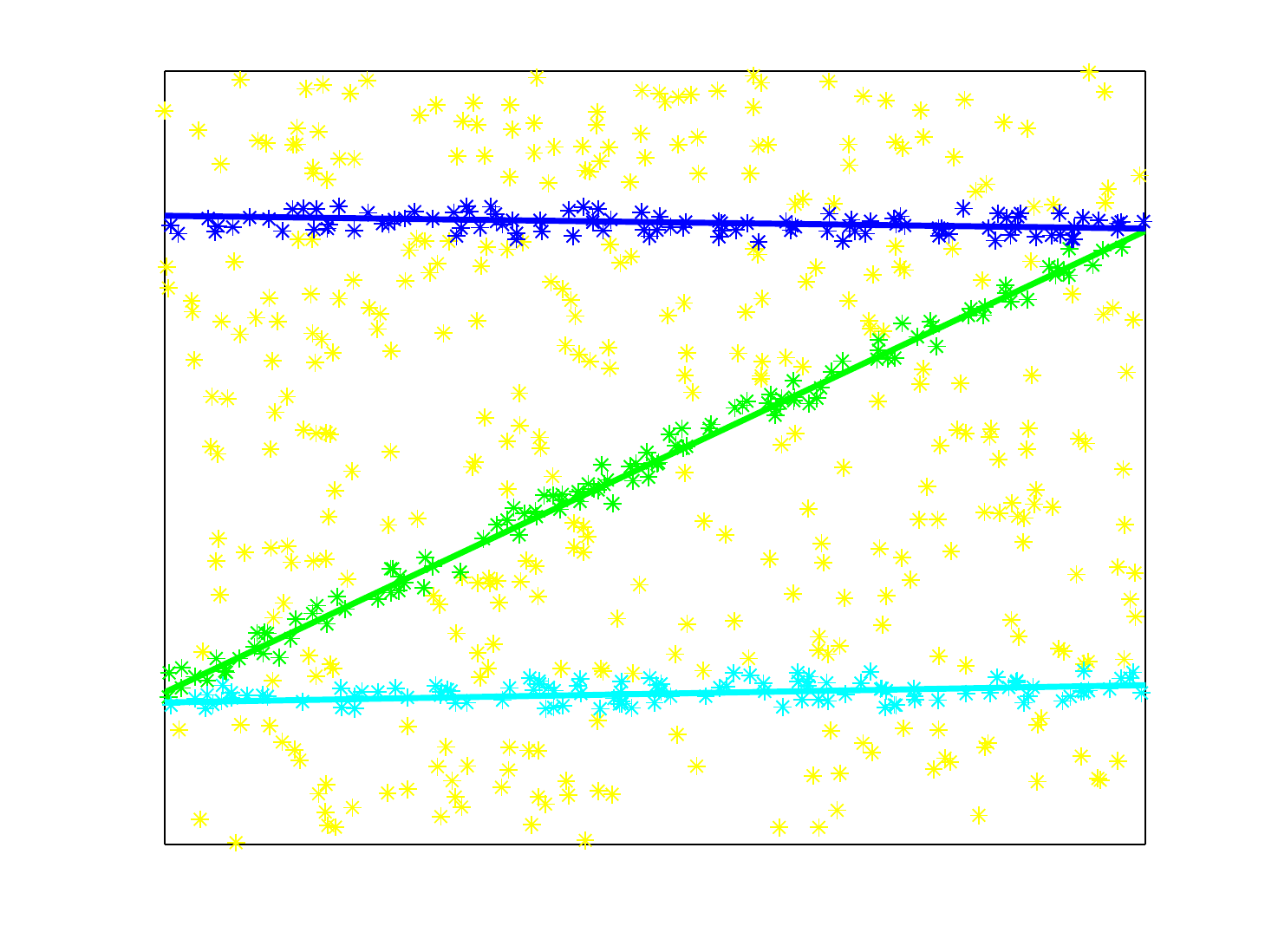}}
  \centerline{\includegraphics[width=1.16\textwidth]{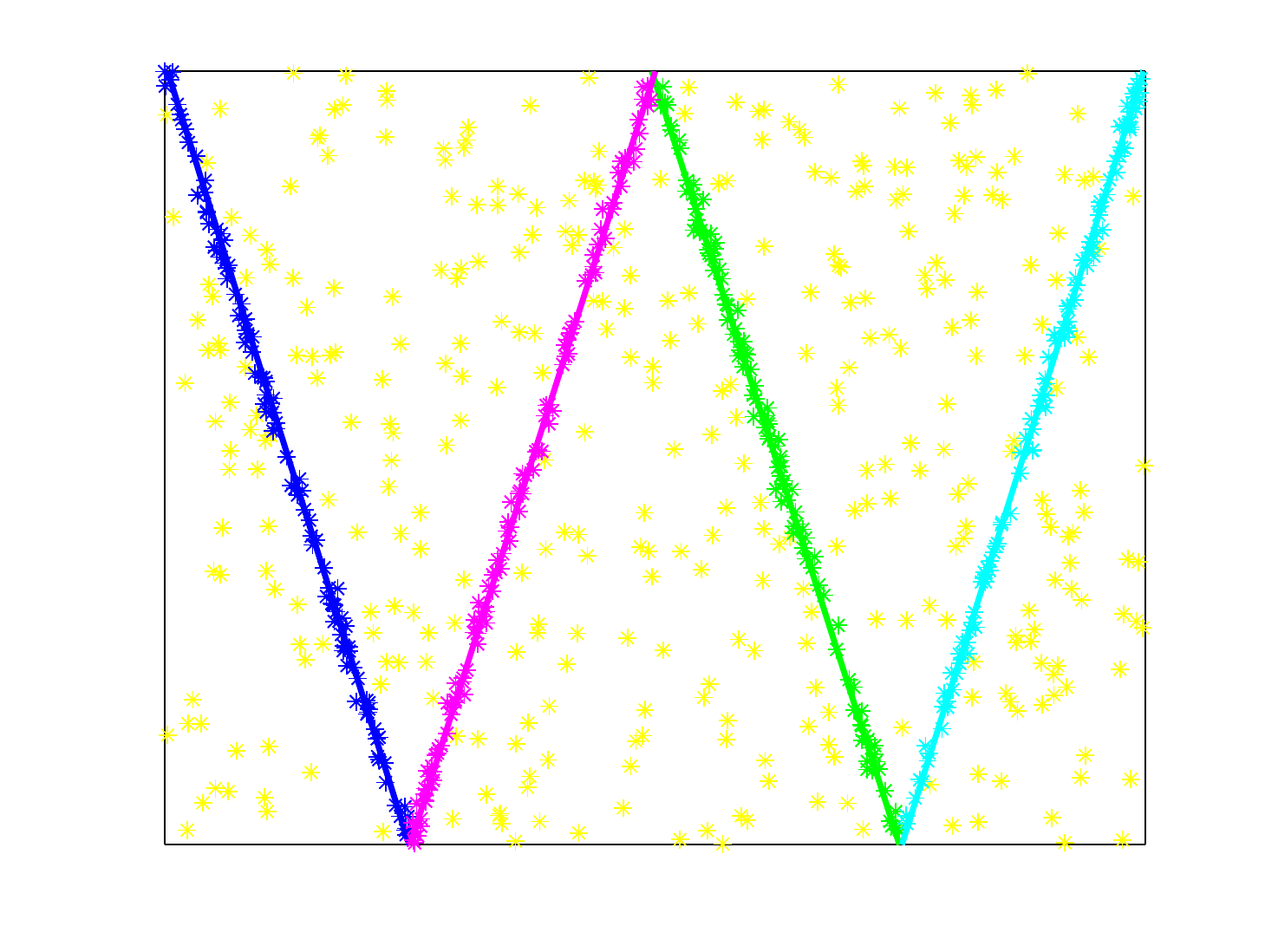}}
  \centerline{\includegraphics[width=1.16\textwidth]{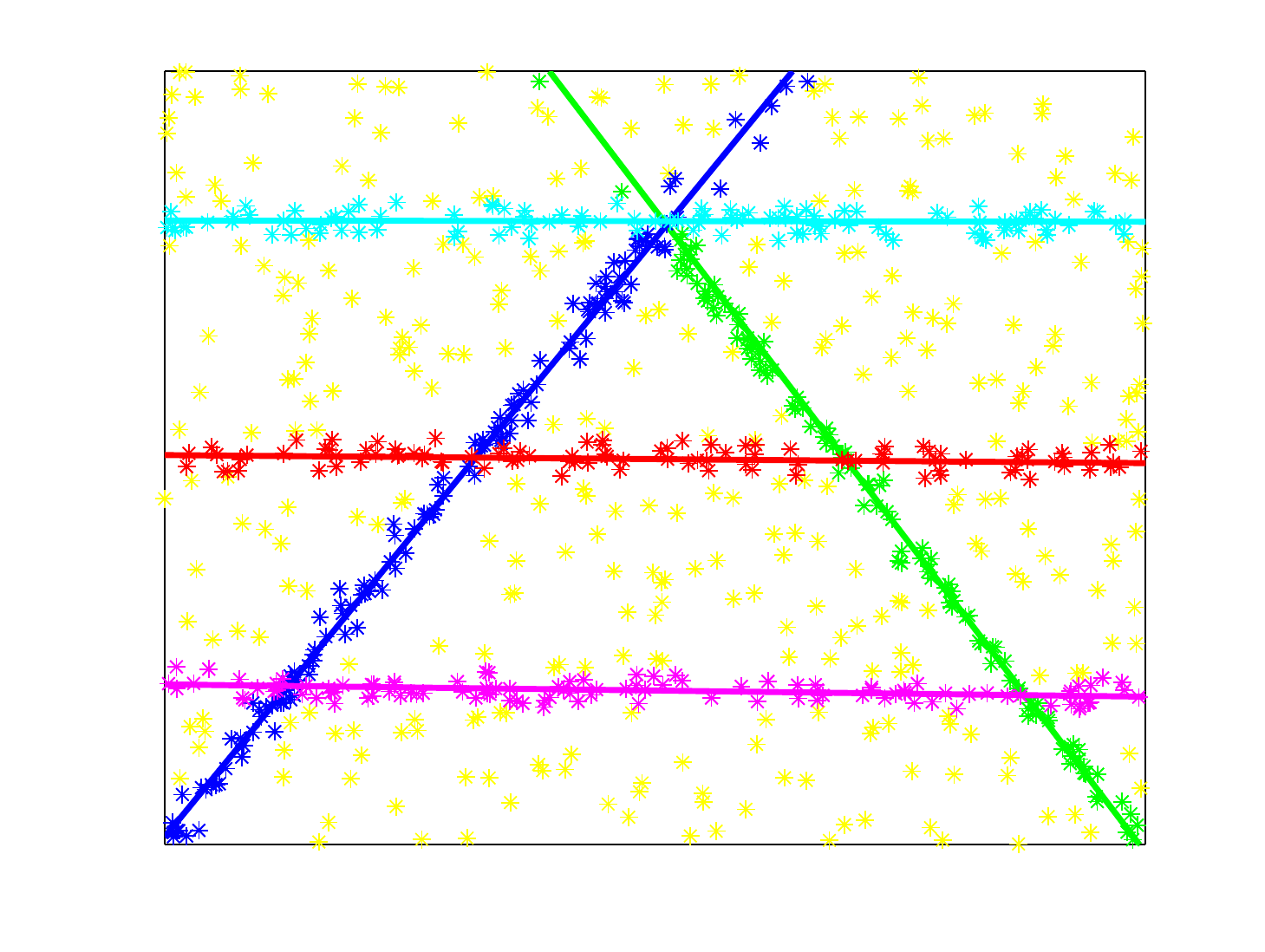}}
  \centerline{\includegraphics[width=1.16\textwidth]{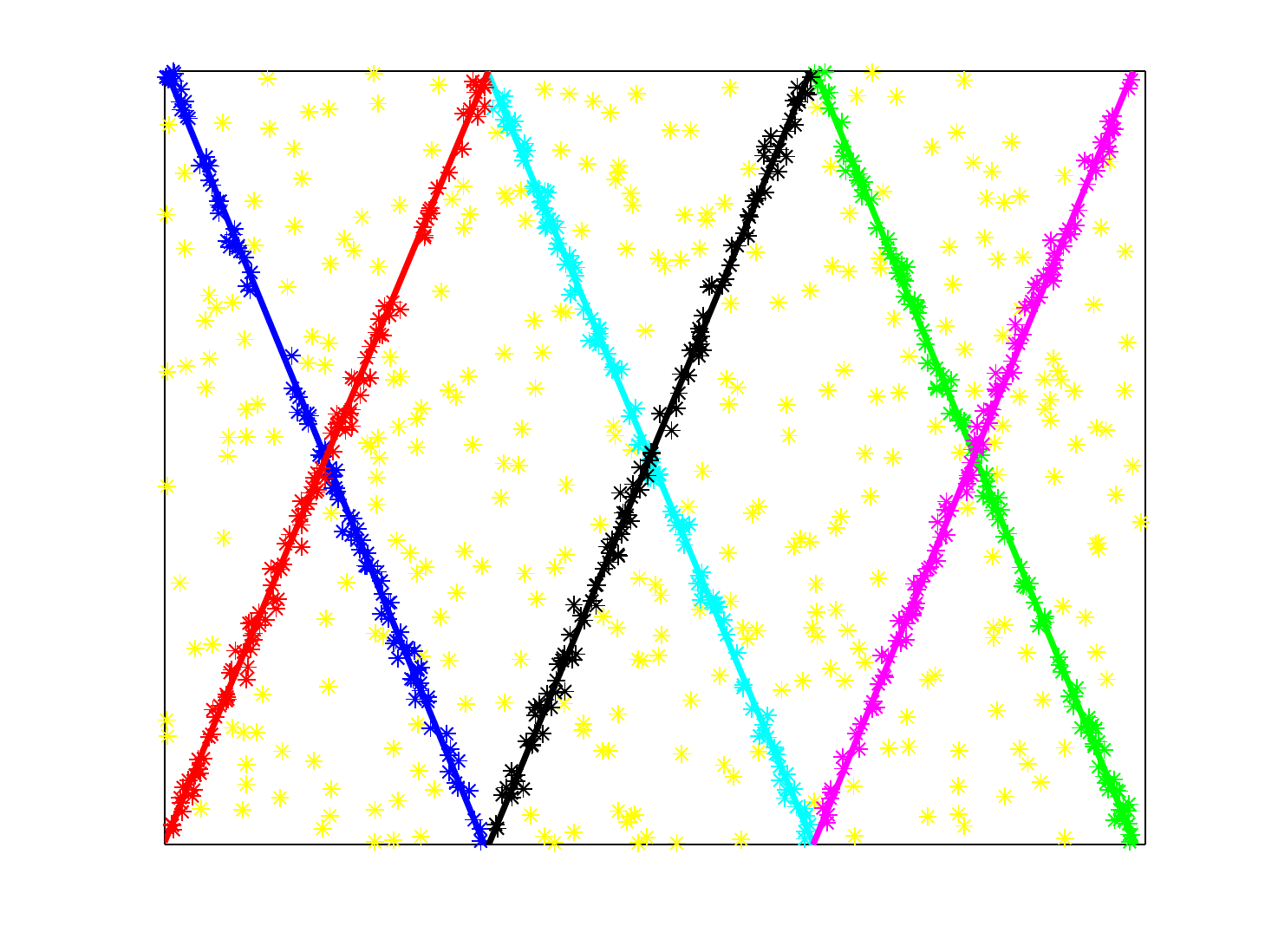}}
  \begin{center} (f)  \end{center}
\end{minipage}
\hfill
\caption{Examples for line fitting and segmentation. $1^{st}$ to $4^{th}$ rows fit and segment three, four, five and six lines, respectively. The corresponding outlier percentages are respectively $86\%$, $88\%$, $89\%$ and $90\%$. The inlier scale is 1.5. (a) The original data with 100 inliers for each line, are distributed in the range of $[0,100]$. (b) to (f) The results obtained by KF, RCG, AKSWH, T-linkage and HF, respectively.}
\label{fig:fivelines}
\end{figure}

Firstly, we evaluate the performance of the five fitting methods on line fitting using four challenging synthetic datasets (see Fig.~\ref{fig:fivelines}). Given the set of ground-truth and estimated line parameters, i.e., $\mathbf{p}=\{p_1,p_2,\ldots,p_{a_0}\}$ and $\mathbf{\tilde{p}}=\{\tilde{p_1},\tilde{p_2},\ldots,\tilde{p_{b_0}}\}$, we compute the error between the pair of parameters as $\|p_i-\tilde{p_j}\|/\sqrt{2}$ {\cite{chin2009robust}}. Then the fitting error between $\mathbf{p}$ and $\mathbf{\tilde{p}}$ is computed as \cite{chin2009robust}:
 \begin{align}
error=|a_0-b_0|+\sum_{i=1}^{\min(a_0,b_0)}\min \varphi_i,
\end{align}
where $\varphi_i$ represents the set of all pairwise errors at the $i$-th summation.

\begin{table}[t]
\centering
\small
\caption{The fitting errors in parameter estimation (and the CPU time in seconds). The smallest fitting errors are boldfaced.}
\centering
\medskip
\begin{tabular}{|c|c|c|c|c|c|}
\hline
   & \small{KF} &\small{RCG} & \small{AKSWH} & \small{T-linkage} & \small{HF} \\
\hline
\multirow{2}{*}{3 lines}& 0.01 &{\bf 0.00} &{\bf 0.00 }  &{\bf 0.00 } & {\bf 0.00 }  \\
& (14.79) & (0.35) &(1.52) & (142.91)& (1.15) \\
\hline
\multirow{2}{*}{4 lines} & 0.48 & 0.32 & 0.32  &  0.34 & {\bf 0.25 }  \\
& (18.45) & (0.42) & (1.59) & (180.57)& (1.32) \\
\hline
\multirow{2}{*}{5 lines} & 1.86 & 2.64 & {\bf 0.27} & 0.44 & 0.32  \\
& (25.56) & (0.58) & (2.85) & (250.27)& (1.18) \\
\hline
\multirow{2}{*}{6 lines} & 0.48 &  3.15 &  0.51 & 0.51 & {\bf 0.47} \\
& (34.70) & (0.60) & (2.98) & (314.94)& (1.24) \\
\hline
\end{tabular}
\label{Table:fivelines}
\end{table}

We repeat each experiment 50 times and show the average results of fitting errors in parameter estimation and the computational speed{, i.e., the CPU time} (we exclude the time used for sampling and generating potential hypotheses which is the same for all the fitting methods) in Table~\ref{Table:fivelines}, and we also show the best fitting results obtained by all the fitting methods in Fig.~\ref{fig:fivelines}.

From Fig.~\ref{fig:fivelines} and Table~\ref{Table:fivelines}, we can see that for the three line data, the five fitting methods succeed in fitting all the lines with low fitting errors. For the four line data, the five methods also succeed in fitting all the lines, but HF achieves the most accurate results among the five methods. For the five line data, KF correctly fits four lines but fails in fitting one line because many inliers are wrongly removed. RCG only correctly fits two lines. Four model instances estimated by RCG overlap to one line. In contrast, AKSWH, T-linkage and HF correctly fit all the lines with the three relatively low fitting errors. The data points are not correctly segmented by T-linkage because T-linkage is a data clustering based fitting method, which cannot effectively deal with the data points near the intersection. For the six line data, RCG only correctly fits two lines although it correctly estimates the number of lines in data by using some used-adjusted thresholds. All of KF, AKSWH, T-linkage and HF succeed in fitting all six lines, but HF achieves the lowest fitting error. Overall, AKSWH and HF have achieved good performance on all the four synthetic datasets, and T-linkage succeeds in fitting all lines with low averaged fitting errors while it wrongly segments many data points for the five line data and the six line data.

For the performance {in terms} of computational time, RCG achieves the fastest speed while it achieves the worst fitting errors for the five line data and the six line data. In contrast, HF achieves the second fastest speed and the lowest or the second lowest fitting errors for all four datasets. HF achieves similar speed with AKSWH for the three line data and the four line data, {but it} is about $140\%$ faster than AKSWH {in} fitting the five line and six line data. HF is faster than KF and T-linkage for all four datasets (about $12.8$-$27.9$ times faster than KF and about $124.2$-$253.9$ faster than T-linkage).

\begin{figure}[t]
\centering
\begin{minipage}{.4\textwidth}
\centering
\centerline{\includegraphics[width=1.02\textwidth]{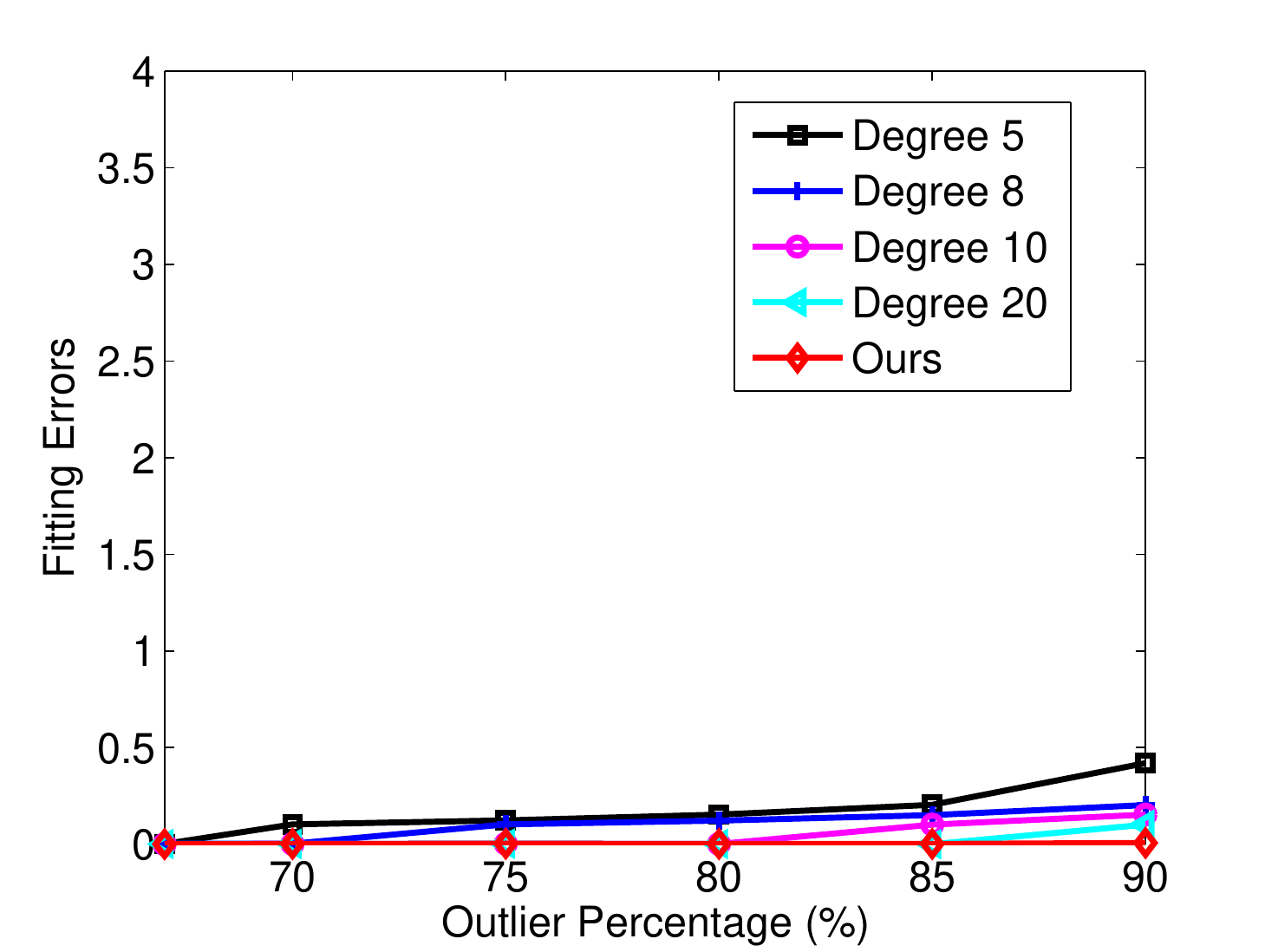}}
  \begin{center} {(a)} \end{center}
\end{minipage}
\begin{minipage}{.4\textwidth}
\centerline{\includegraphics[width=1.02\textwidth]{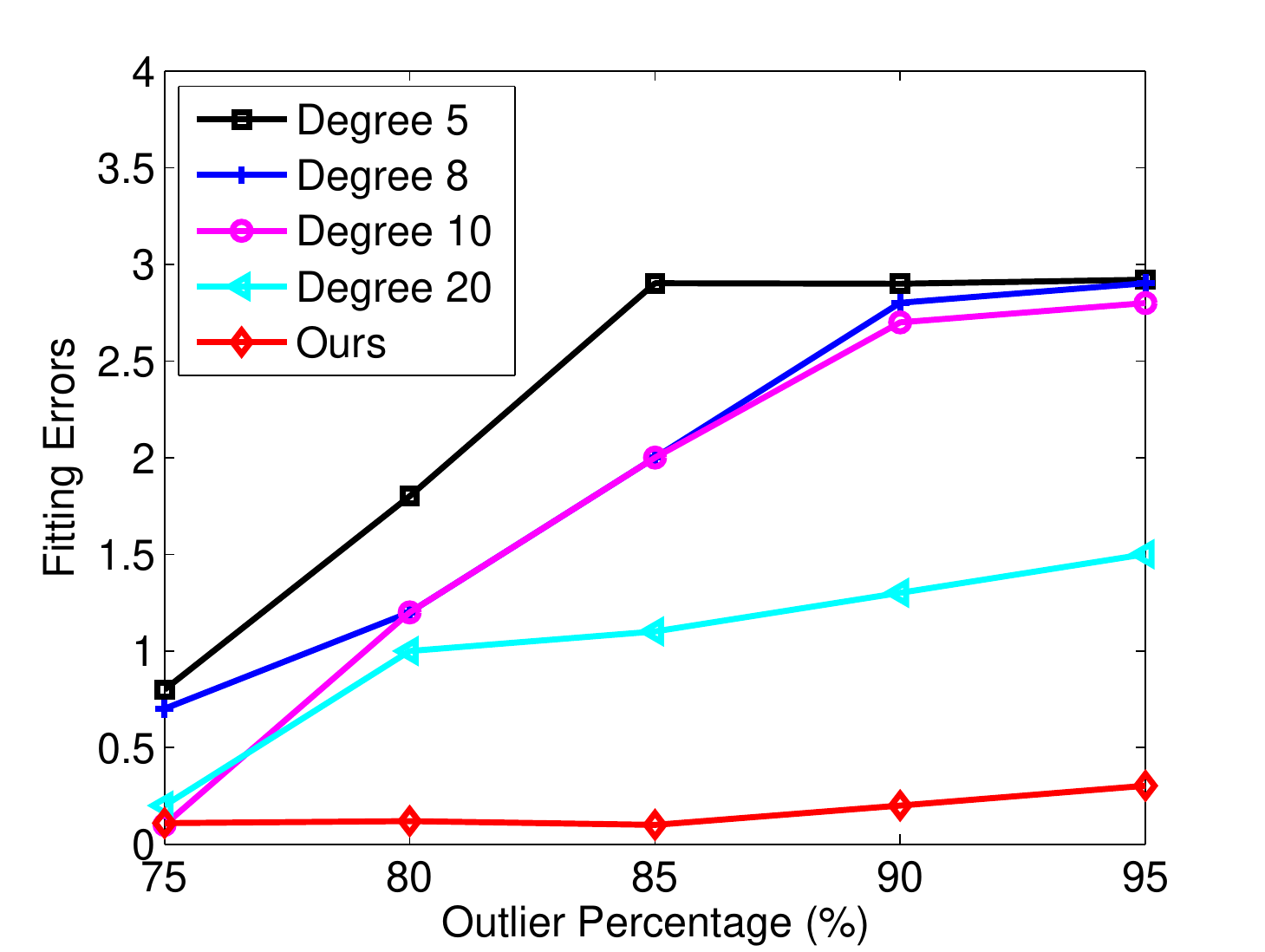}}
  \begin{center} (b) \end{center}
\end{minipage}
\caption{The average fitting errors {obtained by} HF based on {the constructed hypergraph in this paper and different uniform hypergraphs with different fixed degrees in} data with different outlier percentages: (a) and (b) show the performance comparison on the three line data, and the four line data, respectively.}
\label{fig:differentcardinality}
\end{figure}
We also evaluate the performance of HF based on {the constructed hypergraph in this paper, and the different uniform hypergraphs with different fixed degrees} in data with different outlier percentages, which can show the ability of hyperedges with large degrees in dealing with the model fitting problem, as shown in Fig.~\ref{fig:differentcardinality}. We use two datasets (i.e., the three line data and the four line data) in Fig.~\ref{fig:fivelines} for the evaluation. We change the number of outliers to obtain different outlier percentages on the two data. From Fig.~\ref{fig:differentcardinality}(a), we can see that HF based on different hypergraphs {with different degrees} achieves low fitting errors for the three line data with different outlier percentages, but HF based on the proposed hypergraph (recall that the hyperedges have {larger justified degrees} in the hypergraph) obtains more stable fitting results than HF based on the other hypergraphs. For the four line data with different outlier percentages, as shown in Fig.~\ref{fig:differentcardinality}(b), HF achieves better results as the hyperedge degree is increased while HF based on lower hyperedge degree {generally} obtains larger fitting errors{. This} further verifies that using large degrees of hyperedges is beneficial for the hypergraph partition in HF.

\begin{figure}[t]
\centering
\begin{minipage}[t]{.1585\textwidth}
 \centering
 \centerline{\includegraphics[width=1.17\textwidth]{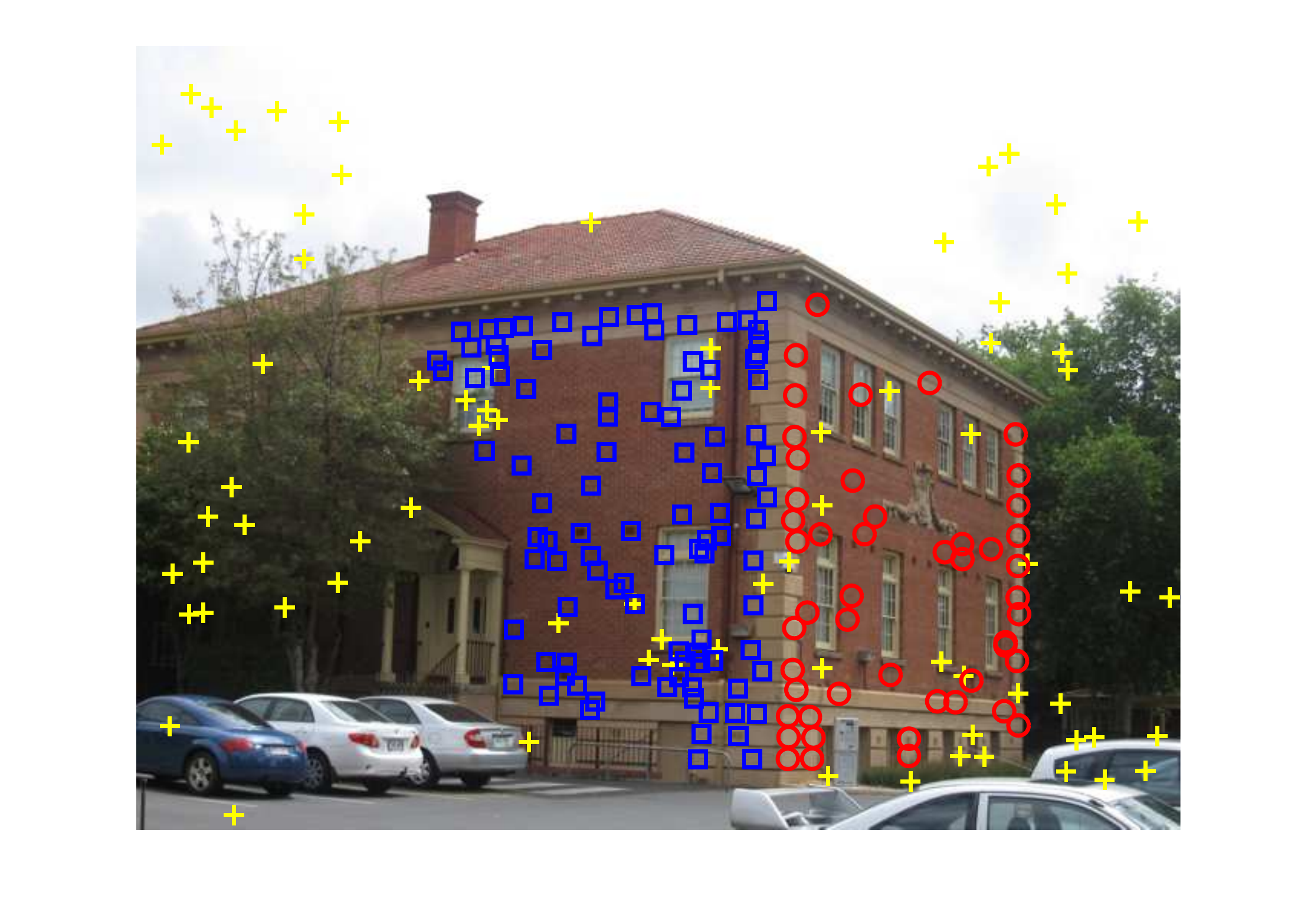}}
  \centerline{\includegraphics[width=1.17\textwidth]{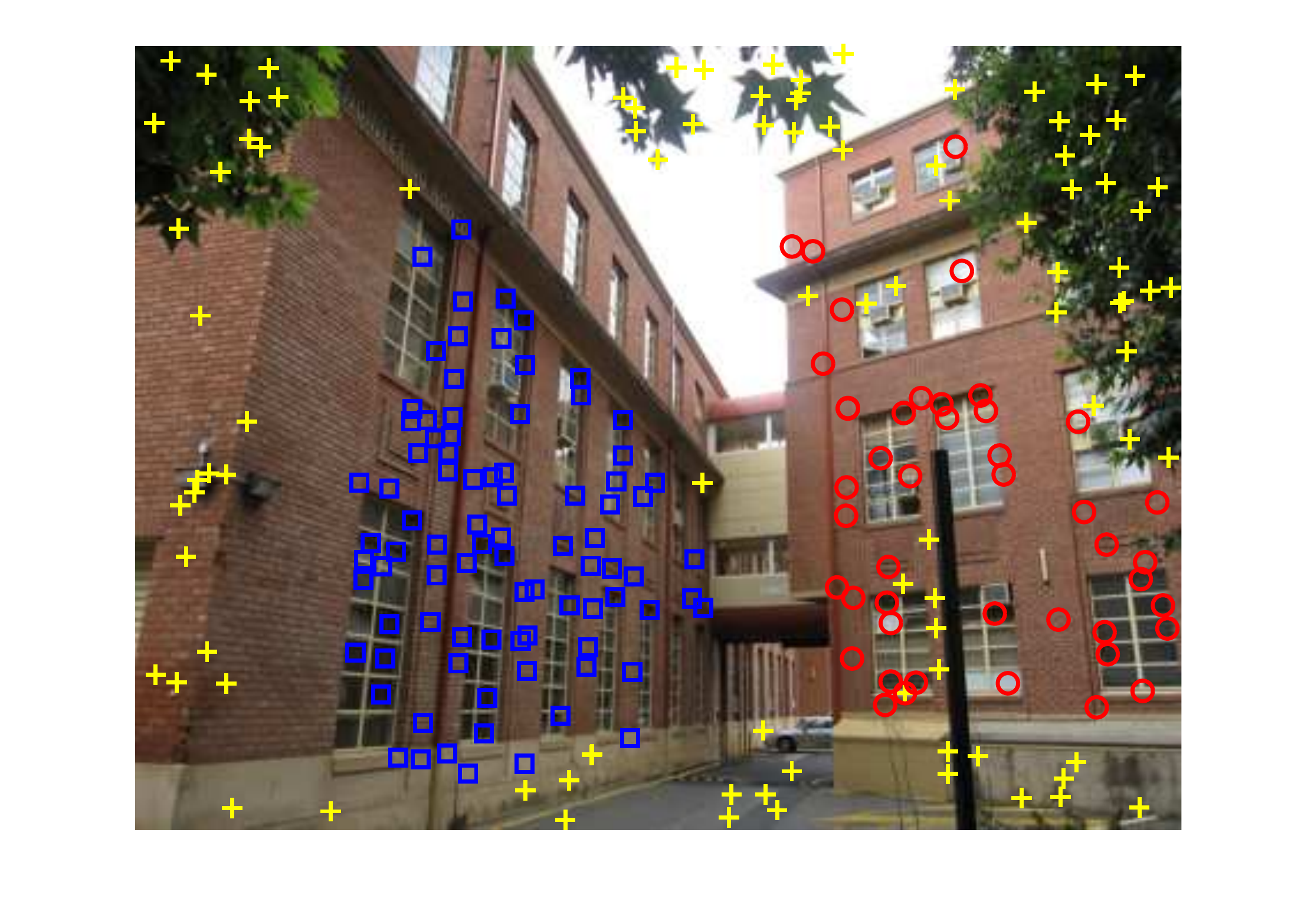}}
  \centerline{\includegraphics[width=1.17\textwidth]{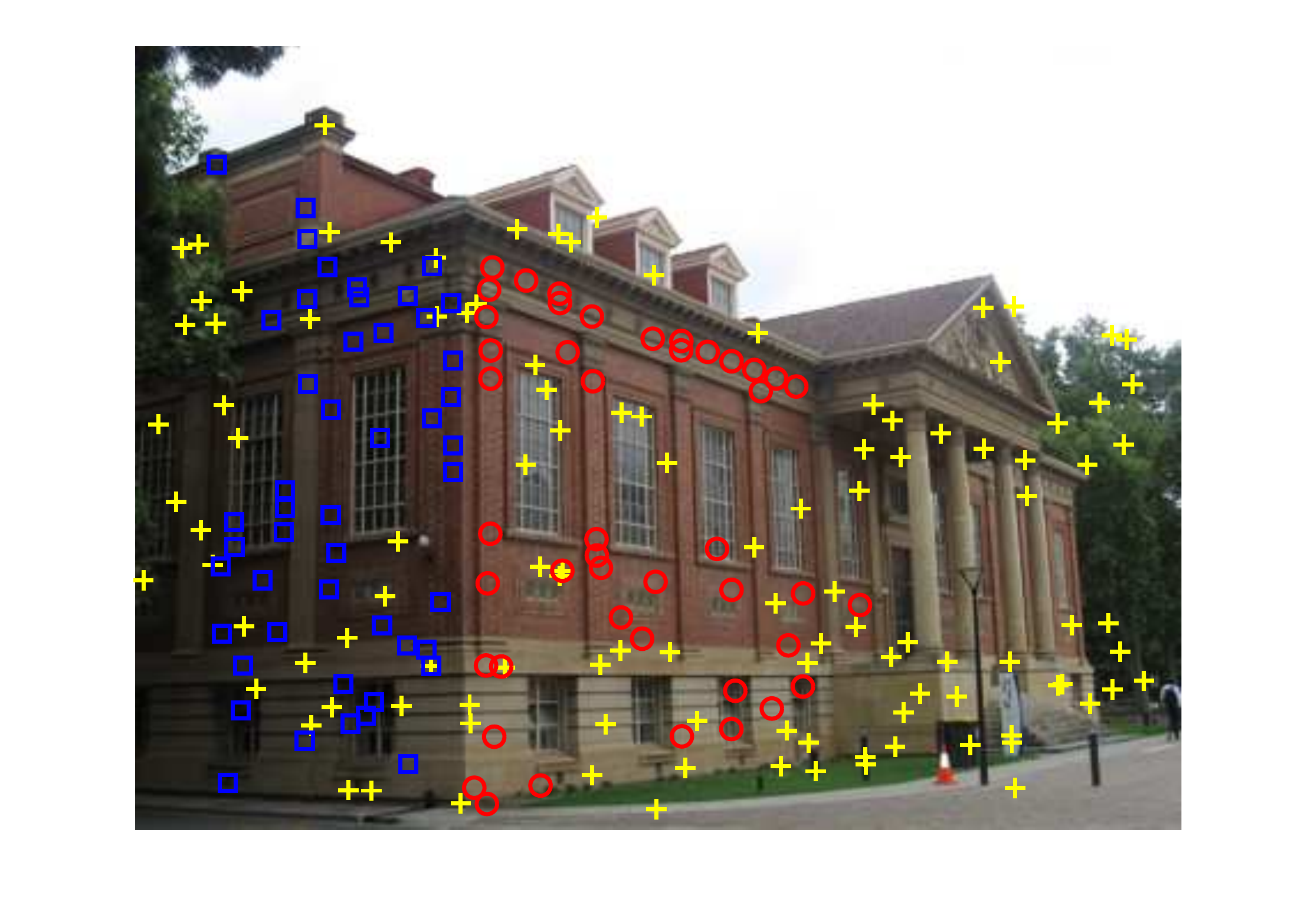}}
\centerline{\includegraphics[width=1.17\textwidth]{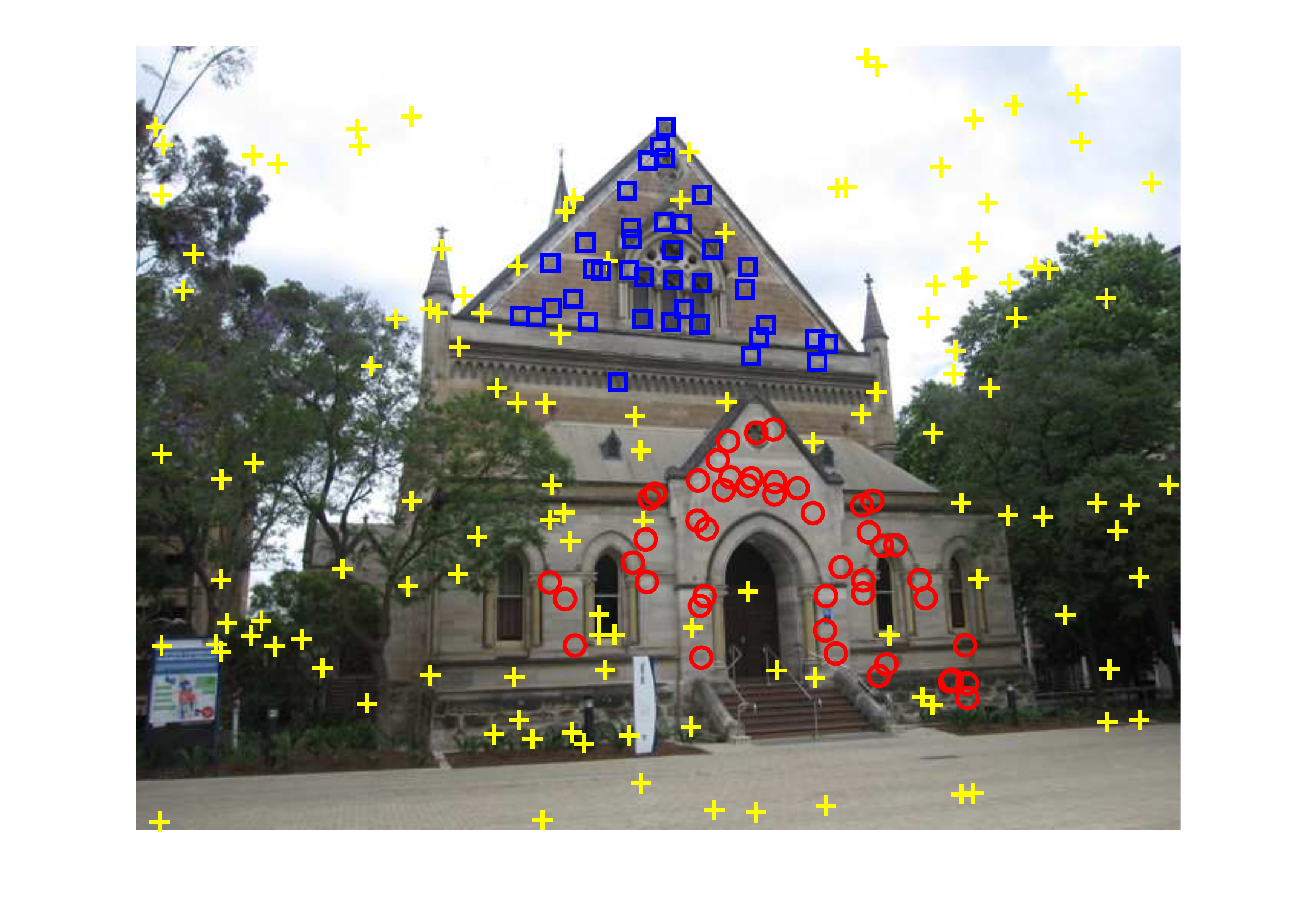}}
  \centerline{\includegraphics[width=1.17\textwidth]{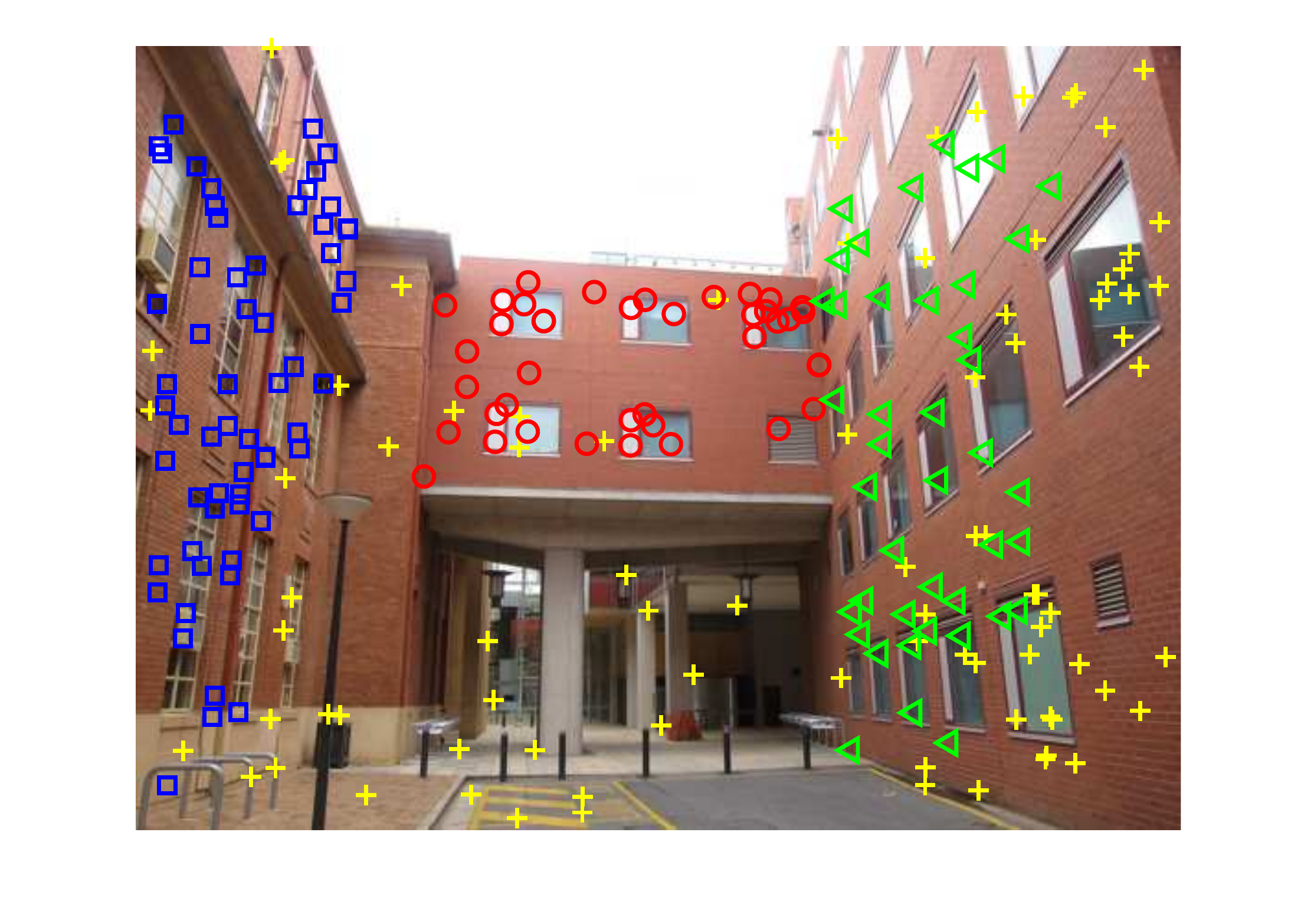}}
 \centerline{\includegraphics[width=1.17\textwidth]{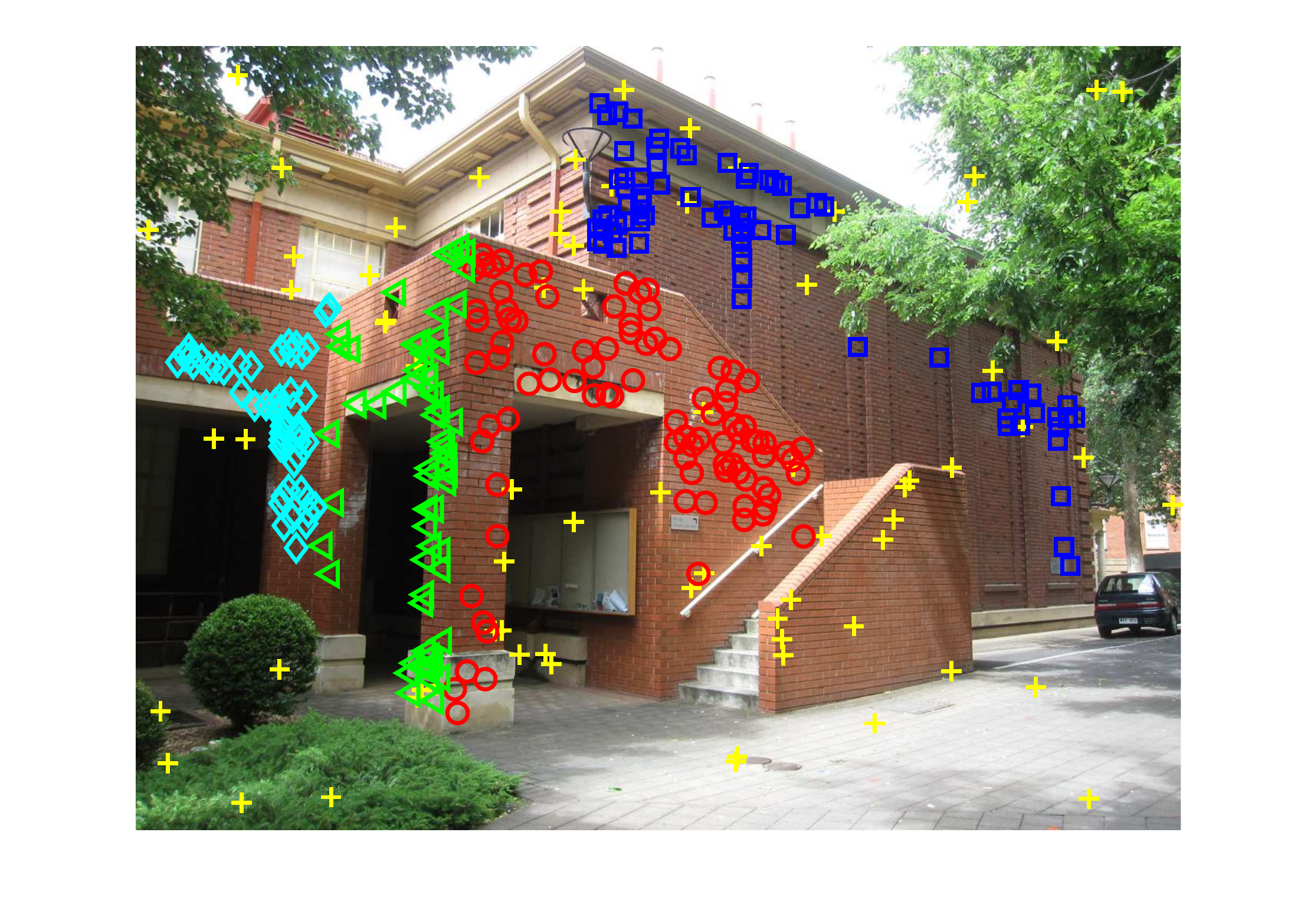}}
  \begin{center} (a) \end{center}
\end{minipage}
\begin{minipage}[t]{.1585\textwidth}
  \centering
 \centerline{\includegraphics[width=1.17\textwidth]{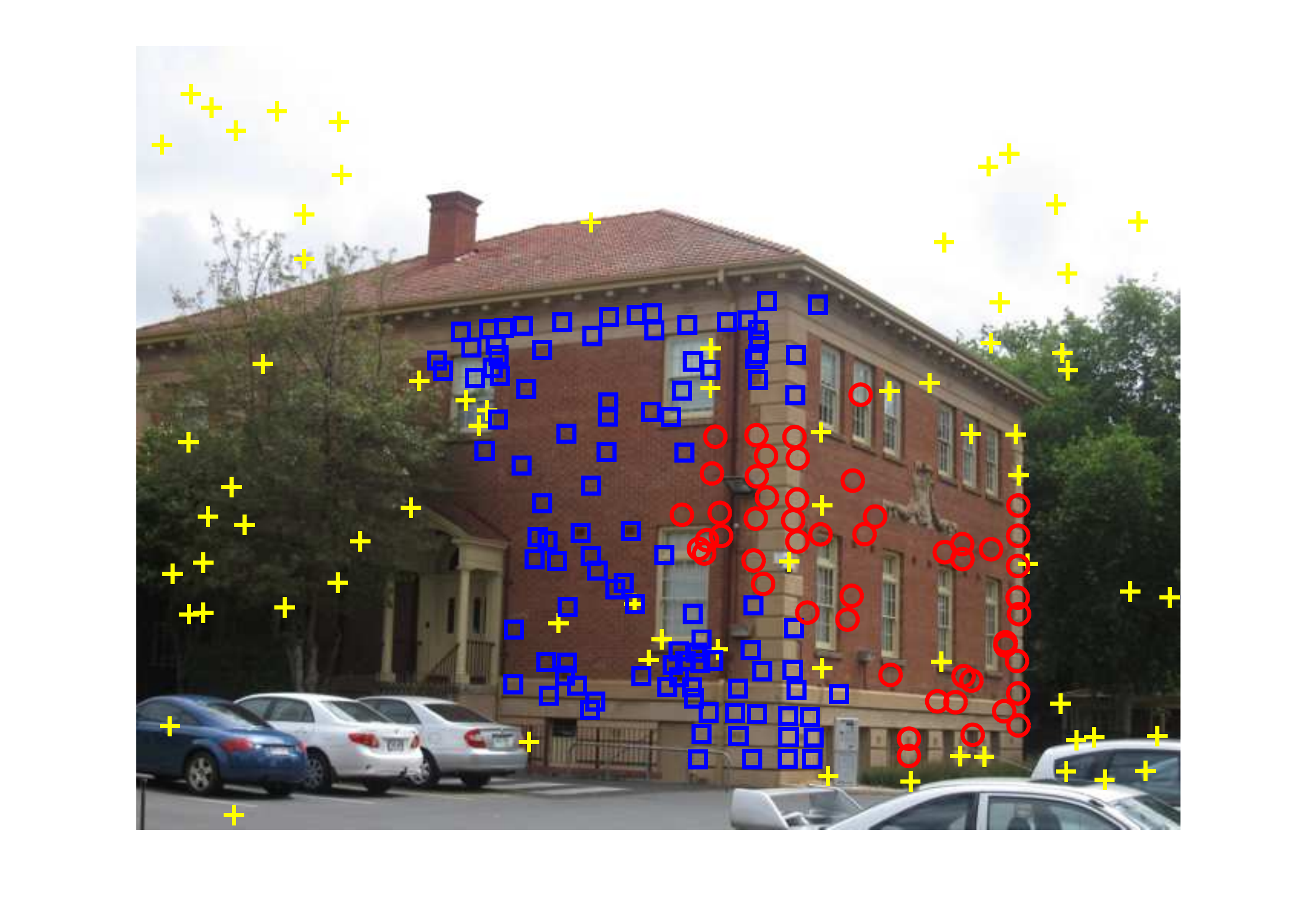}}
  \centerline{\includegraphics[width=1.17\textwidth]{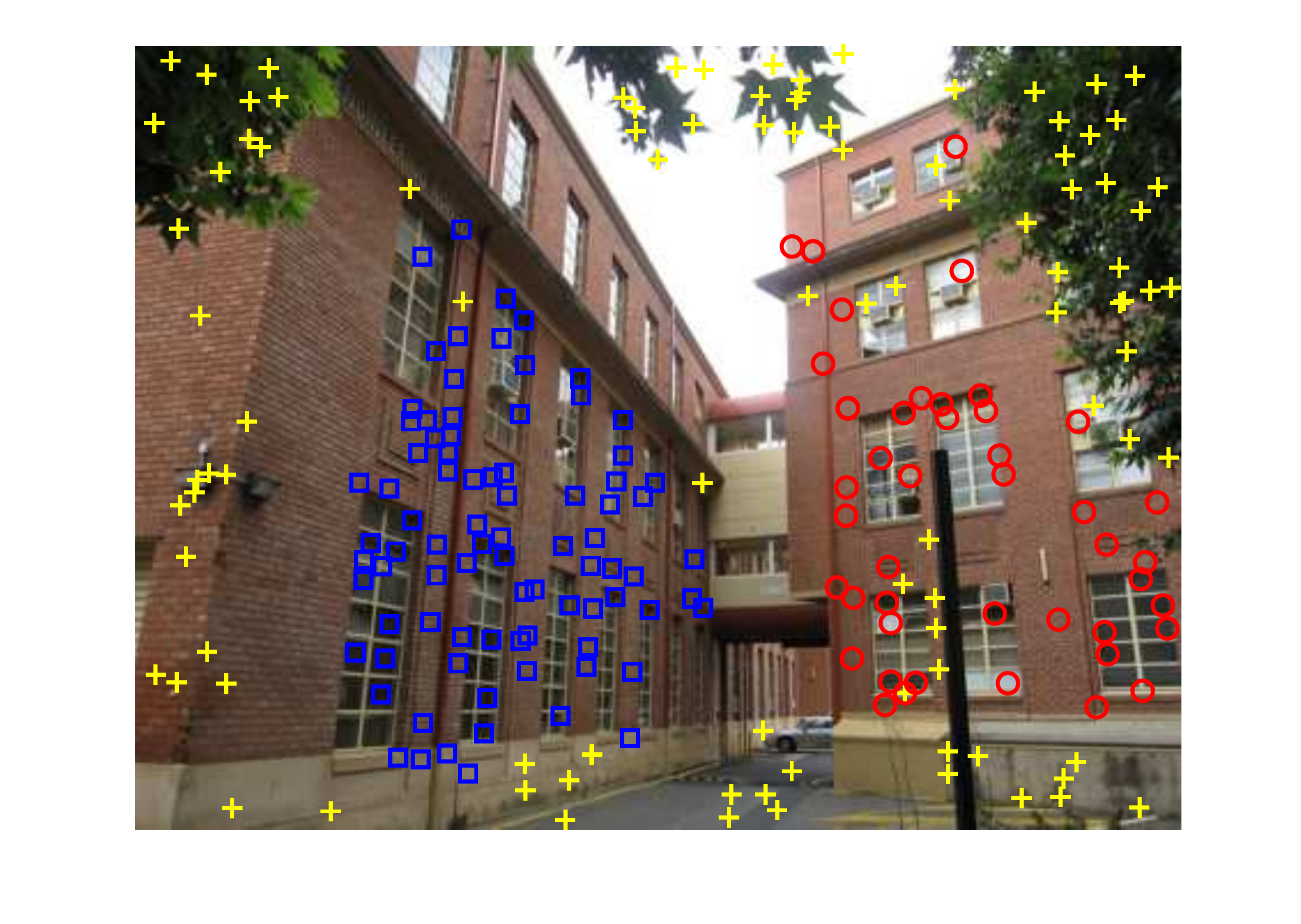}}
  \centerline{\includegraphics[width=1.17\textwidth]{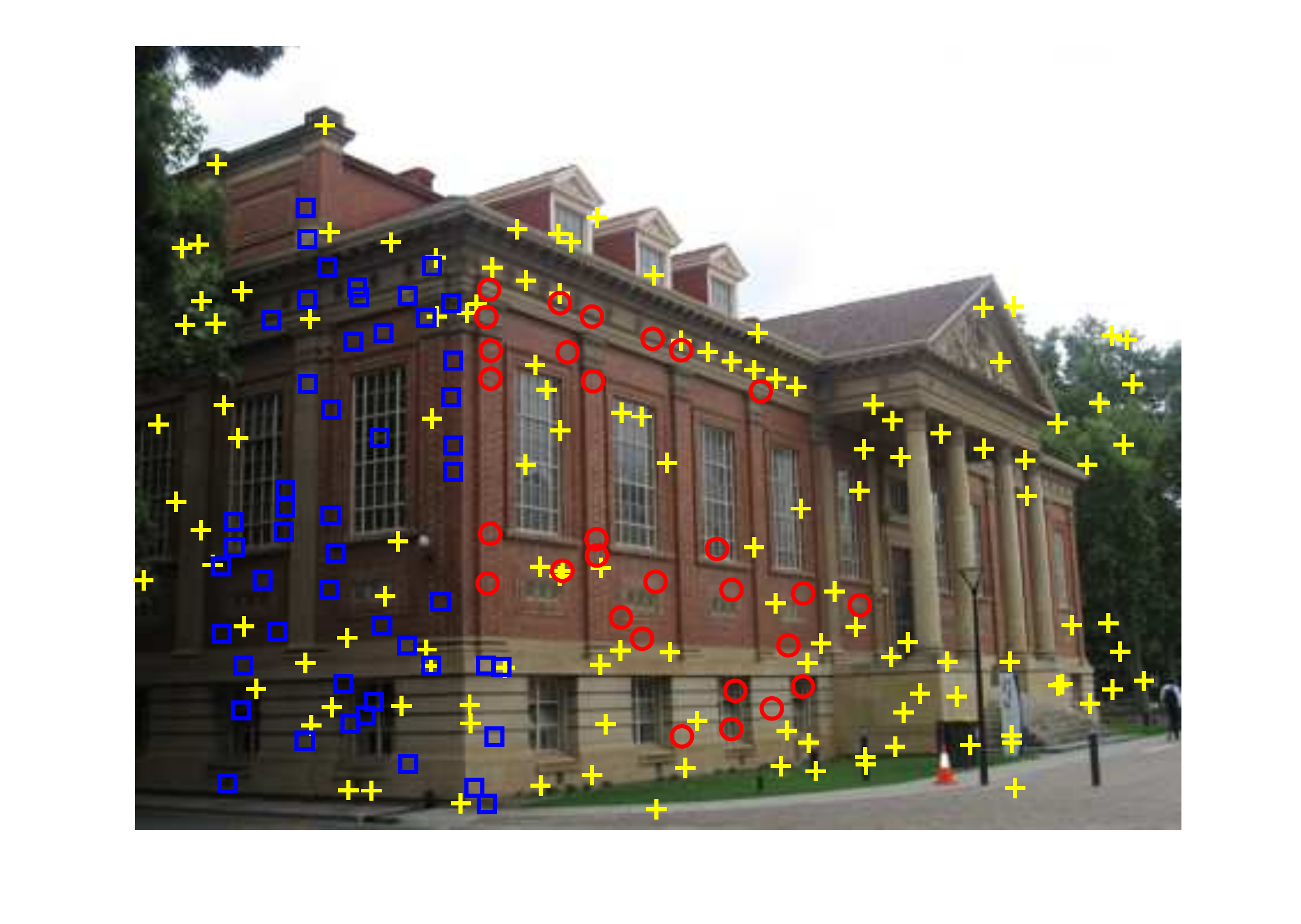}}
\centerline{\includegraphics[width=1.17\textwidth]{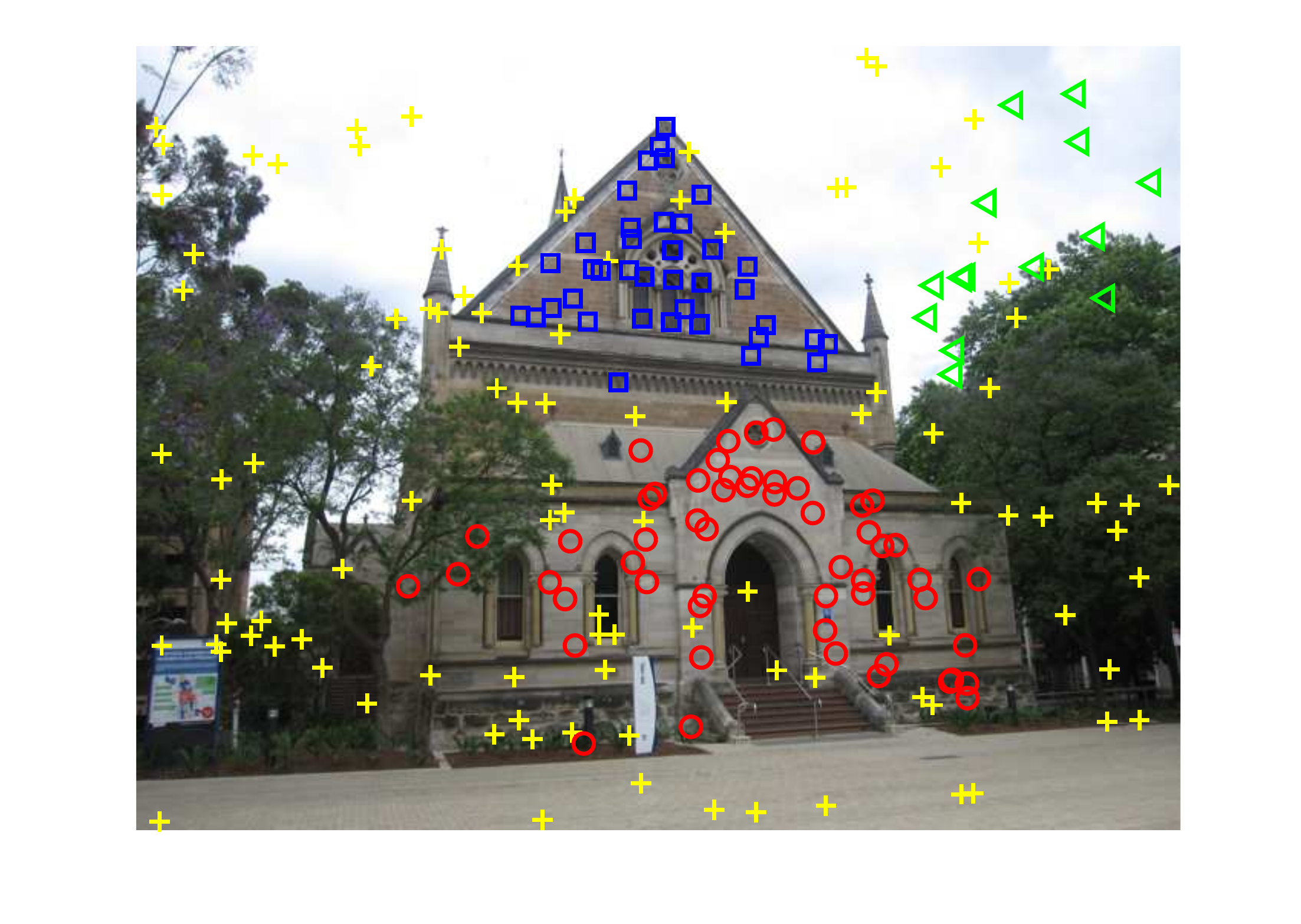}}
  \centerline{\includegraphics[width=1.17\textwidth]{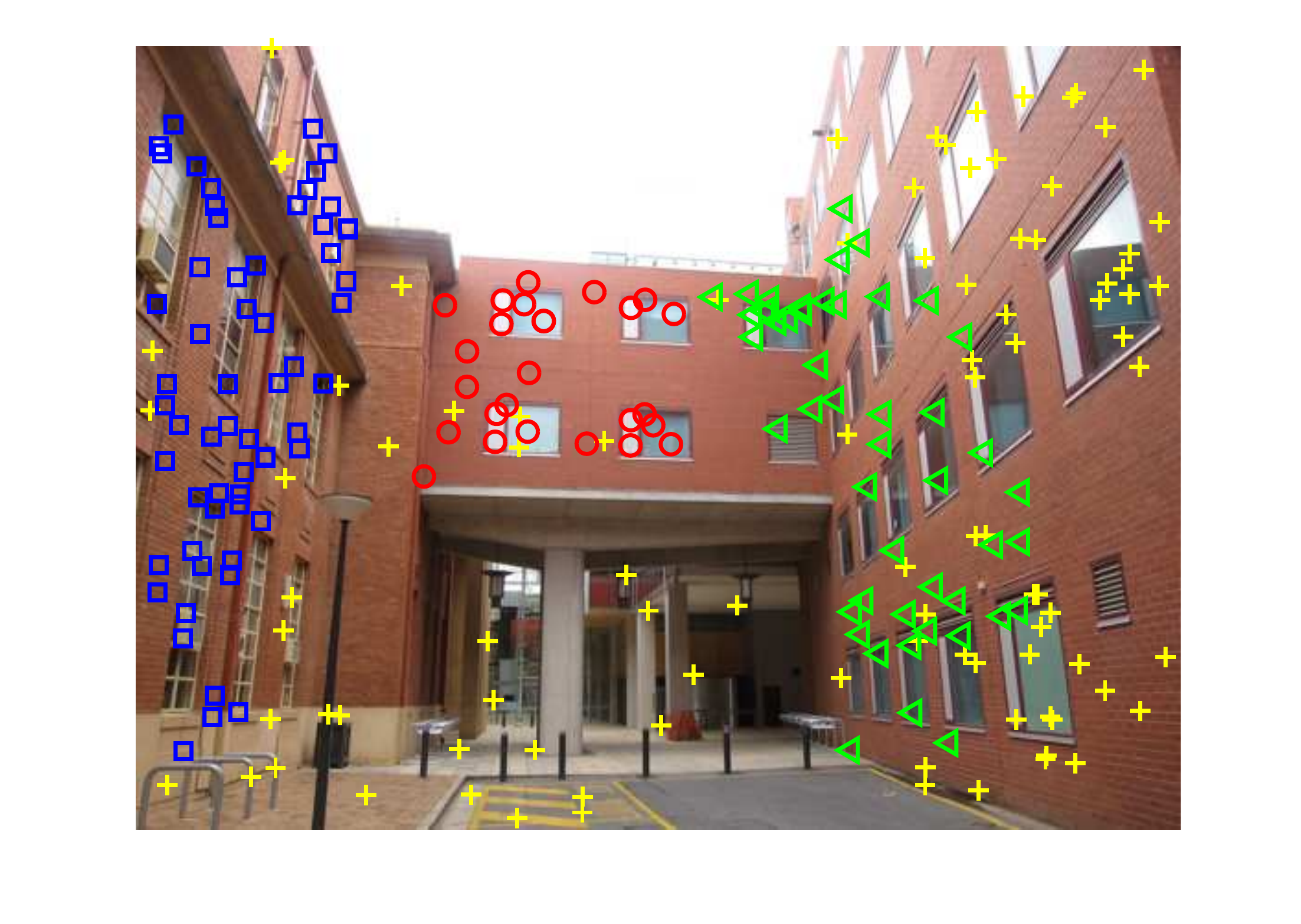}}
 \centerline{\includegraphics[width=1.17\textwidth]{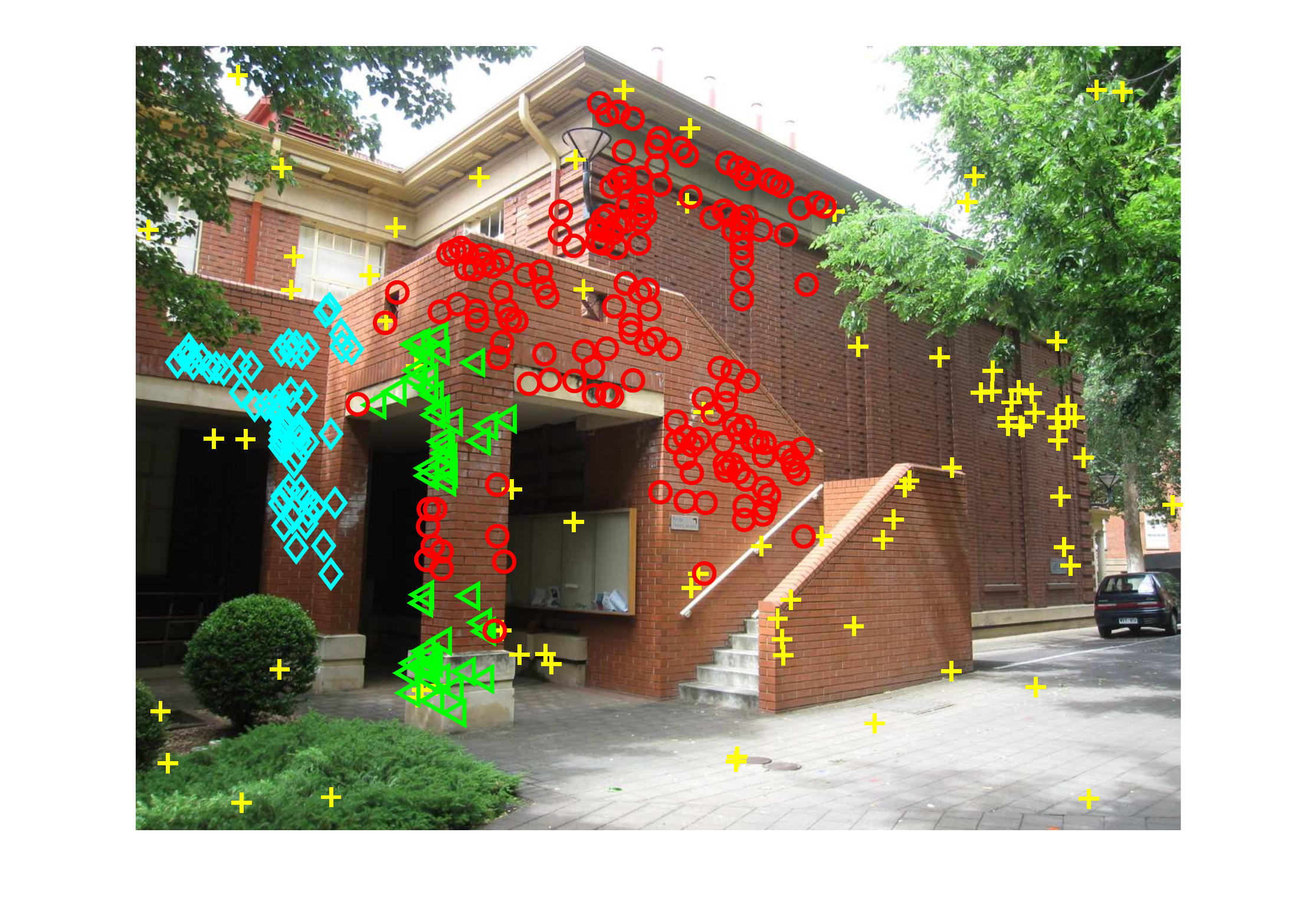}}
  \begin{center} (b)  \end{center}
\end{minipage}
\begin{minipage}[t]{.1585\textwidth}
  \centering
 \centerline{\includegraphics[width=1.17\textwidth]{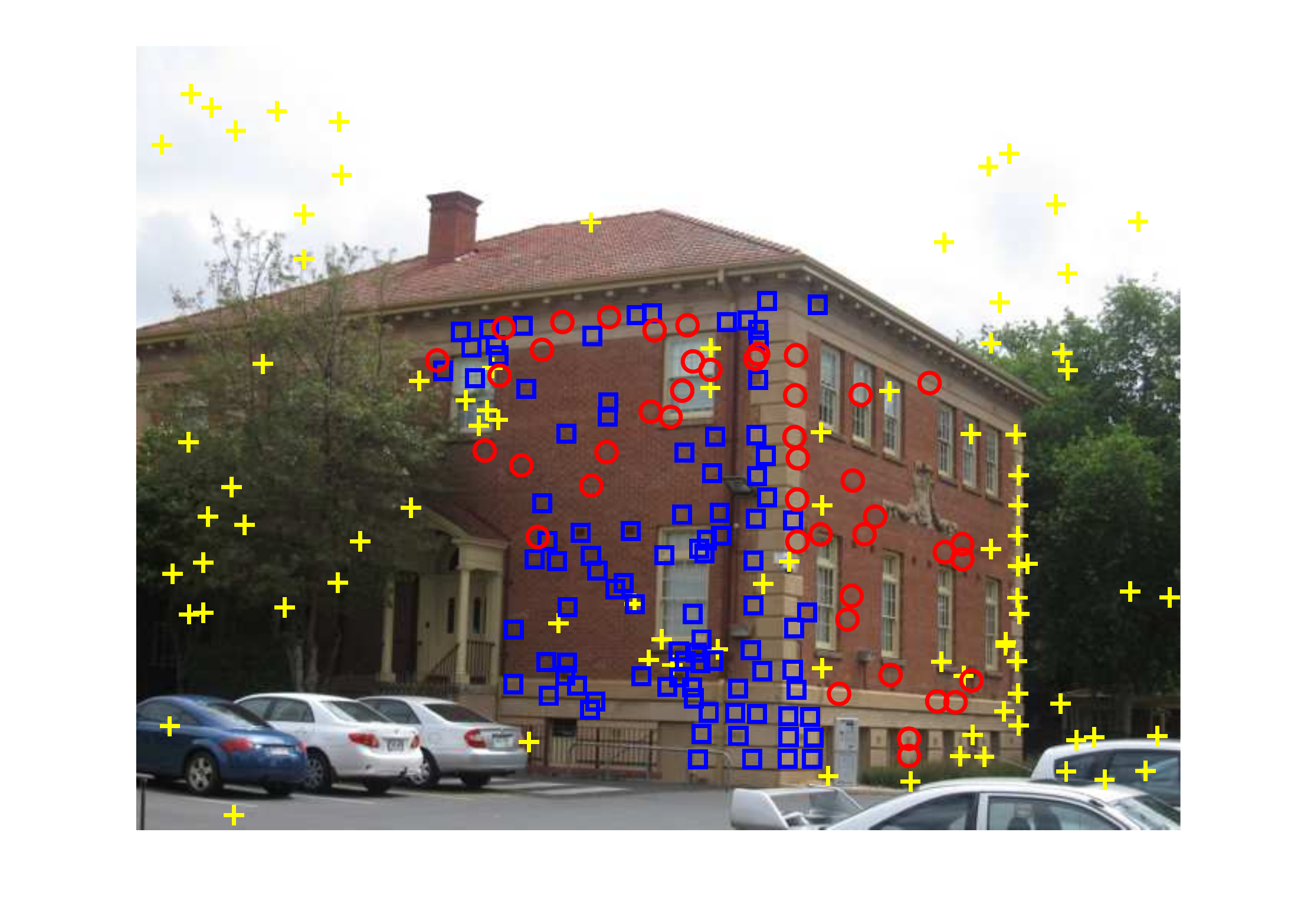}}
  \centerline{\includegraphics[width=1.17\textwidth]{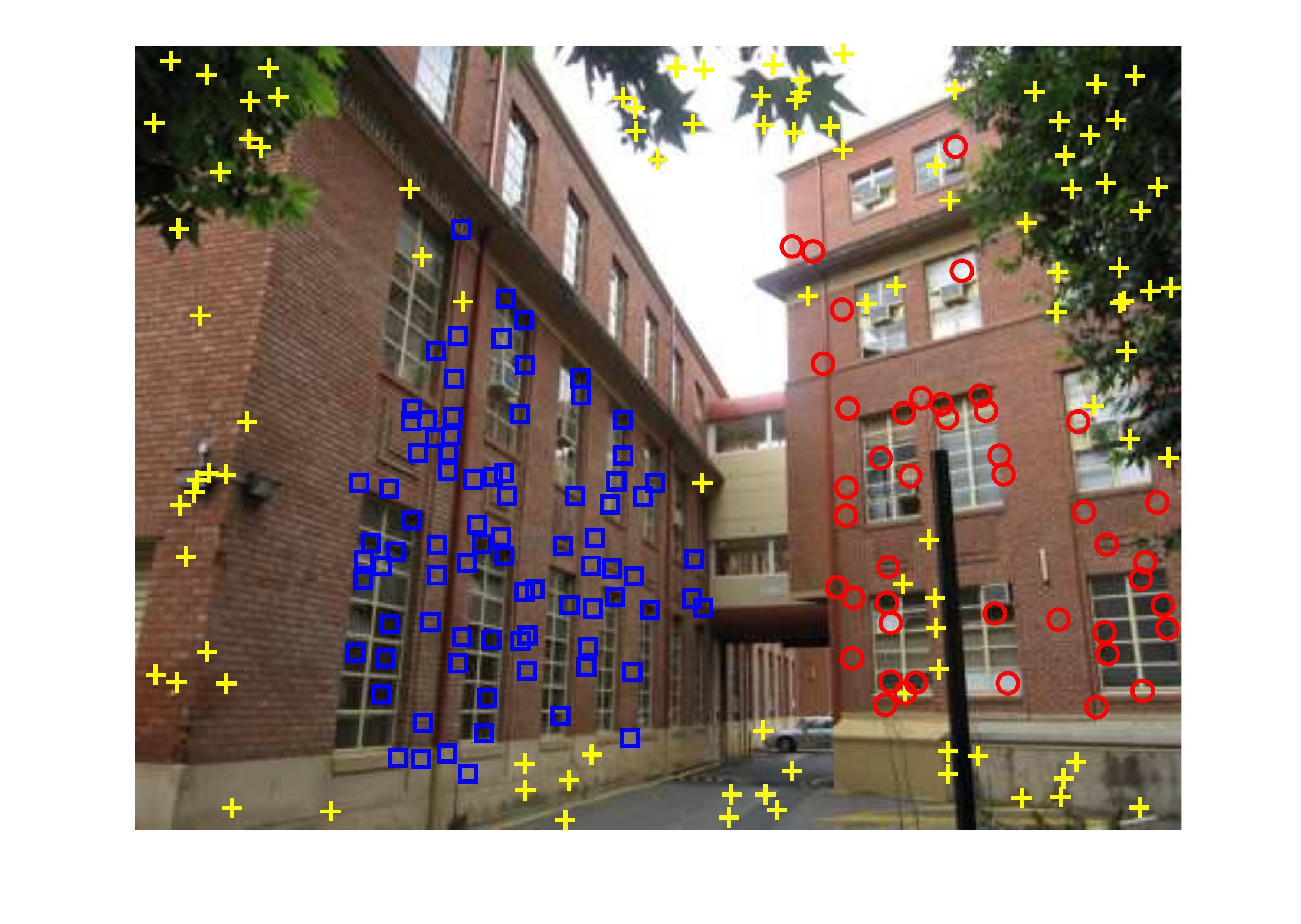}}
  \centerline{\includegraphics[width=1.17\textwidth]{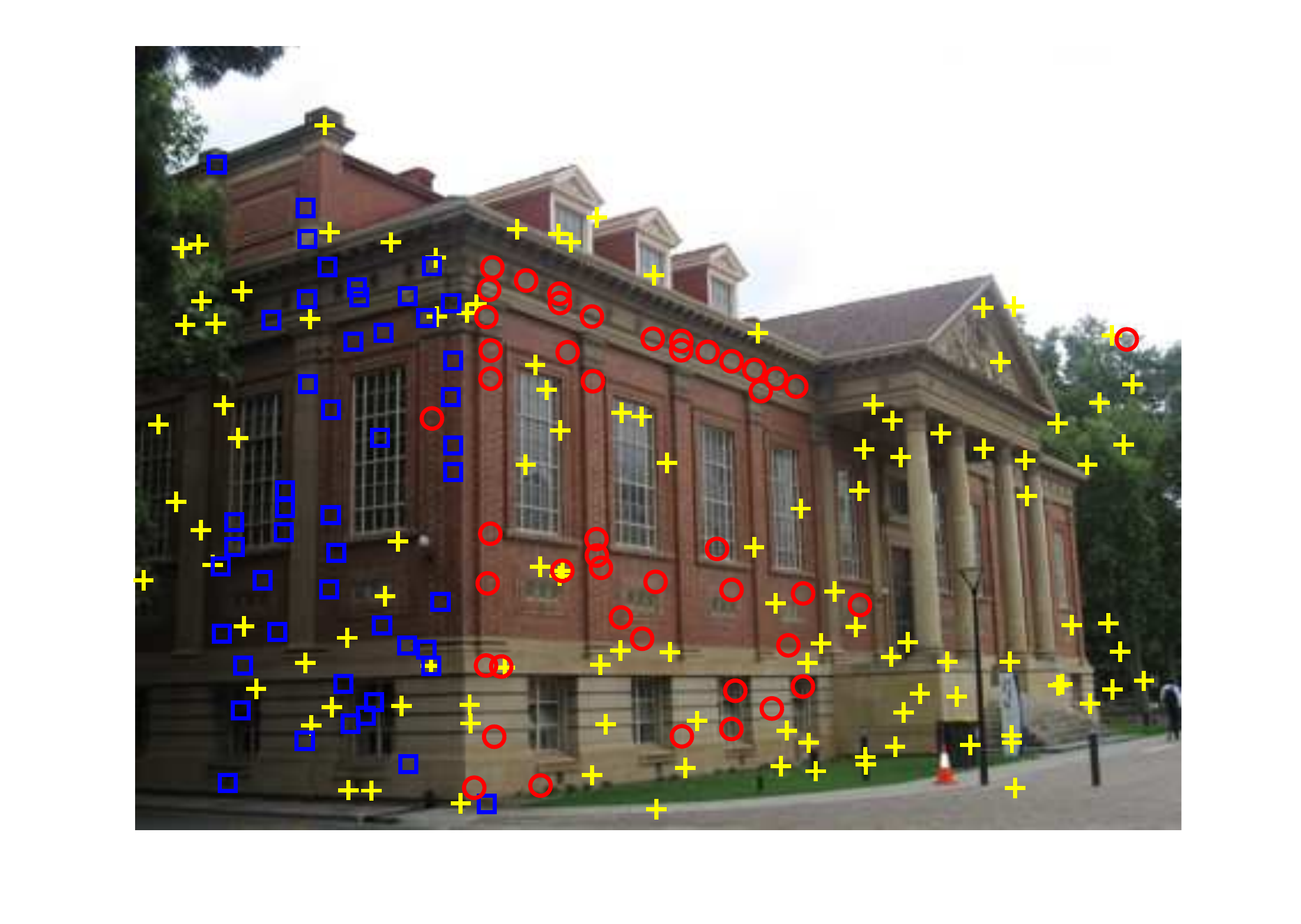}}
\centerline{\includegraphics[width=1.17\textwidth]{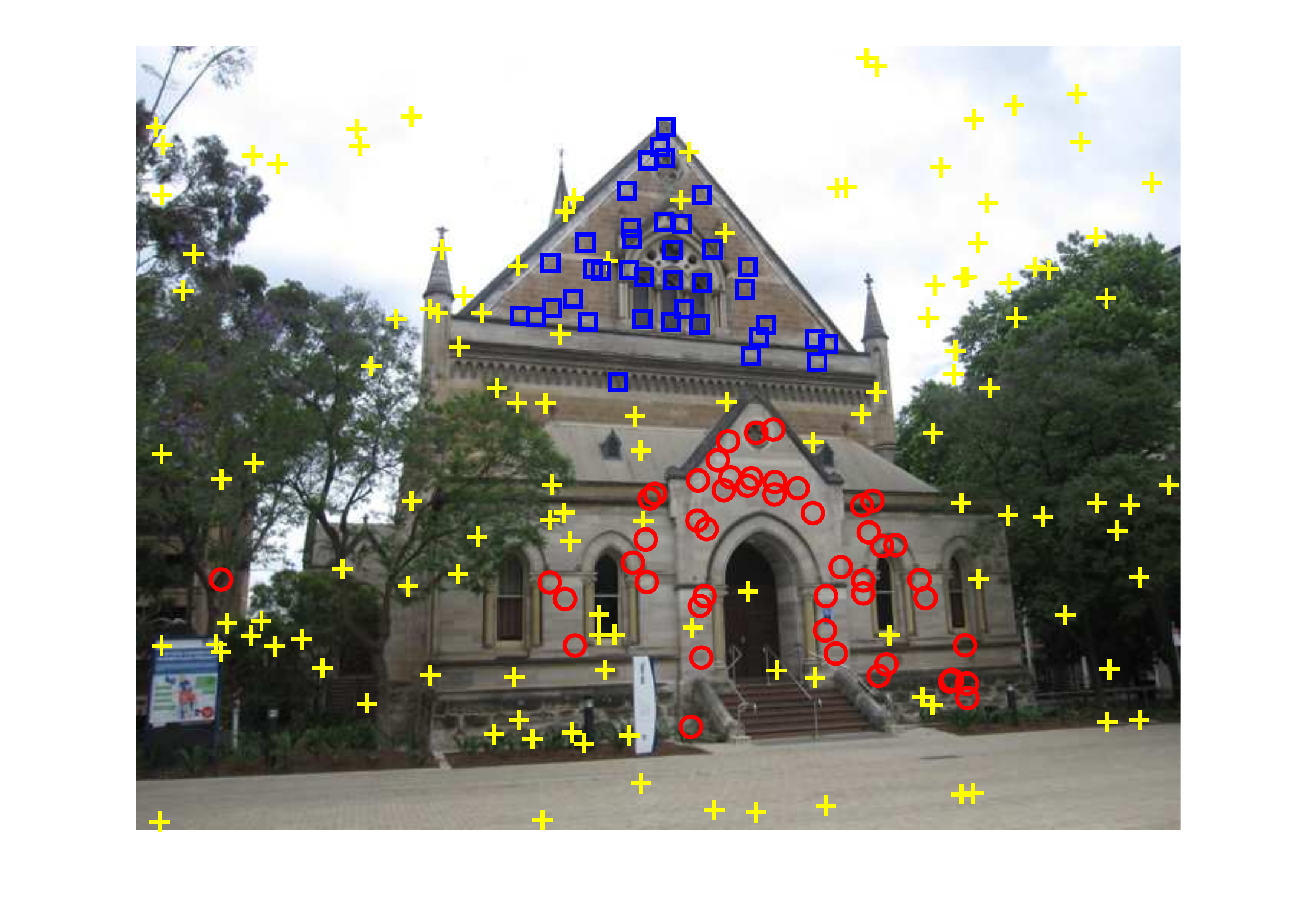}}
  \centerline{\includegraphics[width=1.17\textwidth]{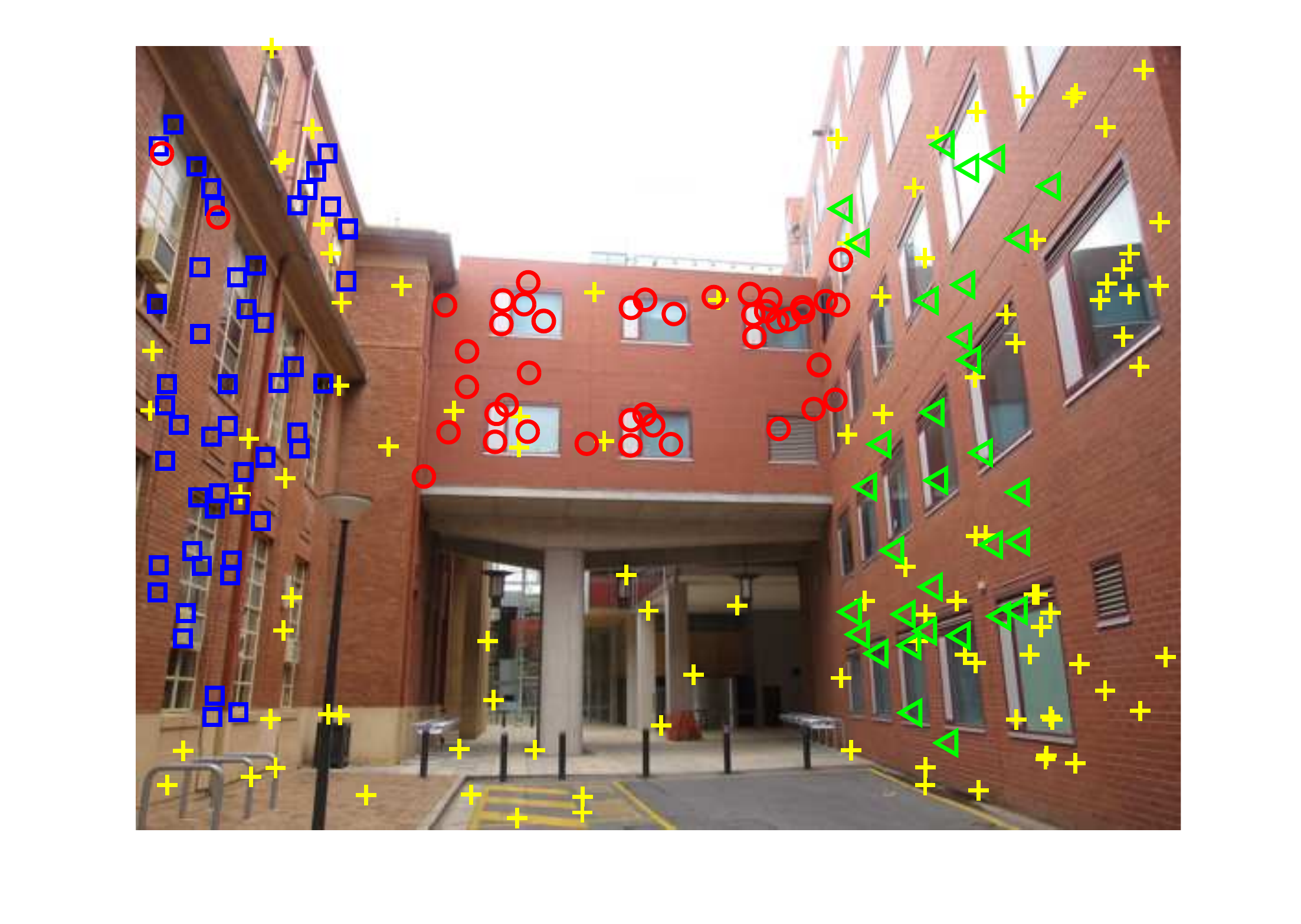}}
 \centerline{\includegraphics[width=1.17\textwidth]{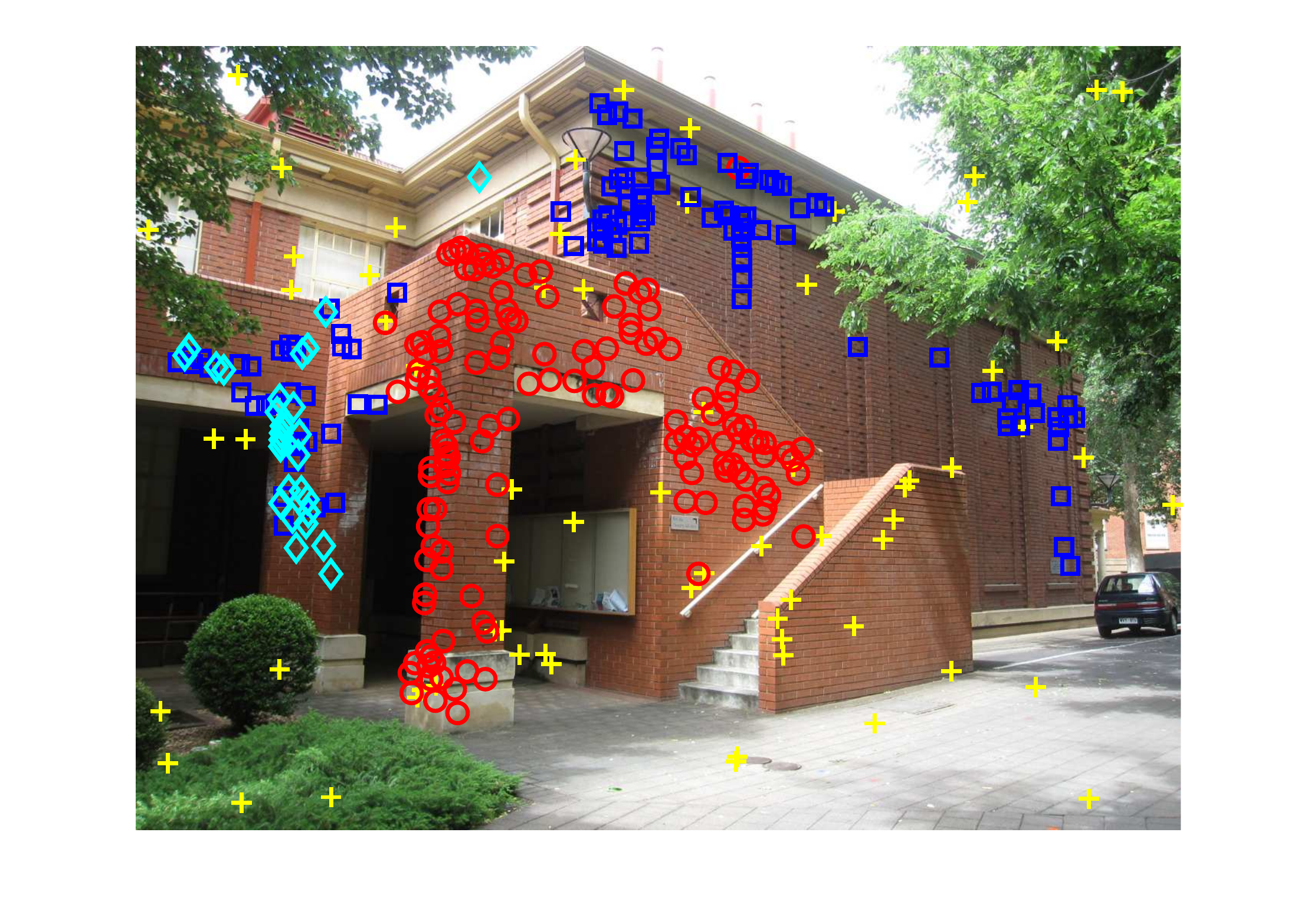}}
  \begin{center} (c) \end{center}
\end{minipage}
\begin{minipage}[t]{.1585\textwidth}
  \centering
 \centerline{\includegraphics[width=1.17\textwidth]{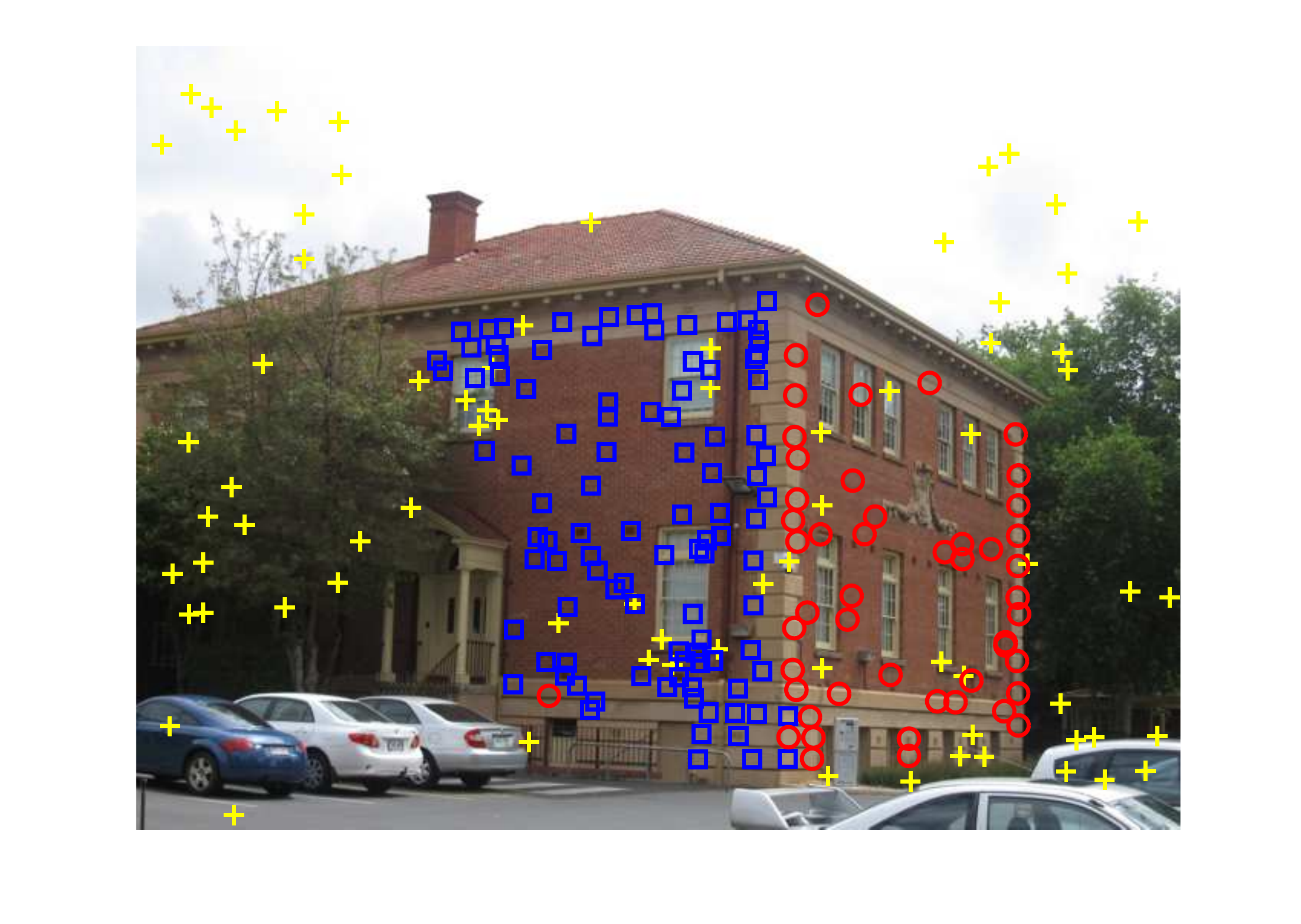}}
  \centerline{\includegraphics[width=1.17\textwidth]{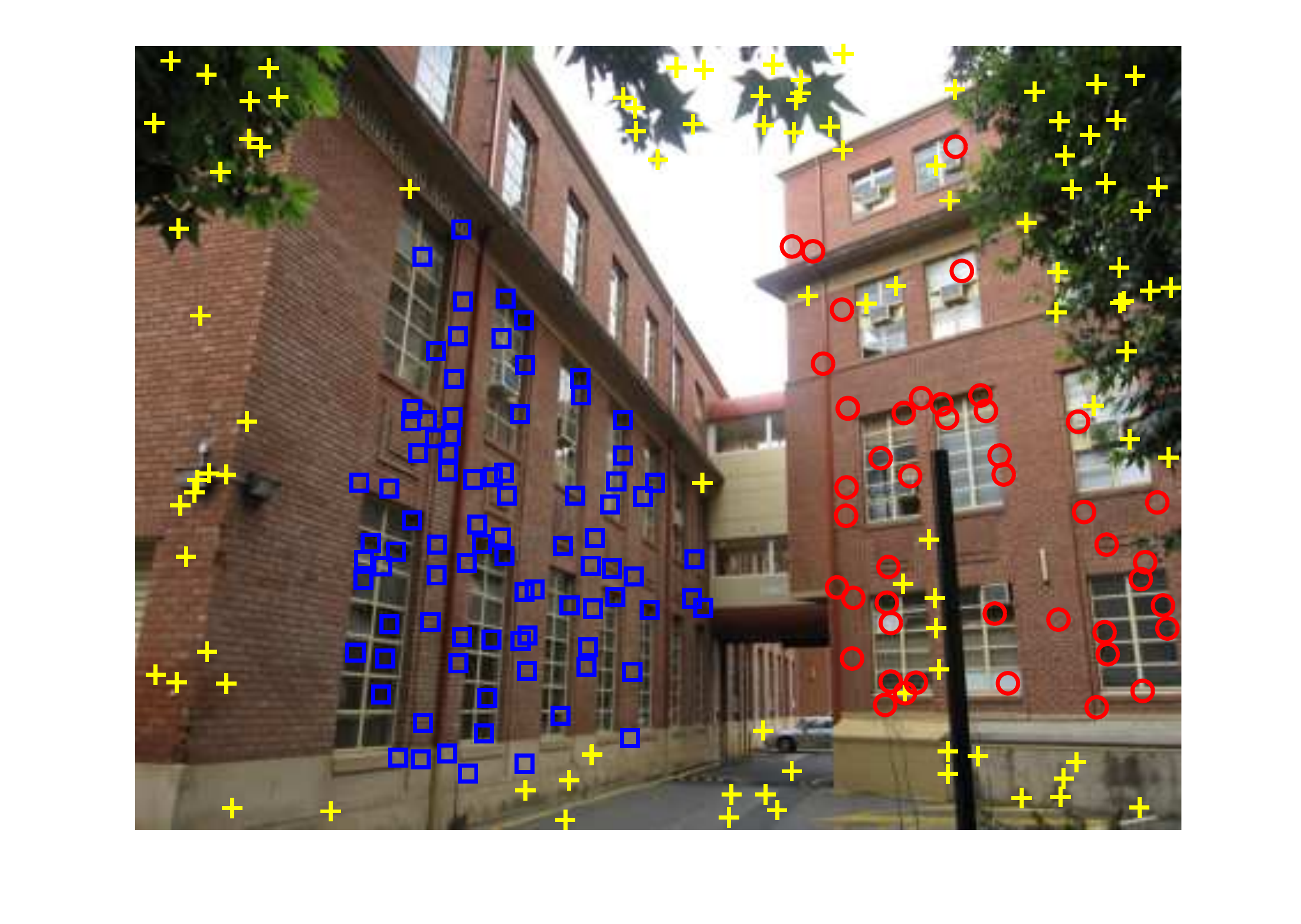}}
  \centerline{\includegraphics[width=1.17\textwidth]{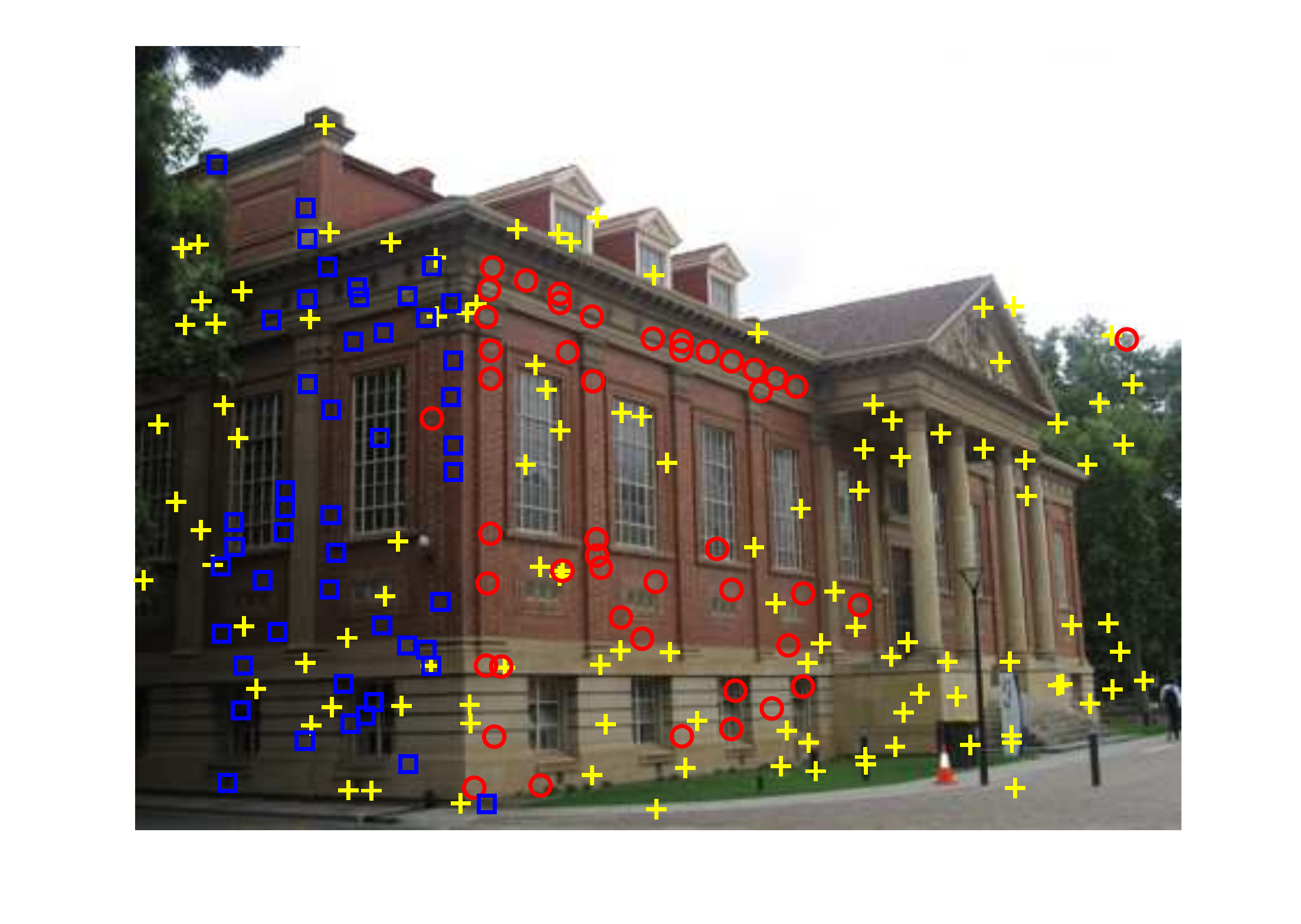}}
\centerline{\includegraphics[width=1.17\textwidth]{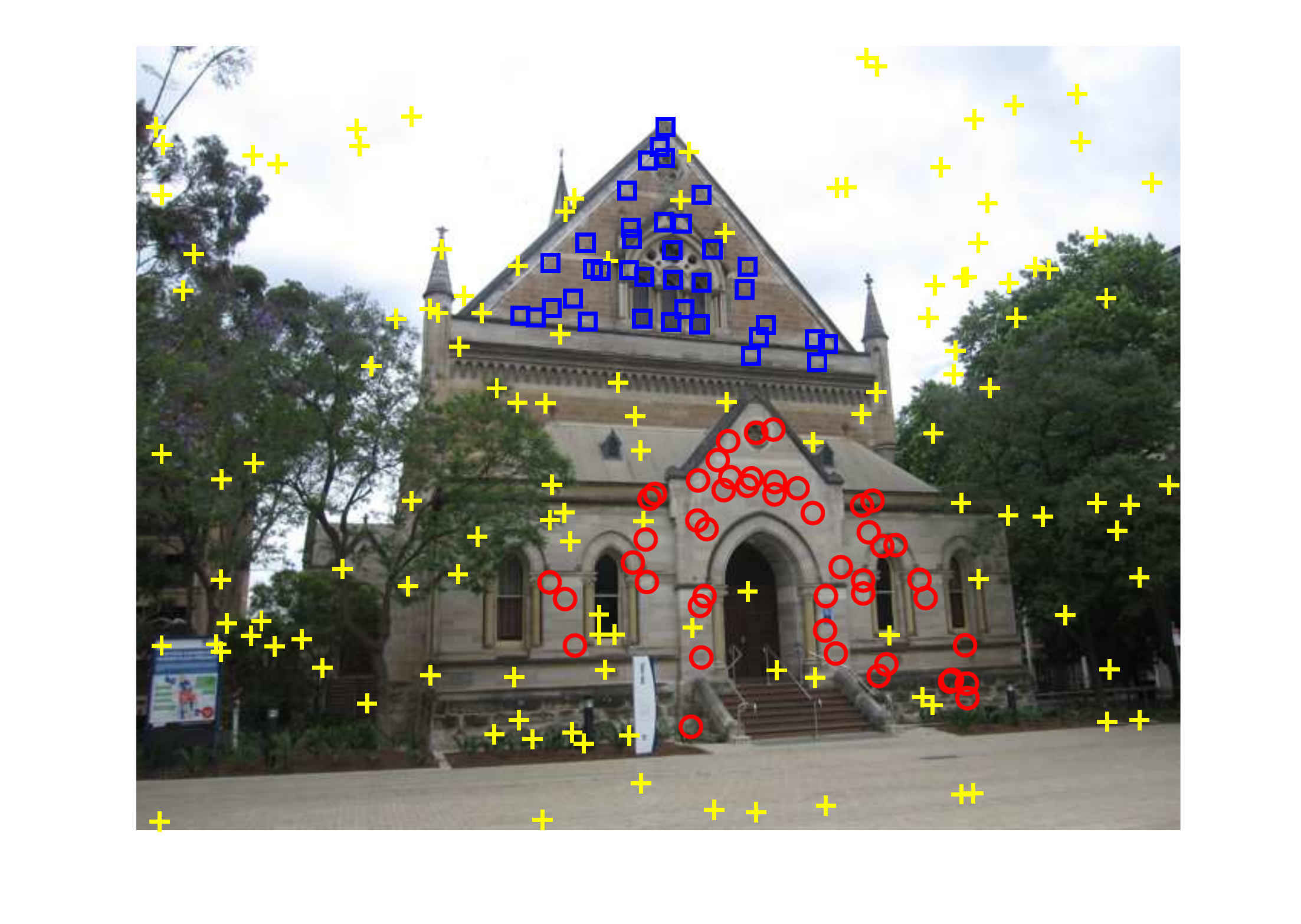}}
  \centerline{\includegraphics[width=1.17\textwidth]{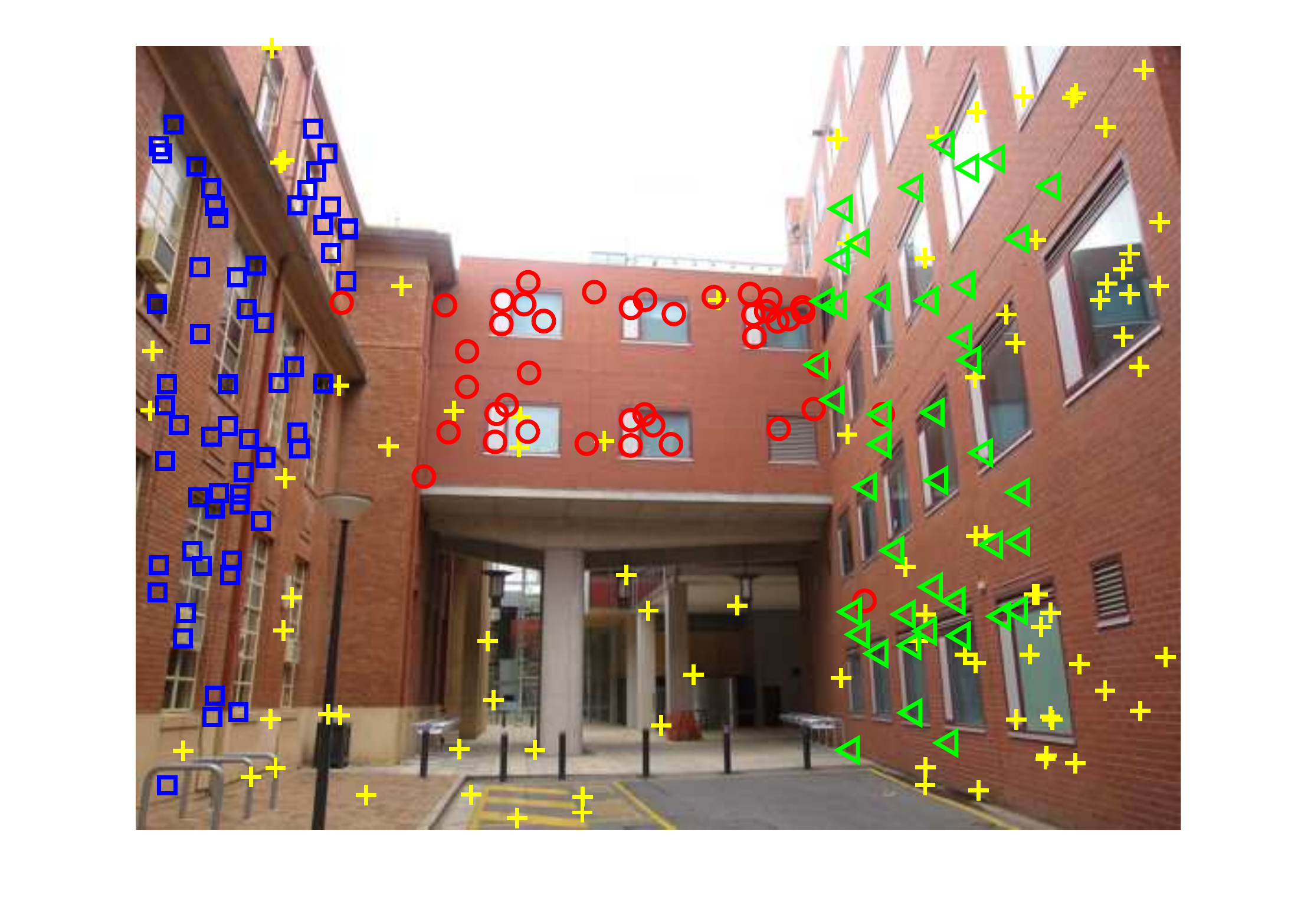}}
 \centerline{\includegraphics[width=1.17\textwidth]{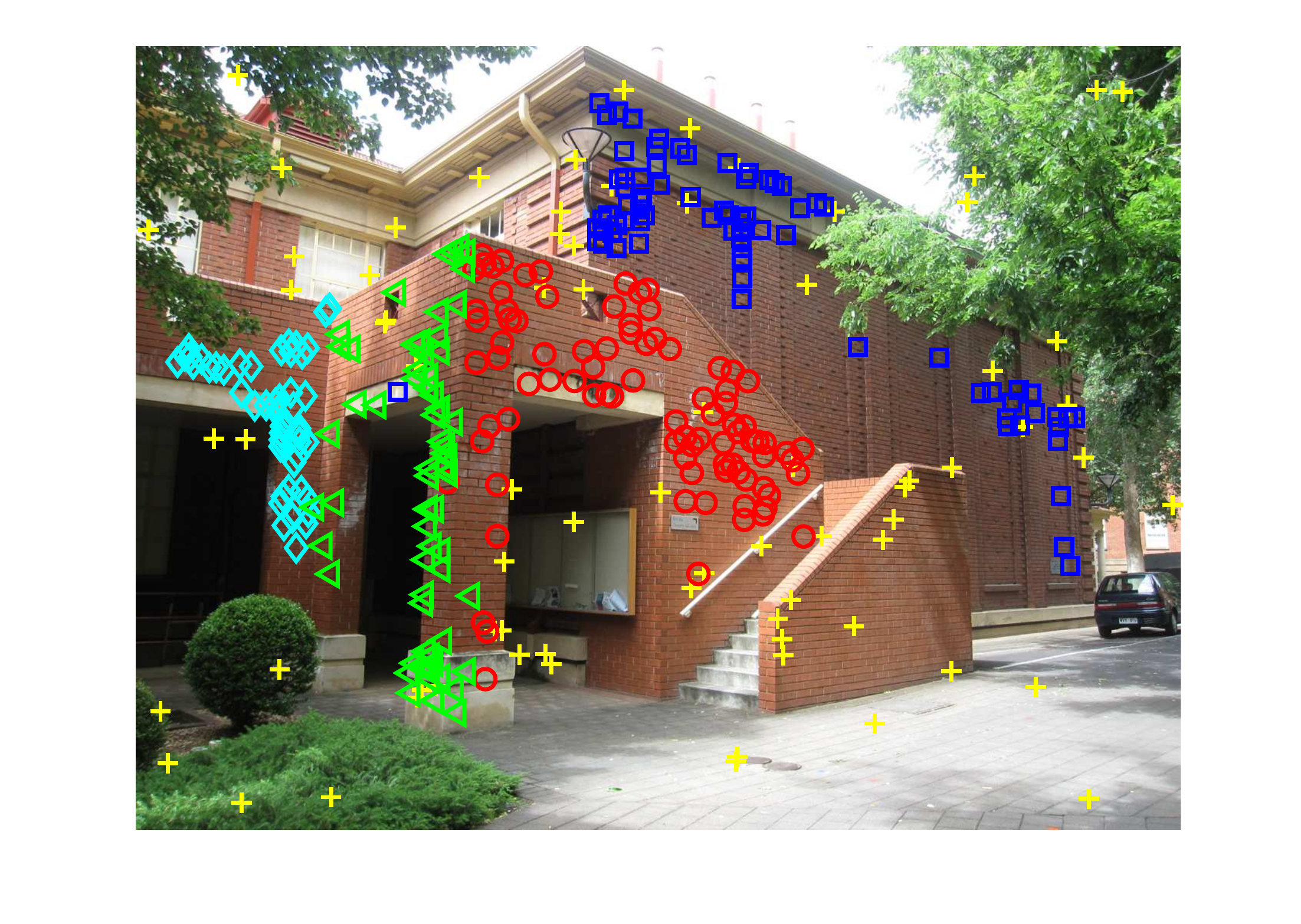}}
  \begin{center} (d) \end{center}
\end{minipage}
\begin{minipage}[t]{.1585\textwidth}
  \centering
 \centerline{\includegraphics[width=1.17\textwidth]{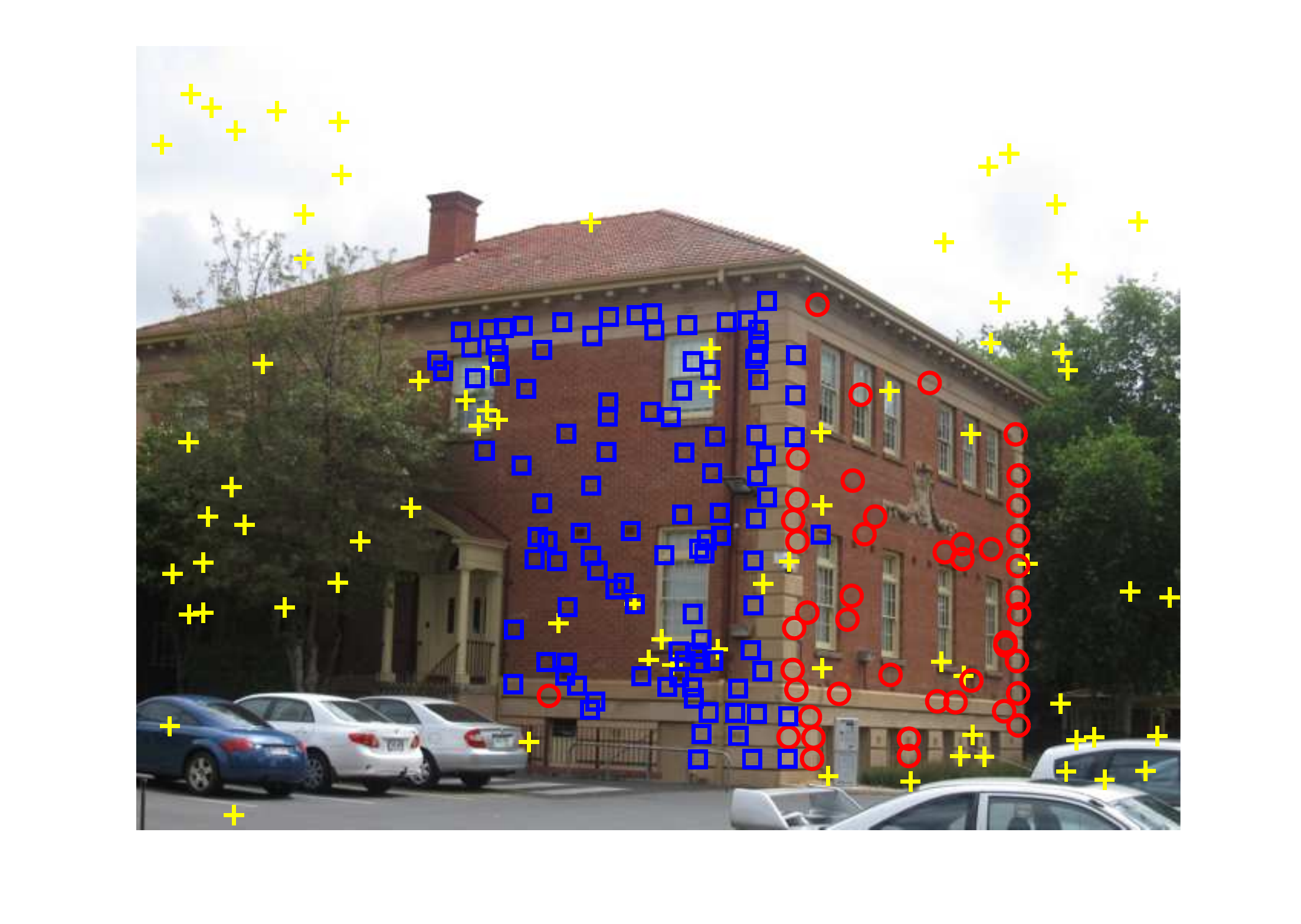}}
  \centerline{\includegraphics[width=1.17\textwidth]{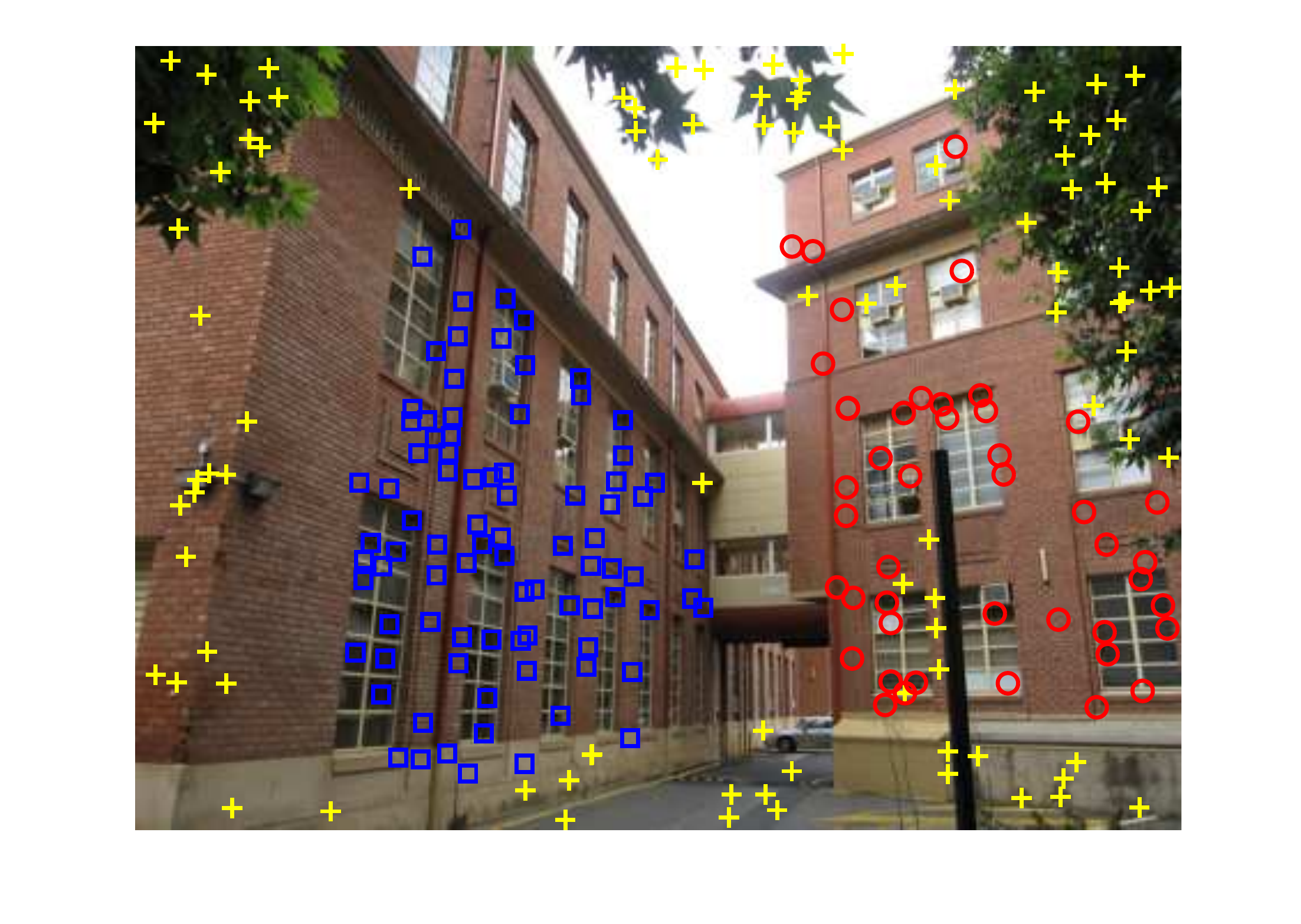}}
  \centerline{\includegraphics[width=1.17\textwidth]{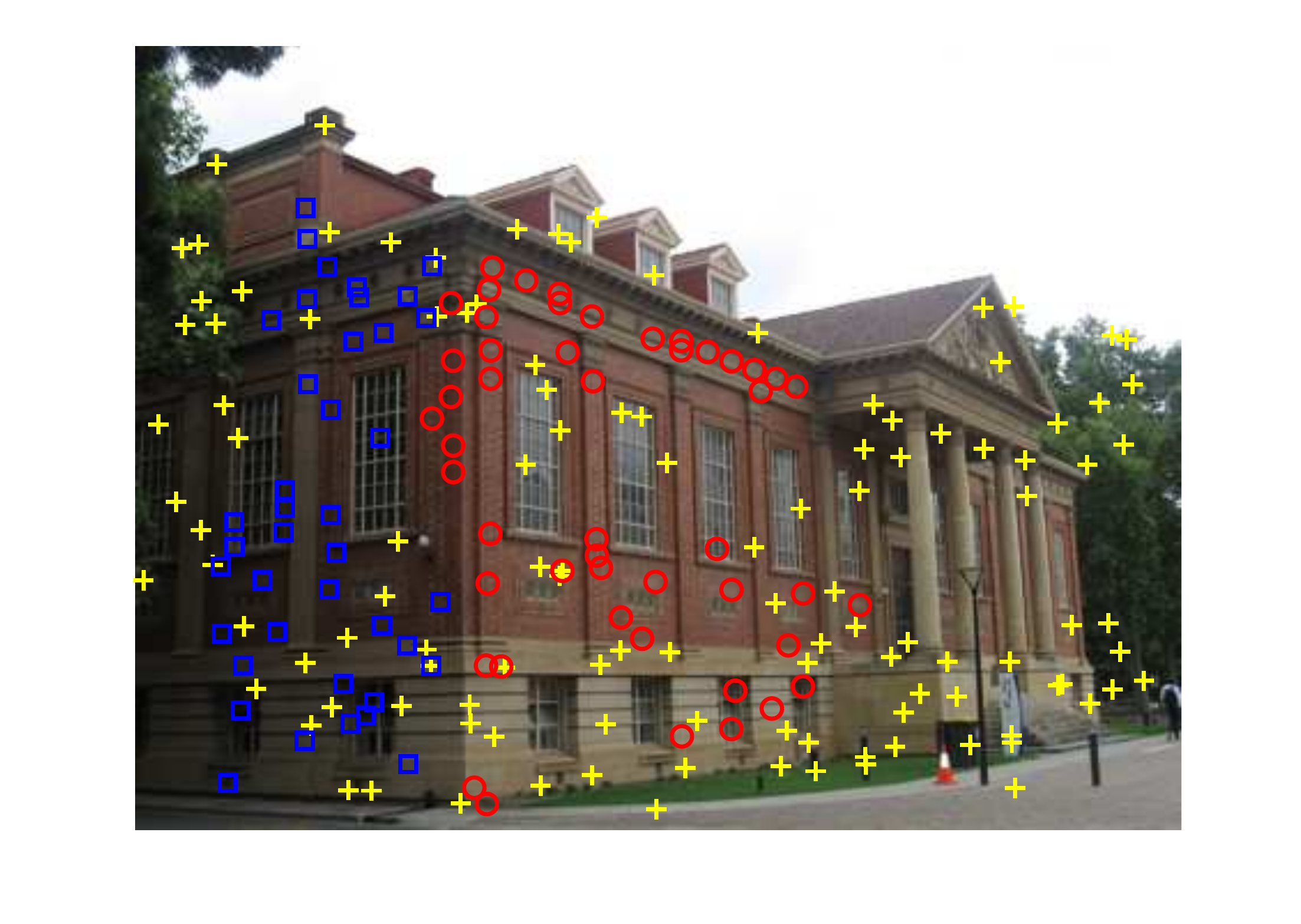}}
\centerline{\includegraphics[width=1.17\textwidth]{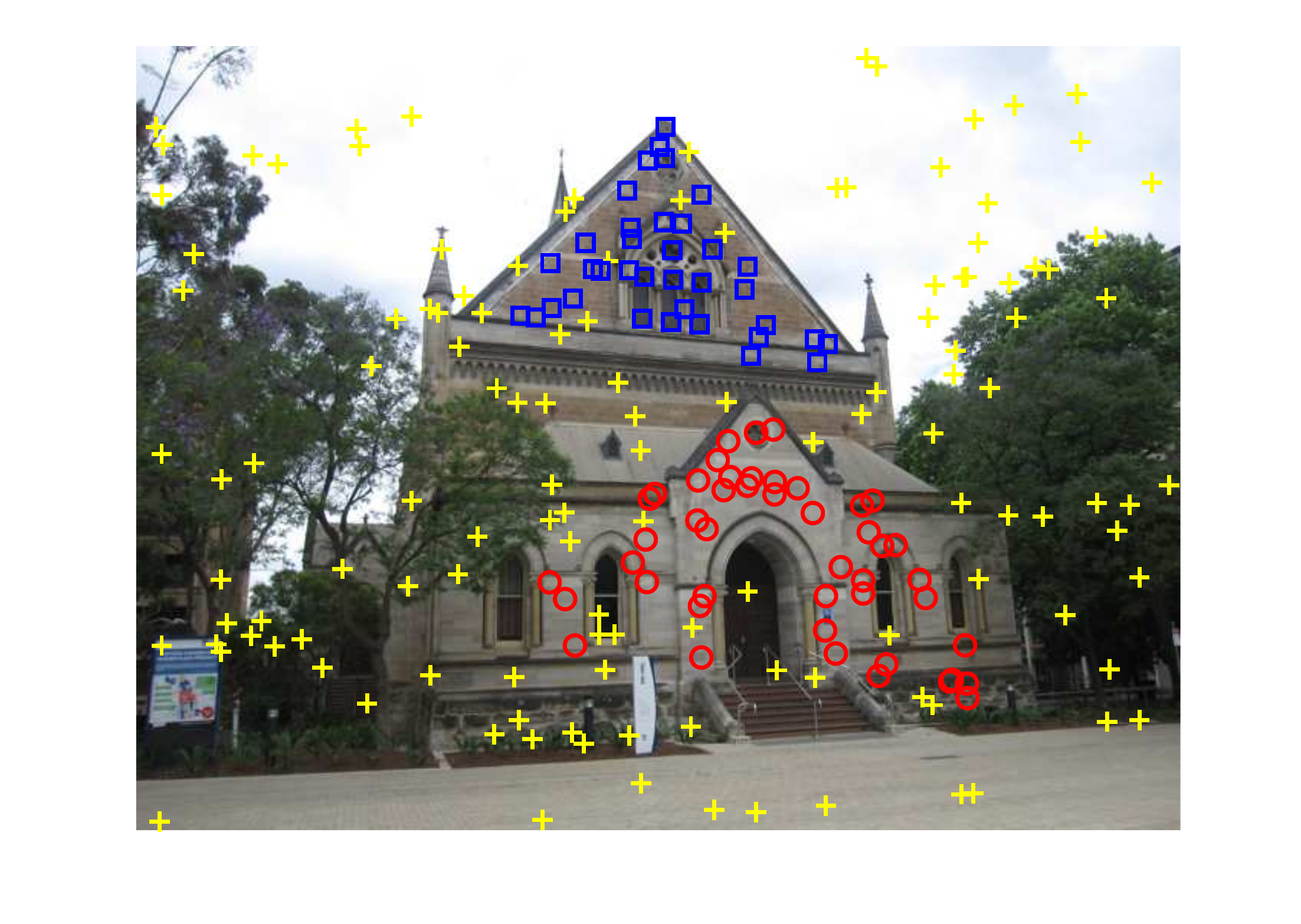}}
  \centerline{\includegraphics[width=1.17\textwidth]{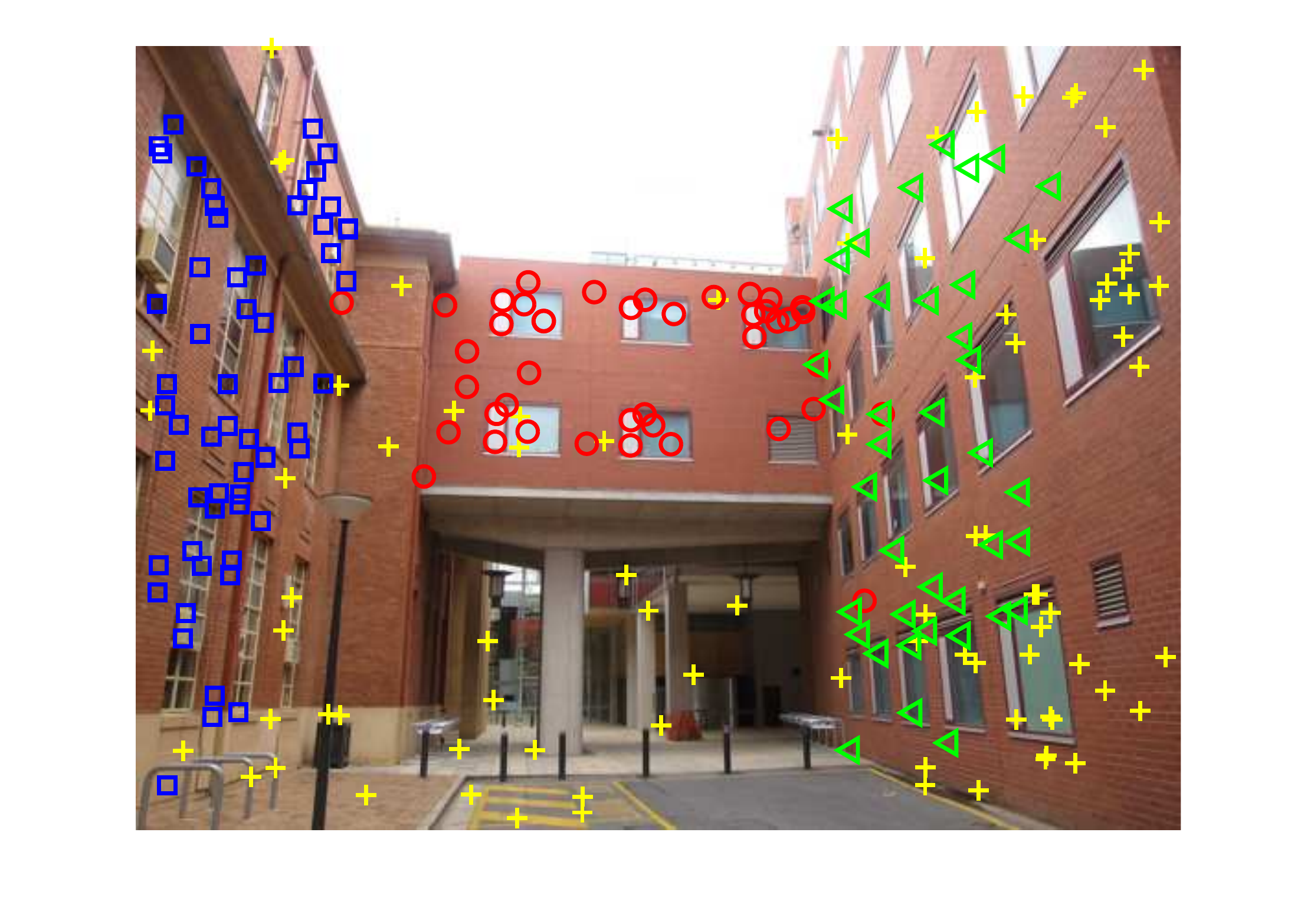}}
 \centerline{\includegraphics[width=1.17\textwidth]{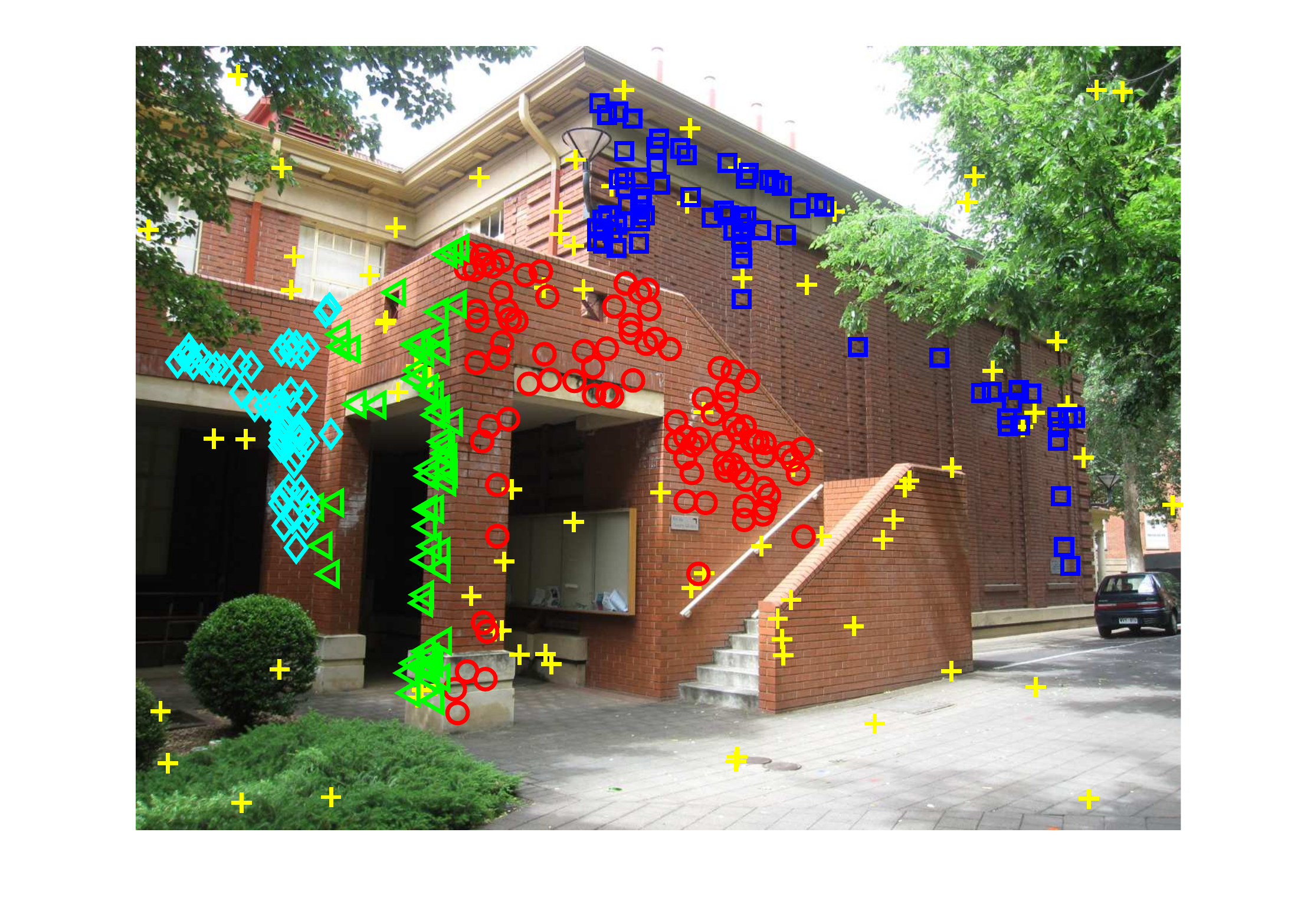}}
  \begin{center} (e)  \end{center}
\end{minipage}
\begin{minipage}[t]{.1585\textwidth}
  \centering
 \centerline{\includegraphics[width=1.17\textwidth]{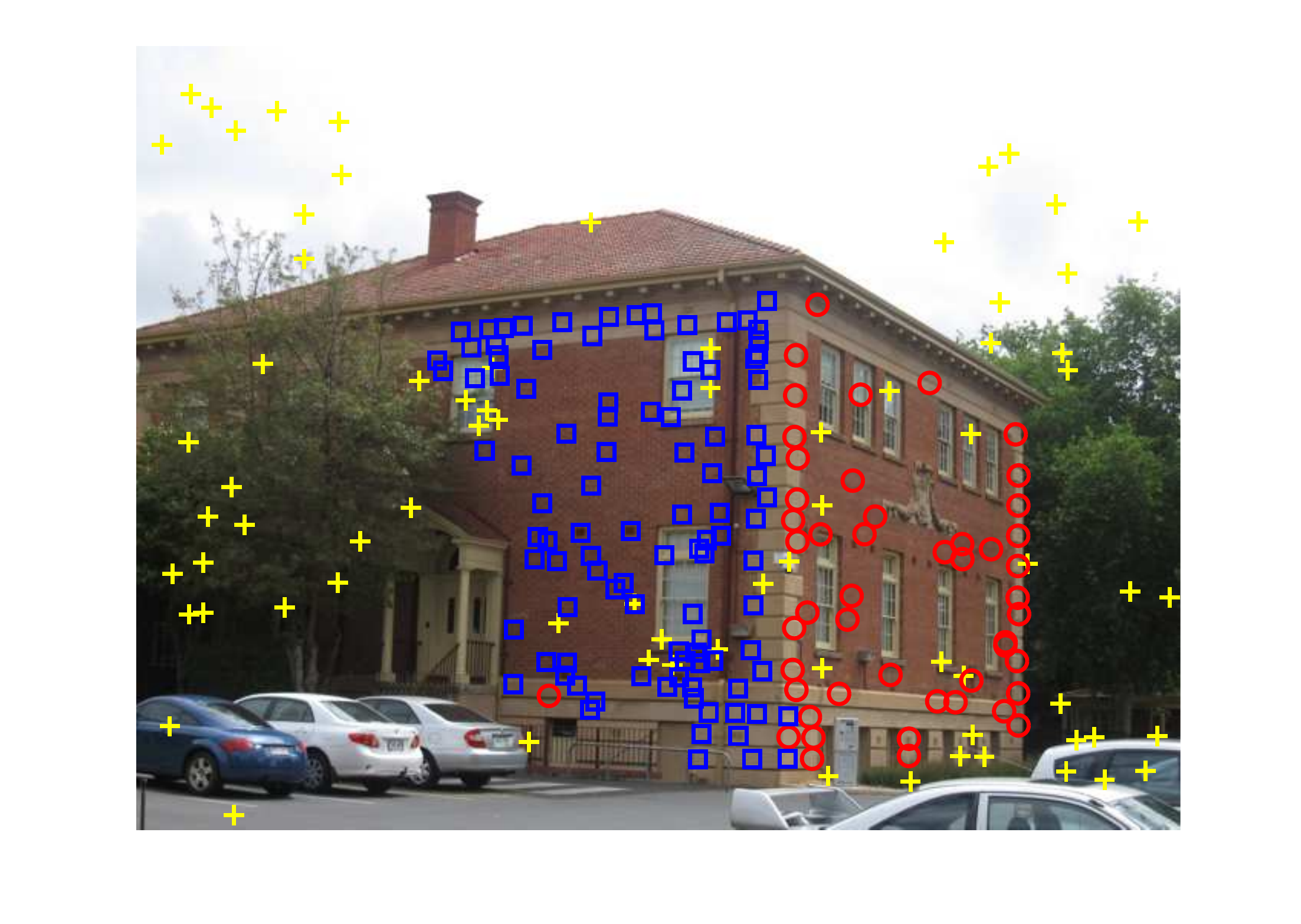}}
  \centerline{\includegraphics[width=1.17\textwidth]{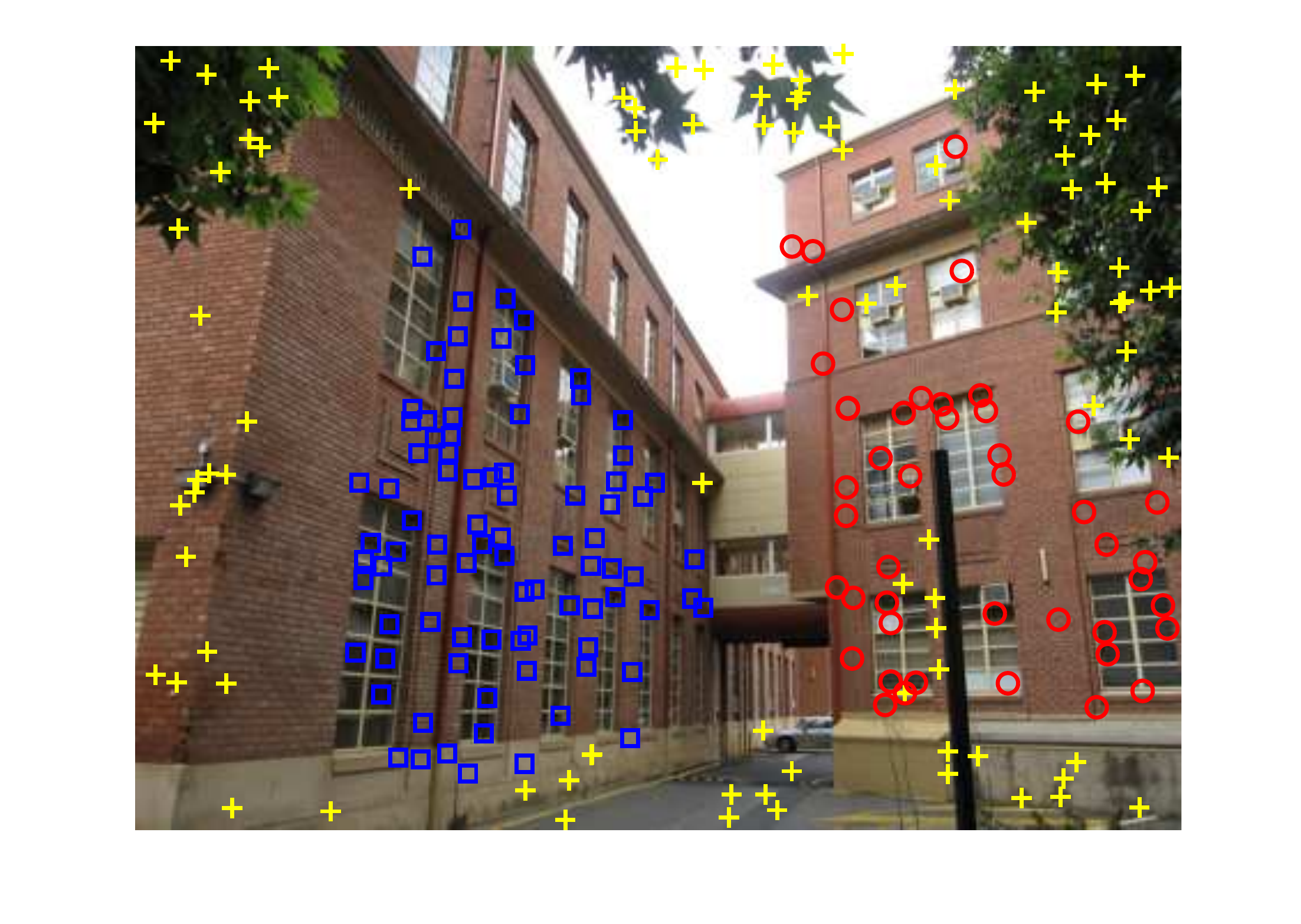}}
  \centerline{\includegraphics[width=1.17\textwidth]{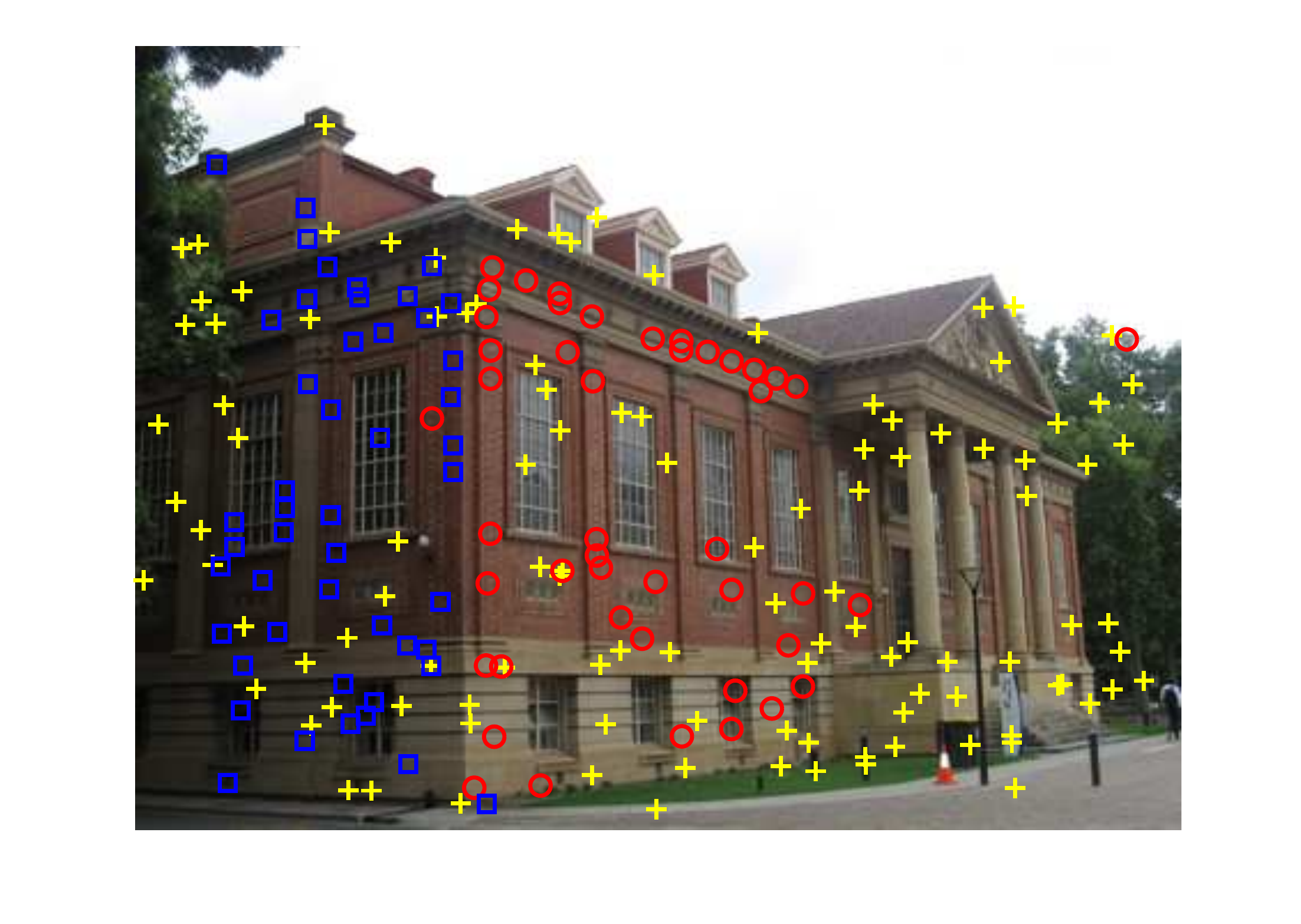}}
\centerline{\includegraphics[width=1.17\textwidth]{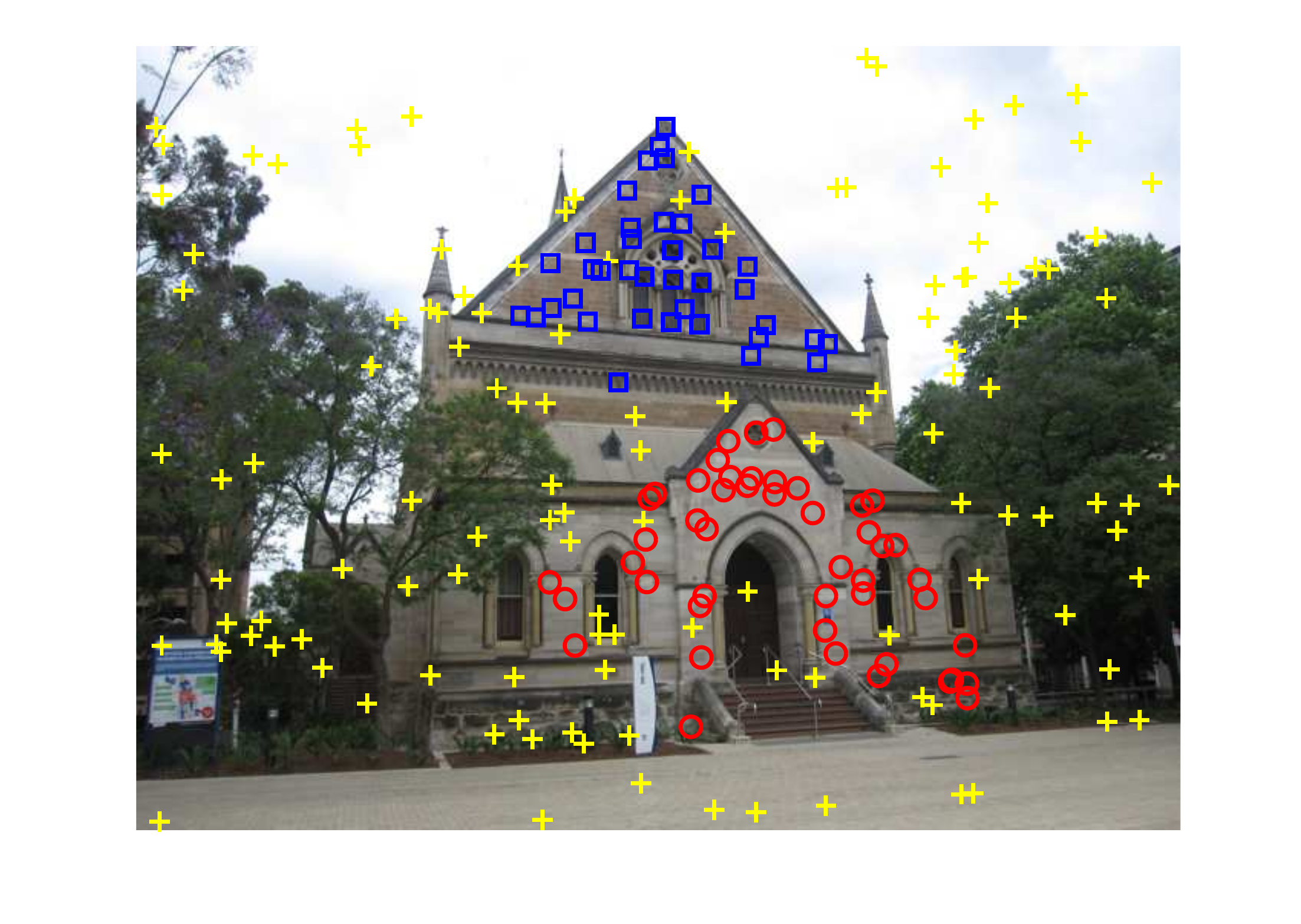}}
  \centerline{\includegraphics[width=1.17\textwidth]{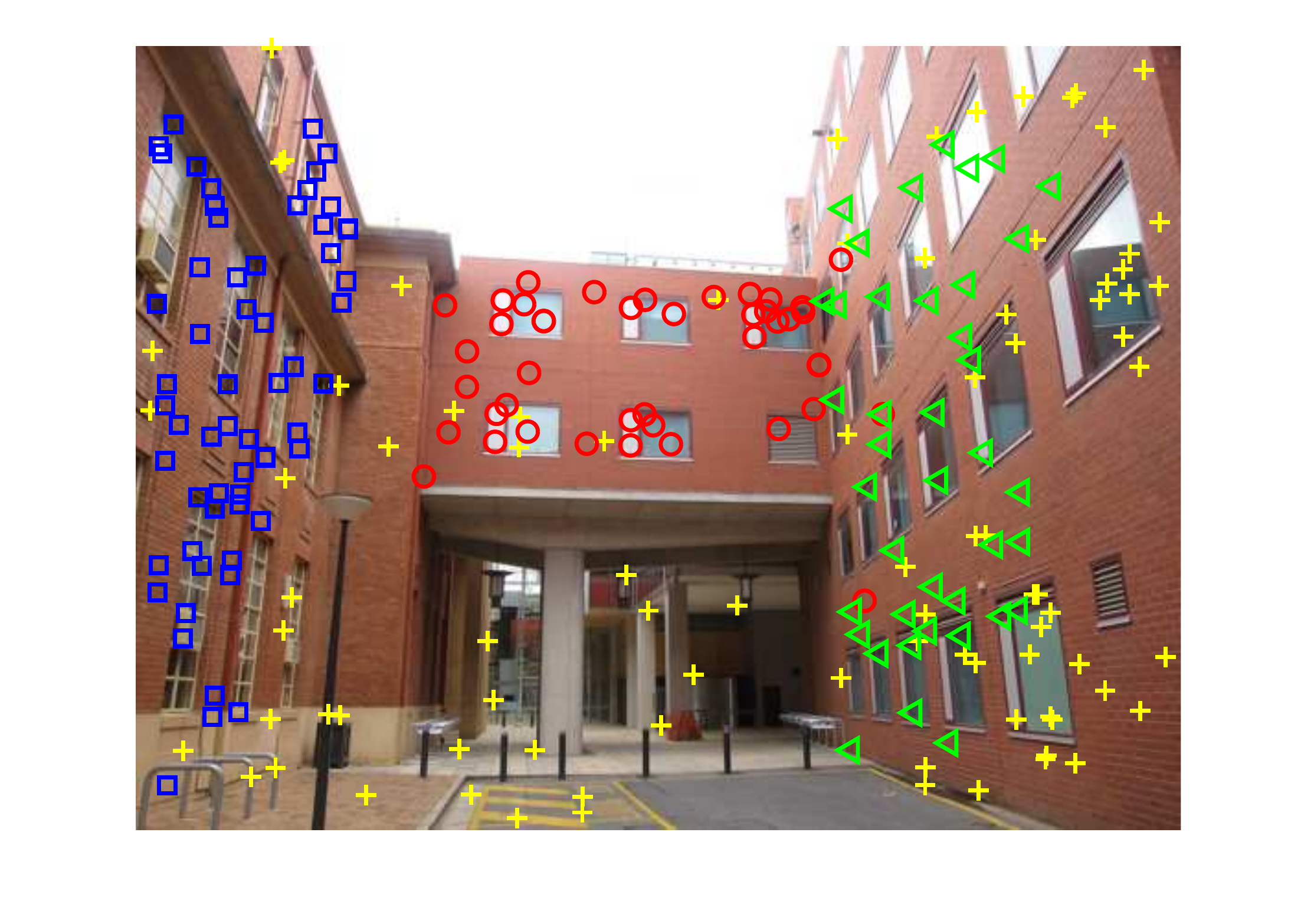}}
 \centerline{\includegraphics[width=1.17\textwidth]{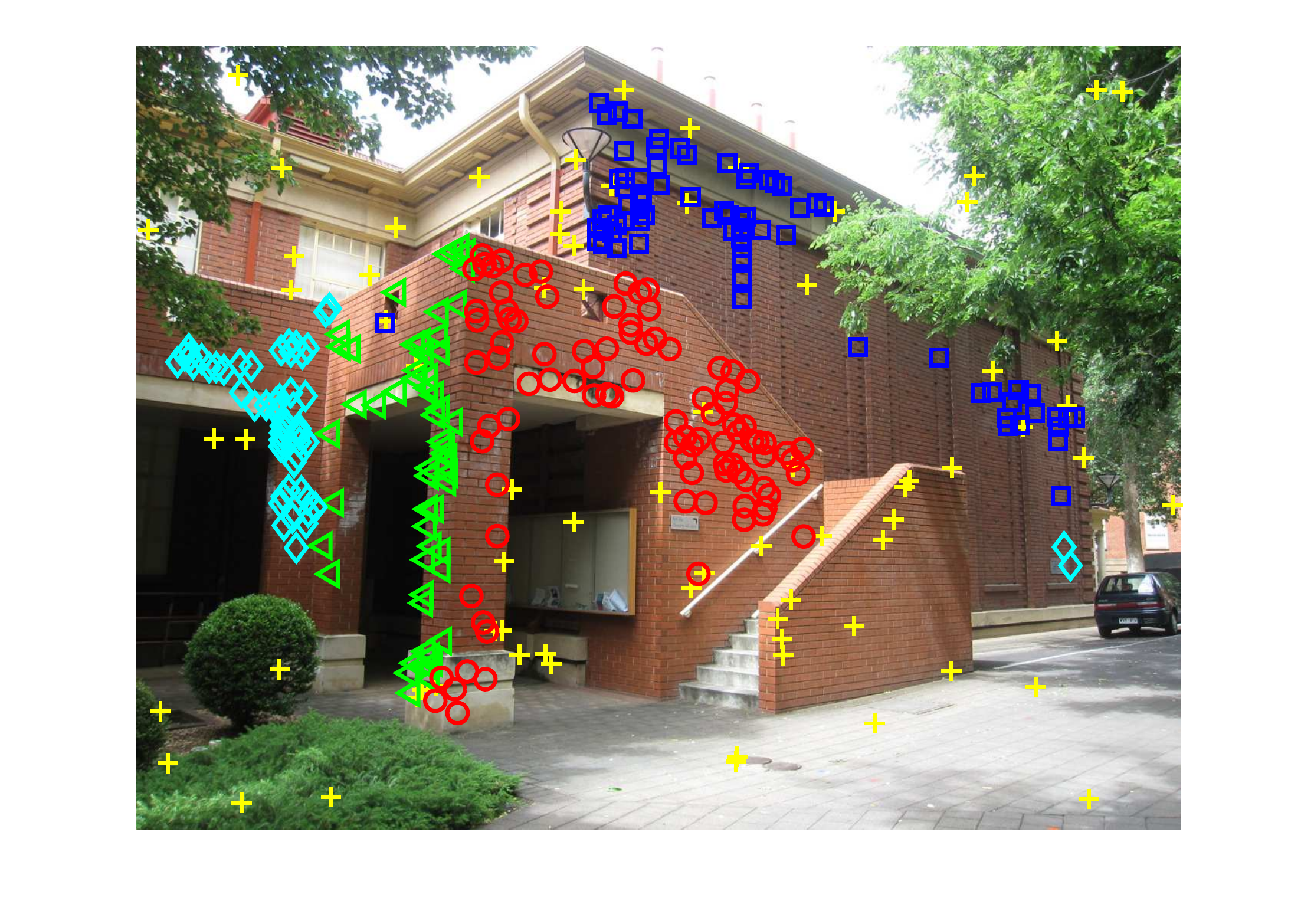}}
   \begin{center} (f)  \end{center}
\end{minipage}
\hfill
\caption{Qualitative comparisons on homography based segmentation using a subset of the data, namely Ladysymon, Sene, Library, Elderhalla, Neem and Johnsona in the top-down order (only one of the two views is shown for each case). (a) The ground truth segmentation results. (b) to (f) The results obtained by KF, RCG, AKSWH, T-linkage and HF, respectively.}
\label{fig:homography}
\end{figure}
\subsection{Real Images}
\begin{figure}[t]
\centering
\begin{minipage}[t]{.1585\textwidth}
 \centering
 \centerline{\includegraphics[width=1.15\textwidth]{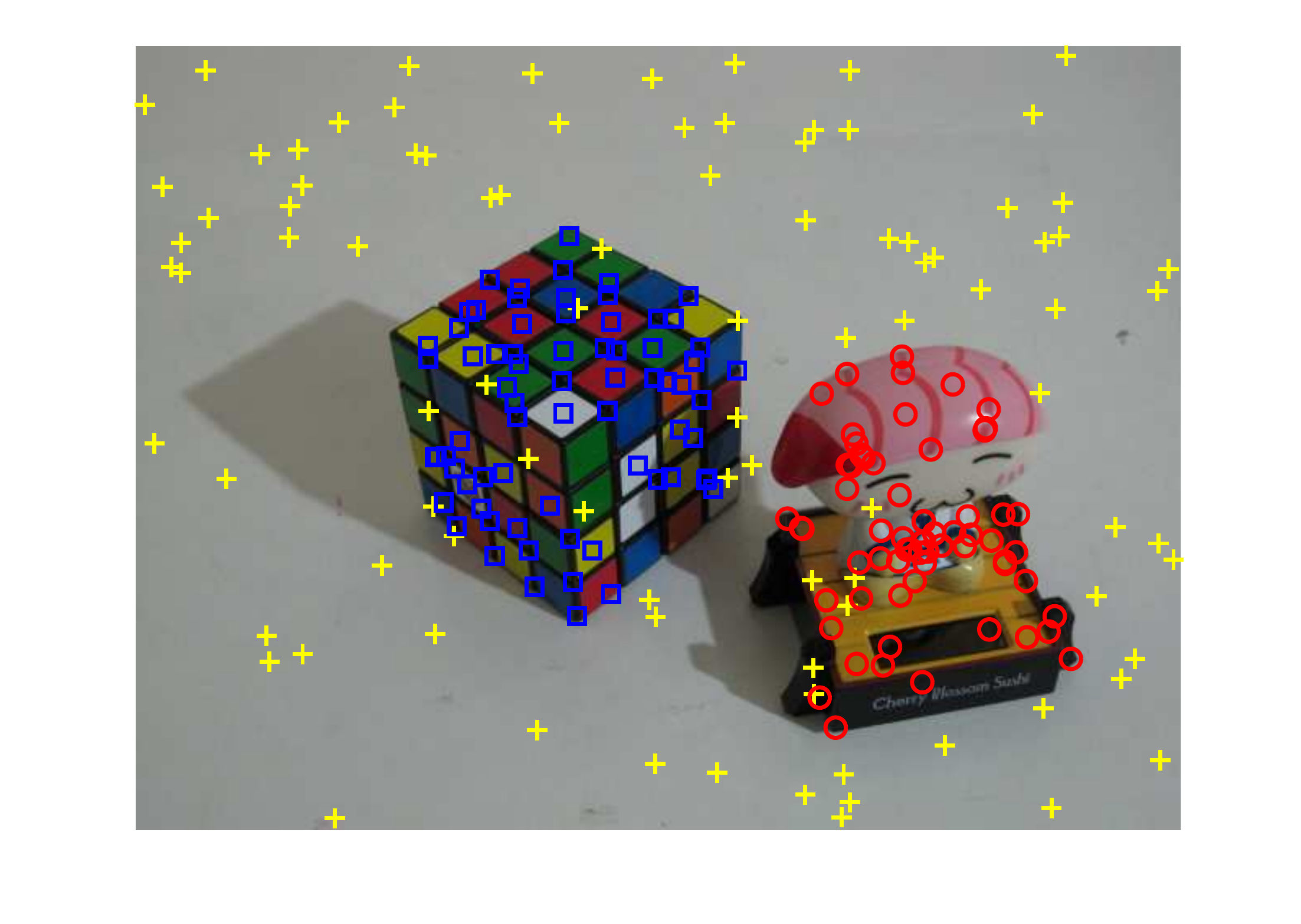}}
  \centerline{\includegraphics[width=1.17\textwidth]{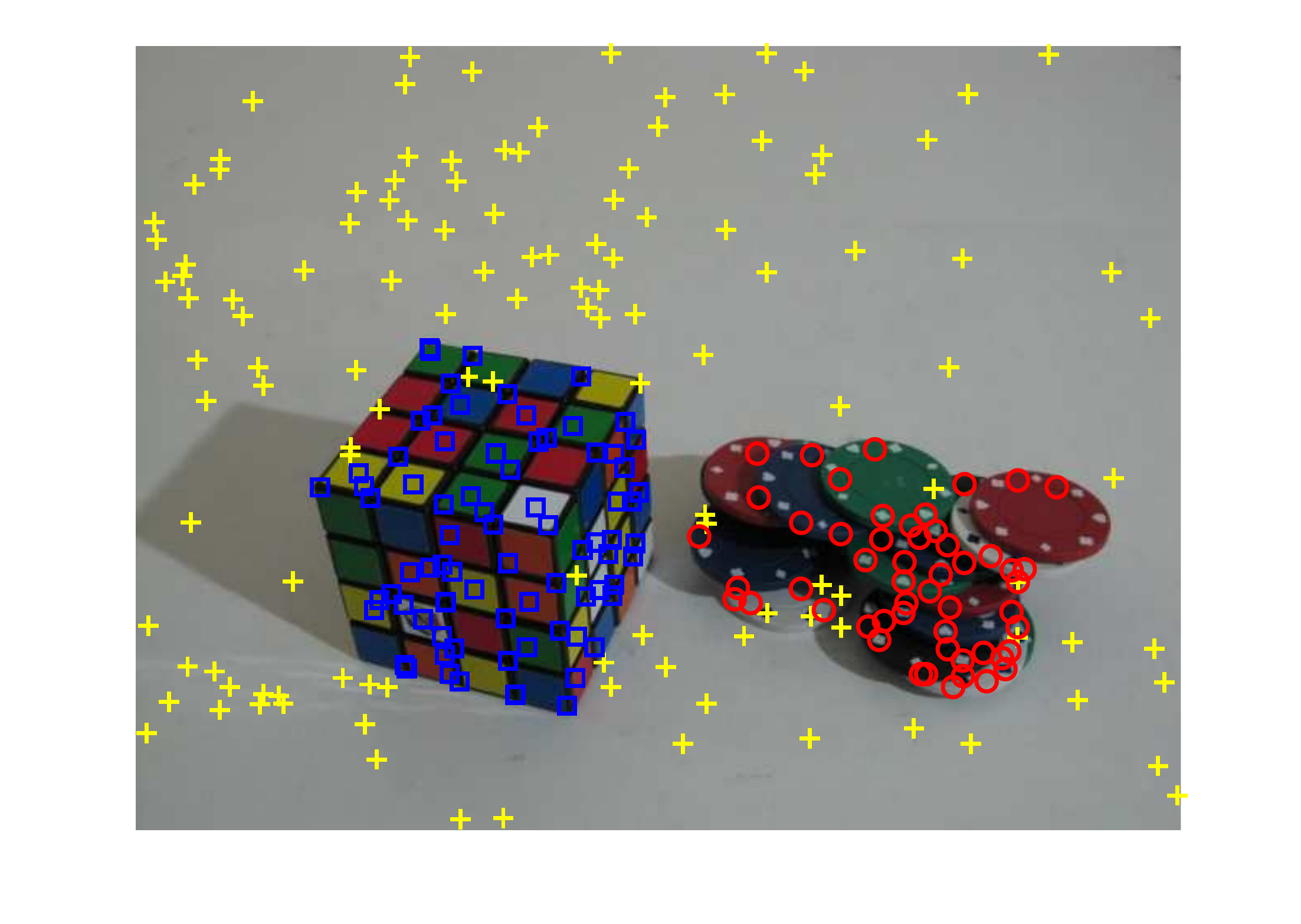}}
\centerline{\includegraphics[width=1.17\textwidth]{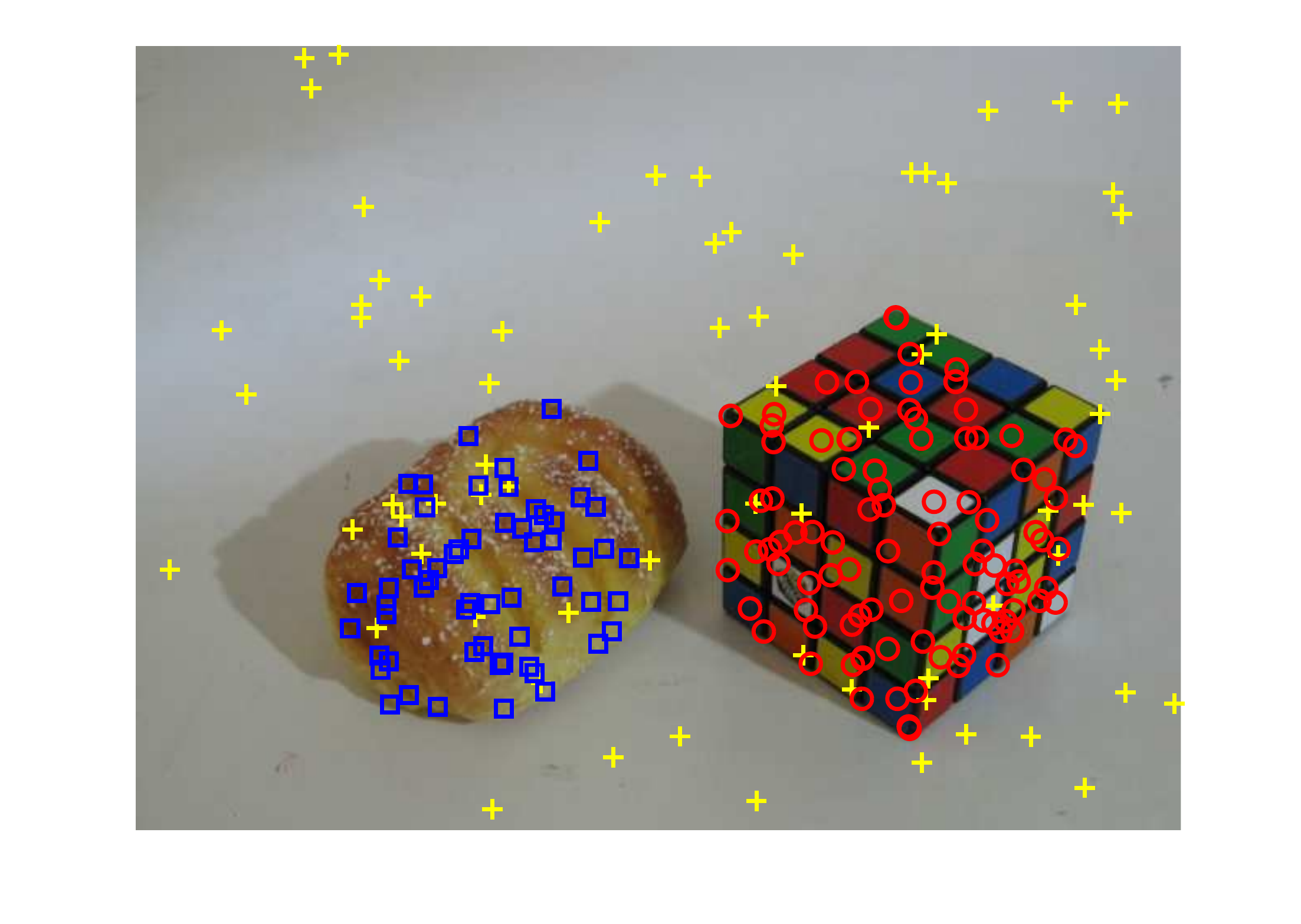}}
\centerline{\includegraphics[width=1.17\textwidth]{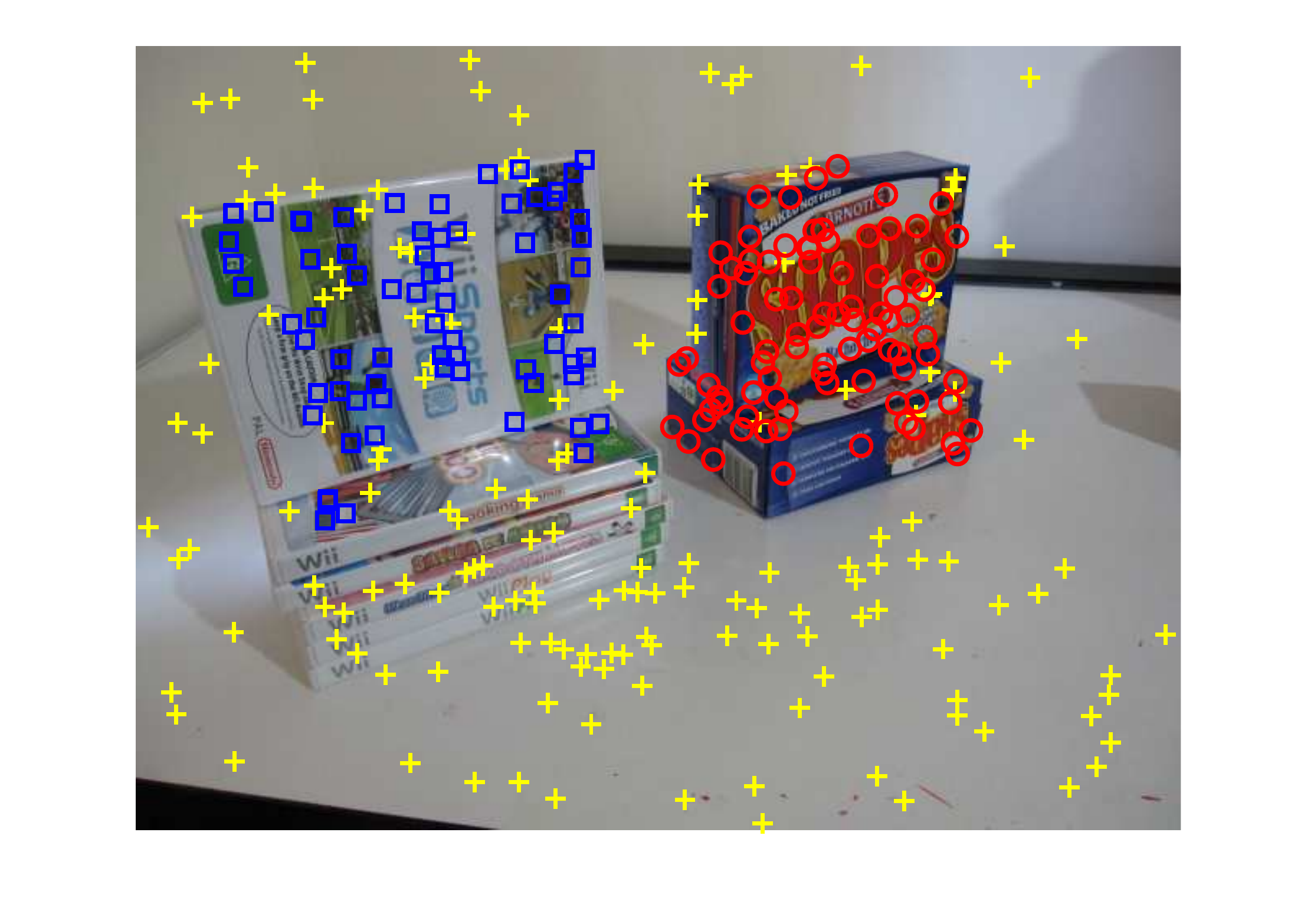}}
\centerline{\includegraphics[width=1.17\textwidth]{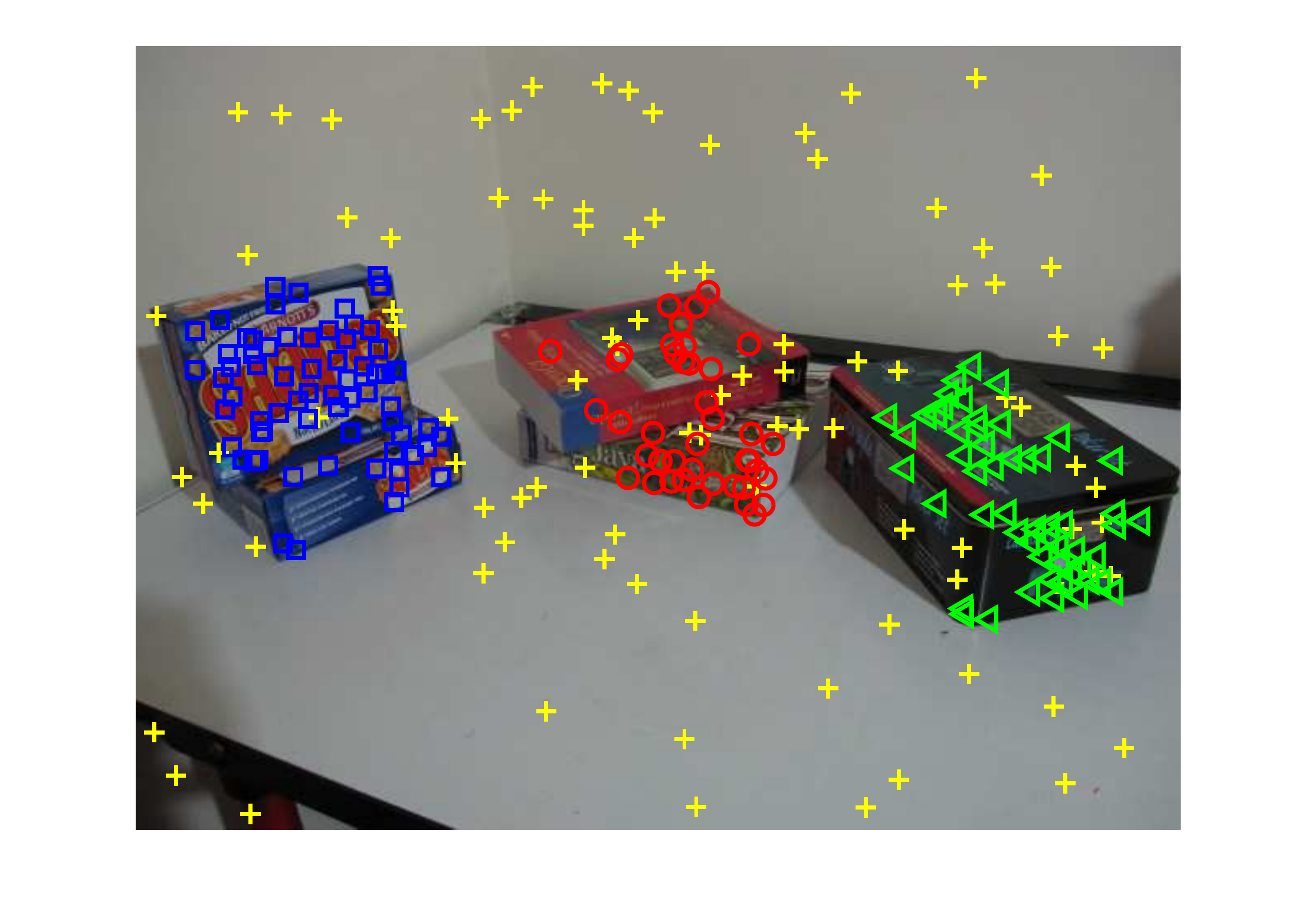}}
\centerline{\includegraphics[width=1.17\textwidth]{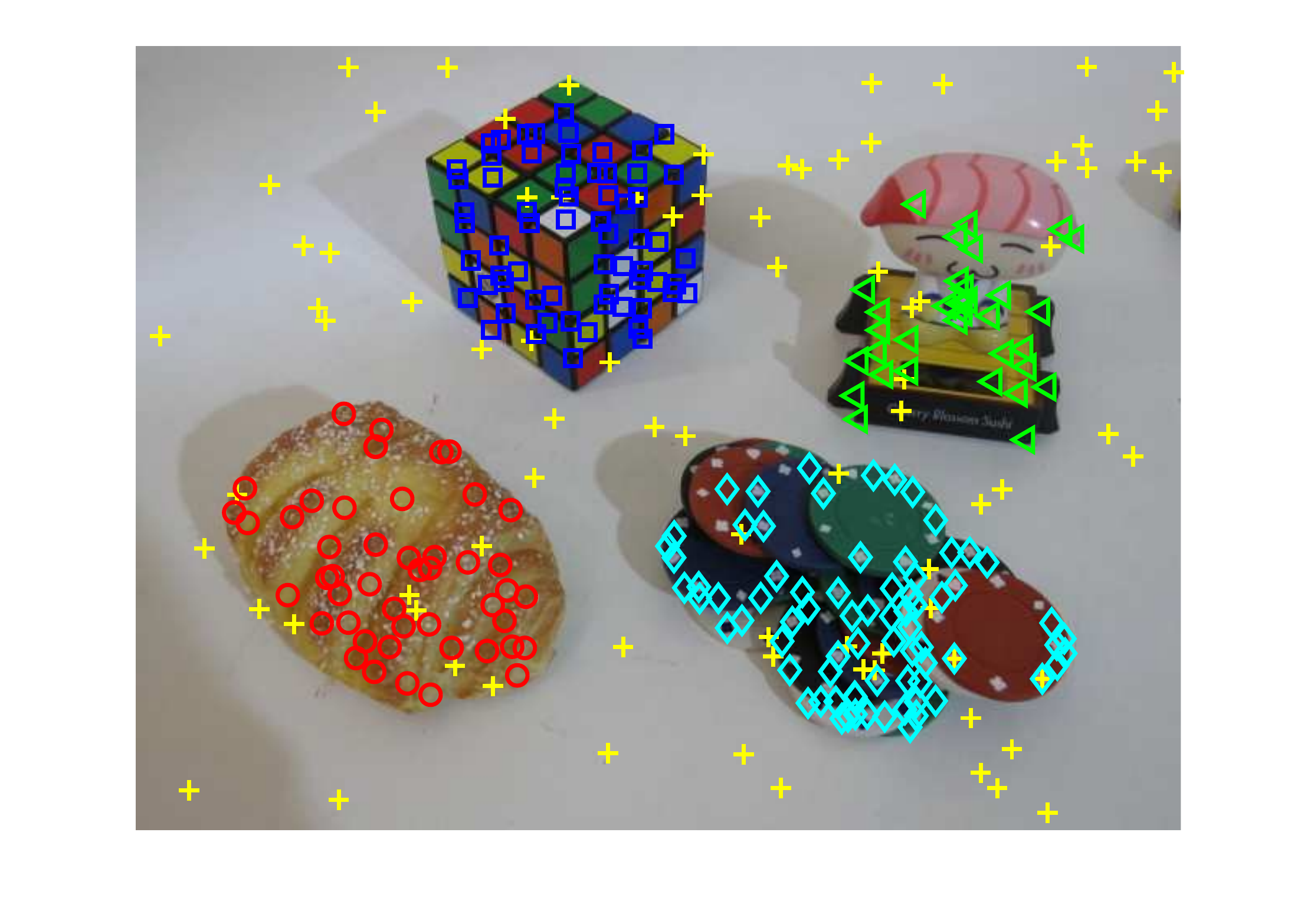}}
  \begin{center} (a) \end{center}
\end{minipage}
\begin{minipage}[t]{.1585\textwidth}
  \centering
 \centerline{\includegraphics[width=1.17\textwidth]{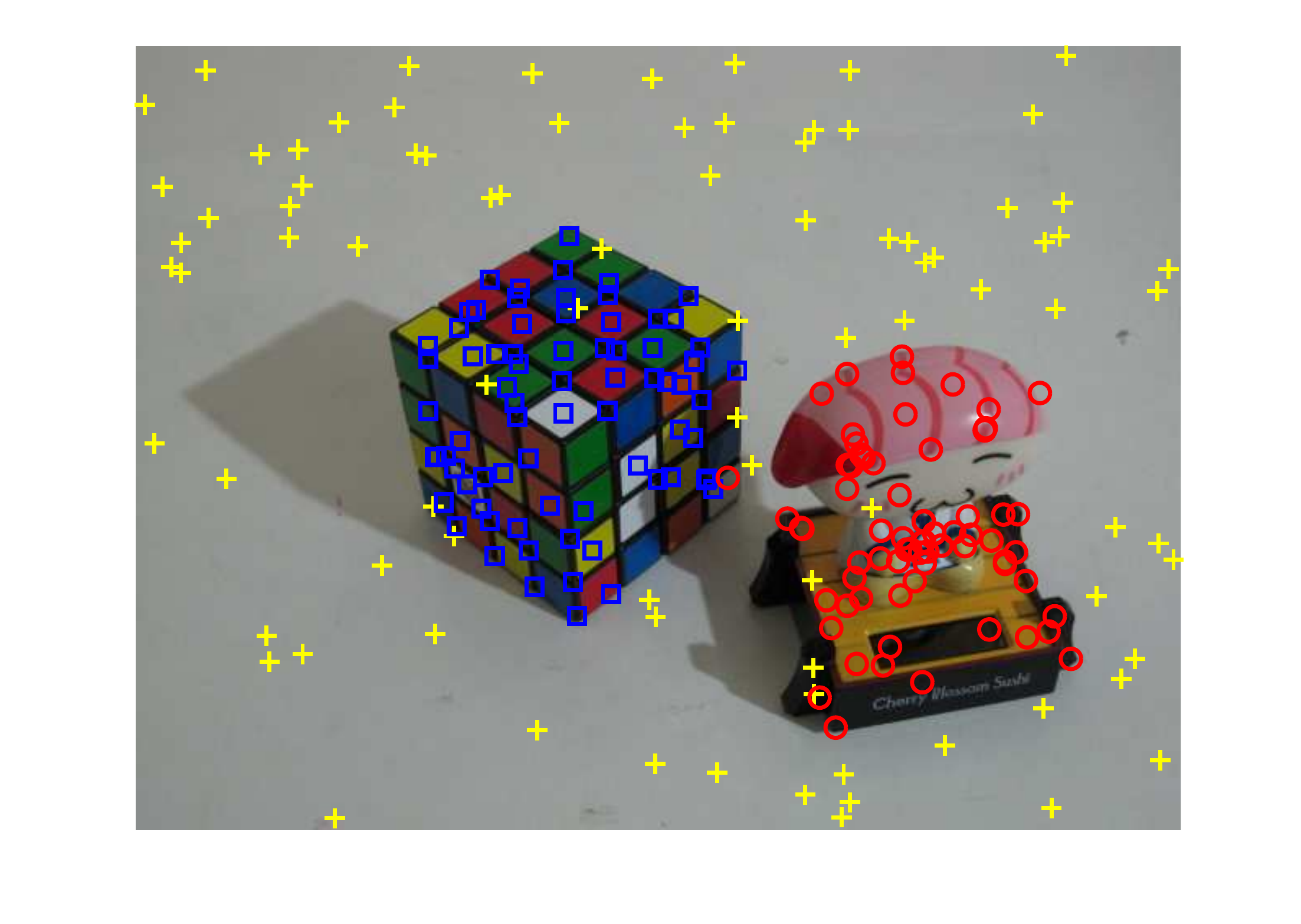}}
  \centerline{\includegraphics[width=1.17\textwidth]{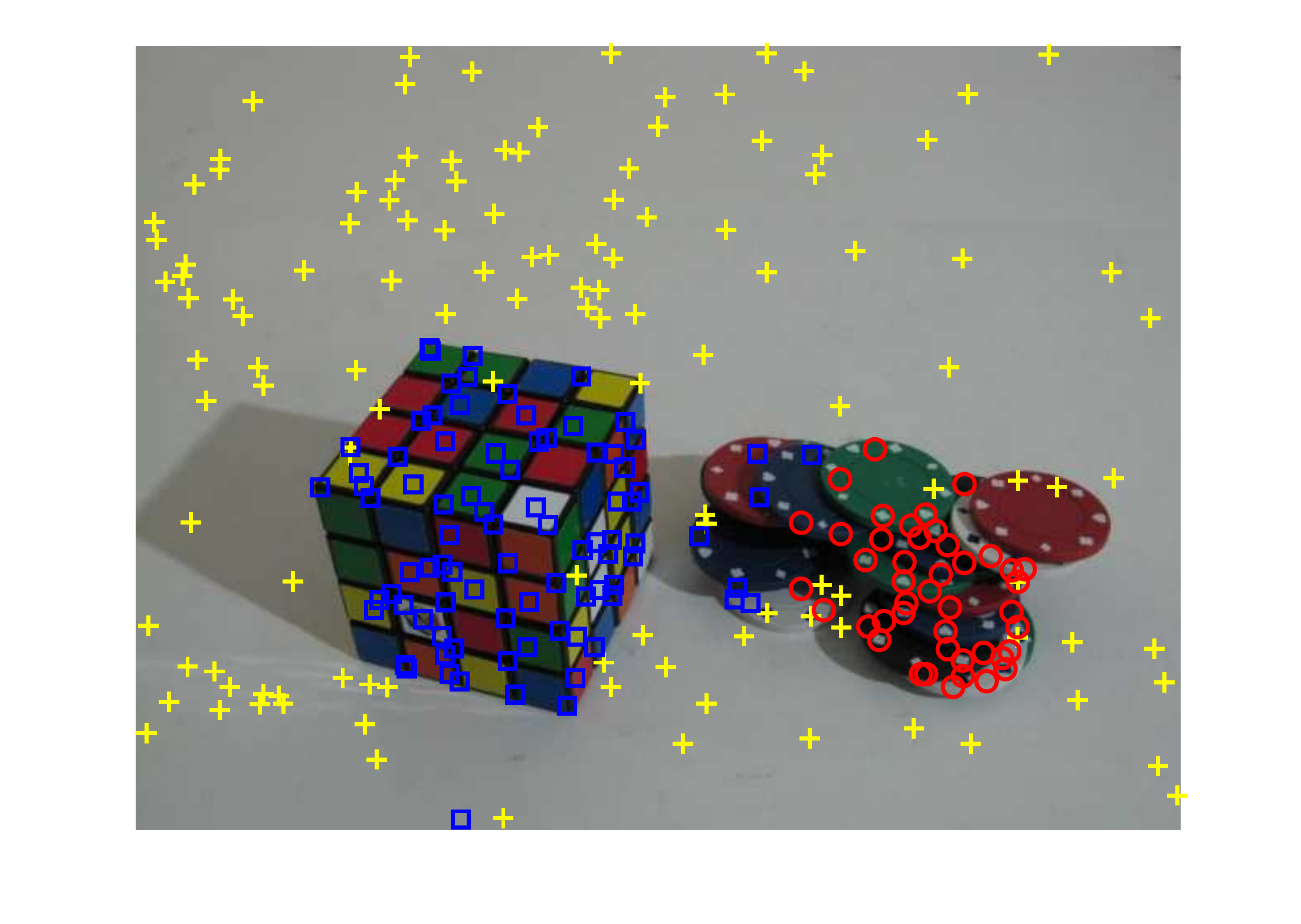}}
\centerline{\includegraphics[width=1.17\textwidth]{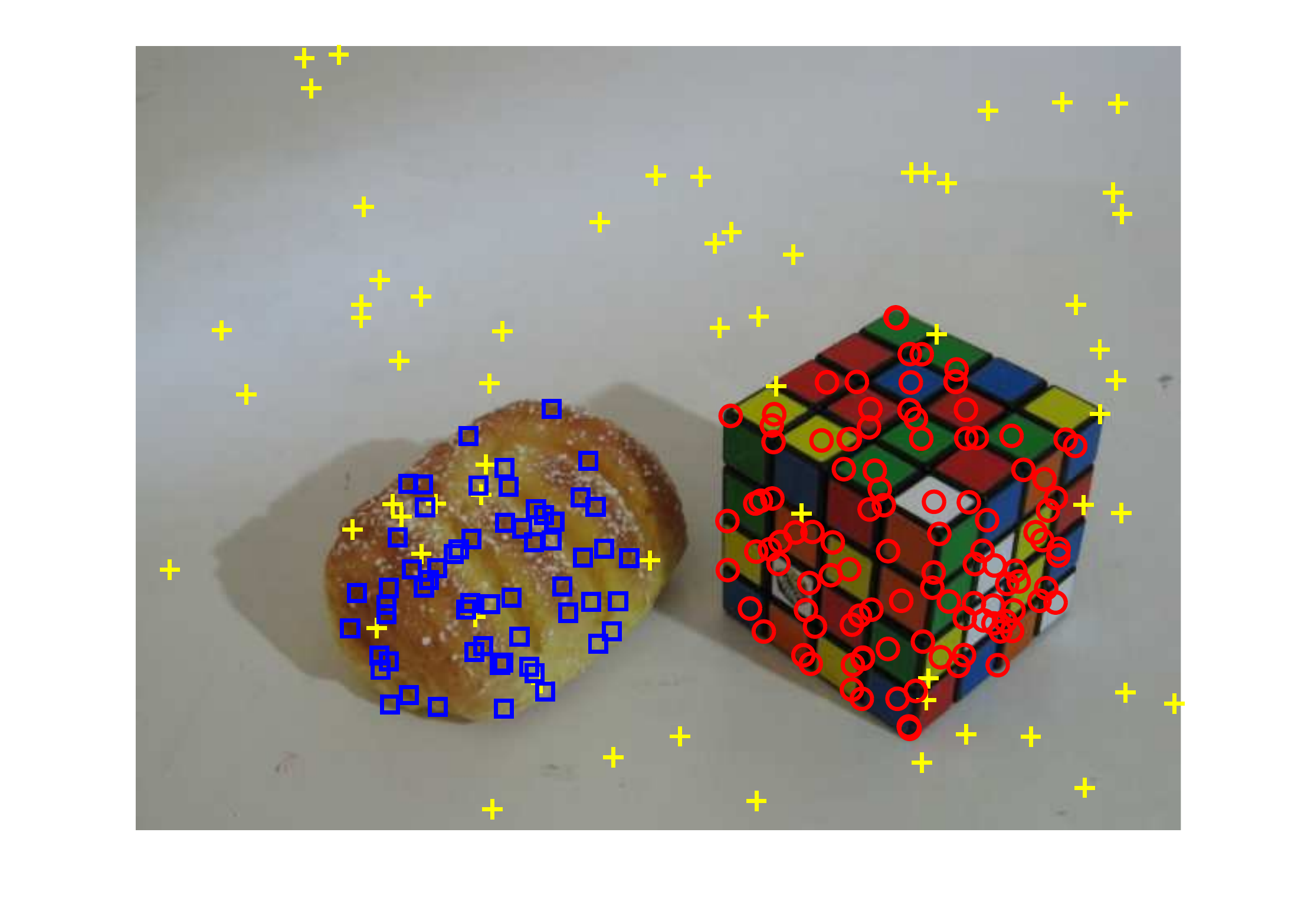}}
\centerline{\includegraphics[width=1.17\textwidth]{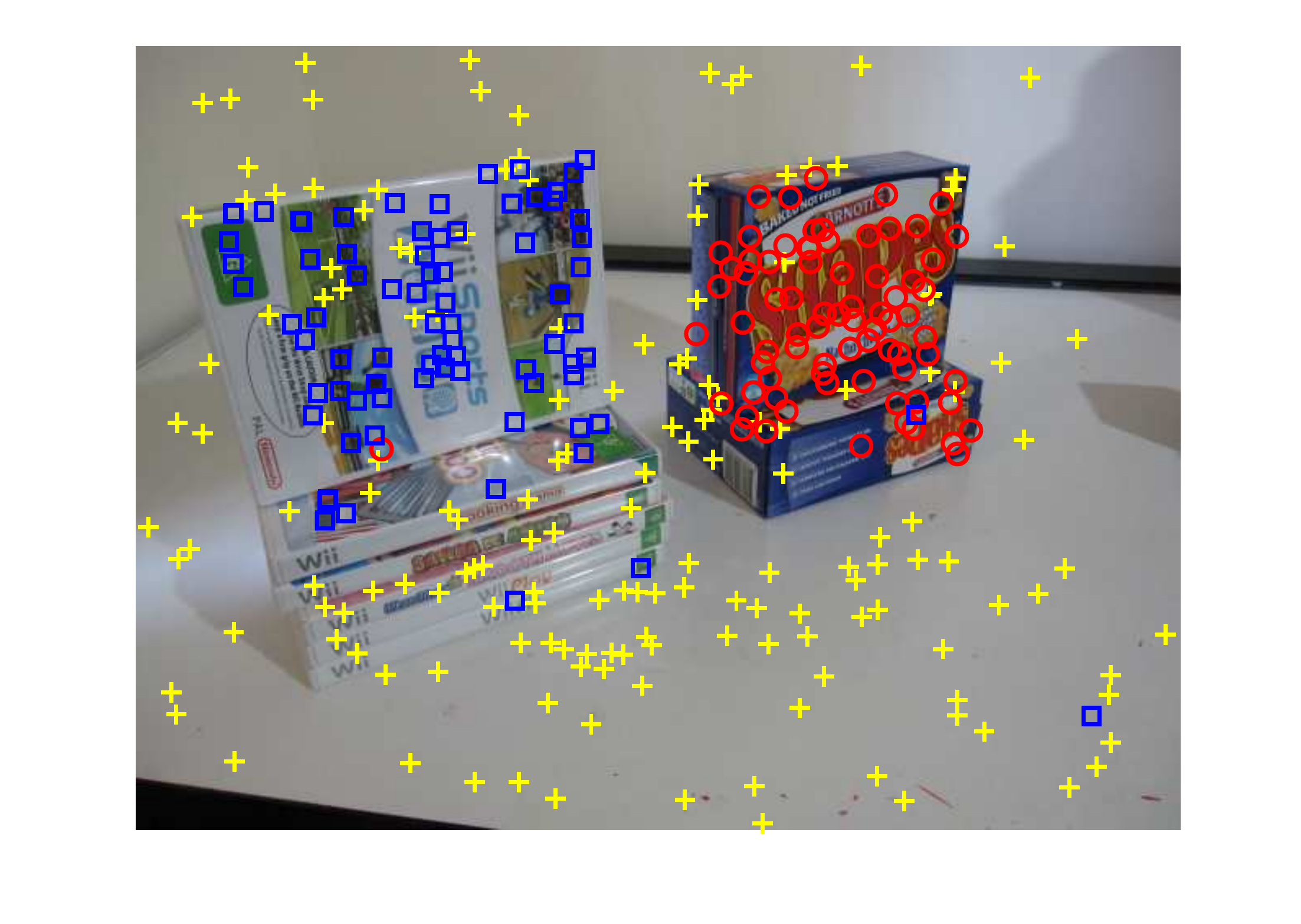}}
\centerline{\includegraphics[width=1.17\textwidth]{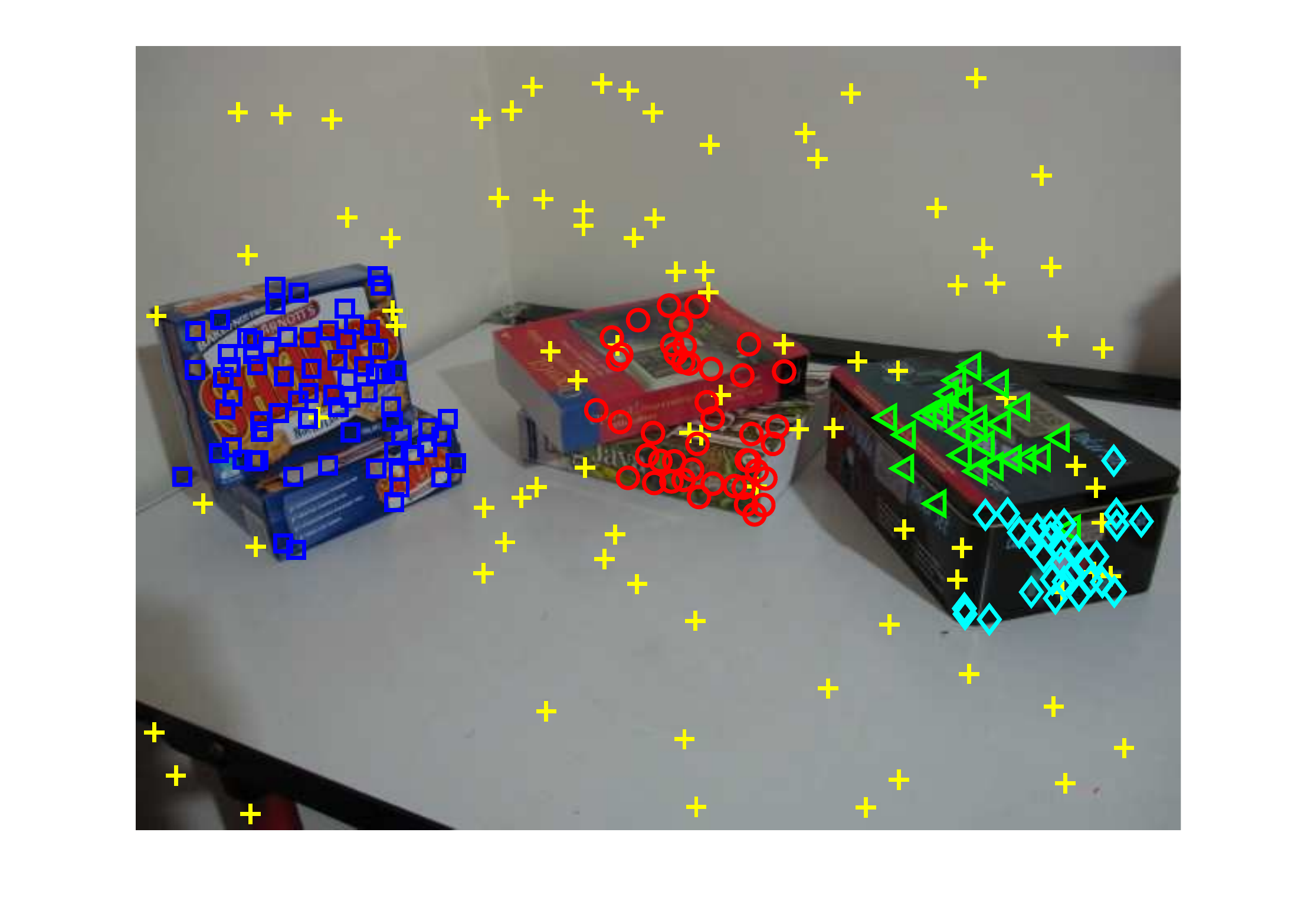}}
\centerline{\includegraphics[width=1.17\textwidth]{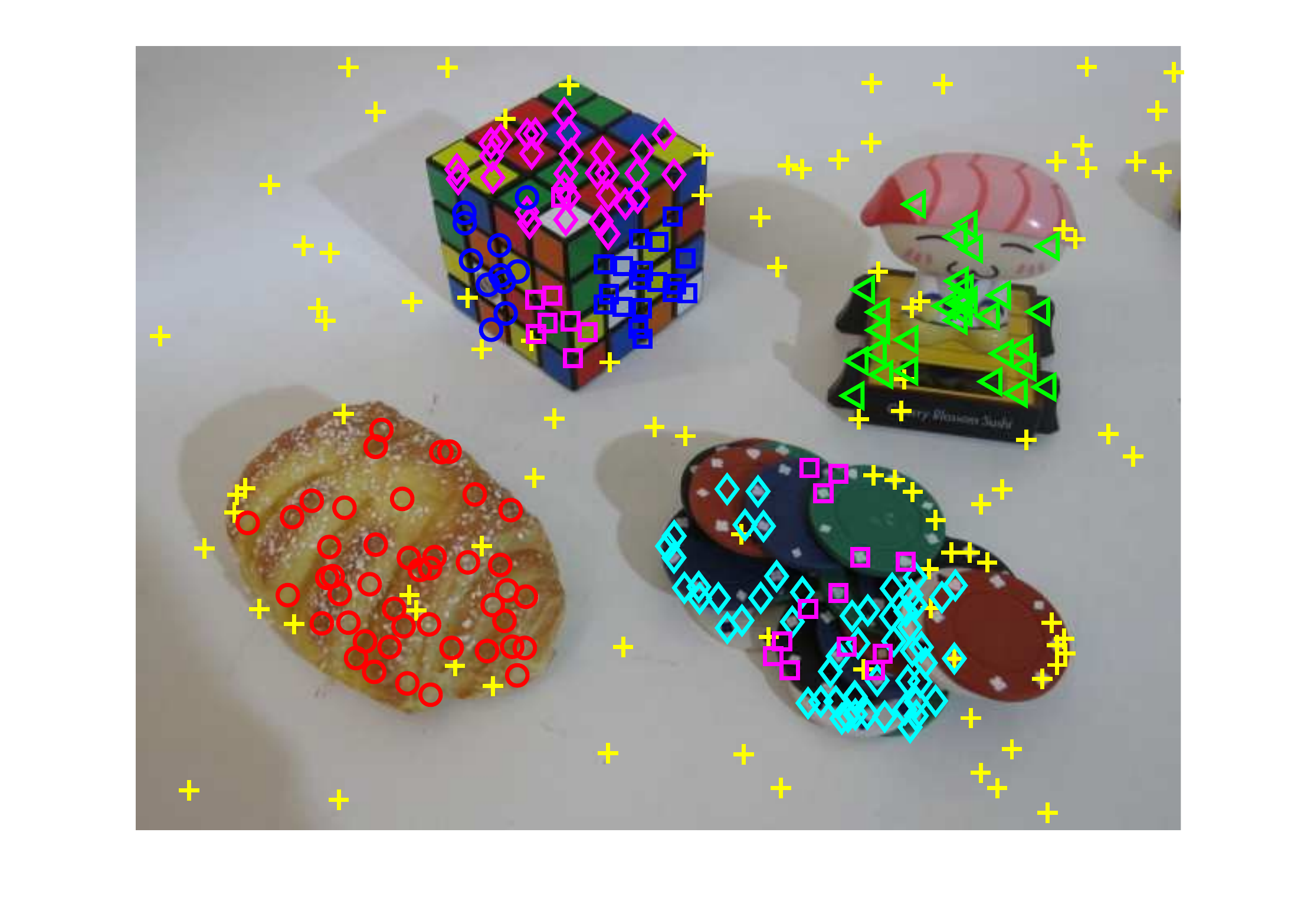}}
  \begin{center} (b)  \end{center}
\end{minipage}
\begin{minipage}[t]{.1585\textwidth}
  \centering
 \centerline{\includegraphics[width=1.17\textwidth]{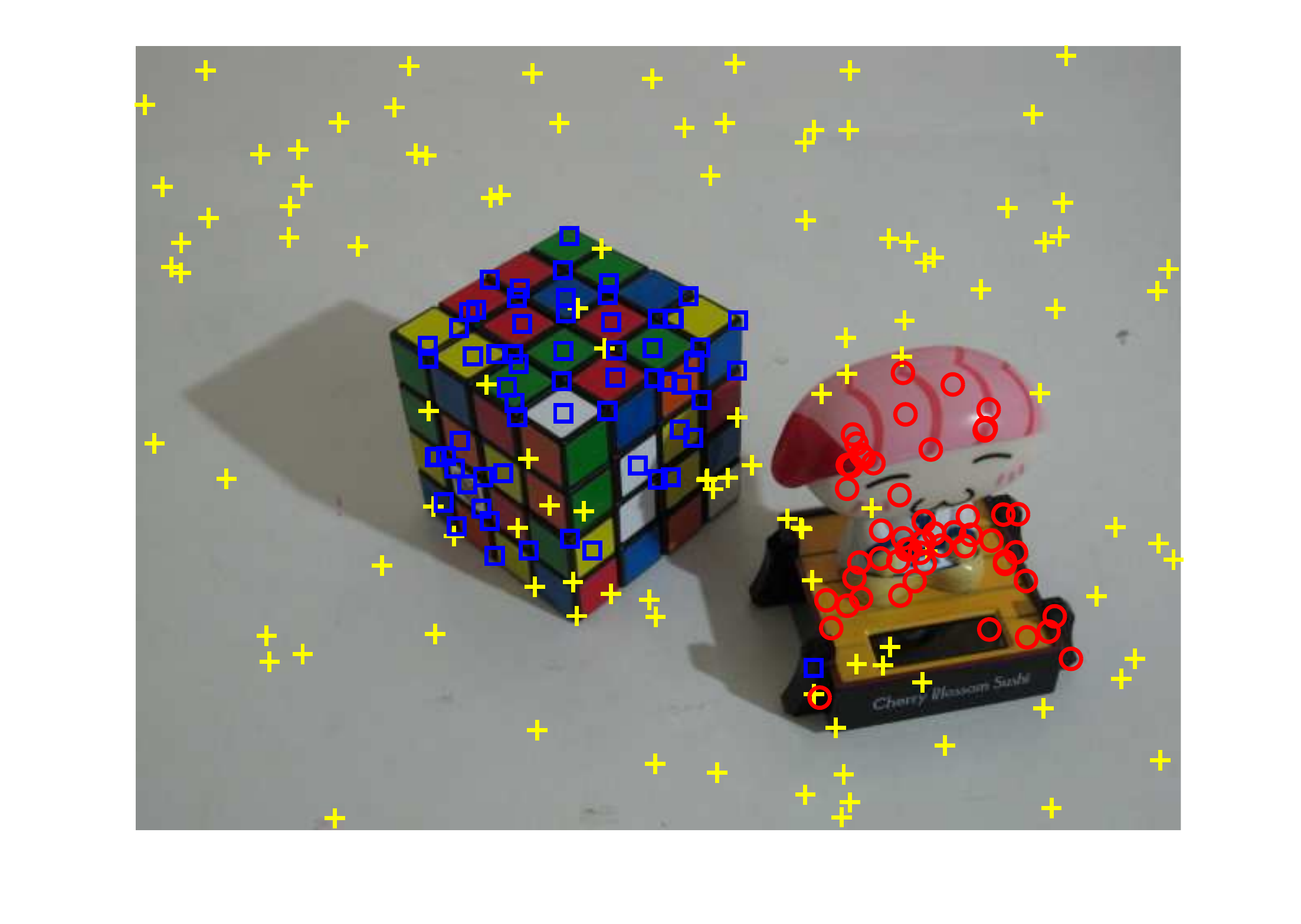}}
  \centerline{\includegraphics[width=1.17\textwidth]{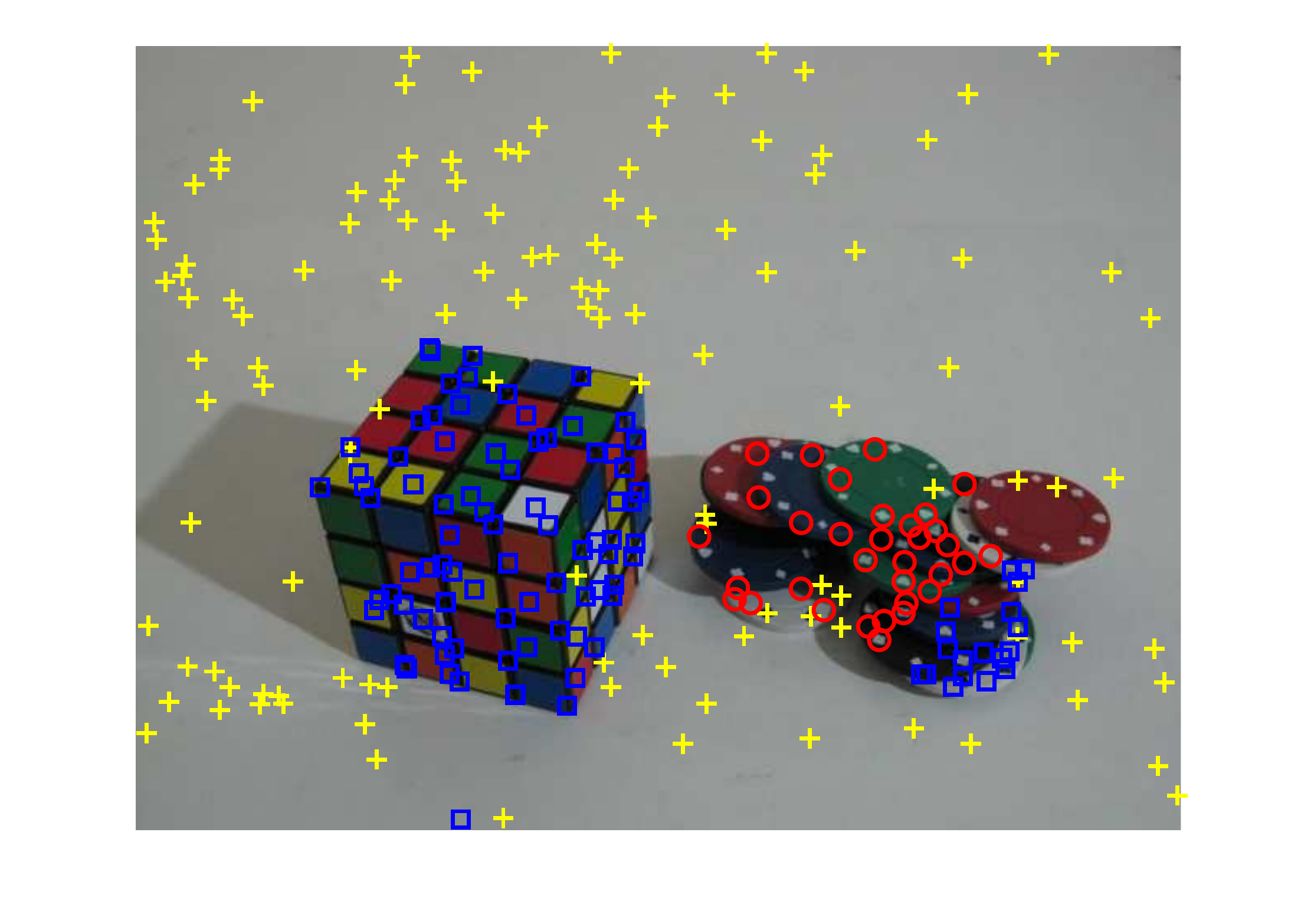}}
\centerline{\includegraphics[width=1.17\textwidth]{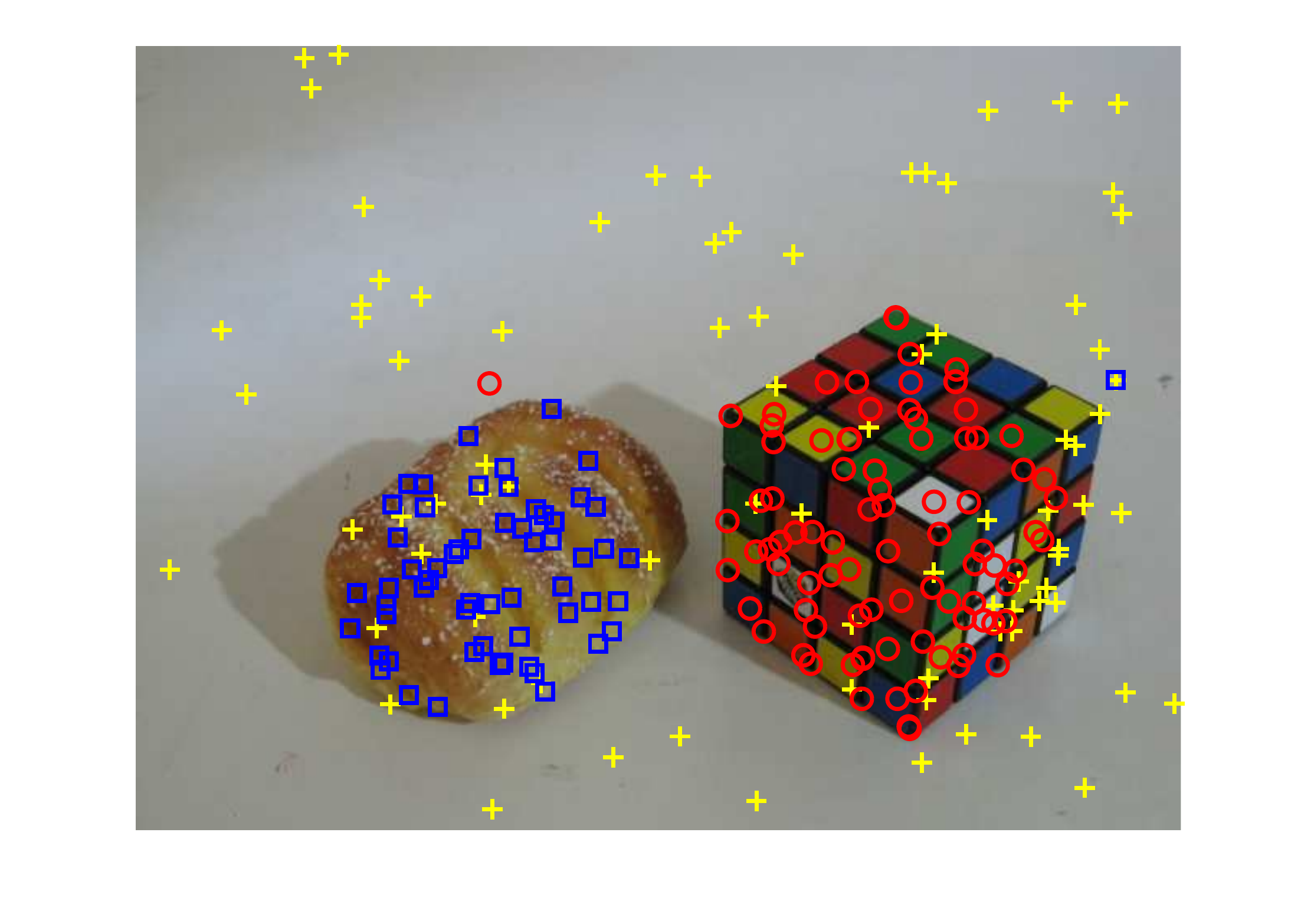}}
\centerline{\includegraphics[width=1.17\textwidth]{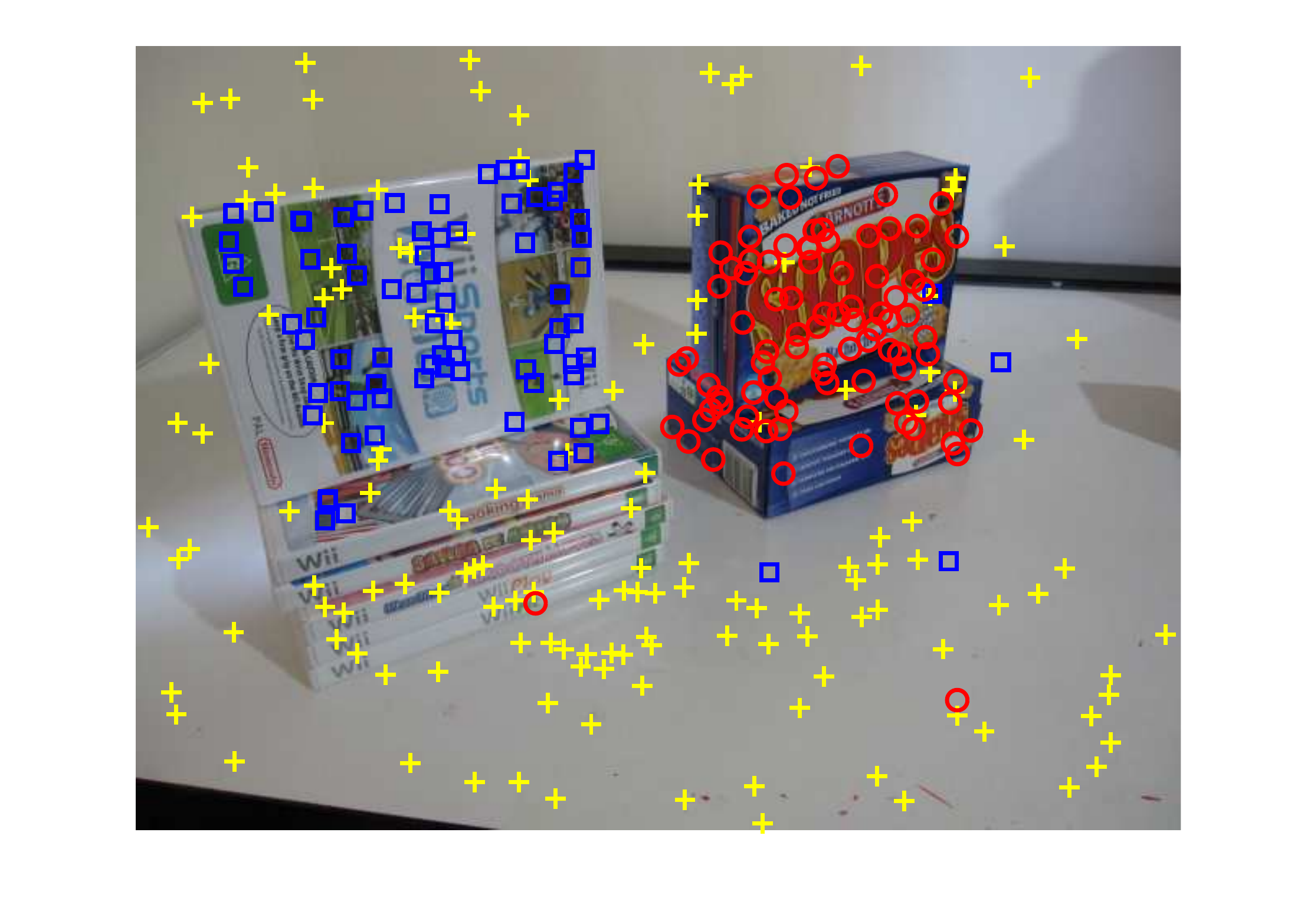}}
\centerline{\includegraphics[width=1.17\textwidth]{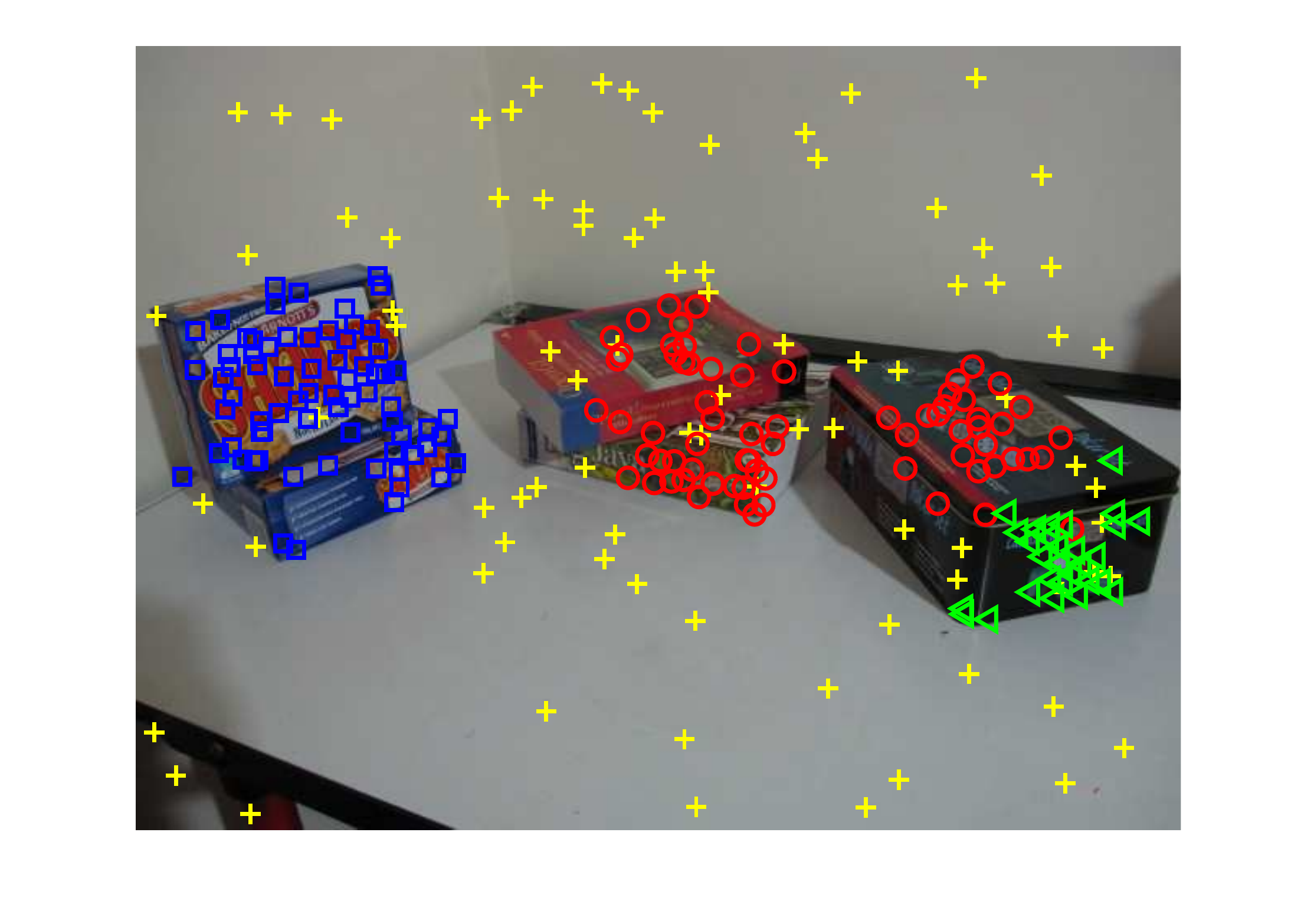}}
\centerline{\includegraphics[width=1.17\textwidth]{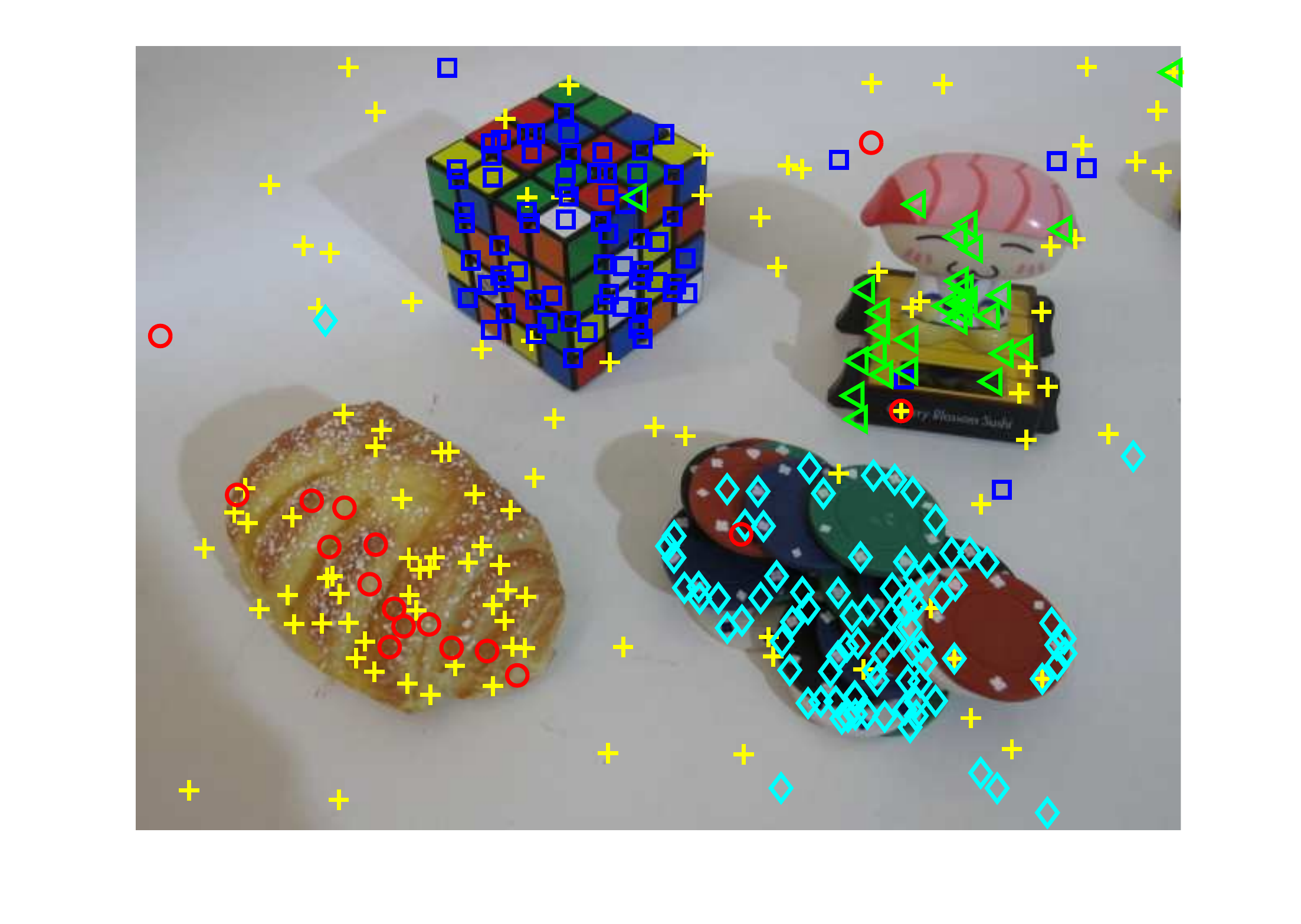}}
  \begin{center} (c) \end{center}
\end{minipage}
\begin{minipage}[t]{.1585\textwidth}
  \centering
 \centerline{\includegraphics[width=1.17\textwidth]{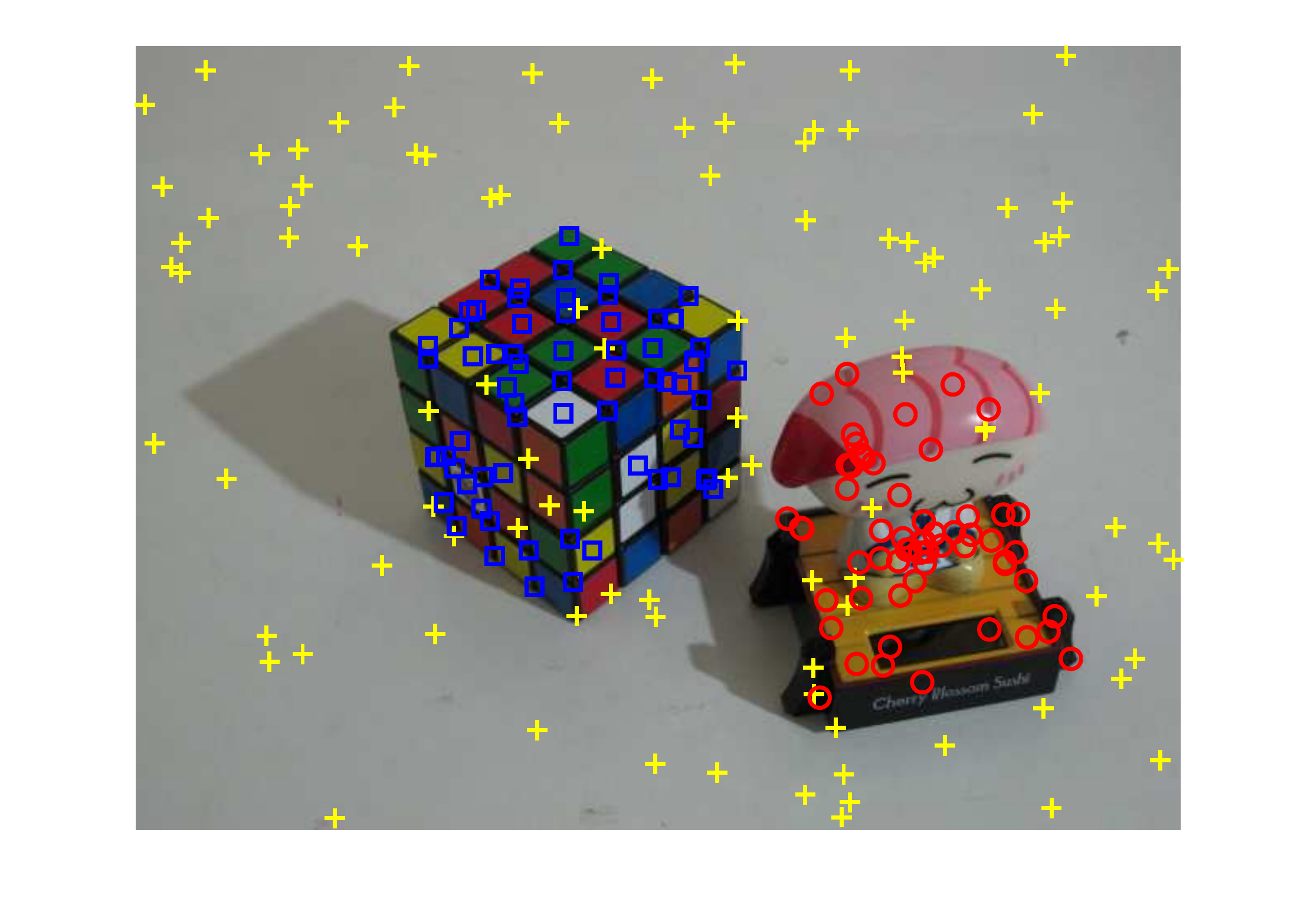}}
  \centerline{\includegraphics[width=1.17\textwidth]{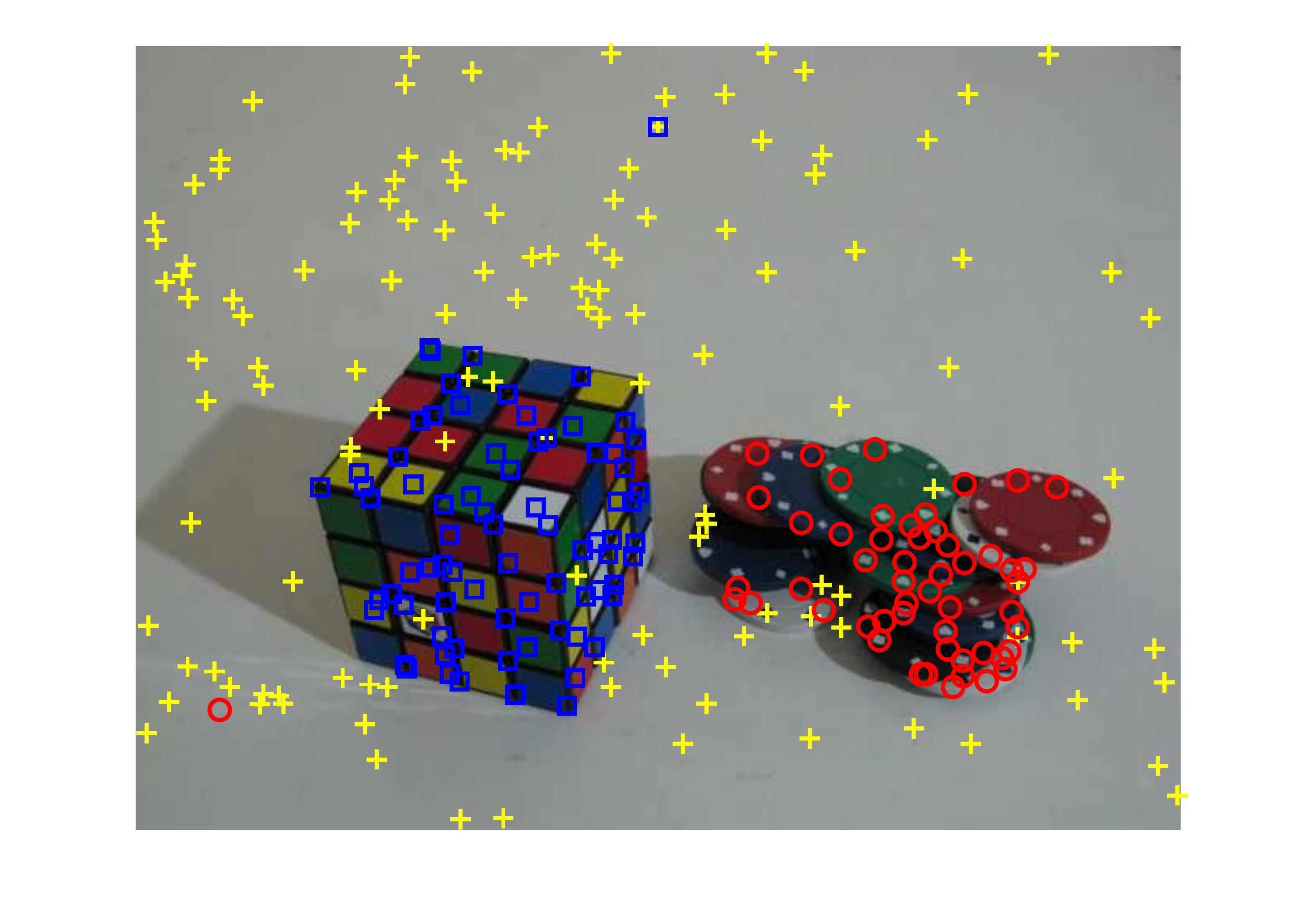}}
\centerline{\includegraphics[width=1.17\textwidth]{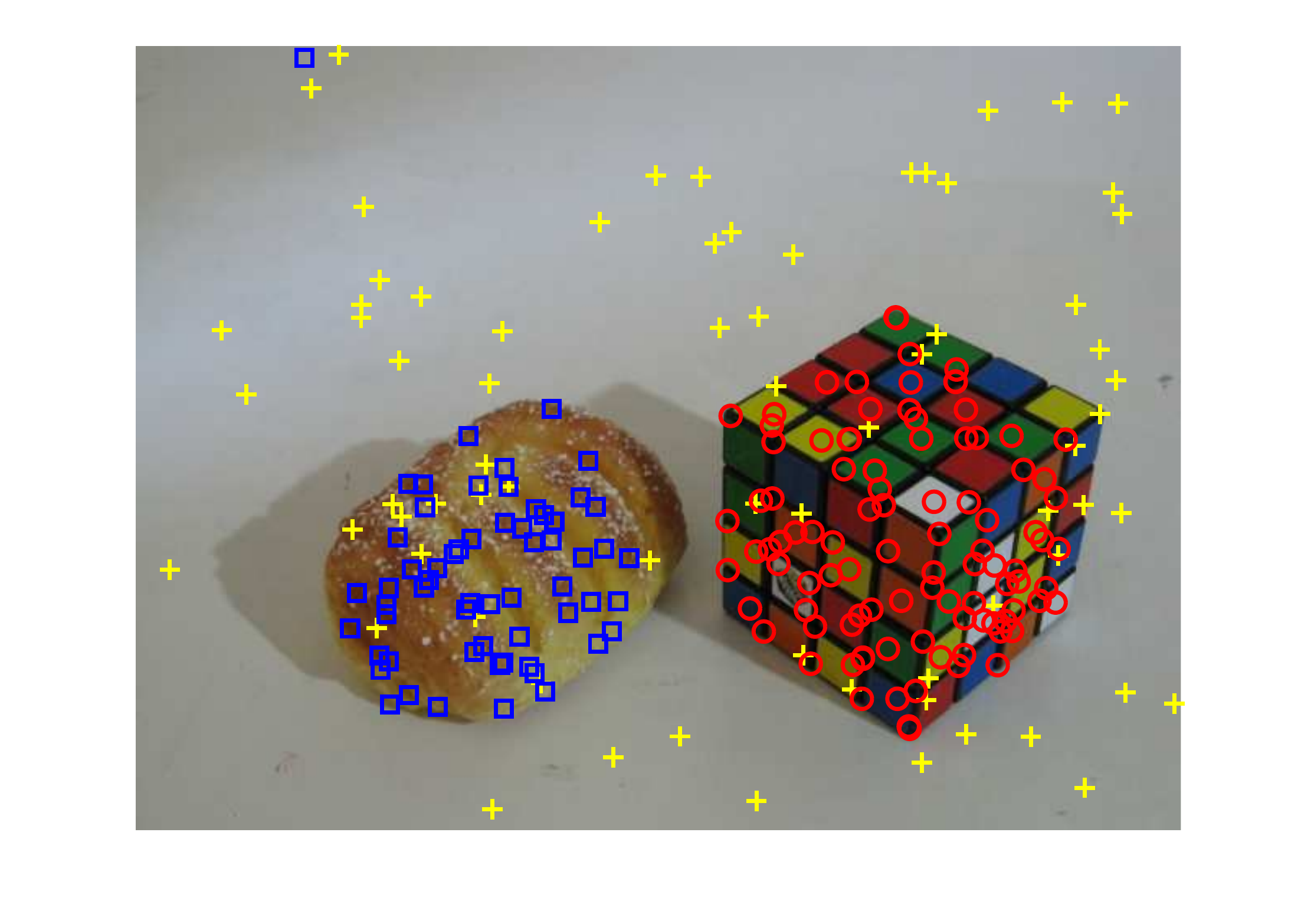}}
\centerline{\includegraphics[width=1.17\textwidth]{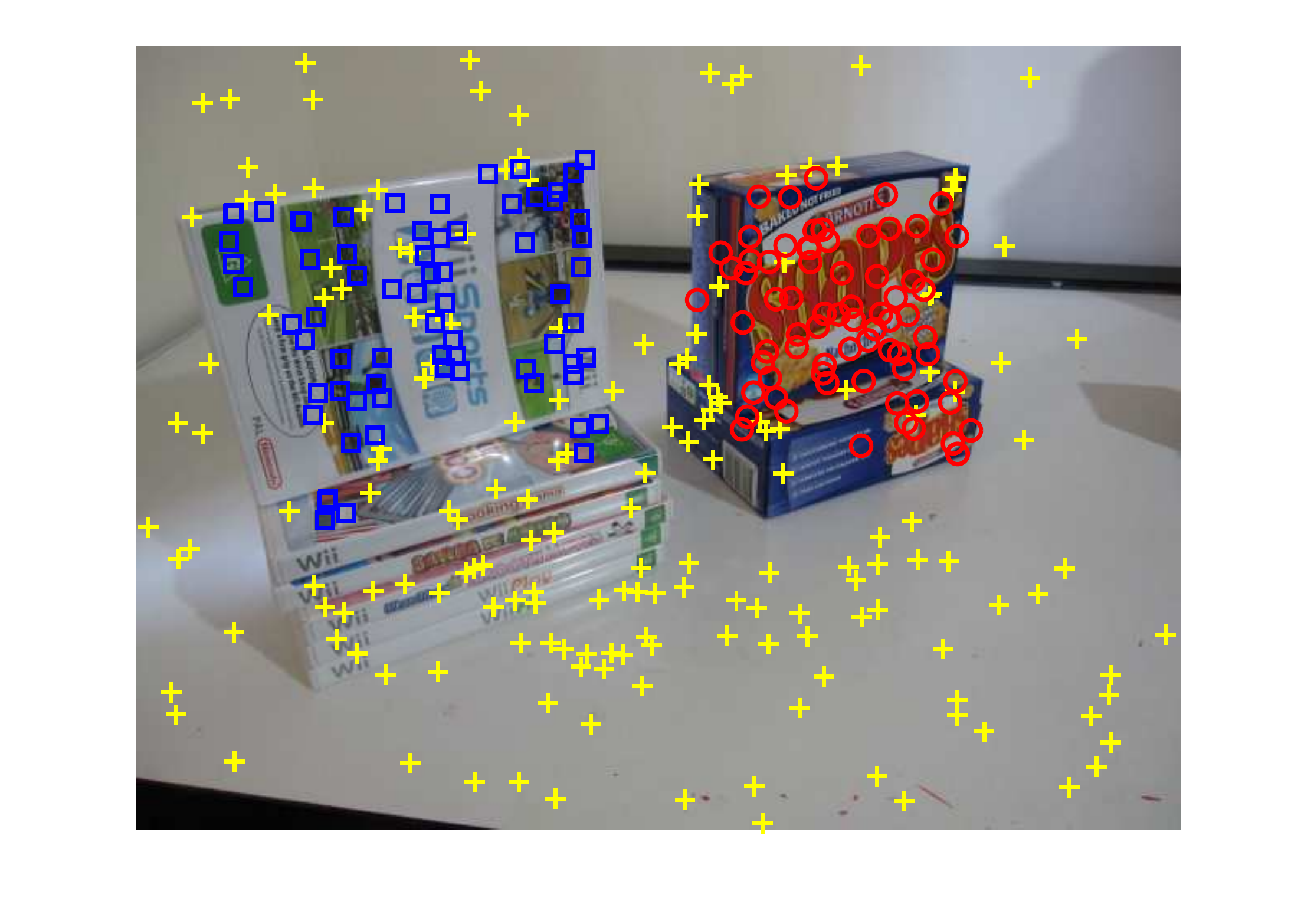}}
\centerline{\includegraphics[width=1.17\textwidth]{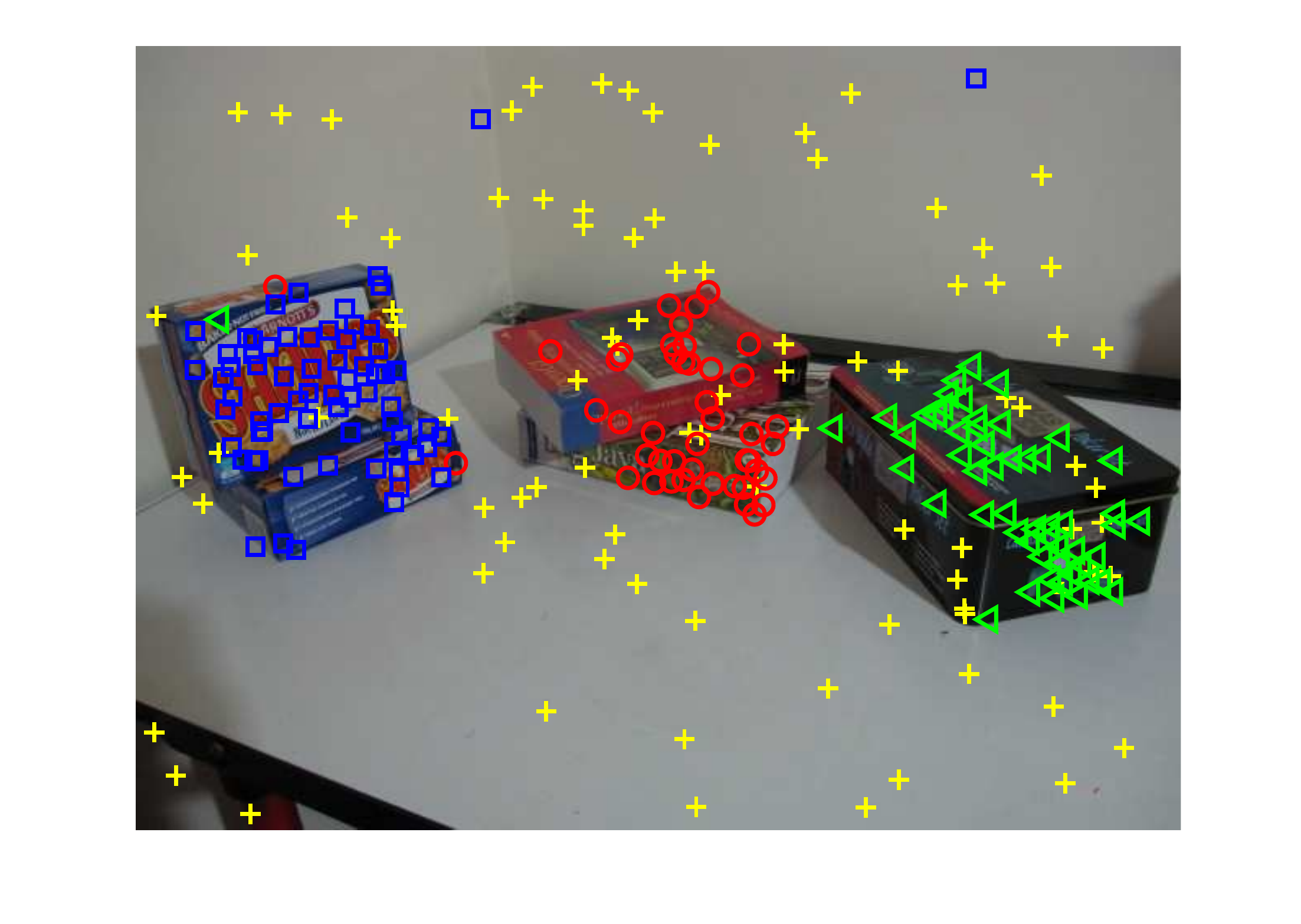}}
\centerline{\includegraphics[width=1.17\textwidth]{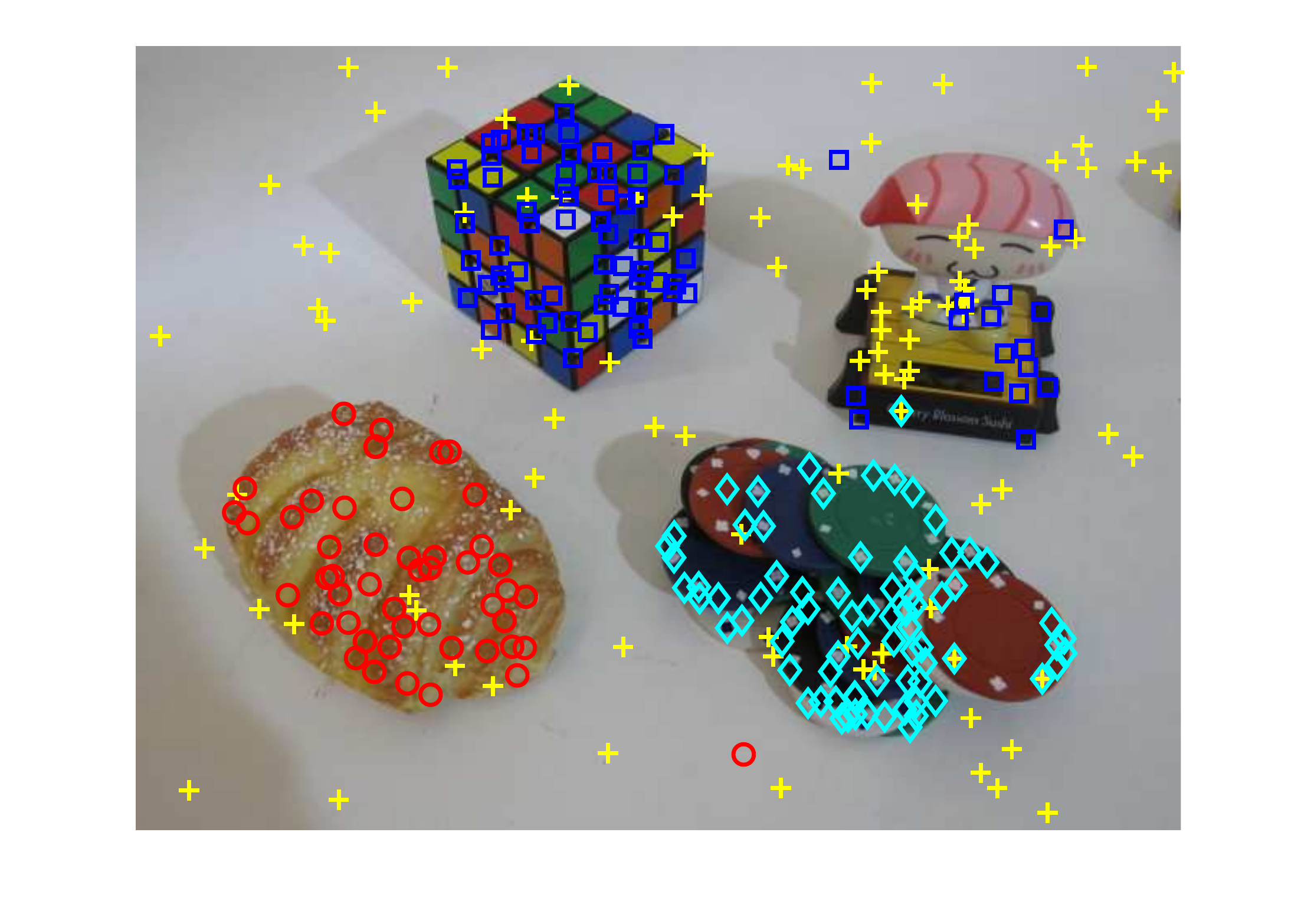}}
  \begin{center} (d)  \end{center}
\end{minipage}
\begin{minipage}[t]{.1585\textwidth}
  \centering
 \centerline{\includegraphics[width=1.17\textwidth]{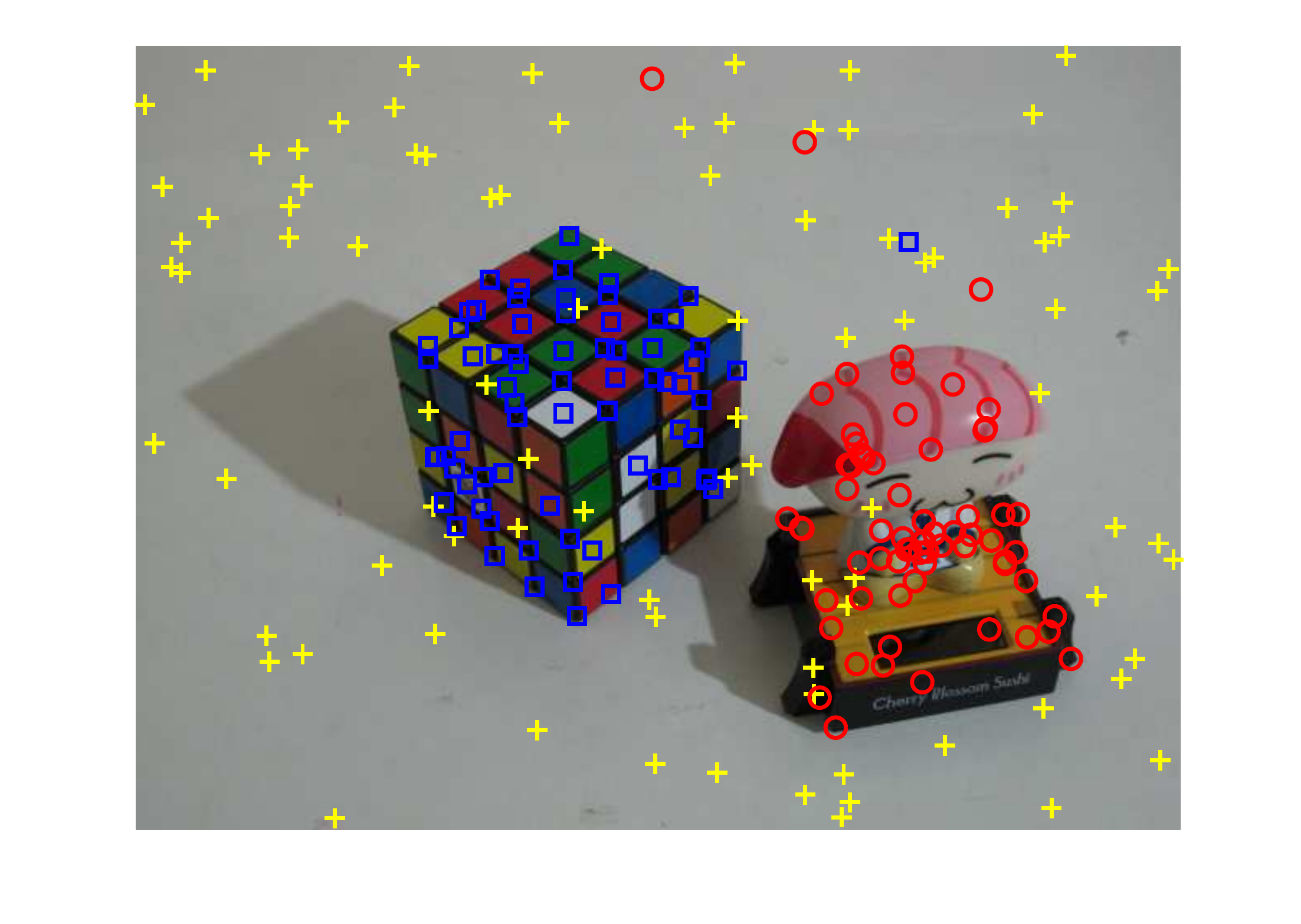}}
  \centerline{\includegraphics[width=1.17\textwidth]{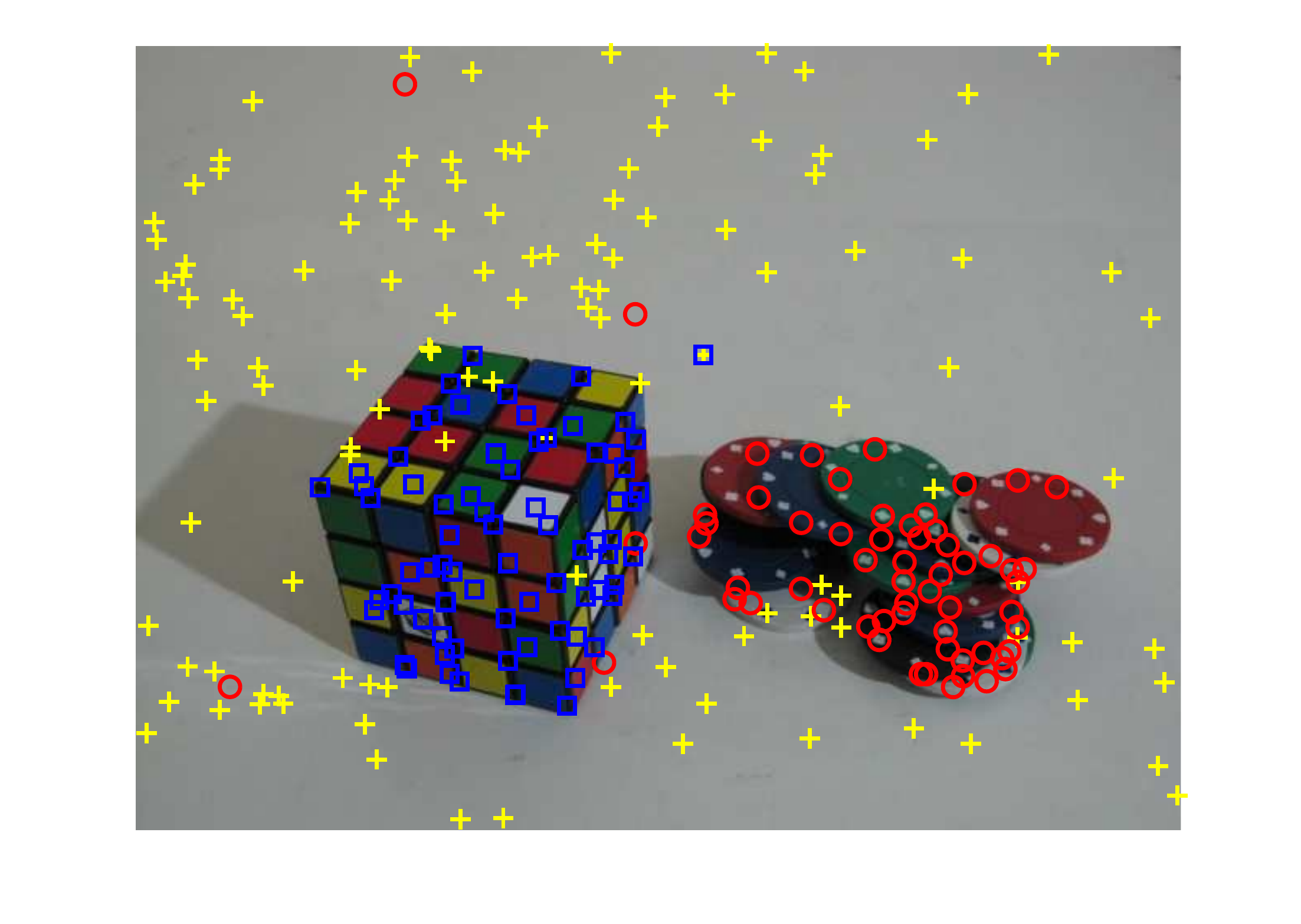}}
\centerline{\includegraphics[width=1.17\textwidth]{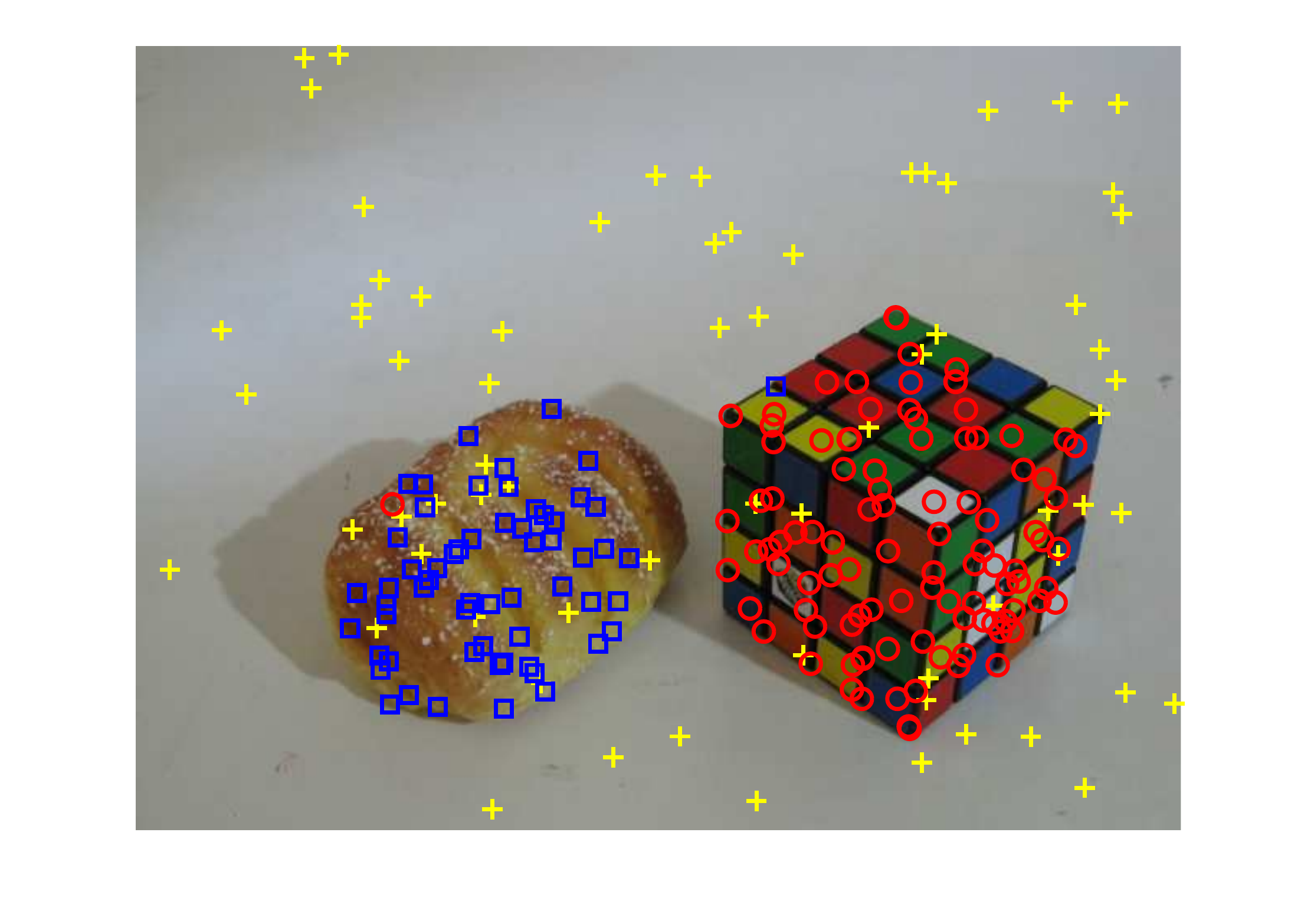}}
\centerline{\includegraphics[width=1.17\textwidth]{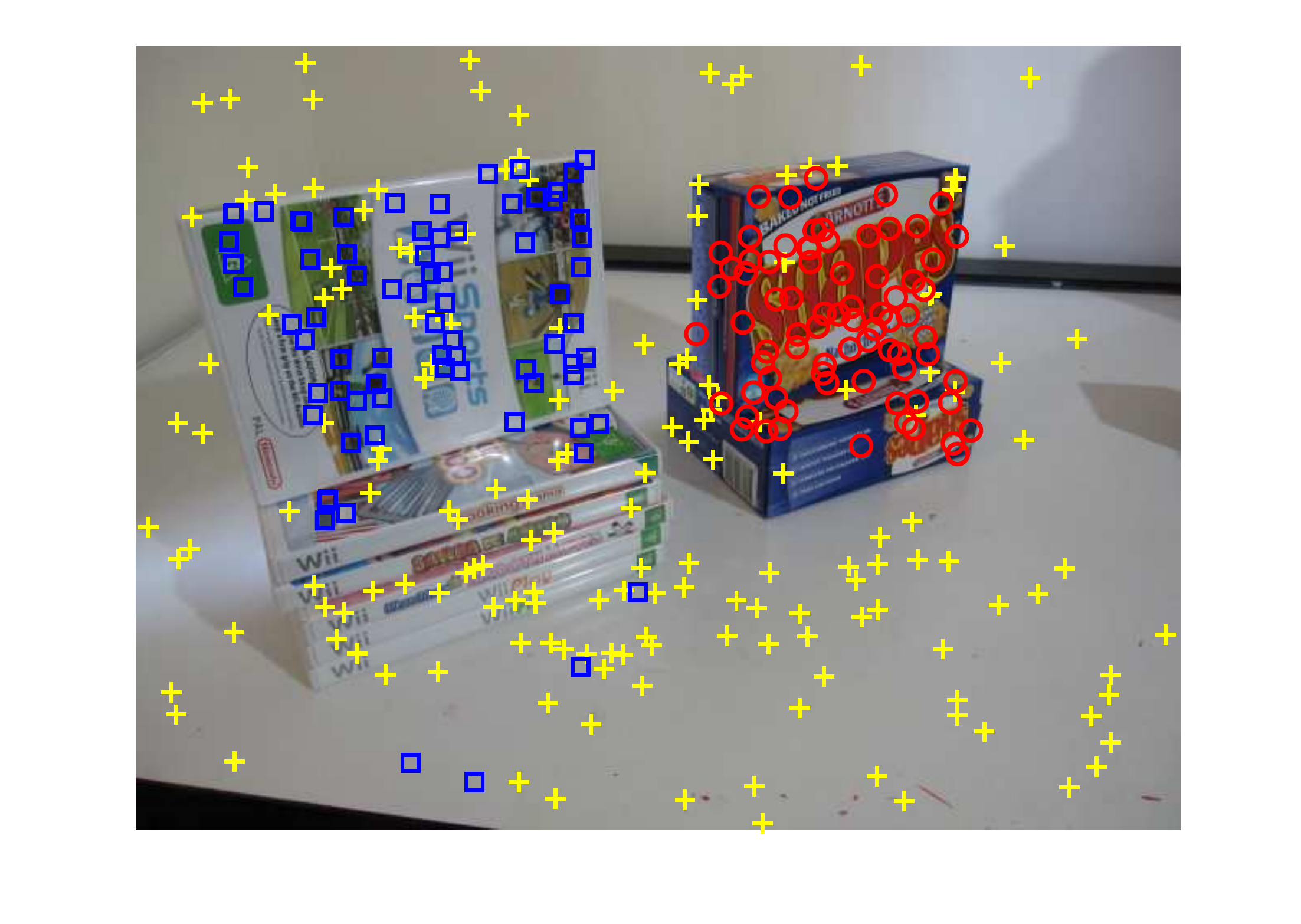}}
\centerline{\includegraphics[width=1.17\textwidth]{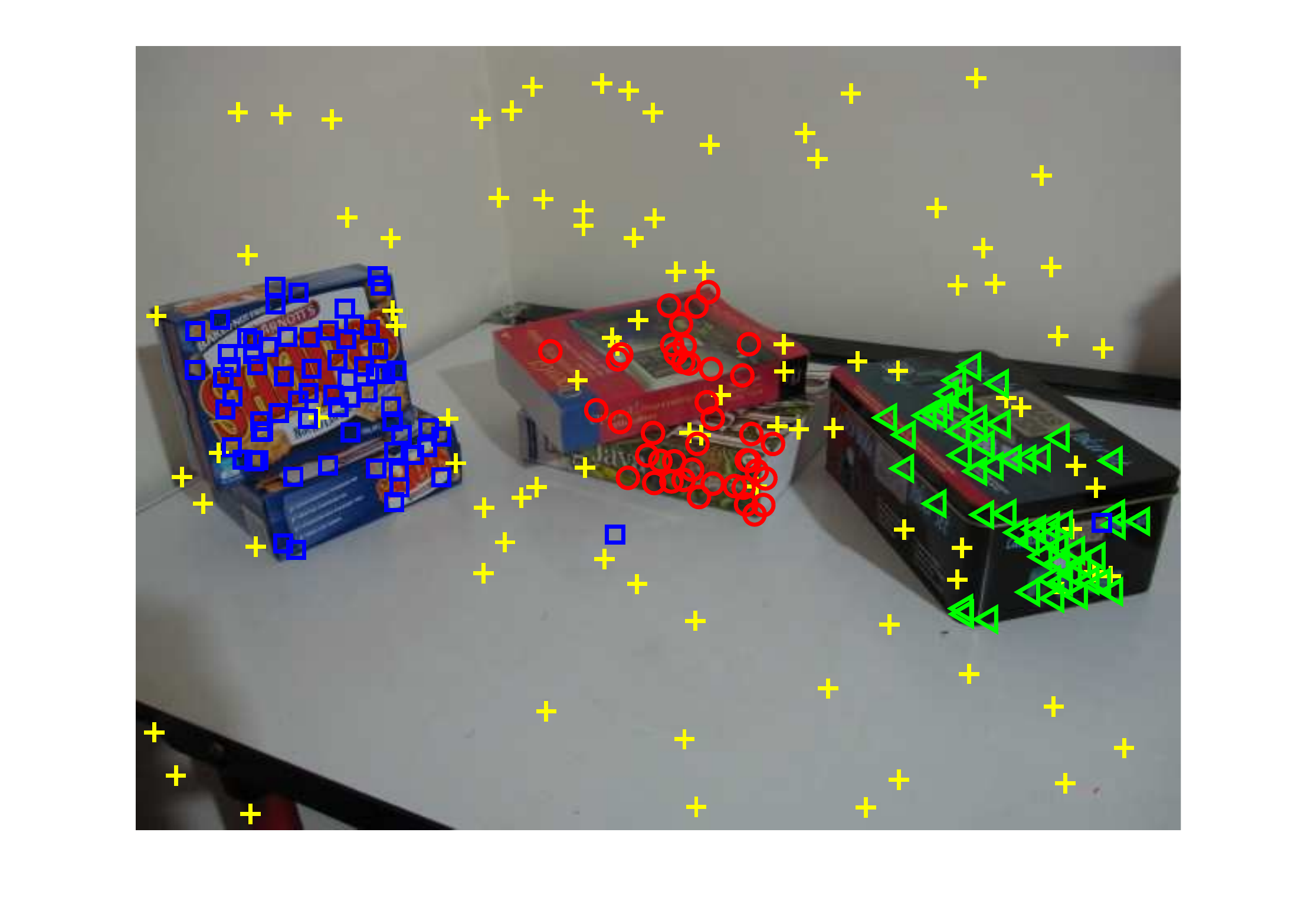}}
\centerline{\includegraphics[width=1.17\textwidth]{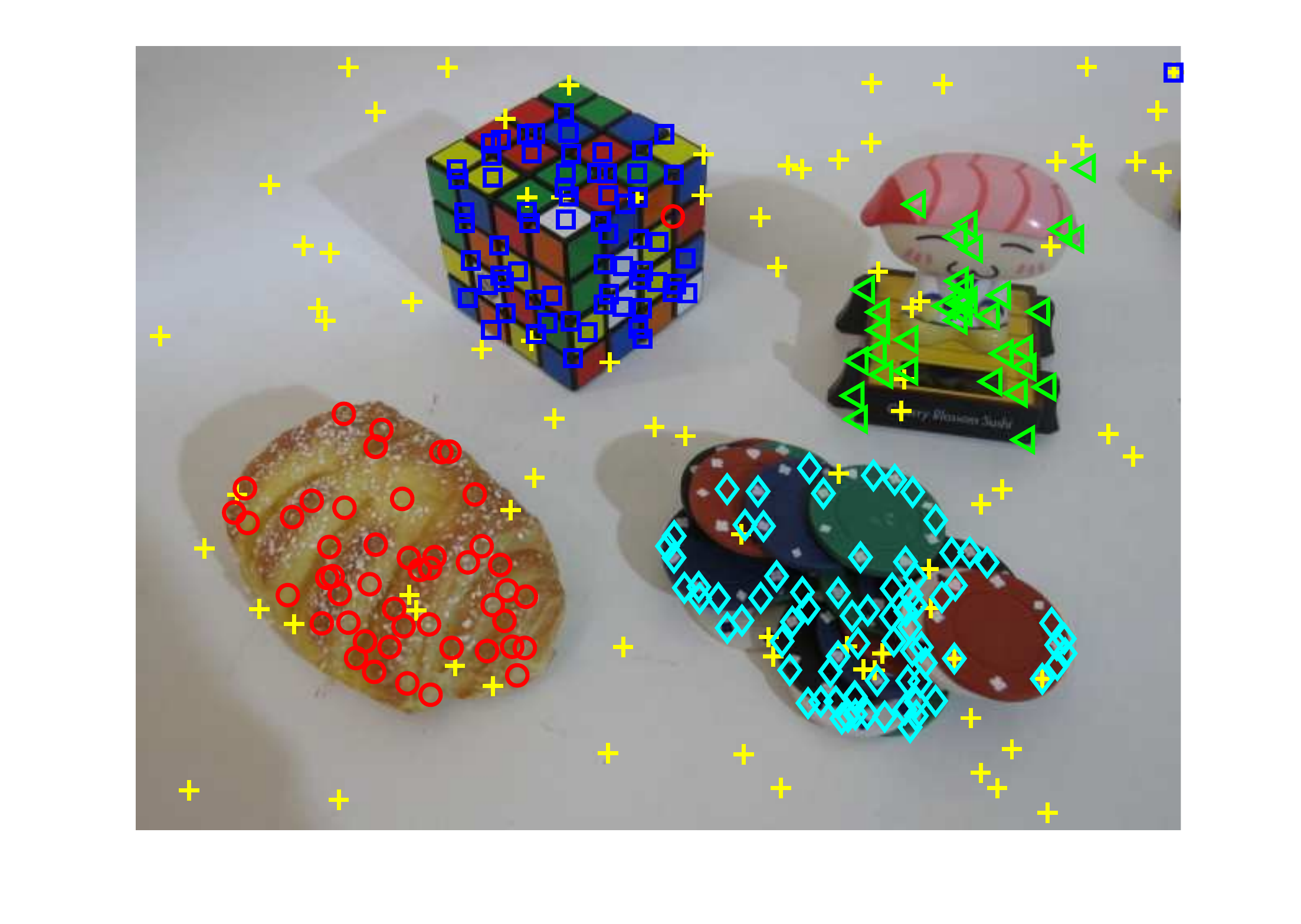}}
  \begin{center} (e)  \end{center}
\end{minipage}
\begin{minipage}[t]{.1585\textwidth}
  \centering
 \centerline{\includegraphics[width=1.17\textwidth]{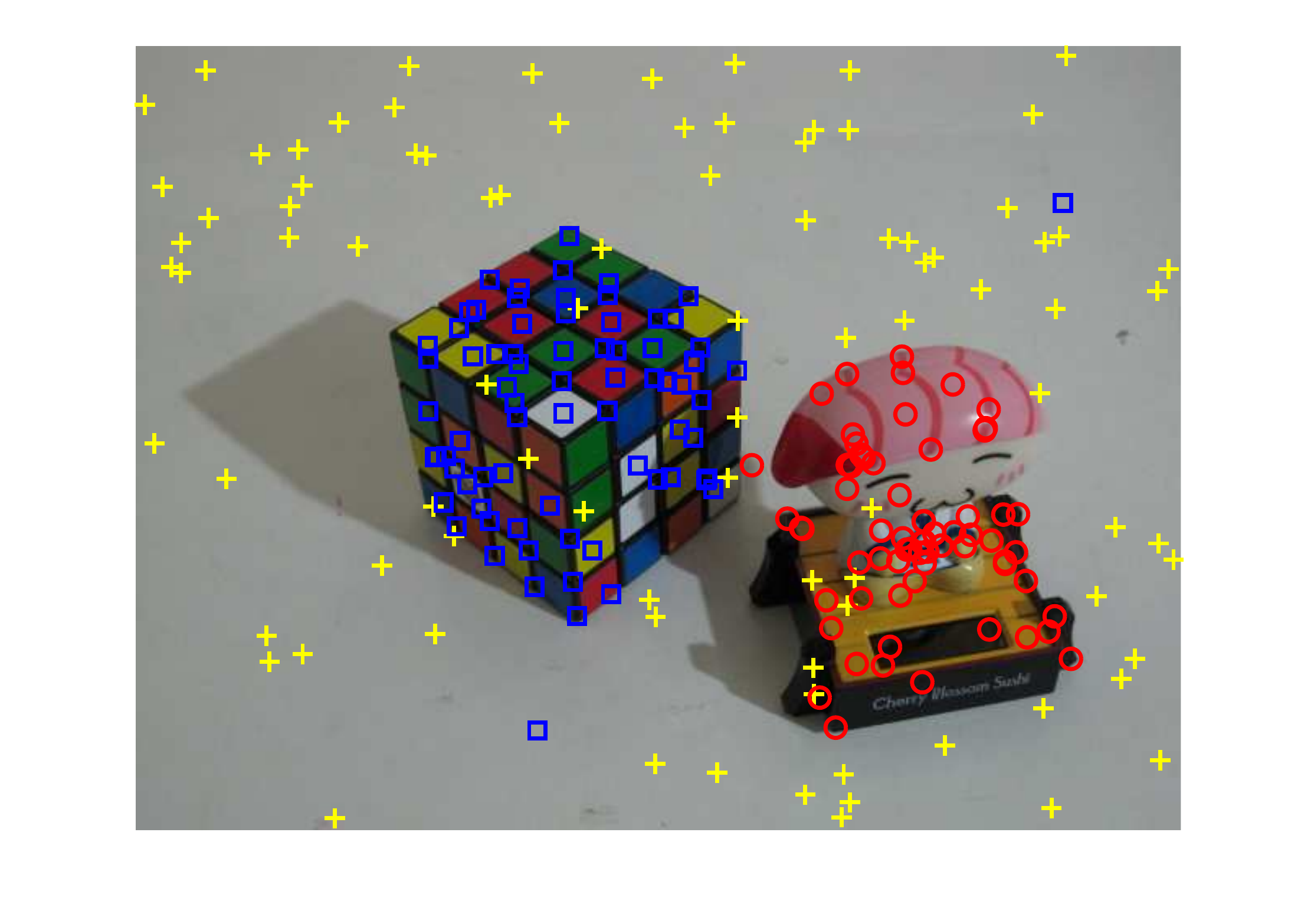}}
  \centerline{\includegraphics[width=1.17\textwidth]{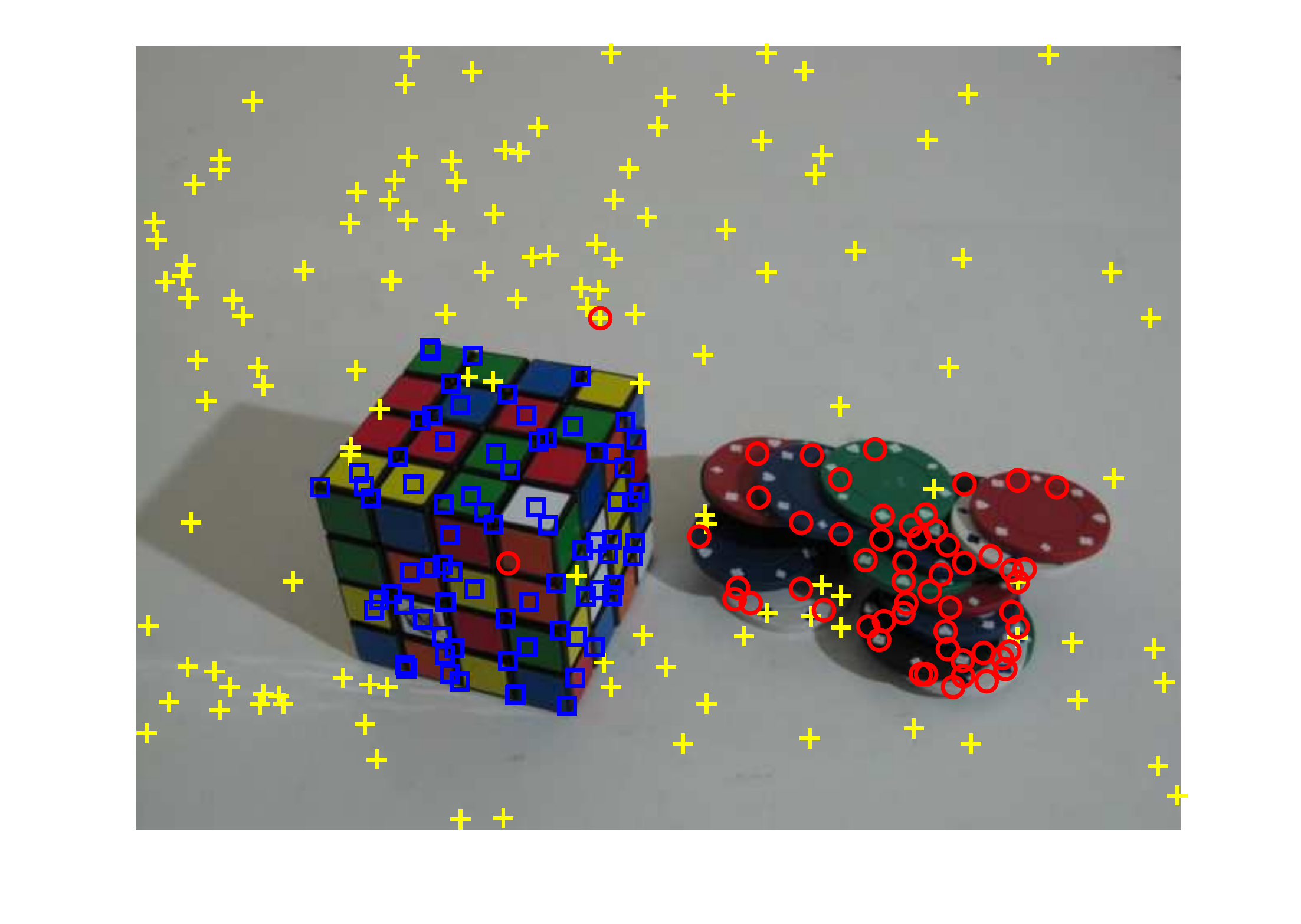}}
\centerline{\includegraphics[width=1.17\textwidth]{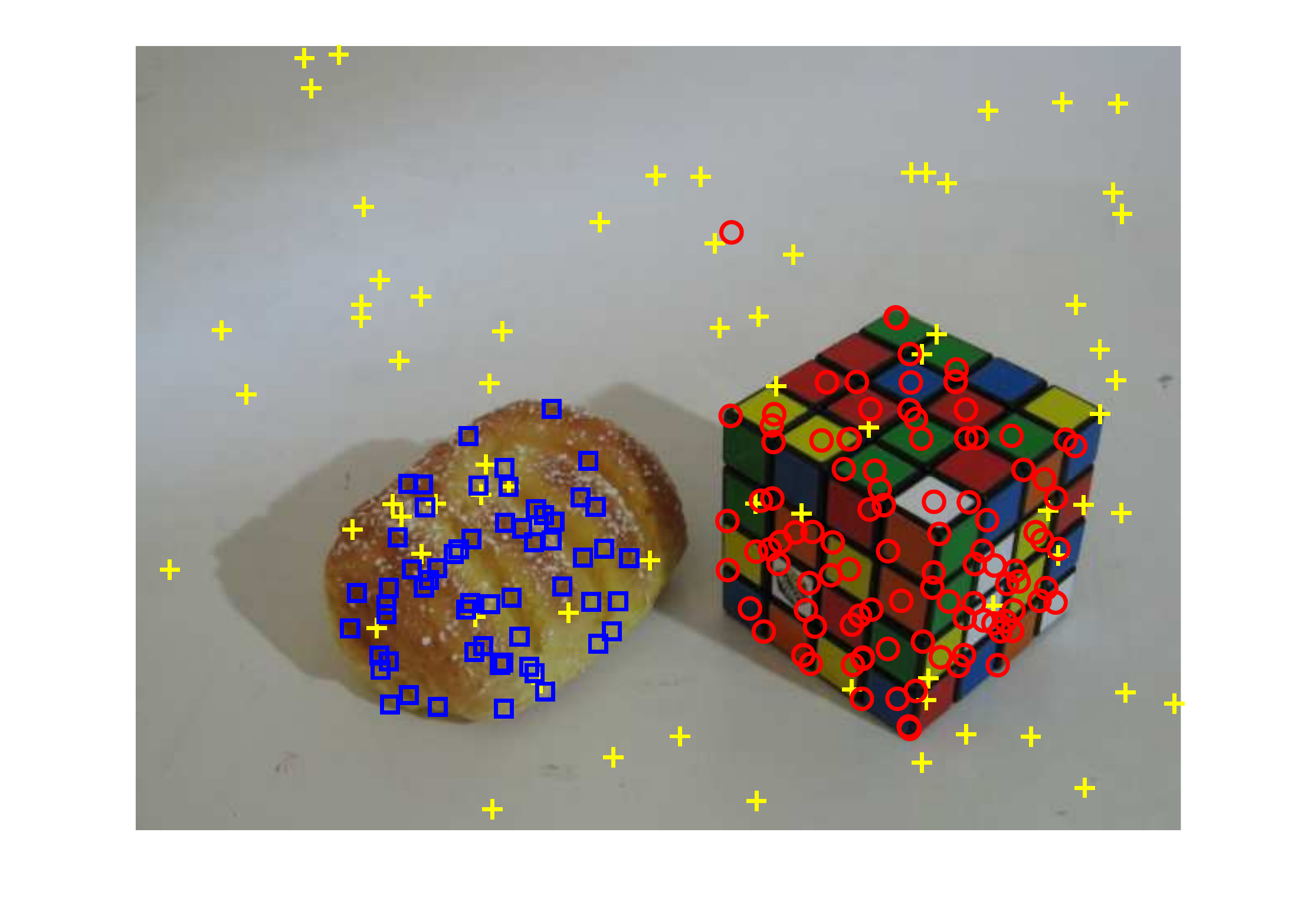}}
\centerline{\includegraphics[width=1.17\textwidth]{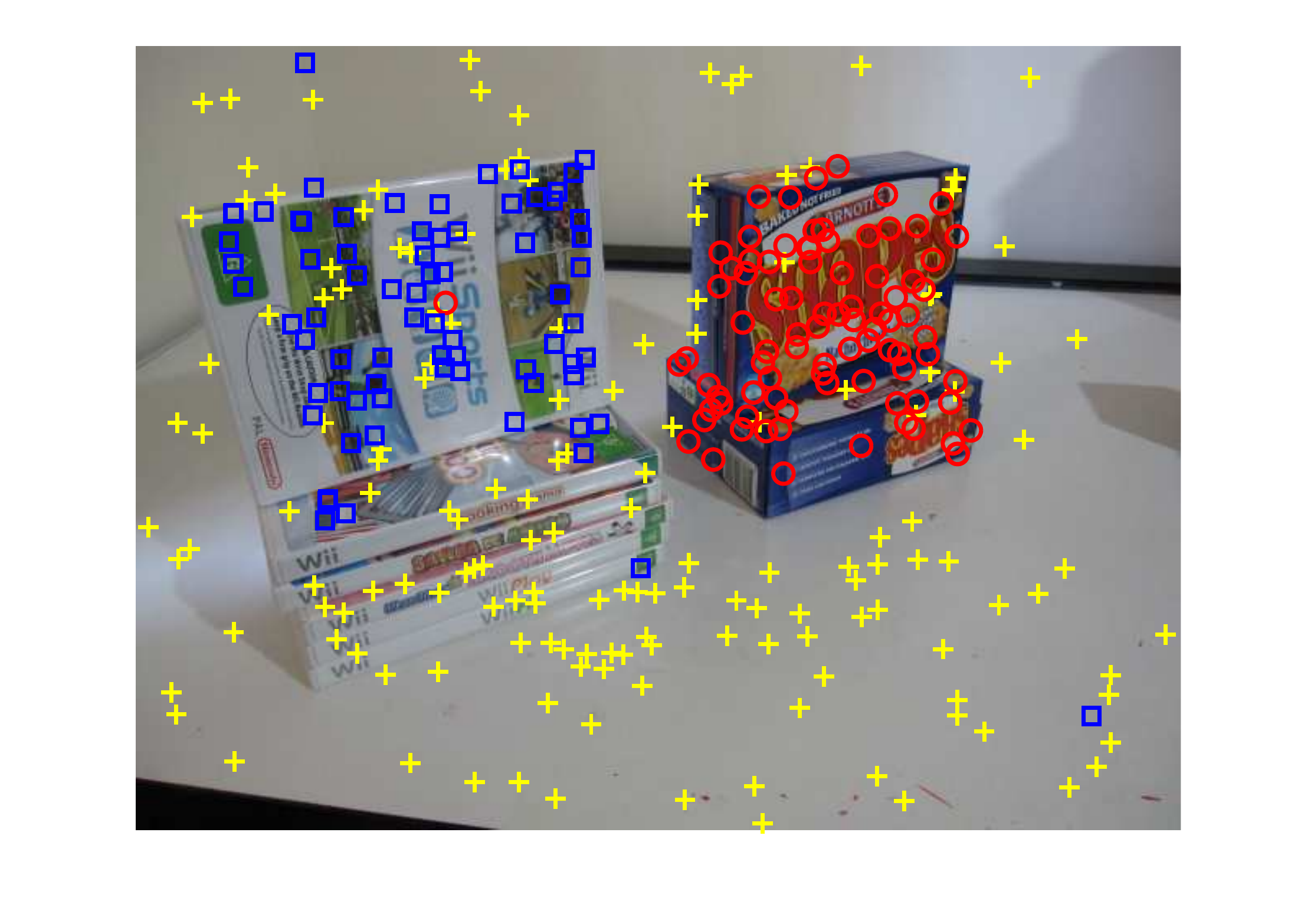}}
\centerline{\includegraphics[width=1.17\textwidth]{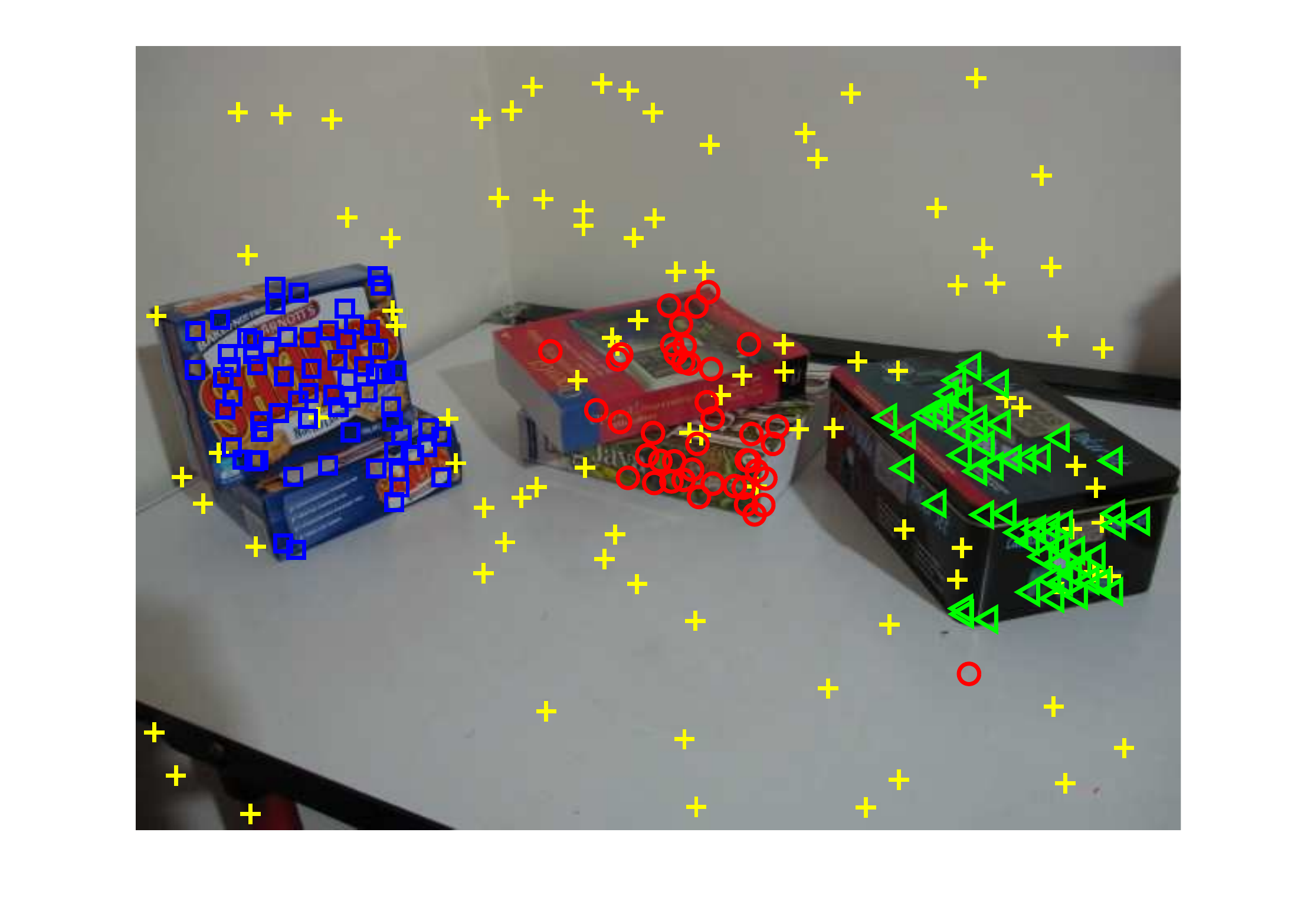}}
\centerline{\includegraphics[width=1.17\textwidth]{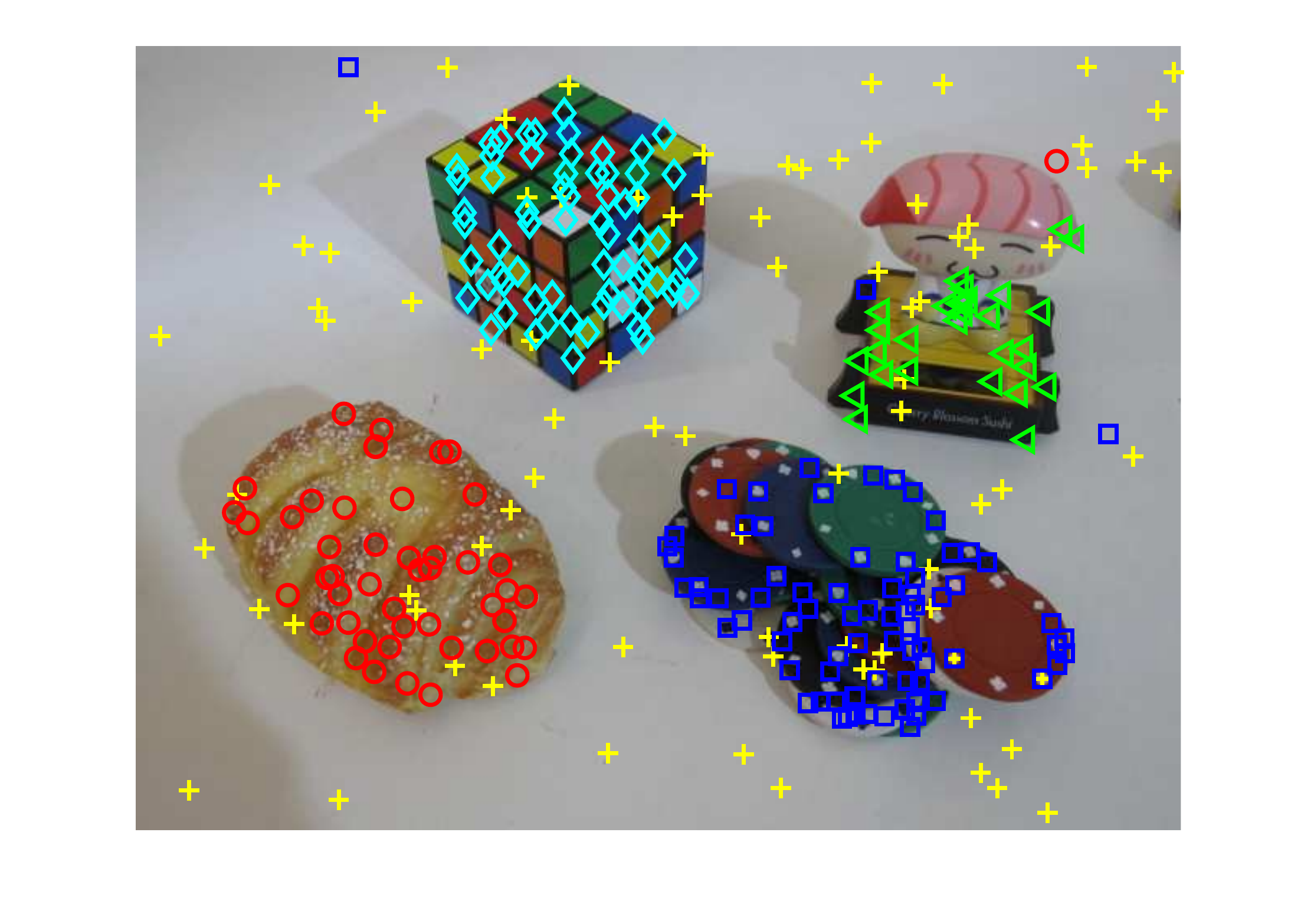}}
  \begin{center} (f)  \end{center}
\end{minipage}
\hfill
\caption{Qualitative comparisons on two-view based motion segmentation using a subset of the data, namely Cubetoy, Cubechips, Breadcube, Gamebiscuit, Biscuitbookbox and Cubebreadtoychips, in the top-down order. (a) The ground truth segmentation results. (b) to (f) The results obtained by KF, RCG, AKSWH, T-linkage and HF, respectively.}
\label{fig:fundamental}
\end{figure}

In this section, we evaluate the performance of the five fitting methods using real images from the AdelaideRMF datasets \cite{wong2011dynamic}\footnote {\url{http://cs.adelaide.edu.au/~hwong/doku.php?id=data}} for homography based segmentation (see Fig.~\ref{fig:homography}) and two-view based motion segmentation (see Fig.~\ref{fig:fundamental}). The ground truth segmentation is also provided by the authors of \cite{wong2011dynamic}. We randomly select six datasets for each of the two applications, respectively.


The segmentation error is computed using the criterion of \cite{mittal2012generalized} and \cite{Magri_2014_CVPR}:
 \begin{align}
error=\frac{number~of~mislabeled~data~points}{total~number~of~data~points}.
\end{align}

\begin{table}[t]
\centering
\small
\caption{The segmentation errors (in percentage) on homography based segmentation (and the CPU time in seconds). The smallest segmentation errors are boldfaced.}
\centering
\medskip
\begin{tabular}{|c|c|c|c|c|c|c|}
\hline
  & Ladysymon & Sene & Library & Elderhalla& Neem & Johnsona  \\
  \hline
\multirow{2}{*}{KF}& 16.46& 12.08& 13.19& 12.15& 10.25& 25.74 \\
& (3.06)& (5.14)& (3.34)& (3.54)& (6.32)& (16.53)\\
\hline
\multirow{2}{*}{RCG}& 22.36& 10.00& 9.77& 10.37& 11.17& 23.06\\
& (0.83)& (0.82)& (0.71)& (1.66)& (0.83)& (1.36)\\
\hline
\multirow{2}{*}{AKSWH}& 8.44 & 2.00& 5.79& 0.98 & 5.56& 8.55\\
& (2.87)& (2.73)& (2.13)& (2.79)& (2.49)& (2.93)\\
\hline
\multirow{2}{*}{T-linkage}& 5.06 & 0.44& 4.65& 1.17& 3.82 & 4.03\\
&(20.86)& (22.78) & (16.04)& (15.28)& (21.40)& (57.11)\\
\hline
\multirow{2}{*}{HF}& {\bf 3.12}& {\bf 0.36}& {\bf2.93}& {\bf 0.84}& {\bf 2.90}& {\bf 3.75}\\
& (2.27)& (2.15) & (1.70) & (1.92)& (2.19)& (2.49)\\
\hline
\end{tabular}
\\
\label{Table:Homography}
\end{table}
We repeat each experiment 50 times. We show the average results of segmentation errors and the computational speed{, i.e., the CPU time (as before}, we exclude the time used for sampling and generating potential hypotheses which is the same for all the fitting methods) in Table~\ref{Table:Homography} and Table~\ref{Table:Fundamental}. The corresponding best segmentation results obtained by the five fitting methods are also shown in Fig.~\ref{fig:homography} and Fig.~\ref{fig:fundamental}.

\begin{table}[t]
\small
\caption{The segmentation errors (in percentage) on two-view based motion segmentation (and the CPU time in seconds). The smallest segmentation errors are boldfaced.}
\centering
\medskip
\begin{tabular}{|c|c|c|c|c|c|c|}
\hline
 & D1 & D2 & D3 & D4 & D5 & D6  \\
\hline
\multirow{2}{*}{KF}& 12.53 & 8.42 & 14.83 & 13.78& 16.06& 31.07\\
& (6.08)& (7.94)& (7.07)& (7.66) & (8.50)& (25.68)\\
\hline
\multirow{2}{*}{RCG}& 13.35& 13.43 & 12.60& 9.94& 16.87& 37.95\\
& (1.34)& (1.69)& (1.53)& (2.36)& (1.71)& (1.83)\\
\hline
\multirow{2}{*}{AKSWH}& 7.23 & 4.72& 5.45& 7.01 & 8.54& 14.95\\
& (4.97)& (5.10)& (6.10)& (6.44)& (5.11)& (5.99)\\
\hline
\multirow{2}{*}{T-linkage}& 5.62 & 5.63  & 4.96& 7.32& {\bf1.93}& {\bf3.11}\\
& (51.65)&(64.87)& (46.17)& (91.49)&(53.44)& (91.05)\\
\hline
\multirow{2}{*}{HF}& {\bf 2.45} & {\bf 4.23}& {\bf2.23}& {\bf 6.59}& {\bf 1.93} & 3.67\\
& (4.87)& (4.98)& (5.42)& (5.59) &(4.98)& (5.56)\\
\hline
\end{tabular}
\\
\medskip
(D1-Cubetoy; D2-Cubechips; D3-Breadcube; D4-Gamebiscuit; D5-Biscuitbookbox; D6-Cubebreadtoychips.)
\label{Table:Fundamental}
\end{table}
\subsubsection{Homography based segmentation}
\label{sec:homographbasedsegmentation}
As shown in Fig.~\ref{fig:homography} and Table~\ref{Table:Homography}, KF succeeds in fitting four datasets (i.e., the ``Ladysymon'', the ``Sene'', the ``Library'' and the ``Neem'' data), but it fails in fitting the other two datasets (i.e., the ``Elderhalla'' and the ``Johnsona" data). And it also achieves high average segmentation errors for all the six datasets. The reason why KF achieves the bad average segmentation errors is that outliers are often clustered with inliers when KF uses the proximity sampling technique, and many inliers are often wrongly removed or outliers are wrongly recognized as inliers by KF. We note that RCG cannot achieve stable fitting results when the model hypotheses contain many bad structures, and it succeeds in fitting $5$ out of $6$ datasets. However it achieves high average segmentation errors for all the six datasets. RCG often wrongly estimates the potential structures in data during {detection of} the dense subgraphs. AKSWH obtains good performance {and succeeds in} fitting all the six data. However, sometimes two potential structures in {the} data are clustered together by AKSWH ({i.e.,} in the ``Ladysymon'' and the ``Johnsona" data), which {increases} the average fitting errors. In contrast, both T-linkage and HF obtain good performance, but HF achieves the lowest average segmentation errors for all the six datasets.

For the computational efficiency, HF takes similar computational time as AKSWH. However, HF is about $1.3$-$6.6$ times faster than KF and it is more than one order faster than T-linkage for all six datasets. We note that the gap of computational efficiency between HF and T-linkage is larger for the datasets that contain a larger number of data points, e.g., the ``Johnsona" data (HF is about $22.9$ times faster than T-linkage). This is because that the agglomerative clustering step in T-linkage {takes} more time to deal with {a} large number of data points. RCG is faster than HF for all the six datasets but it yields {much} larger average segmentation errors.

\subsubsection{Two-view based motion segmentation}
\label{sec:motionsegmentation}

From Fig.~\ref{fig:fundamental} and Table~\ref{Table:Fundamental}, we can see that KF achieves high average segmentation errors in all the six datasets, although it can succeed in fitting some datasets (i.e., the ``Cubetoy'', the ``Cubechips'', the ``Breadcube'' and the ``Gamebiscuit'' data). The model instances estimated by KF often overlap to one structure in some datasets (i.e., the ``Biscuitbookbox'' and the ``Cubebreadtoychips'' data). RCG achieves the worst results (achieving {the highest} average segmentation errors in fitting {four} of the six datasets). RCG cannot effectively detect dense subgraphs (representing potential structures in data) for model fitting when there exists a large proportion of bad model hypotheses because {the} bad model hypotheses may lead to an inaccurate similarity measure between data points (the similarity measure plays an important role in RCG). AKSWH achieves better fitting results than RCG and KF, and it can obtain low segmentation errors in $5$ out of $6$ datasets.
However, AKSWH misses one structure in the ``Cubebreadtoychips'' data. {In this case, most} of model hypotheses (generated for the structure with a small number of inlier data points) are removed when AKSWH selects significant model hypotheses, and the few remained model hypotheses are wrongly clustered. Both T-linkage and HF can effectively estimate the model instances and achieve low segmentation errors in all the six datasets. However, HF obtains the lowest segmentation errors in $5$ out of $6$ datasets. Moreover, HF is also very efficient, i.e., HF achieves the second fastest speed among the five fitting methods in all six datasets (RCG achieves the fastest speed while it cannot effectively estimate the model instances in most cases).

\subsubsection{{3D-motion segmentation}}
\label{sec:3Dmotionsegmentation}
\begin{figure}[t]
\centering
\begin{minipage}[t]{.243\textwidth}
  \centering
 \centerline{\includegraphics[width=1.27\textwidth]{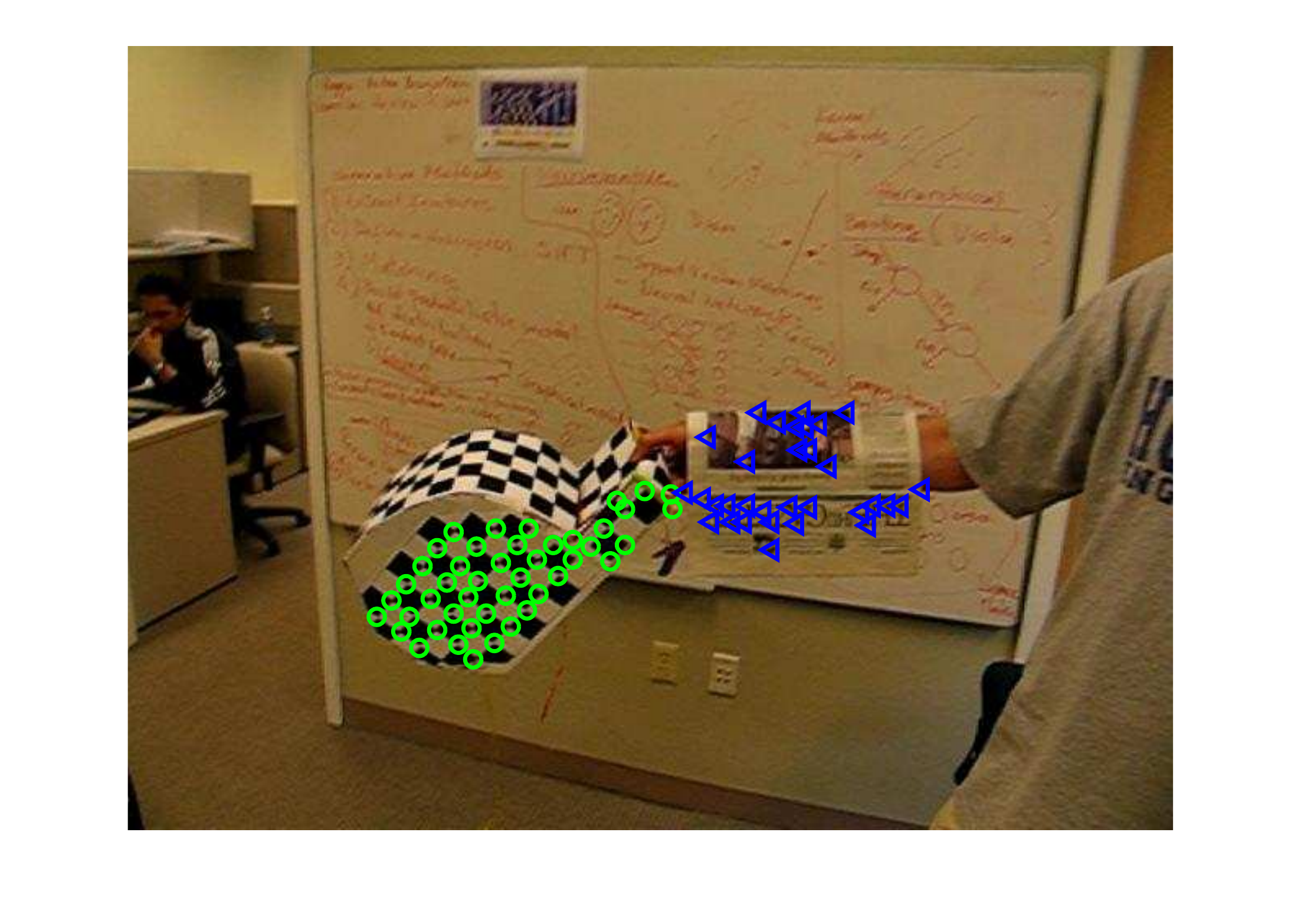}}
  \begin{center}{ (a) Arm} \end{center}
\end{minipage}
\begin{minipage}[t]{.243\textwidth}
  \centering
 \centerline{\includegraphics[width=1.27\textwidth]{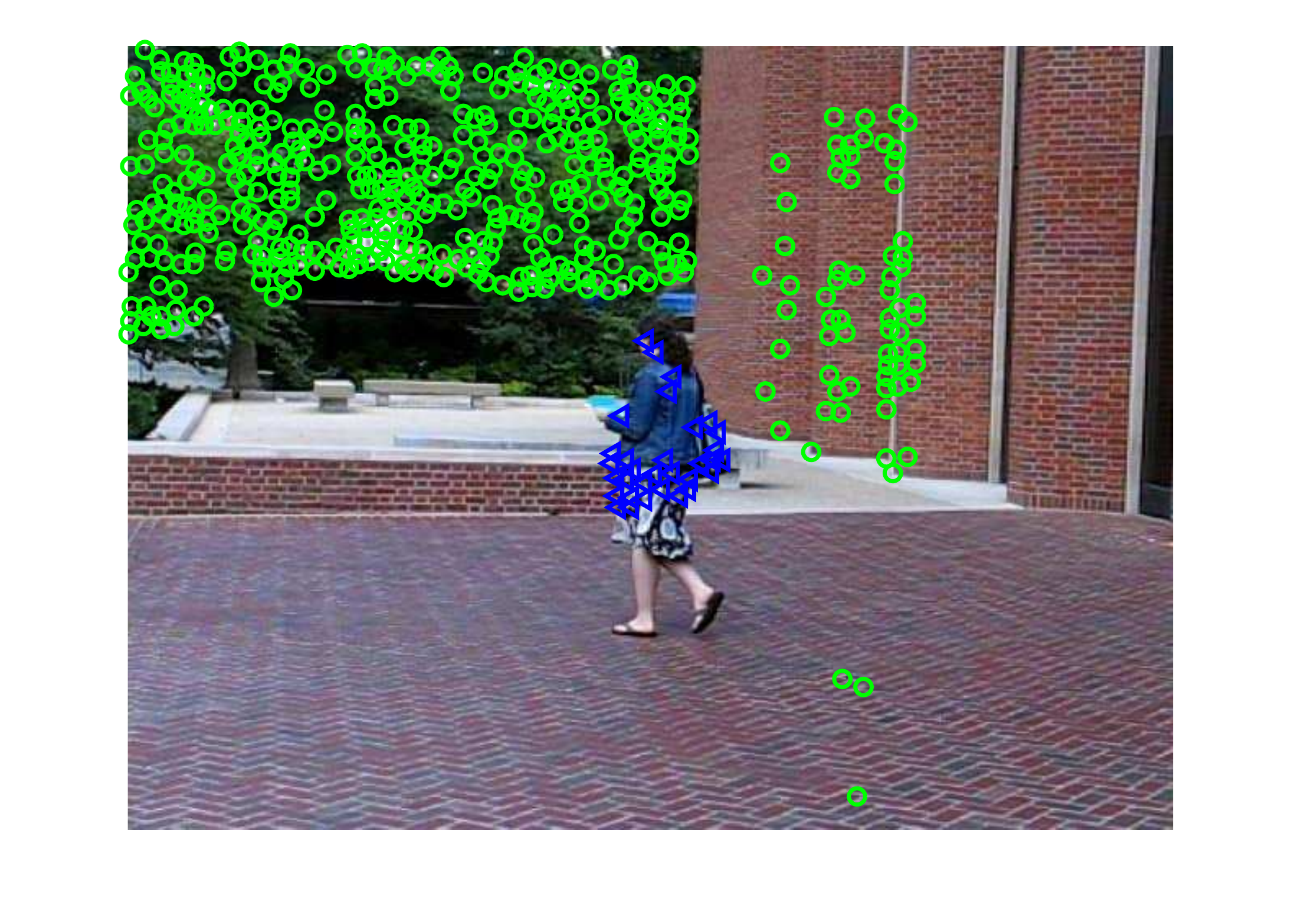}}
  \begin{center} {(b) People1} \end{center}
\end{minipage}
\begin{minipage}[t]{.243\textwidth}
 \centering
 \centerline{\includegraphics[width=1.27\textwidth]{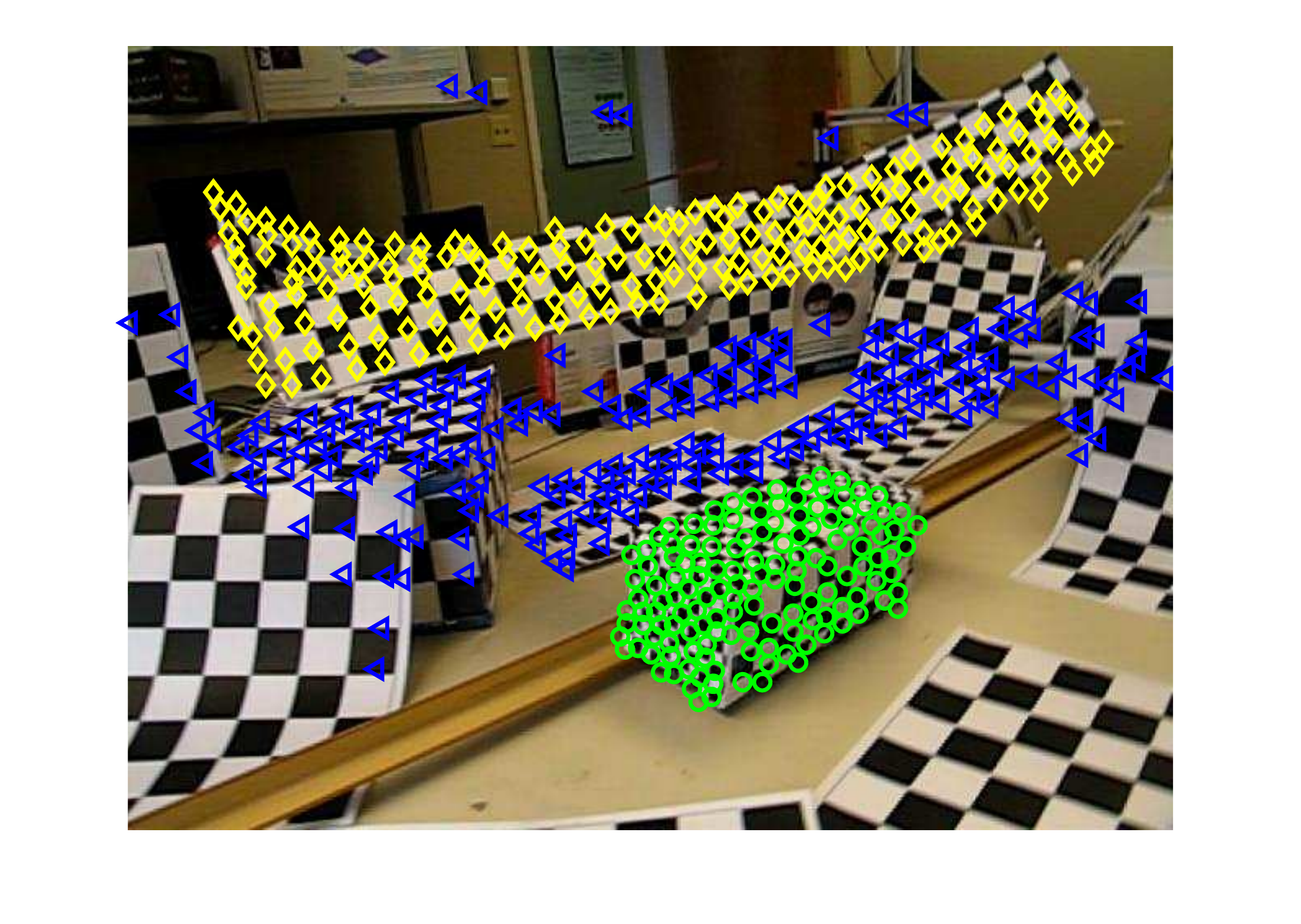}}
  \begin{center}{ (c) 2T3RTCR}  \end{center}
\end{minipage}
\begin{minipage}[t]{.243\textwidth}
  \centering
 \centerline{\includegraphics[width=1.27\textwidth]{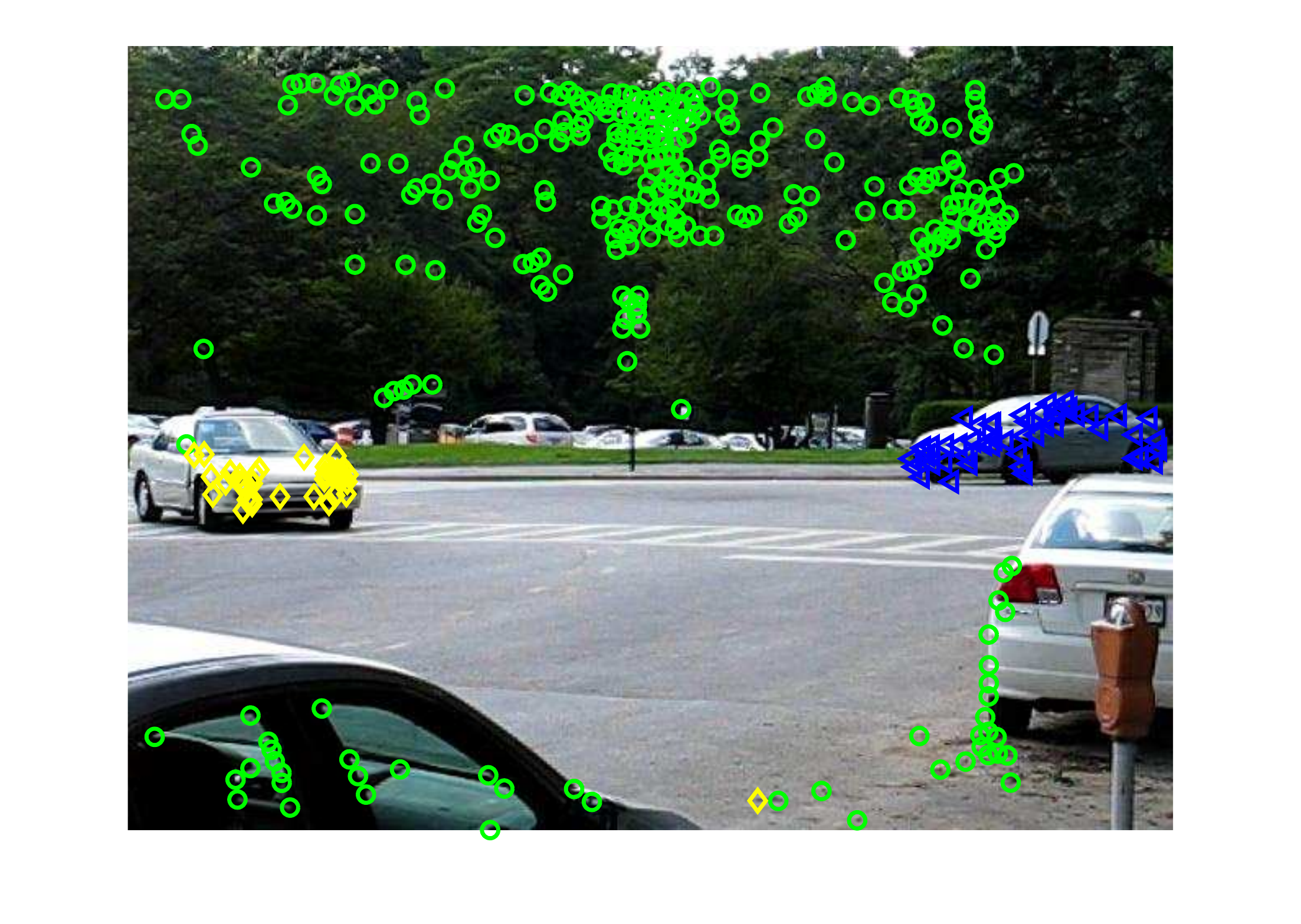}}
  \begin{center} {(d) Cars5}  \end{center}
\end{minipage}
\hfill
\caption{{Some results of 3D-motion segmentation obtained by the proposed method for several sequences in the Hopkins 155 dataset are shown.}}
\label{fig:3Dexample}
\end{figure}

{
In this sub-section, we evaluate the proposed method on 3D-motion segmentation. Similar to \cite{tennakoon2015robust,Magri_2014_CVPR}, we formulate the problem of 3D-motion segmentation as a subspace clustering problem. We evaluate HF on the Hopkins 155 motion dataset~\cite{tron2007benchmarkc}\footnote {{\url{http://www.vision.jhu.edu/data/hopkins155}}}, and we show some results obtained by HF on Fig.~\ref{fig:3Dexample}. We can see that HF successfully estimates the subspace on the motion datasets with two motions (i.e., ``Arm" and ``People1") and three motions (i.e., ``2T3RTCR" and ``Cars5").}

{
To provide a qualitative measure of the performance of the proposed fitting method, we also use the ``checkerboard" image sequence as \cite{tennakoon2015robust}, and we are able to compare directly to PM~\cite{tennakoon2015robust} and indirectly to RANSAC, Energy minimization~\cite{delong2012fast}, QP-MF~\cite{yu2011global} and SSC~\cite{elhamifar2013sparse}\footnote {{Here, we use the same input datasets as PM,  and thus we can compare the proposed method to PM directly. The results obtained by RANSAC, Energy minimization, QP-MF and SSC are taken from \cite{tennakoon2015robust}, by which we can compare the proposed HF with them indirectly.}}. Table~\ref{Table:3Dmotion} shows the results obtained by all the competing methods. The results (except the results obtained by AKSWH and HF) are taken from \cite{tennakoon2015robust} (here, AKSWH is performed by formulating the problem of 3D-motion segmentation as a subspace clustering problem). For the datasets with two motions, as shown in Table~\ref{Table:3Dmotion}, we can see that HF achieves the second lowest average segmentation error (which is only inferior to that obtained by SSC) in all eight fitting methods. However, HF achieves a zero median segmentation error as SSC does. In contrast, AKSWH cannot achieve good results. This is because inliers and outliers are hard to be distinguished when the residual values from the data points to the hypotheses are small and confusingly close to each other, which causes that AKSWH cannot effectively cluster hypotheses based on the information derived from inliers. For the datasets with three motions, HF achieves the second best average result and the fourth lowest median error in all eight fitting methods. In contrast, AKSWH achieves the second worst average result (which is only superior to that obtained by RANSAC) and the eighth median error in all eight fitting methods. Therefore, the superiority of HF over AKSWH on 3D-motion segmentation is obvious. Compared with SSC which shows better performance, HF can estimate the number of model instances and is more robust to outliers.
}

\begin{table}[t]
\footnotesize
\centering
\caption{{The segmentation errors (in percentage) on 3D-motion segmentation.}}
\centering
\medskip
\begin{tabular}{|c|c|c|c|c|c|c|c|c|}
\hline
 & {RANSAC} & {Enargy} & {QP-MF} & {SSC} & {PM T1} & {PM T2} & {AKSWH} & {HF} \\
\hline
\multicolumn{9}{|c|}{{2 Motions}}\\
\hline
 {Mean}& {6.52}  & {5.28 }& {9.98 }& {2.23 }&{3.98 }&{ 3.88 }&{ 10.96 }&{ 2.57} \\
\hline
 {Median} &{ 1.75 }&{ 1.83 }&{ 1.38  }&{0.00 }&{ 0.00 }&{ 0.00 }&{ 4.64 }&{ 0.00} \\
\hline
\multicolumn{9}{|c|}{{3 Motions}}\\
\hline
{ Mean} &{ 25.78 }&{ 21.38 }&{ 15.61 }&{ 5.77 }&{ 11.06 }&{ 6.81 }&{ 22.28 }&{ 6.75} \\
\hline
 {Median}&{ 26.01 }&{ 21.14 }&{ 8.82 }&{ 0.95 }&{ 1.20 }&{ 1.04 }&{ 26.06 }&{ 4.91} \\
\hline
\end{tabular}
\\
\label{Table:3Dmotion}
\end{table}

\subsection{Failure cases}
\begin{figure}[t]
\centering
\begin{minipage}[t]{.243\textwidth}
 \centering
 \centerline{\includegraphics[width=1.27\textwidth]{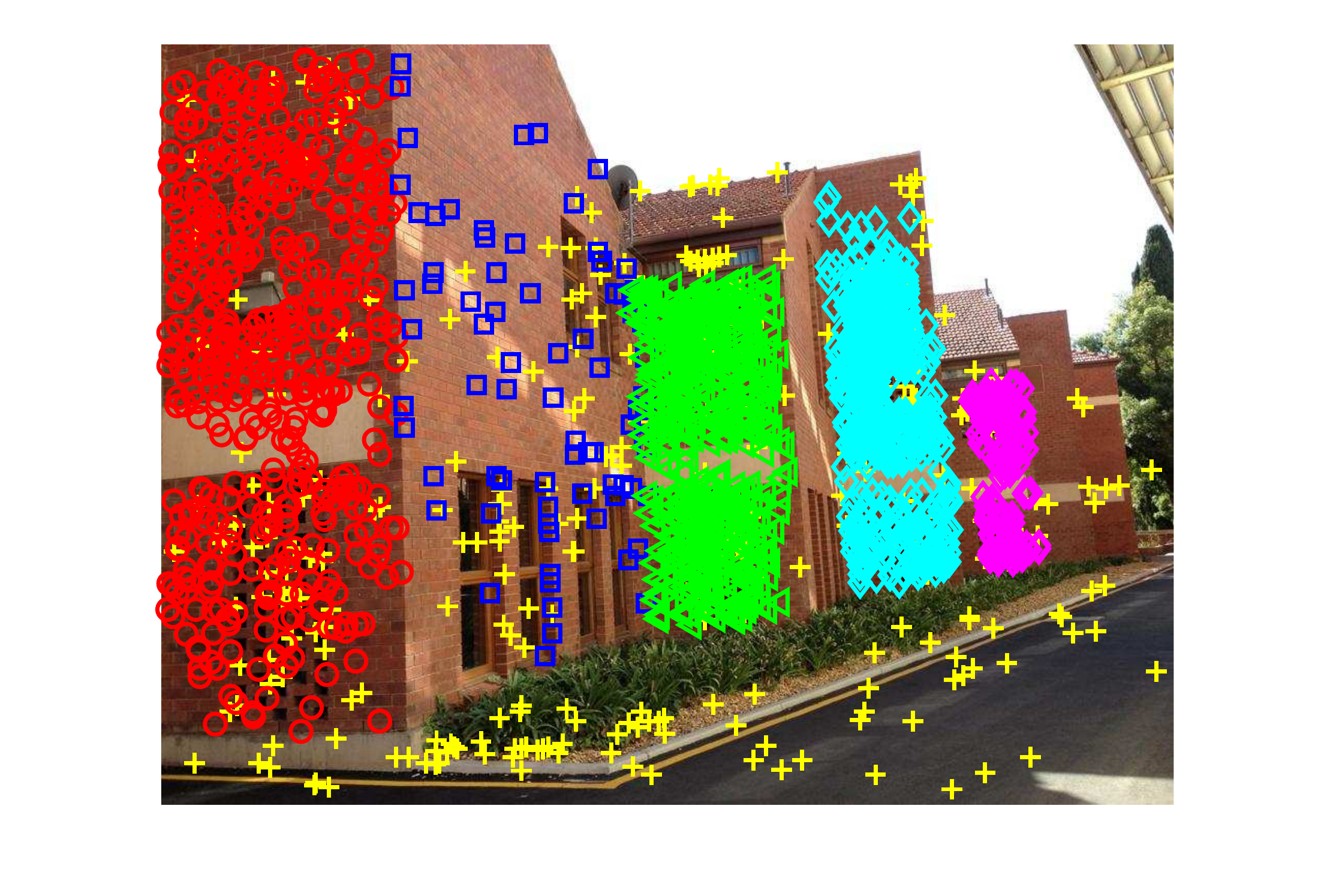}}
  \begin{center} (a)  \end{center}
\end{minipage}
\begin{minipage}[t]{.243\textwidth}
  \centering
 \centerline{\includegraphics[width=1.27\textwidth]{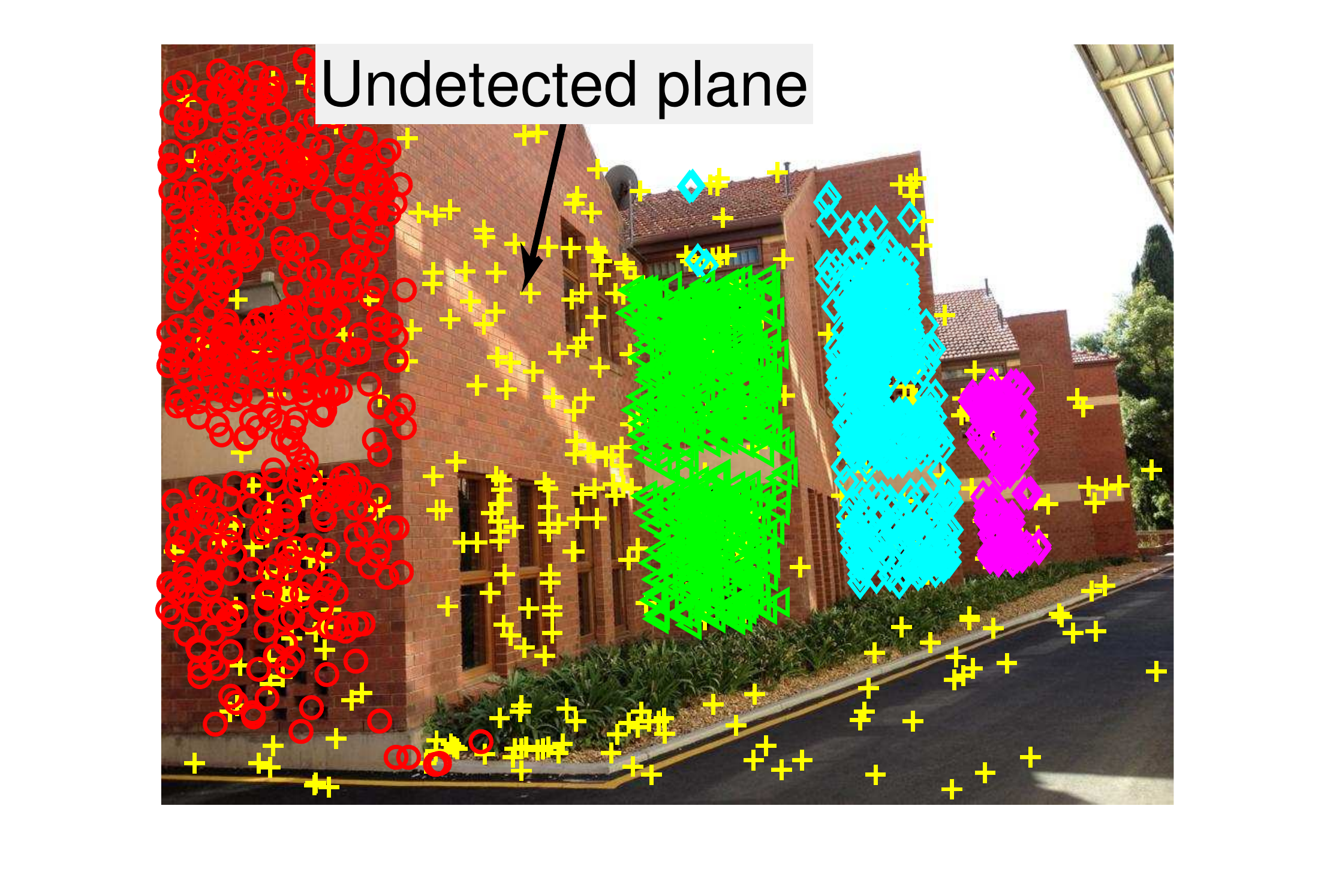}}
  \begin{center} (b)  \end{center}
\end{minipage}
\begin{minipage}[t]{.243\textwidth}
  \centering
 \centerline{\includegraphics[width=1.27\textwidth]{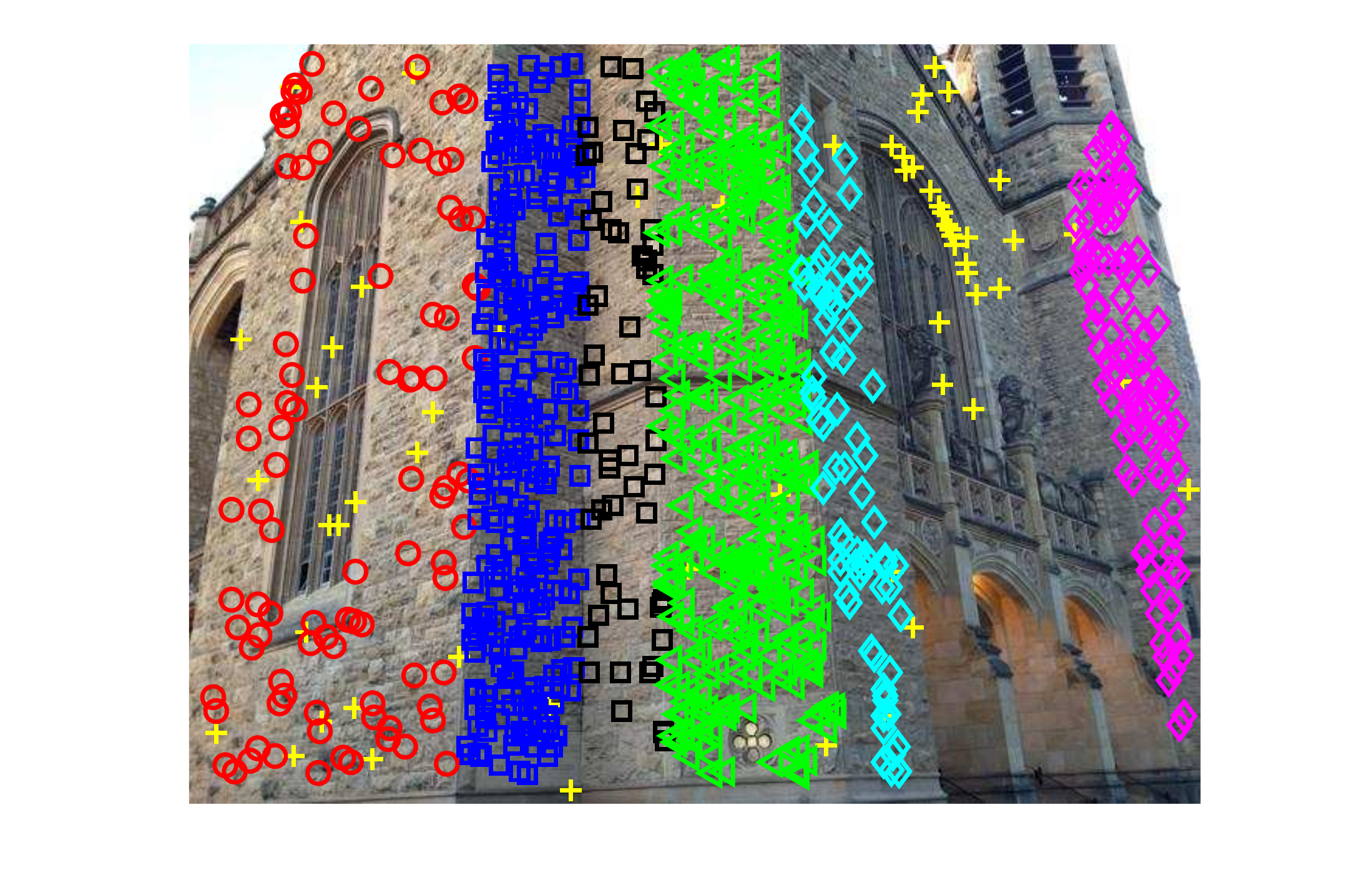}}
  \begin{center} (c) \end{center}
\end{minipage}
\begin{minipage}[t]{.243\textwidth}
  \centering
 \centerline{\includegraphics[width=1.27\textwidth]{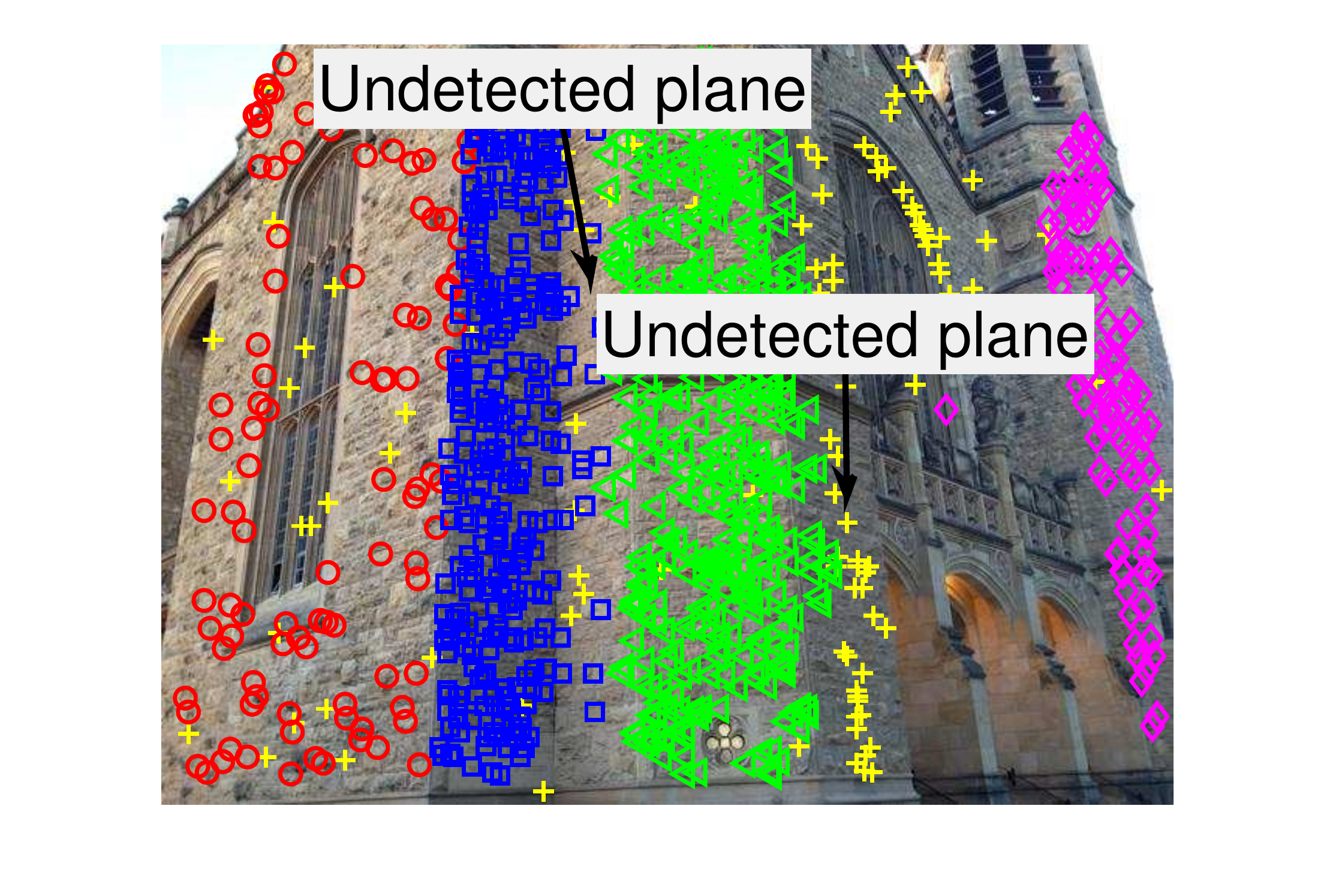}}
  \begin{center} (d)  \end{center}
\end{minipage}
\hfill
\caption{Two examples show that the proposed method fails to estimate the number of model instances in homography based segmentation (only one of the two views is shown for each case). (a) and (c) The ground truth segmentation results on the ``Unionhouse'' data and the ``Bonhall'' data. (b) and (d) The results obtained by HF. The undetected planes are pointed by black arrows.}
\label{fig:failcase}
\end{figure}

In this sub-section, we investigate the circumstances in which the proposed method cannot correctly estimate the number of the model instances in data. Figs.~\ref{fig:failcase}(b) and \ref{fig:failcase}(d) show the failures of HF on the ``Unionhouse" and the ``Bonhall" data. The model instances corresponding to the undetected planes on the ``Unionhouse" and the ``Bonhall" data have very few inliers, which {is} due to the small physical size or lack of textures on the surface of the missing structure~\cite{pham2014random}.
The number of inliers belonging to different model instances in these datasets is extremely unbalanced, which have great influence on the effectiveness of fitting methods. Note that the proposed method selects significant hyperedges by the step of hypergraph pruning to improve the effectiveness and efficiency of sub-hypergraph detection, but this will increase the challenge due to the fact that this step may also remove most of good model hypotheses that correspond to the model instances with few inliers in unbalanced data. In addition, one more challenge involved in these examples comes from the spatial smoothness assumption. The spatial smoothness assumption is too simple to correctly handle complex situations where the key-point data from a plane is broken into separate clusters. These challenges also affect the performance of the other competing fitting methods.

\begin{figure}[t]
\centering
\begin{minipage}[t]{.243\textwidth}
  \centering
 \centerline{\includegraphics[width=1.27\textwidth]{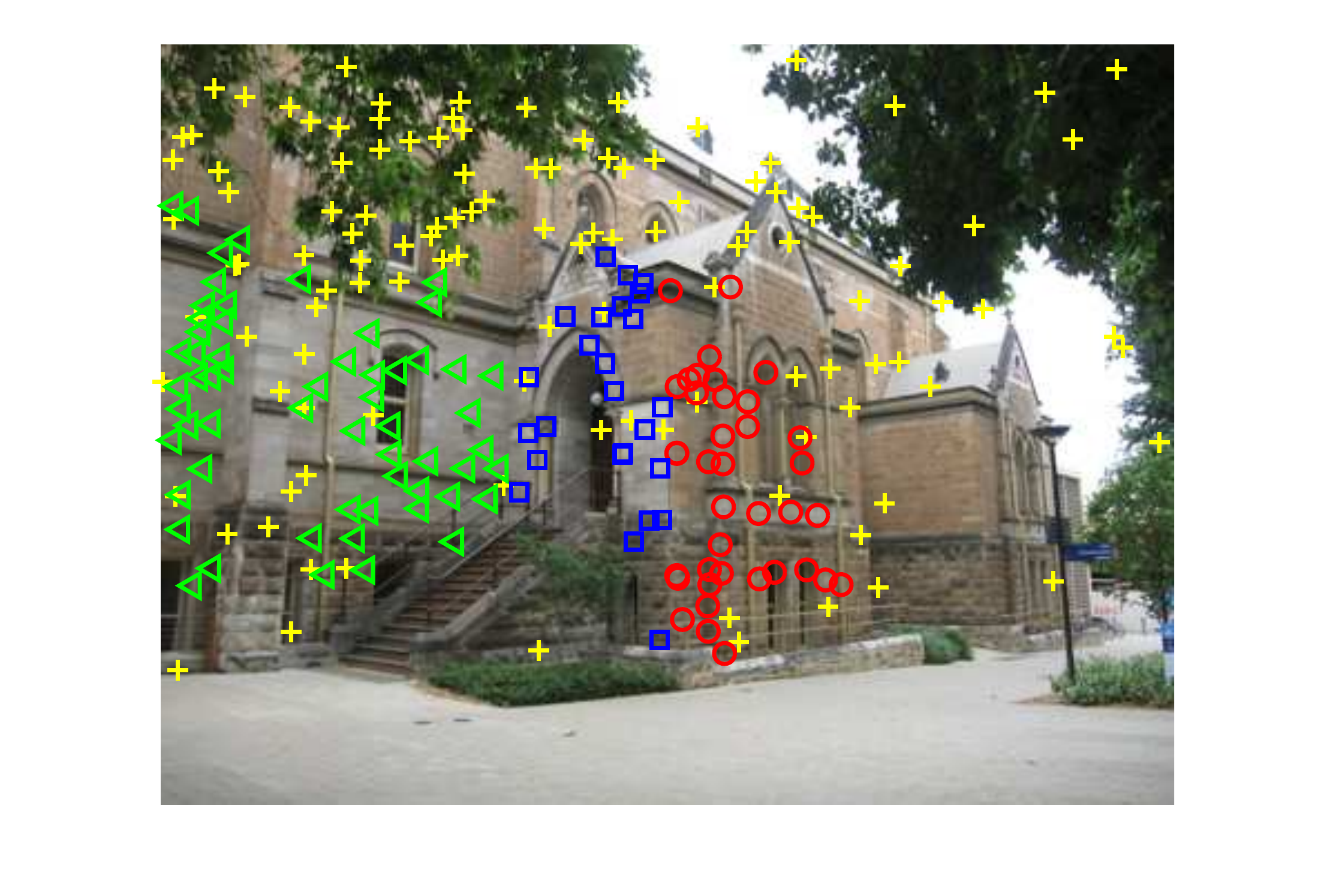}}
  \begin{center} (a)  \end{center}
\end{minipage}
\begin{minipage}[t]{.243\textwidth}
  \centering
 \centerline{\includegraphics[width=1.27\textwidth]{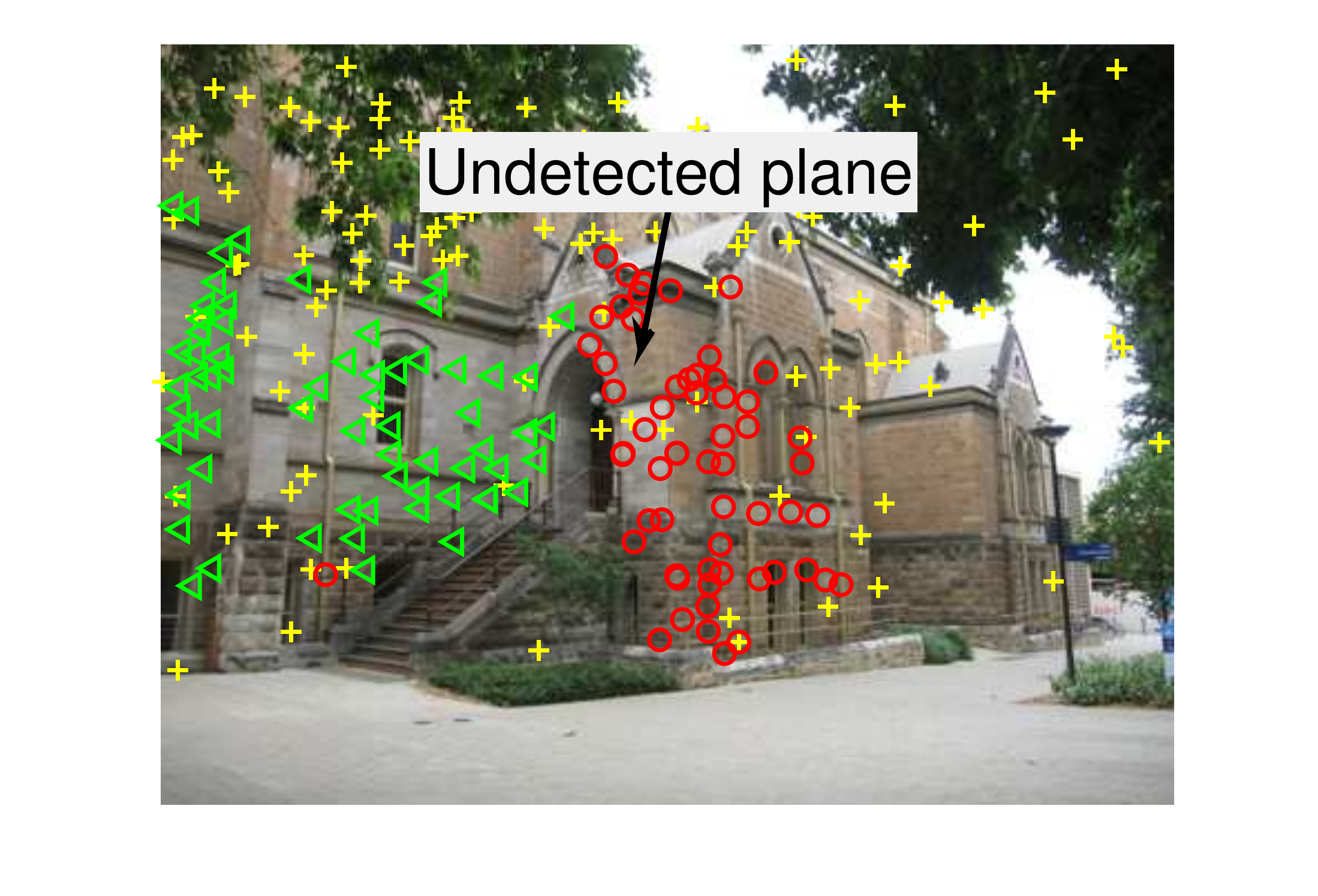}}
  \begin{center} (b)\end{center}
\end{minipage}
\begin{minipage}[t]{.243\textwidth}
  \centering
 \centerline{\includegraphics[width=1.27\textwidth]{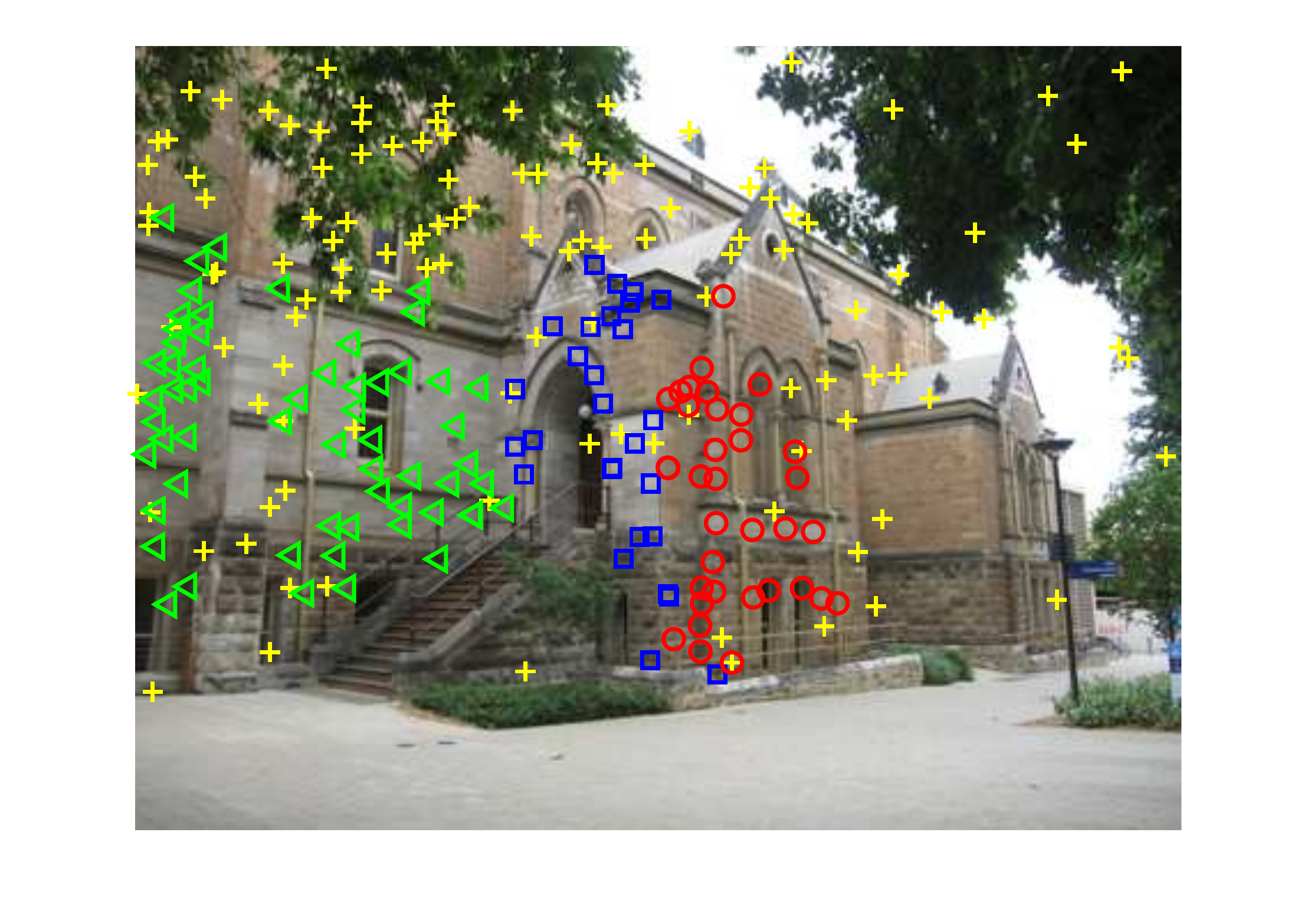}}
  \begin{center} (c) \end{center}
\end{minipage}
\hfill
\caption{One example shows that the proposed method deals with the ``Elderhallb" data in which the inliers belonging to two model instances are not separated well in homography based segmentation (only one of the two views is shown). (a) The ground truth segmentation results on the ``Elderhallb" data. (b) and (c) The results obtained by HF without and with providing the correct number of model instances beforehand, respectively.}
\label{fig:succeedexample}
\end{figure}

We also consider the circumstance in which the proposed method deals with the data where the inliers belonging to two model instances are not separated well. Fig.~\ref{fig:succeedexample}(a) shows an example of this circumstance on the ``Elderhallb'' data. We note that the proposed method improves the stability to effectively estimate the number of model instances in the data by eliminating duplicate hyperedges (by the step $6$ in Algorithm~\ref{alg:summarize}). This step can effectively help the proposed method improve the average fitting accuracy in most cases. However, this step may also fuse two model instances when their inliers are not separated well, as shown in Fig.~\ref{fig:succeedexample}(b). This is because the data points near the intersection of two model instances wrongly increase the value of the mutual information shared by the two corresponding model hypotheses. Note that the other fitting methods (such as KF and AKSWH) cannot effectively deal with data points near the intersection of two model instances as well, while it can be counterbalanced by the proposed method if the number of model instances in data is provided in advance (i.e., replacing the step $4$ of Algorithm~\ref{alg:subhypergraphdection} by using a user-specified model instance number) according to the ground truth, as shown in Fig.~\ref{fig:succeedexample}(c).

%
%
\section{Conclusions}
\label{sec:conclusion}
In this paper, we have presented a novel hypergraph based fitting method (HF) for geometric model fitting. HF formulates the geometric model fitting problem as a hypergraph partition problem based on a hypergraph model, in which each vertex represents a data point and each hyperedge denotes a model hypothesis. The hypergraph is derived from model hypotheses and the corresponding inlier data points, and it contains large and ``data-determined'' degrees of hyperedges, which can effectively express the relationships {between} model hypotheses and data points. We have also developed a robust hypergraph partition algorithm for effective sub-hypergraph detection. The proposed method can adaptively estimate the number of model instances in data and in parallel estimate the parameters of each model instance.

HF deals with the partitioning problem with a variant of the hypergraph partition algorithm introduced in \cite{zhou2007learning}. The {previous} hypergraph partition algorithms are often not suitable for model estimation problems, since {those} algorithms tend to find a balanced cut~\cite{parag2011supervised}. However, HF separates data points into $k$ clusters simultaneously instead of a two-way partition. In addition, the {hypergraph} constructed in this paper {contains hyperedges of large degrees and removes} a large number of outliers in data by hypergraph pruning (which are beneficial to the global clustering). Overall, HF can effectively solve the partitioning problem for model fitting. {In terms of} computational time, HF is very efficient {because it adopts} the effective hypergraph pruning and {sub-hypergraph detection}. The experimental results have shown that HF generally performs better than the other competing methods on both synthetic data and real images, and HF is faster than most of the state-of-the-art fitting methods.

\section*{Acknowledgment}
\indent This work was supported by the National Natural Science Foundation of China under Grants 61472334 {and 61170179}, and supported by the Fundamental Research Funds for the Central Universities under Grant 20720130720. David Suter acknowledged funding under ARC DPDP130102524.
\section*{References}
{\small
\bibliographystyle{elsarticle-num}

}


\end{document}